\documentclass[letterpaper,10pt]{article} 

\usepackage{opticameet3} 

\usepackage{bm}
\usepackage{subcaption} 
\usepackage{amsmath,amssymb}
\usepackage{booktabs}
\usepackage{mathptmx}
\usepackage{newtxmath}    
\usepackage[colorlinks=true,bookmarks=false,citecolor=blue,urlcolor=blue]{hyperref} 

\DeclareCaptionLabelFormat{empty}{}
\begin{document}

\pagestyle{plain}

\title{Flow Matching and Diffusion Models via PointNet for Generating Fluid Fields on Irregular Geometries}

\author{Ali Kashefi}
\address{Stanford University, Stanford, 94305, CA, USA}
\email{kashefi@stanford.edu}

\vspace{-0.6em}
\begin{center}
\footnotesize
\url{https://github.com/Ali-Stanford/Diffusion_Flow_Matching_PointNet_CFD}
\end{center}

\begin{abstract}
We present two novel generative geometric deep learning frameworks, termed Flow Matching PointNet and Diffusion PointNet, for predicting fluid flow variables on irregular geometries by incorporating PointNet into flow matching and diffusion models, respectively. In these frameworks, a reverse generative process reconstructs physical fields from standard Gaussian noise conditioned on unseen geometries. The proposed approaches operate directly on point-cloud representations of computational domains (e.g., grid vertices of finite-volume meshes) and therefore avoid the limitations of pixelation used to project geometries onto uniform lattices, as is common in U-Net-based flow matching and diffusion models. In contrast to graph neural network–based diffusion models, Flow Matching PointNet and Diffusion PointNet do not exhibit high-frequency noise artifacts in the predicted fields. Moreover, unlike such approaches, which require auxiliary intermediate networks to condition geometry, the proposed frameworks rely solely on PointNet, resulting in a simple and unified architecture. The performance of the proposed frameworks is evaluated on steady incompressible flow past a cylinder, using a geometric dataset constructed by varying the cylinder's cross-sectional shape and orientation across samples. The results demonstrate that Flow Matching PointNet and Diffusion PointNet achieve more accurate predictions of velocity and pressure fields, as well as lift and drag forces, and exhibit greater robustness to incomplete geometries compared to a vanilla PointNet with the same number of trainable parameters.
\end{abstract}


\section{Introduction and motivation}
\label{Sect1}

Flow matching \cite{lipman2023flow}, diffusion models \cite{ho2020denoising}, and their advanced variants \cite{liu2022flow,rombach2022high,song2020score} are generative deep learning frameworks that have demonstrated success in computer graphics and computer vision, particularly for image and video generation \cite{ho2022video}, outperforming earlier generative approaches such as generative adversarial networks (GANs) \cite{goodfellow2014generative,dhariwal2021diffusion} and variational autoencoders (VAEs) \cite{KingmaW13}. At a high level, in supervised flow matching \cite{lipman2023flow} and diffusion \cite{ho2020denoising} frameworks, a forward process adds noise to the training data, while an inverse process predicts the clean fields. Several formulations of flow matching and diffusion models operate in a conditional setting, where generation is guided by auxiliary inputs such as textual prompts \cite{rombach2022high,nichol2022point} or low-resolution images \cite{saharia2022image,DiffusionReview}. Motivated by their success in computer graphics, these models have been extended to computational physics, enabling a range of applications including predictions conditioned on material properties \cite{liu2024uncertainty,zeni2025generative,kartashov2025large}, high-fidelity field reconstruction from low-resolution data \cite{shu2023physicsDiffusion,erichson2025flex}, and prediction of physical fields conditioned on the geometry of the computational domain \cite{liu2024uncertainty,wang2025aerodit,hu2025generative}. In this article, we focus on the latter case.

Flow matching and diffusion models conditioned on the geometry of the computational domain fall within the category of geometric deep learning. These approaches are critical in computational physics, as they accelerate shape-optimization studies that require exploration of high-dimensional geometric parameter spaces \cite{allaire2021shape}. Generative flow matching and diffusion models conditioned on geometry have been shown to outperform regression-based surrogate models for predicting physical fields (see e.g., Refs. \cite{jadhav2023stressd,zhou2024text2pde,li2025generativeFlowMatch}), such as convolutional neural networks (CNNs) \cite{bhatnagar2019prediction}, Fourier neural operators (FNOs) \cite{li2020fourier}, and deep operator neural networks (DeepONets) \cite{lu2021learning}, which directly map geometry to physical fields. In particular, traditional CNNs \cite{bhatnagar2019prediction}, FNOs \cite{li2020fourier}, and DeepONets \cite{lu2021learning} produce a single output for a given geometry, whereas flow matching and diffusion models can generate multiple realizations for the same geometry, thereby quantifying model uncertainty. Moreover, diffusion models have demonstrated superior performance compared to GANs and VAEs in predicting physical field variables (see, e.g., Refs. \cite{jadhav2023stressd,drygala2024comparison}). In geometric deep learning using flow matching and diffusion models, two steps are particularly important. First, how to represent irregular geometries; and second, how to condition flow matching and diffusion models on geometry from a computer science perspective.

 For the representation of irregular geometries, two common approaches are used. The first is a pixelation-based strategy, in which the geometry is projected onto uniform Cartesian grids. In this setting, Q. Liu and N. Thuerey \cite{liu2024uncertainty} proposed a diffusion model conditioned on the geometry of airfoils to predict velocity and pressure fields. In their work \cite{liu2024uncertainty}, the airfoil geometry was represented using the pixelation approach, and a U-Net architecture \cite{ronneberger2015u,ho2020denoising} was employed as the backbone neural network. The authors \cite{liu2024uncertainty} performed geometric conditioning by concatenating the binary image of the airfoil with the velocity and pressure fields as the input to the U-Net, while noise was added only to the velocity and pressure fields during training (see Fig. 3 of Ref. \cite{liu2024uncertainty}). A similar approach was taken by Zheng et al. \cite{wang2025aerodit}, where the authors employed an encoder that maps a two-dimensional binary image of the geometry to a latent feature representation, which was then used in a diffusion transformer to condition the diffusion model on the geometry (see Fig. 1 of Ref. \cite{wang2025aerodit}). Similarly, J. Hu et al. \cite{hu2025generative} proposed a diffusion model conditioned on the geometry of obstacles to predict incompressible flow fields. The authors \cite{hu2025generative} conditioned the diffusion model on geometry by encoding the obstacle as a binary mask through a U-Net–style encoder and injecting this geometry representation into the diffusion process via cross-attention blocks (see Fig. 1 of Ref. \cite{hu2025generative}). Their results demonstrated that the proposed diffusion model outperformed CNNs and VAEs \cite{hu2025generative}.

However, pixelation strategies lose smooth boundary information (see e.g., Fig. 17 of Ref. \cite{liu2024uncertainty} and Fig. 3 of Ref. \cite{wang2025aerodit}) and require high global resolution to capture local geometric details (e.g., the trailing edge of an airfoil), leading to increased computational cost. In addition, training data must be interpolated onto uniform grids, introducing additional errors \cite{kashefi2021PointNet,kashefi2025kolmogorov}. In contrast, graph and point cloud representations enable precise geometry modeling with adaptive resolution, eliminate the need for interpolation, and thereby reduce both approximation errors and computational cost \cite{kashefi2021PointNet,kashefi2025kolmogorov}. To avoid the limitations of the first approach, the second approach has been proposed, which represents geometries using unstructured meshes or point clouds, and leverages graph neural networks \cite{scarselli2008graph,xu2018how,wu2020comprehensive} or PointNet \cite{qi2017pointnet,qi2017pointnet++,kashefi2021PointNet,kashefi2022physics,kashefi2023PIPNelasticity,kashefi2025pointnetKAN}. M. Lino et al. \cite{valencia2025learning} proposed a diffusion graph network and a flow matching graph network for predicting velocity and pressure fields around airfoils with a variety of geometries \cite{valencia2025learning}. However, the authors \cite{valencia2025learning} faced shortcomings due to the limited performance of message passing for propagating features across the graph, as this led to high-frequency noise in the velocity and pressure field predictions. To overcome this challenge, the authors \cite{valencia2025learning} introduced an alternative version in which noise was added during training in the latent space after encoding the physical fields and the geometry of the domain by a graph neural network. Consequently, their neural network architecture became more complicated than a pure graph neural network and required auxiliary neural networks, such as a variational autoencoder at intermediate levels (see Fig. 2 of Ref. \cite{valencia2025learning}). In the same line of research, J. Kim et al. \cite{kim2025point} introduced a point-wise diffusion model in which the geometry of the physical domain was represented as a set of points; however, each point was treated independently, with its own forward and inverse diffusion process, an independent positional encoding, and an independent diffusion transformer block. Although the method yielded successful results, the neural network architecture was complex (see Fig. 3 of Ref. \cite{kim2025point}) and did not account for the fact that physical field predictions (e.g., velocity and pressure) depended on global geometric features, since noise addition and denoising were not functions of the entire geometry. Moreover, the independent treatment of points made it unclear how the reverse process could be used to generate multiple samples and quantify model uncertainty, which is a key advantage of diffusion models over other surrogate approaches. This trend motivates us to introduce a simple and elegant neural network based on flow matching and diffusion models for irregular geometries.

We propose, for the first time, Flow Matching PointNet and Diffusion PointNet, in which the segmentation branch of PointNet \cite{qi2017pointnet} is employed as the denoising network within flow matching \cite{lipman2023flow} and diffusion \cite{luo2021diffusion} frameworks. PointNet \cite{qi2017pointnet} and its extensions \cite{qi2017pointnet,qi2017pointnet++,zhao2021point,qi2024shapellm,kashefi2025pointnetKAN} are used for classification and segmentation of point clouds in computer graphics. Kashefi et al. \cite{kashefi2021PointNet} were the first to adapt PointNet \cite{qi2017pointnet} for supervised deep learning of incompressible flow. Subsequent studies extended their framework to compressible flows \cite{CompressiblePointNet}, elasticity \cite{DeepONetPointNetElasticity}, physics-informed machine learning \cite{kashefi2022physics,kashefi2023PIPNelasticity,kashefi2023PIPNporous,kashefi2025physicsKAN}, and Kolmogorov–Arnold networks \cite{kashefi2025kolmogorov}.

The Flow Matching PointNet and Diffusion PointNet frameworks consist of a forward process and an inverse denoising process. In Diffusion PointNet, the forward process incrementally corrupts ground-truth velocity and pressure fields in fluid flows by adding standard Gaussian noise. PointNet is then trained to predict the noise added at each diffusion step \cite{ho2020denoising}. This prediction is conditioned on the spatial coordinates of the point cloud, which represent the domain geometry, as well as on a sinusoidal time embedding \cite{vaswani2017attention}. The conditioning is implemented by concatenating the spatial-coordinate vectors and the sinusoidal-embedding vector with the corrupted velocity and pressure fields. In the inverse process, generation begins from samples drawn from a standard Gaussian noise distribution and is conditioned on the spatial coordinates of the point cloud and the sinusoidal embedding. During the inverse procedure, the trained PointNet predicts the noise at each stage and progressively reconstructs the velocity and pressure fields in a manner consistent with the imposed conditioning. In the Flow Matching PointNet model, PointNet is trained to predict a target vector field that corresponds to the temporal derivative of a continuously perturbed field with respect to an auxiliary time variable \cite{lipman2023flow}. In the inverse process, this learned vector field is used to numerically integrate an ordinary differential equation, progressively reconstructing the velocity and pressure fields from an initial standard Gaussian noise distribution \cite{lipman2023flow}, using the same conditioning mechanism as in Diffusion PointNet. To investigate the performance of Flow Matching PointNet and Diffusion PointNet, the classical fluid dynamics problem of incompressible flow past a cylinder is considered, in which the cross-sectional shape of the cylinder varies across the dataset.

Note that combinations of diffusion models and point cloud neural networks in the computer graphics literature have been used in which, during the forward process, noise is added to the spatial coordinates of shapes and, during the inverse process, new three-dimensional (3D) geometries are generated \cite{luo2021diffusion,vahdat2022lion}. In contrast, in the present study, noise is added to and removed from the physical fields, while the spatial coordinates remain clean and unchanged. Additionally, Y. He et al. \cite{he2025pointdiffuse} proposed a framework for semantic segmentation of 3D point clouds using a diffusion model (see Fig. 1 of Ref. \cite{he2025pointdiffuse}). In their approach, the diffusion process is conditioned on two components: first, the geometry of the input represented by the point cloud, and second, the semantic context, which provides a prior estimate of the possible label types for the points in the point cloud. Although this framework is successful in computer graphics, it is not applicable to predicting physical quantities such as velocity and pressure fields, since these quantities are continuous rather than discrete integer labels, as in semantic segmentation.


We summarize the key features and advantages of the proposed Flow Matching PointNet and Diffusion PointNet frameworks as follows. First, both Flow Matching PointNet and Diffusion PointNet predict velocity and pressure fields on unseen geometries more accurately than the baseline (vanilla) PointNet while using the same number of trainable parameters. Second, the proposed frameworks naturally handle irregular geometries and avoid the limitations of CNN-based flow matching and diffusion models that rely on pixelation techniques. Third, in contrast to graph neural network–based diffusion models, the proposed frameworks do not exhibit high-frequency noise artifacts in the predicted physical fields. Fourth, the proposed frameworks employ a simple PointNet architecture and condition the geometry through a straightforward concatenation strategy, eliminating the need for auxiliary neural networks at intermediate levels (e.g., variational autoencoders) for geometric conditioning. Finally, the proposed frameworks use a single, unified forward and inverse process for the entire point cloud, rather than independent diffusion processes for individual points, resulting in more coherent and robust frameworks.

The structure of the rest of this paper is as follows. Section \ref{Sect2} describes how the labeled data are generated for training the proposed deep learning configurations. Section \ref{Sect3} discusses the architectures of the Flow Matching PointNet, Diffusion PointNet, and baseline PointNet frameworks, along with their training strategies. Quantitative and visual results and predictions of these three models, along with a comprehensive comparison from different aspects such as velocity and pressure field predictions, lift and drag computations, and robustness in the presence of distorted point clouds, are presented in Sect. \ref{Sect4}. Lastly, Section \ref{Sect5} concludes the paper with a summary and proposes several ideas for future research directions.

\begin{figure}[h]
  \centering 
      \begin{subfigure}[b]{0.49\textwidth}
        \centering
        \includegraphics[width=\textwidth]{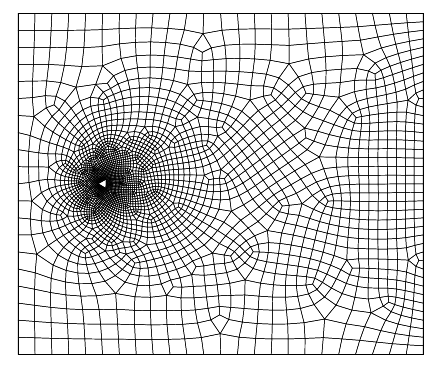}
    \end{subfigure}
    \begin{subfigure}[b]{0.49\textwidth}
        \centering
        \includegraphics[width=\textwidth]{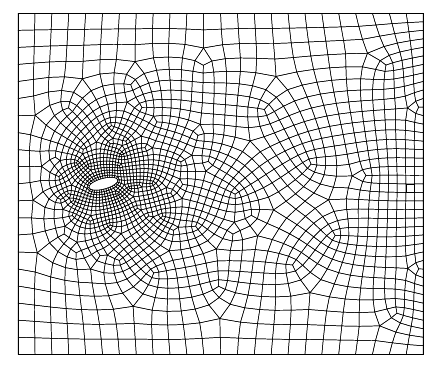}
    \end{subfigure}

    \caption{Finite volume meshes used for the numerical simulation of flow over a cylinder with a triangular cross section, with 2775 vertices on the left panel, and an elliptical cross section, with 2672 vertices on the right panel.}
  
  \label{Fig0}
\end{figure}


\section{Data preparation}
\label{Sect2}

To train the Flow Matching PointNet and Diffusion PointNet models and evaluate their prediction accuracy, labeled data are necessary. For this purpose, we use the well-known steady-state incompressible flow past a cylinder problem as a benchmark example. The motion of an incompressible, viscous, Newtonian fluid is governed by the laws of mass and momentum conservation, together with suitable boundary conditions, as follows:

\begin{equation}
\nabla \cdot \boldsymbol{u} = 0 \quad \text{in } V,
\label{Eq1}
\end{equation}

\begin{equation}
\rho (\boldsymbol{u} \cdot \nabla)\boldsymbol{u} - \mu \Delta \boldsymbol{u} + \nabla p = \boldsymbol{0} \quad \text{in } V,
\label{Eq2}
\end{equation}

\begin{equation}
\boldsymbol{u} = \boldsymbol{u}_{\Gamma_\text{D}} \quad \text{on } \Gamma_\text{D},
\label{Eq3}
\end{equation}

\begin{equation}
-p\boldsymbol{n} + \mu \nabla \boldsymbol{u} \cdot \boldsymbol{n} = \boldsymbol{t}_{\Gamma_\text{Ne}} \quad \text{on } \Gamma_\text{Ne},
\label{Eq4}
\end{equation}
where $\boldsymbol{u}$ and $p$ are the velocity vector and the pressure field in the domain $V$, respectively. The constants $\rho$ and $\mu$ denote the fluid density and dynamic viscosity. The Dirichlet and Neumann boundaries, $\Gamma_\text{D}$ and $\Gamma_\text{Ne}$, are distinct and non-overlapping. The outward unit normal vector on $\Gamma_\text{Ne}$ is $\boldsymbol{n}$, and the traction vector there is $\boldsymbol{t}_{\Gamma_\text{Ne}}$. The velocity components in the $x$ and $y$ directions are represented by $u$ and $v$.

We seek to solve Eqs.~(\ref{Eq1})--(\ref{Eq2}) for flow past an infinite cylinder with different cross-sectional geometries. The computational domain is set as $V = [0, 38~\text{m}] \times [0, 32~\text{m}]$. The cylinder cross-section, centered at $(8~\text{m}, 16~\text{m})$, is modeled as a rigid body with a no-slip boundary condition. The inflow, top, and bottom boundaries are assigned a uniform velocity $u_\infty$ parallel to the $x$-axis. To simulate far-field behavior \cite{KashefiCoarse3}, a stress-free condition is applied at the outflow boundary (i.e., $\boldsymbol{t}_{\Gamma_\text{Ne}} =\boldsymbol{0}$). All physical quantities are expressed in SI units, with $\rho = 1.00~\text{kg/m}^3$, $u_\infty = 1.00~\text{m/s}$, and $\mu = 0.05~\text{Pa}\cdot\text{s}$.



The computational domain $V$ is discretized into unstructured finite volume meshes using Gmsh \cite{geuzaine2009gmsh}. Figure \ref{Fig0} displays two sample meshes corresponding to cylinders with triangular and elliptical cross-sections. The governing Eqs. \ref{Eq1}--\ref{Eq2} are numerically solved using the OpenFOAM package \cite{weller1998tensorial}, which applies the Semi-Implicit Method for Pressure Linked Equations (SIMPLE) \cite{caretto1973two}. The iteration process continues until the $L^2$ norm of the residuals becomes smaller than $10^{-3}$, at which point the flow is assumed steady.

To create labeled data, seven common geometric shapes are used for the cross-section, each with various scales and orientations, as summarized in Table~\ref{Table1}. The selected shapes include circle \cite{behr1995incompressible,KashefiCoarse1,KashefiCoarse3}, square \cite{sen2011flow}, triangle \cite{kumar2006numerical}, rectangle \cite{zhong2019flow}, ellipse \cite{mittal1996direct}, pentagon, and hexagon \cite{abedin2017simulation}. The characteristic length of the geometry is denoted by $L$, and the Reynolds number is defined as
\begin{equation}
    \text{Re} = \frac{\rho L u_\infty}{\mu}.
    \label{Eq6}
\end{equation}
As shown in Table~\ref{Table1}, for circular, square, equilateral triangle, pentagon, and hexagon cross-sections, $L = a$, whereas for rectangles, ellipses, and non-equilateral triangles, $L = b$. The Reynolds number in the dataset ranges from 20.0 to 76.0, with a total of 2235 labeled cases. The dataset is randomly divided into training (1772 samples), validation (241 samples), and test (222 samples) subsets. To prepare point-cloud representations, we use the mesh vertices as points, following the approach of Ref. \cite{kashefi2021PointNet}. To emphasize the regions around the cylinder and its wake, the first $N = 1024$ nearest vertices to the cylinder's center are chosen to form each point cloud. Because the grid vertices are unevenly distributed in the unstructured meshes, the coordinate ranges differ between samples: $x_\text{min} \in [0, 5.46~\text{m}]$, $x_\text{max} \in [10.55~\text{m}, 22.14~\text{m}]$, $y_\text{min} \in [0, 13.41~\text{m}]$, and $y_\text{max} \in [18.57~\text{m}, 32~\text{m}]$.

We define the dimensionless variables $u^{*}$, $v^{*}$, and $p^{*}$ for the velocity and pressure as
\begin{equation}
    u^{*} = \frac{u}{u_\infty},
    \label{Eq7}
\end{equation}
\begin{equation}
    v^{*} = \frac{v}{u_\infty},
    \label{Eq8}
\end{equation}
\begin{equation}
    p^{*} = \frac{p - p_0}{\rho u_\infty^2},
    \label{Eq9}
\end{equation}
where $p_0$ denotes the atmospheric pressure. Next, we normalize the spatial coordinates of the training data to the range $[-1, 1]$ using
\begin{equation}
    \{\phi'\} = 2\left(\frac{\{\phi\} - \min(\{\phi\})}{\max(\{\phi\}) - \min(\{\phi\})}\right) - 1,
    \label{Eq10}
\end{equation}
where $\{\phi\}$ includes $\{x^{*}, y^{*}\}$. Similarly, we normalize the field variables of training data to the range of $[0, 1]$ using

\begin{equation}
    \{\psi'\} = \frac{\{\psi\} - \min(\{\psi\})}{\max(\{\psi\}) - \min(\{\psi\})},
    \label{Eq100}
\end{equation}
where $\{\psi\}$ includes $\{u^{*}, v^{*}, p^{*}\}$.
The scaled data are denoted by $\{x', y', u', v', p'\}$. The predictions are later rescaled back to their physical domain values. The reason for adopting a different scaling for the velocity and pressure fields compared to the spatial coordinates $x$ and $y$ is that our machine learning experiments indicate that this scaling yields more accurate and stable results than other possible choices.


\begin{table}
\caption{Details of the generated data}\label{Table1}
\begin{tabular}{lllllll}
\toprule
 Shape & 
\vtop{\hbox{\strut Schematic}\hbox{\strut figure}}
 & \vtop{\hbox{\strut Variation in}\hbox{\strut orientation}} & \vtop{\hbox{\strut Variation in}\hbox{\strut length scale}} & \vtop{\hbox{\strut Number}\hbox{\strut of data}}\\
\hline
Circle & \includegraphics[width=0.08\linewidth]{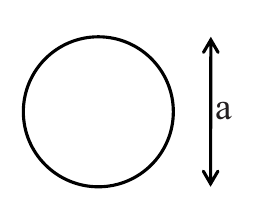} & - & $a=1$ m & 1 \\
Equilateral hexagon & \includegraphics[width=0.08\linewidth]{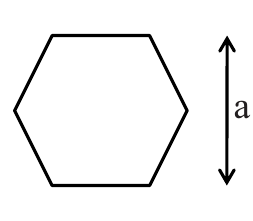} & $3^\circ$, $6^\circ$, \ldots, $60^\circ$ & $a=1$ m & 20 \\
Equilateral pentagon & \includegraphics[width=0.08\linewidth]{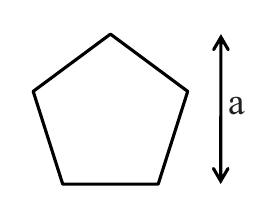} & $3^\circ$, $6^\circ$, \ldots, $72^\circ$ & $a=1$ m & 24 \\
Square & \includegraphics[width=0.08\linewidth]{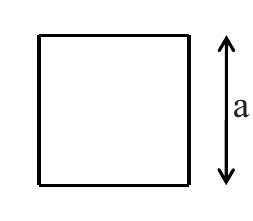} &  $3^\circ$, $6^\circ$, \ldots, $90^\circ$ & $a=1$ m & 30 \\
Equilateral triangle & \includegraphics[width=0.08\linewidth]{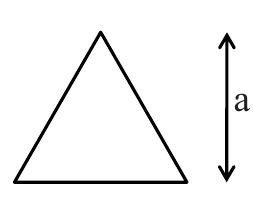} & $3^\circ$, $6^\circ$, \ldots, $180^\circ$ & $a=1$ m & 60\\
Rectangle & \includegraphics[width=0.08\linewidth]{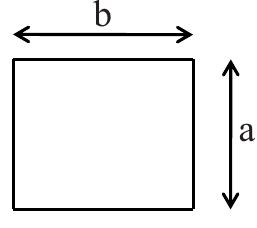} &  $3^\circ$, $6^\circ$, \ldots, $180^\circ$ & $a=1$ m; $b/a=$ 1.2, 1.4, \ldots, 3.6 & 780 \\
Ellipse & \includegraphics[width=0.08\linewidth]{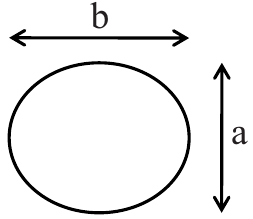} &  $3^\circ$, $6^\circ$, \ldots, $180^\circ$ & $a=1$ m; $b/a=$ 1.2, 1.4, \ldots, 3.8 & 840\\
Triangle & \includegraphics[width=0.08\linewidth]{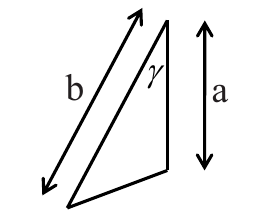} &  $3^\circ$, $6^\circ$, \ldots, $360^\circ$ & \vtop{\hbox{\strut $a=1$ m; $b/a=$ 1.5, 1.75}\hbox{\strut $\gamma=$  60$^\circ$, 80$^\circ$}} & 480 \\
\bottomrule
\end{tabular}
\end{table}


\begin{figure}[h]
  \centering 
      \begin{subfigure}[b]{1.0\textwidth}
        \centering
        \includegraphics[width=\textwidth]{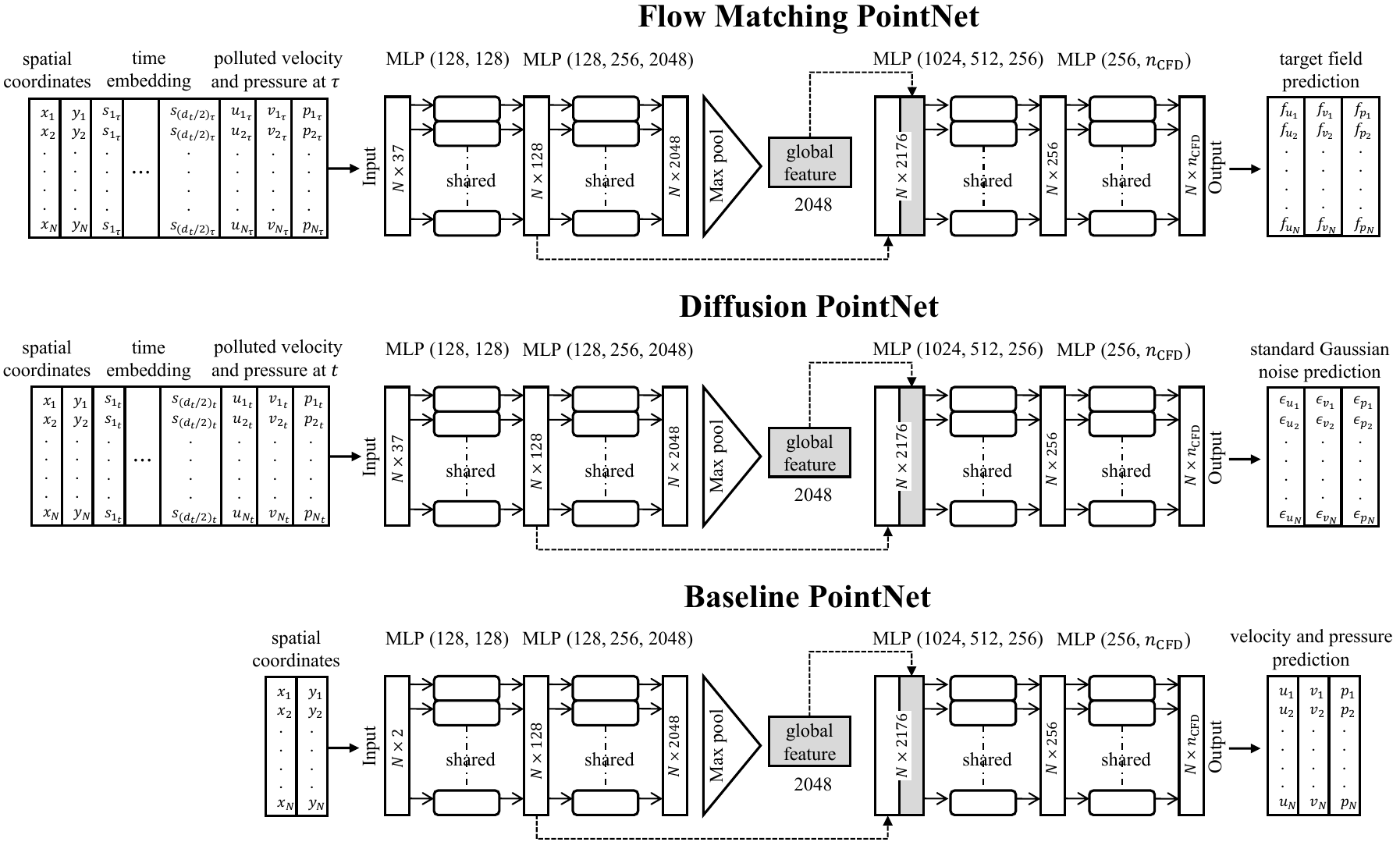}
    \end{subfigure}

    \caption{Architecture of Flow Matching PointNet, Diffusion PointNet, and Baseline PointNet. Labels in the form $(\mathcal{B}_1,\mathcal{B}_2)$ and $(\mathcal{B}_1,\mathcal{B}_2,\mathcal{B}_3)$ are used to denote shared MLPs, as illustrated in Sect. \ref{Sect34}. $n_{\mathrm{CFD}}$ denotes the number of flow variables and $N$ represents the number of points in the point clouds. In the Flow Matching PointNet architecture, the subscript $\tau$ is used to indicate $\boldsymbol{s}_\tau$ and $\boldsymbol{y}_\tau$ (see Sect. \ref{Sect31}). In the Diffusion PointNet architecture, the subscript $t$ is used to show $\boldsymbol{s}_t$ and $\boldsymbol{y}_t$ (see Sect. \ref{Sect32}). In Flow Matching PointNet and Diffusion PointNet, the outputs are, respectively, the prediction of the target field ($\boldsymbol{f}_\text{target}$) and the noisy field ($\boldsymbol{y}_\text{noisy}$), given the polluted velocity and pressure fields at steps $\tau$ and $t$, and conditioned on the geometry of the physical domain represented as a point cloud.}

  \label{ArchitectureOfAll}
\end{figure}


\section{Deep learning architectures}
\label{Sect3}

In this section, we elaborate on the architecture, training procedure, and inference process for the three models of Flow Matching PointNet, Diffusion PointNet, and the baseline PointNet. The neural network of each model is PointNet \cite{qi2017pointnet}; however, in each case, PointNet is used for a different purpose. All three models employ the same PointNet architecture; however, their inputs and outputs differ (see Fig. \ref{ArchitectureOfAll}). The architecture of PointNet \cite{qi2017pointnet} itself remains unchanged unless explicitly stated, for instance, when the activation functions differ among the models. To avoid redundancy, we first describe the algorithmic formulation and the corresponding input and output definitions for each model, and then present the architecture of PointNet at the end of this section (Sect. \ref{Sect34}).



\subsection{Generative Flow Matching PointNet}
\label{Sect31}

We describe the concept and architecture of the proposed Flow Matching PointNet model in this subsection. At a high level, the objective is to predict the desired physical fields such as velocity and pressure, conditioned on a given geometry represented by scattered spatial coordinates. The Flow Matching PointNet is obtained by combining PointNet \cite{qi2017pointnet} with the flow matching formulation \cite{lipman2023flow}, as shown in the top panel of Fig. \ref{ArchitectureOfAll}. Let $\boldsymbol{y}_\text{clean} \in \mathbb{R}^{N \times n_\text{CFD}}$ denote the clean desired physical fields, where $n_\text{CFD}$ represents the number of field variables. In this study, we focus on predicting the velocity and pressure variables in two dimensions, namely $u'$, $v'$, and $p'$, hence $n_\text{CFD} = 3$. Let $\boldsymbol{y}_\text{noisy} \in \mathbb{R}^{N \times n_\text{CFD}}$ denote a standard Gaussian noise field. We define a temporal variable $\tau$. At an intermediate $\tau \in [0,1]$, a linear interpolation is given by
\begin{equation}
\boldsymbol{y}_\tau = (1 - \tau)\boldsymbol{y}_\text{clean} + \tau\boldsymbol{y}_\text{noisy}.
\label{eq:xt}
\end{equation}
Notably, at $\tau = 0$, we recover $\boldsymbol{y}_\text{clean}$, while at $\tau = 1$, we obtain $\boldsymbol{y}_\text{noisy}$. The derivative of $\boldsymbol{y}_\tau$ with respect to $\tau$ defines the target field as
\begin{equation}
\frac{d\boldsymbol{y}_\tau}{d \tau} = \boldsymbol{f}_\text{target},
\label{eq:ode}
\end{equation}
where
\begin{equation}
\boldsymbol{f}_\text{target} = \boldsymbol{y}_\text{noisy} - \boldsymbol{y}_\text{clean},
\label{eq:vtarget}
\end{equation}
and $\boldsymbol{f}_\text{target} \in \mathbb{R}^{N \times n_\text{CFD}}$.

The input of the Flow Matching PointNet consists of the spatial coordinates of each point cloud, representing the geometry of the computational domain, concatenated with the $u'$, $v'$, and $p'$ field variables that are perturbed by a certain level of noise according to Eq. \ref{eq:xt} at a temporal variable $\tau$. In addition to the spatial coordinates (e.g., $x'$ and $y'$) and the perturbed variables, the network must also be informed of the temporal variable, which indicates the level of noise contamination. To achieve this, we construct sinusoidal temporal embedding vectors \cite{vaswani2017attention} and concatenate them with the input vector of the Flow Matching PointNet. The output of the network is the predicted target field (i.e., $\boldsymbol{f}_{\text{target}}$). Let us mathematically formulate this procedure. Let $\boldsymbol{x} \in \mathbb{R}^{N \times d}$ denote the spatial coordinates of the points in a point cloud consisting of $N$ points that represent a computational domain, where $d$ is the number of spatial dimensions. In this study, we set $d = 2$. At the temporal variable $\tau$, a sinusoidal time-embedding layer ($\boldsymbol{s}_\tau$) with dimension $d_\tau$ is defined as
\begin{equation}
\boldsymbol{s}_\tau =
\big[
\sin(\omega_k \tau),\,\cos(\omega_k \tau)
\big]_{k=1}^{d_\tau/2},
\label{eq:timeemb1}
\end{equation}
where, for each $k$,
\begin{equation}
\omega_k
=
\exp\!\left(
-\frac{k-1}{\frac{d_\tau}{2}-1}\ln \left(10^4 \right)
\right).
\label{eq:timeemb2}
\end{equation}
In this study, we set $d_\tau = 32$, and our machine learning experiments indicate that this choice provides sufficient accuracy. The constant $10^{4}$ in Eq. \ref{eq:timeemb2} controls the range of frequencies, ensuring that the encoded time values span several orders of magnitude. This logarithmic scaling enables the network to capture both slow and fast temporal variations effectively across different noise levels \cite{vaswani2017attention}.

Note that $\boldsymbol{s}_\tau \in \mathbb{R}^{1 \times d_\tau}$, simply because each point cloud at a given $\tau$ has one defined vector $\boldsymbol{s}_\tau$. To ensure dimensional consistency, we repeat the same vector $\boldsymbol{s}_\tau$ for every pair of $(x', y')$ in the point cloud, thereby constructing the tensor $\boldsymbol{S}_\tau \in \mathbb{R}^{N \times d_\tau}$. Hence, for each point cloud, the embedded tensor $\boldsymbol{S}_\tau$ is concatenated with the contaminated field $\boldsymbol{y}_\tau$ and the spatial coordinates $\boldsymbol{x}$, forming an input tensor $\mathcal{X} \in \mathbb{R}^{N \times (d + d_\tau + n_\text{CFD})}$ for the Flow Matching PointNet. The output of the Flow Matching PointNet is a tensor $\mathcal{Y} \in \mathbb{R}^{N \times n_\text{CFD}}$, representing the prediction of $\boldsymbol{f}_\text{target}$. The loss function is defined as the mean squared error that measures the discrepancy between $\mathcal{Y}$ and $\boldsymbol{f}_\text{target}$ and can be written as
\begin{equation}
\mathcal{L}_{\text{Flow Matching PointNet}}
=
\frac{1}{N \times n_\text{CFD}}
\sum_{i=1}^{N}
\left(
\left(\tilde{f}_{u_i} - \left(\epsilon_{u_i} - u_i'\right)\right)^2
+
\left(\tilde{f}_{v_i} - \left(\epsilon_{v_i} - v_i'\right)\right)^2
+
\left(\tilde{f}_{p_i} - \left(\epsilon_{p_i} - p_i'\right)\right)^2
\right),
\label{LFMP}
\end{equation}
where $\tilde{f}_{u_i}$, $\tilde{f}_{v_i}$, and $\tilde{f}_{p_i}$ denote the predicted components of the target field $\boldsymbol{f}_{\text{target}}$ produced by the Flow Matching PointNet, with the subscripts $u_i$, $v_i$, and $p_i$ corresponding to the $u$-velocity, $v$-velocity, and pressure components, respectively. $\epsilon_{u_i}$, $\epsilon_{v_i}$, and $\epsilon_{p_i}$ denote the corresponding components of $\boldsymbol{y}_{\text{noisy}}$ and the subscripts are similarly defined.

Flow Matching PointNet is trained using the Adam optimizer \cite{kingma2014adam} with a constant learning rate of 0.001 and a batch size of 256. For each training sample, a temporal variable $\tau$ is uniformly sampled from the interval $[0, 1]$, and three independent standard Gaussian noise fields are generated for the $u'$, $v'$, and $p'$ components. Note that each point cloud in the batch gets its own random $\tau$ value, and given $\tau$, the noisy input $\boldsymbol{y}_\tau$ and the target field $\boldsymbol{f}_{\text{target}}$ are computed using Eq. \ref{eq:xt} and Eq. \ref{eq:vtarget}.

After training, Flow Matching PointNet generates the velocity and pressure fields conditioned on a geometry from the test set. Given $\boldsymbol{x}$ as the point cloud representing the target geometry from the test set, we generate three independent standard Gaussian fields representing $\boldsymbol{y}_\text{noisy}$ and integrate Eq. \ref{eq:ode} from $\tau = 1$ to $\tau = 0$ using the explicit Euler method with $N_\tau$ steps, such that $\Delta \tau = 1/N_\tau$. At each step,
\begin{equation}
\boldsymbol{y}_n = \boldsymbol{y}_{n+1} - \Delta \tau \mathcal{Y}_{n+1 \rightarrow n},
\label{Euler}
\end{equation}
where $\mathcal{Y}_{n+1 \rightarrow n}$ denotes the prediction of the Flow Matching PointNet given the concatenation of $\boldsymbol{x}$, $\boldsymbol{y}_n$, and $\boldsymbol{S}_\tau$ at $\tau = 1.0 - n \Delta \tau$. The final output at $\tau = 0$ corresponds to $\boldsymbol{y}_\text{clean}$, which represents the predicted velocity and pressure fields ($u'$, $v'$, $p'$) associated with the given geometry. In the present setup, we set $N_\tau = 1000$.


\begin{figure}[t]
  \centering 
      \begin{subfigure}[b]{1.0\textwidth}
        \centering
        \includegraphics[width=\textwidth]{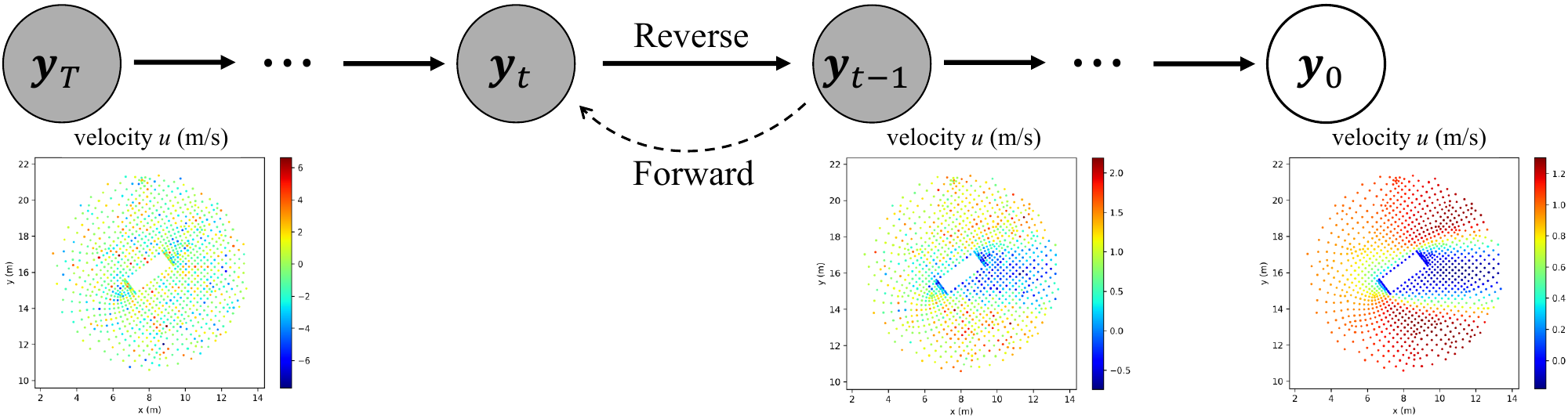}
    \end{subfigure}

    \caption{Schematic of the reverse process in Diffusion PointNet. As an example, the denoising process for the $u$-velocity field is shown, while the network jointly predicts the noise for all velocity and pressure field variables.}

  \label{FigDiffusionSchem}
\end{figure}


\subsection{Generative Diffusion PointNet}
\label{Sect32}

In this subsection, we present the underlying concept and architectural design of the proposed Diffusion PointNet model. Similar to the previous part, the primary goal is to predict physical fields such as velocity and pressure, conditioned on a geometry represented by scattered spatial coordinates. The Diffusion PointNet framework integrates the PointNet architecture \cite{qi2017pointnet} with a denoising diffusion probabilistic model \cite{ho2020denoising}, as shown in the central panel of Fig. \ref{ArchitectureOfAll}.


Similar to Sect. \ref{Sect31}, let $\boldsymbol{y}_\text{clean} \in \mathbb{R}^{N \times n_\text{CFD}}$ represent the clean field, where $N$ and $n_\text{CFD}$ are defined as before, since our objective is to predict the two-dimensional velocity and pressure variables $u'$, $v'$, and $p'$. Figure \ref{FigDiffusionSchem} illustrates the schematic of the forward and inverse processes in Diffusion PointNet. In the forward diffusion process, standard Gaussian noise is incrementally added to $\boldsymbol{y}_\text{clean}$ over $T$ discrete steps using a noise scheduler, producing a sequence of noisy fields $\{\boldsymbol{y}_t\}_{t=1}^{T}$, where $\boldsymbol{y}_t \in \mathbb{R}^{N \times n_\text{CFD}}$. In this study, we use a cosine noise scheduler defined as 

\begin{equation}
\beta_t = 1 - 
\frac{\cos^2\!\left(\frac{(t/T)+r}{1+r}\frac{\pi}{2}\right)}
{\cos^2\!\left(\frac{r}{1+r}\frac{\pi}{2}\right)},
\label{eq:beta}
\end{equation}
where $r$ is the small offset parameter introduced in Ref. \cite{nichol2021improved} to regularize the noise schedule, and set to 0.008 in our machine learning experiments. At any diffusion step $t$, a noisy sample is generated as
\begin{equation}
\boldsymbol{y}_t = 
\sqrt{\bar{\alpha}_t}\,\boldsymbol{y}_\text{clean} +
\sqrt{1-\bar{\alpha}_t}\,\boldsymbol{y}_\text{noisy},
\label{eq:forward}
\end{equation}
where
\begin{equation}
\alpha_t = 1-\beta_t,
\end{equation}
and
\begin{equation}
\bar{\alpha}_t = \prod_{i=1}^{t}\alpha_i,
\end{equation}
and, as mentioned in Sect. \ref{Sect31}, $\boldsymbol{y}_\text{noisy}$ is standard Gaussian noise, and $\boldsymbol{y}_\text{noisy} \in \mathbb{R}^{N \times n_\text{CFD}}$.

The input to the Diffusion PointNet comprises the spatial coordinates of the point cloud, which define the geometry of the computational domain, concatenated with the field variables $u'$, $v'$, and $p'$ after being perturbed by noise prescribed by Eq. \ref{eq:forward} at the diffusion step $t$. Similar to the Flow Matching PointNet, the Diffusion PointNet must be provided with the step $t$, along with the spatial coordinates (e.g., $x'$ and $y'$) and the perturbed velocity and pressure variables. The step $t$ identifies the associated level of noise contamination imposed through Eq. \ref{eq:forward}. To this end, we use the sinusoidal time-embedding layer \cite{vaswani2017attention} introduced in Eqs. \ref{eq:timeemb1}--\ref{eq:timeemb2}. In terms of notation, instead of $\tau$, we now simply use $t$. For completeness, we rewrite the time-embedding layer $\boldsymbol{s}_t$ with dimension $d_t$, defined as
\begin{equation}
\boldsymbol{s}_t =
\big[
\sin(\omega_k t),\,\cos(\omega_k t)
\big]_{k=1}^{d_t/2},
\label{eq:timeemb10}
\end{equation}
where for every index $k$,
\begin{equation}
\omega_k
=
\exp\!\left(
-\frac{k-1}{\frac{d_t}{2}-1}\ln \left(10^4 \right)
\right),
\label{eq:timeemb20}
\end{equation}
and we set $d_t=32$. Analogous to $\boldsymbol{S}_\tau$ used in the Flow Matching PointNet framework, we define $\boldsymbol{S}_t$ for the Diffusion PointNet model, where $\boldsymbol{S}_t \in \mathbb{R}^{N \times d_t}$. In this sense, for a given point cloud at diffusion step $t$, the embedded tensor $\boldsymbol{S}_t$ is concatenated with the noisy field $\boldsymbol{y}_t$ and the spatial coordinates of the point cloud $\boldsymbol{x} \in \mathbb{R}^{N \times d}$, resulting in the input tensor $\mathcal{X} \in \mathbb{R}^{N \times (d + d_t + n_\text{CFD})}$ for the Diffusion PointNet. Accordingly, the Diffusion PointNet outputs a tensor $\mathcal{Y} \in \mathbb{R}^{N \times n_\text{CFD}}$, which represents the prediction of the noise $\boldsymbol{y}_\text{noisy}$. The loss function is defined as the mean squared error between $\mathcal{Y}$ and $\boldsymbol{y}_\text{noisy}$ and can be written as
\begin{equation}
\mathcal{L}_{\text{Diffusion PointNet}}
=
\!
\frac{1}{N \times n_\text{CFD}}
\sum_{i=1}^{N}
\left(
\left(\tilde{\epsilon}_{u_i} - \epsilon_{u_i}\right)^2
+
\left(\tilde{\epsilon}_{v_i} - \epsilon_{v_i}\right)^2
+
\left(\tilde{\epsilon}_{p_i} - \epsilon_{p_i}\right)^2
\right),
\label{LDP}
\end{equation}
where $\epsilon_{u_i}$, $\epsilon_{v_i}$, and $\epsilon_{p_i}$ denote the ground-truth standard Gaussian noise, as similarly defined in Sect. \ref{Sect31}, and $\tilde{\epsilon}_{u_i}$, $\tilde{\epsilon}_{v_i}$, and $\tilde{\epsilon}_{p_i}$ denote the corresponding noise components predicted by the Diffusion PointNet. Diffusion PointNet is optimized with the Adam algorithm \cite{kingma2014adam} using a constant learning rate of 0.001 and a batch size of 256. For each training iteration, a random step $t$ is sampled independently for every point cloud in the training batch, and three independent standard Gaussian noises are generated, corresponding to the $u'$, $v'$, and $p'$ variables, to construct the tensor $\boldsymbol{y}_\text{noisy}$ presented in Eq.~\ref{eq:forward}. The corresponding noisy tensor $\boldsymbol{y}_t$ is then computed using Eq.~\ref{eq:forward}.

Once training is complete, the Diffusion PointNet generates the corresponding velocity and pressure variables for a geometry selected from the test set by simulating the reverse diffusion process, as shown in Fig. \ref{FigDiffusionSchem}. For a point cloud $\boldsymbol{x}$ representing a geometry from the test set, the reconstruction begins by generating three independent standard Gaussian noise fields to construct the tensor $\boldsymbol{y}_{T}$. The fields are then iteratively denoised from $t = T$ to $t = 0$ using
\begin{equation}
\boldsymbol{y}_{t-1}
=
\frac{1}{\sqrt{\alpha_t}}
\Big(
\boldsymbol{y}_t
-
\frac{\beta_t}{\sqrt{1-\bar{\alpha}_t}}
\, \mathcal{Y}_{t \rightarrow t-1}
\Big)
+
\sqrt{\beta_t}\,\boldsymbol{z},
\label{eq:reverse}
\end{equation}
where $\boldsymbol{z}$ denotes three independent standard Gaussian noise fields, and $\boldsymbol{z} \in \mathbb{R}^{N \times n_\text{CFD}}$. For $t = 0$, the term $\sqrt{\beta_t} \, \boldsymbol{z}$ is omitted from Eq.~\ref{eq:reverse} in accordance with the denoising diffusion probabilistic formulation \cite{ho2020denoising}. $\mathcal{Y}_{t \rightarrow t-1}$ denotes the prediction of the Diffusion PointNet given the concatenation of $\boldsymbol{x}$, $\boldsymbol{y}_t$, and $\boldsymbol{S}_t$ at $t$. The final output $\boldsymbol{y}_0$ at $t=0$ yields the predicted velocity and pressure fields $(u',v',p')$ consistent with the input geometry. In our experiments, we set $T = 1000$.

\subsection{Baseline PointNet}
\label{Sect33}

Now that the Flow Matching PointNet and Diffusion PointNet architectures have been described, the baseline PointNet can be introduced in a straightforward manner. The baseline PointNet for predicting fluid flow variables was first proposed by Kashefi et al. \cite{kashefi2021PointNet} and has since been adopted in several subsequent frameworks (see e.g., Refs. \cite{tejero2024point,kashefi2025kolmogorov,hwang2025point}). The version considered in this work is identical to that originally presented by Kashefi et al \cite{kashefi2021PointNet}. In the baseline PointNet, the input consists solely of the spatial coordinates of the point cloud representing the geometry of the domain, and the output corresponds to the physical fields of interest, as shown in the bottom panel of Fig. \ref{ArchitectureOfAll}. Following the notation introduced in Sect. \ref{Sect31}, the input is given by $\boldsymbol{x} \in \mathbb{R}^{N \times d}$, and consequently, the input tensor is $\mathcal{X} \in \mathbb{R}^{N \times d}$. The output consists of the velocity components and pressure field, forming an output tensor $\mathcal{Y} \in \mathbb{R}^{N \times n_{\mathrm{CFD}}}$. Hence, PointNet learns a direct mapping from $\mathcal{X}$ to $\mathcal{Y}$. The loss function is the mean squared error and can be written as

\begin{equation}
\mathcal{L}_{\text{Baseline PointNet}}
=
\frac{1}{N \times n_\text{CFD}}
\sum_{i=1}^{N}
\left(
\left(\tilde{u}_i' - u_i'\right)^2
+
\left(\tilde{v}_i' - v_i'\right)^2
+
\left(\tilde{p}_i' - p_i'\right)^2
\right),
\label{LBP}
\end{equation}
where $\tilde{u}_i'$, $\tilde{v}_i'$, and $\tilde{p}_i'$ denote, respectively, the predicted $u$-velocity, $v$-velocity, and pressure fields obtained using the baseline PointNet. Training is performed using a batch size of 256 and the Adam optimizer \cite{kingma2014adam} with a constant learning rate of 0.001.


\subsection{PointNet}
\label{Sect34}

In Sections \ref{Sect31}--\ref{Sect33}, we explained the role of PointNet \cite{qi2017pointnet} in each of the three proposed models, namely Flow Matching PointNet, Diffusion PointNet, and the baseline PointNet. In this part, we describe the architecture of PointNet itself. In general, the pair $\mathcal{X}$ and $\mathcal{Y}$ denote the input and output (or the quantity to be predicted) of PointNet \cite{qi2017pointnet}, respectively. In the earlier sections, we clarified what $\mathcal{X}$ and $\mathcal{Y}$ represent for each of the three models. In summary, in Flow Matching PointNet, $\mathcal{X} \in \mathbb{R}^{N \times (d + d_\tau + n_\text{CFD})}$ and $\mathcal{Y} \in \mathbb{R}^{N \times n_\text{CFD}}$; in Diffusion PointNet, $\mathcal{X} \in \mathbb{R}^{N \times (d + d_t + n_\text{CFD})}$ and $\mathcal{Y} \in \mathbb{R}^{N \times n_\text{CFD}}$; and in the baseline PointNet, $\mathcal{X} \in \mathbb{R}^{N \times d}$ and $\mathcal{Y} \in \mathbb{R}^{N \times n_\text{CFD}}$.

The original PointNet model developed for computer graphics applications consists of two branches: a classification branch and a segmentation branch \cite{qi2017pointnet}. In this work, we employ the segmentation branch of PointNet \cite{qi2017pointnet}. Figure \ref{ArchitectureOfAll} illustrates the overall architecture of the segmentation branch of PointNet \cite{qi2017pointnet,kashefi2021PointNet}. At a high level, the architecture can be viewed as an encoder–decoder framework. A key advantage of PointNet \cite{qi2017pointnet} and its variants (see e.g., Ref. \cite{qi2017pointnet++}) is their ability to process unordered sets of points that constitute a point cloud. In particular, PointNet is invariant under the $N!$ permutations of an input vector representing a point cloud with $N$ points \cite{qi2017pointnet}. This permutation invariance is achieved through two main mechanisms. First, in both the encoder and decoder, each neural network layer uses a shared weight matrix and a shared bias vector across all neurons, forming what is known as a shared layer. Second, a symmetric function, one that is itself invariant to input ordering, is applied to encode global features from the input. As shown in Fig. \ref{ArchitectureOfAll}, PointNet uses the max function as its symmetric function, as it has been shown to outperform alternatives such as the average function \cite{qi2017pointnet}.

In Fig. \ref{ArchitectureOfAll}, the notation \((\mathcal{B}_1,\mathcal{B}_2)\) refers to two consecutive hidden shared layers with \(\mathcal{B}_1\) and \(\mathcal{B}_2\) neurons, whereas the triplet \((\mathcal{B}_1,\mathcal{B}_2,\mathcal{B}_3)\) follows the same convention. As shown in Fig. \ref{ArchitectureOfAll}, the initial feature extraction stage consists of two shared MLP blocks with layer sizes \((128,128)\) and \((128,256,2048)\). A symmetric max pooling operation is then applied to obtain a global feature vector of size \(2048\). This global feature is concatenated with an intermediate tensor of size \(N \times 128\), producing a combined representation of dimension \(N \times 2176\). The subsequent processing stage employs two additional shared MLP blocks with sizes \((1024,512,256)\) and \((256,n_{\mathrm{CFD}})\). Consequently, the final output of PointNet is a tensor of size \(N \times n_{\mathrm{CFD}}\), as depicted in Fig. \ref{ArchitectureOfAll}. After each shared MLP layer, a rectified linear unit activation function,
\begin{equation}
\sigma(\lambda) = \max(0,\lambda),
\label{relu}
\end{equation}
is applied and is followed by batch normalization, except for the final layer. In Flow Matching PointNet and Diffusion PointNet, the last layer does not use any activation function or batch normalization. In contrast, the baseline PointNet applies a sigmoid activation function in the final layer, defined as
\begin{equation}
\sigma(\lambda) = \frac{1}{1 + e^{-\lambda}}.
\label{sigmoid}
\end{equation}
Further details about the architecture of PointNet can be found in Refs. \cite{qi2017pointnet,kashefi2021PointNet,kashefi2025kolmogorov}.


\begin{table}[h]
 \centering
\caption{Comparison between the performance of the Flow Matching PointNet, Diffusion PointNet, and baseline PointNet models for prediction of the velocity and pressure fields for the test set containing 222 unseen geometries. Number of sampling for the Flow Matching PointNet and Diffusion PointNet is 100. In numbers reported in the form $\alpha_1 \pm \alpha_2$, $\alpha_1$ denotes the mean error and $\alpha_2$ represents the corresponding standard deviation. Note that $||\cdots||$ indicates the $L^2$ norm.}\label{Table5}
\begin{tabular}{llll}
\toprule
 &  Flow Matching PointNet & Diffusion PointNet & Baseline PointNet \\
\midrule
Average $||\Tilde{u}-u||/||u||$ &  1.39556E$-$2 $\pm$ 1.52319E$-$4 & 1.25420E$-$2 $\pm$ 1.14052E$-$4 &  3.19833E$-$2\\
Maximum $||\Tilde{u}-u||/||u||$ &  1.70648E$-$1 $\pm$ 9.32820E$-$3 & 1.70626E$-$1 $\pm$ 4.10404E$-$3 &  1.22576E$-$1\\
Minimum $||\Tilde{u}-u||/||u||$ &  5.86007E$-$3 $\pm$ 3.41877E$-$4 & 3.94060E$-$3 $\pm$ 2.04423E$-$4 &  1.12399E$-$2\\
\midrule
Average $||\Tilde{v}-v||/||v||$ &  4.80994E$-$2 $\pm$ 3.42805E$-$4 & 6.16094E$-$2 $\pm$ 3.59036E$-$4 &  1.32724E$-$1\\
Maximum $||\Tilde{v}-v||/||v||$ &  5.44083E$-$1 $\pm$ 2.99690E$-$2 & 5.31338E$-$1 $\pm$ 1.79378E$-$2 &  4.25843E$-$1\\
Minimum $||\Tilde{v}-v||/||v||$ &  2.29497E$-$2 $\pm$ 1.92305E$-$3 & 3.19955E$-$2 $\pm$ 1.40590E$-$3 &  6.49750E$-$2\\
\midrule
Average $||\Tilde{p}-p||/||p||$ &  3.66295E$-$2 $\pm$ 3.52680E$-$4 & 3.28203E$-$2 $\pm$ 2.33396E$-$4 &  1.08271E$-$1\\
Maximum $||\Tilde{p}-p||/||p||$ &  1.82688E$-$1 $\pm$ 7.92950E$-$3 & 1.65267E$-$1 $\pm$ 4.97312E$-$3 &  2.53676E$-$1\\
Minimum $||\Tilde{p}-p||/||p||$ &  1.63747E$-$2 $\pm$ 1.79350E$-$3 & 1.51713E$-$2 $\pm$ 6.72365E$-$4 &  3.41443E$-$2\\
\midrule
Number of trainable &  3554179 & 3554179 & 3554179 \\
parameters &  &  & \\
\bottomrule
\end{tabular}
\end{table}


\begin{figure}[h]
  \centering 
      \begin{subfigure}[b]{0.32\textwidth}
      \caption{Flow Matching PointNet}
        \centering
        \includegraphics[width=\textwidth]{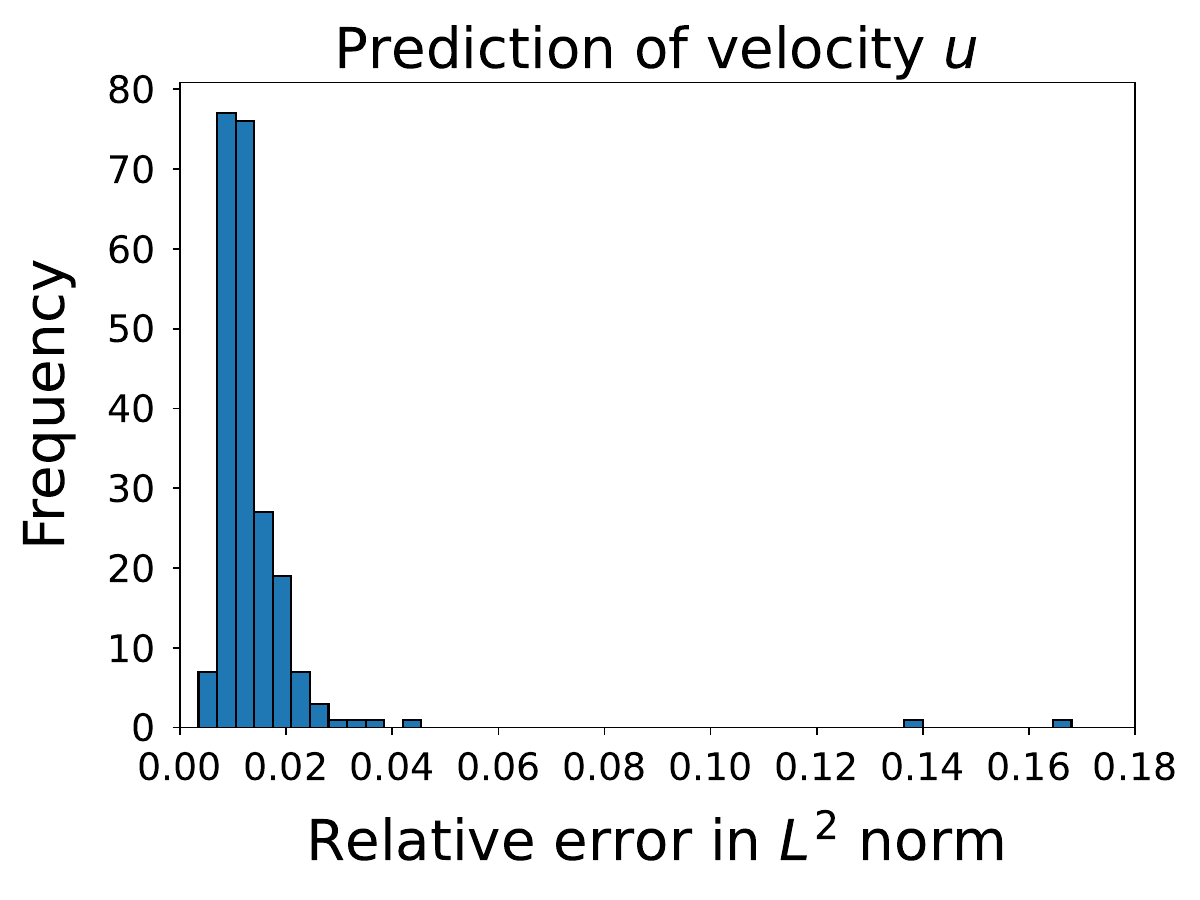}
    \end{subfigure}
    \begin{subfigure}[b]{0.32\textwidth}
     \caption{Flow Matching PointNet}
        \centering
        \includegraphics[width=\textwidth]{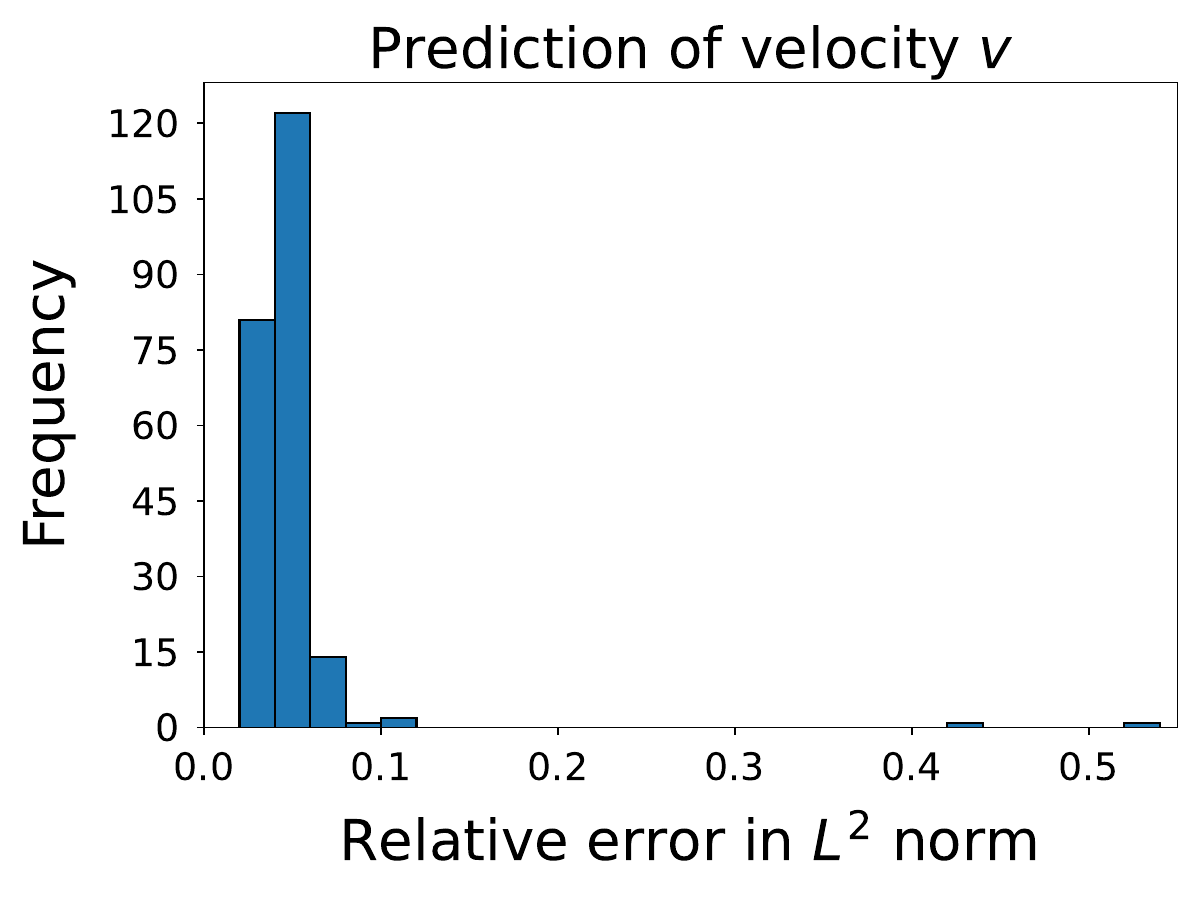}
    \end{subfigure}
    \begin{subfigure}[b]{0.32\textwidth}
    \caption{Flow Matching PointNet}
        \centering
        \includegraphics[width=\textwidth]{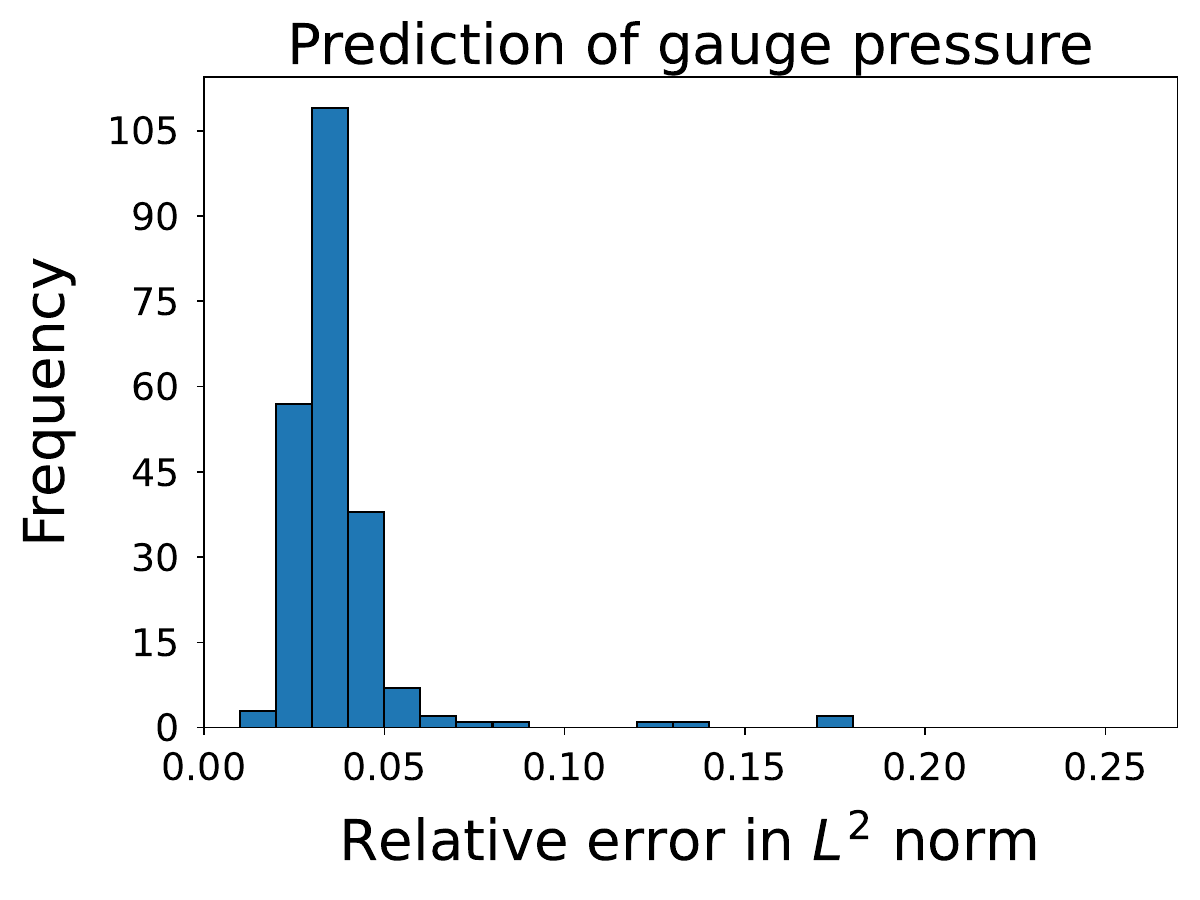}
    \end{subfigure}

    \begin{subfigure}[b]{0.32\textwidth}
    \caption{Diffusion PointNet}
        \centering
        \includegraphics[width=\textwidth]{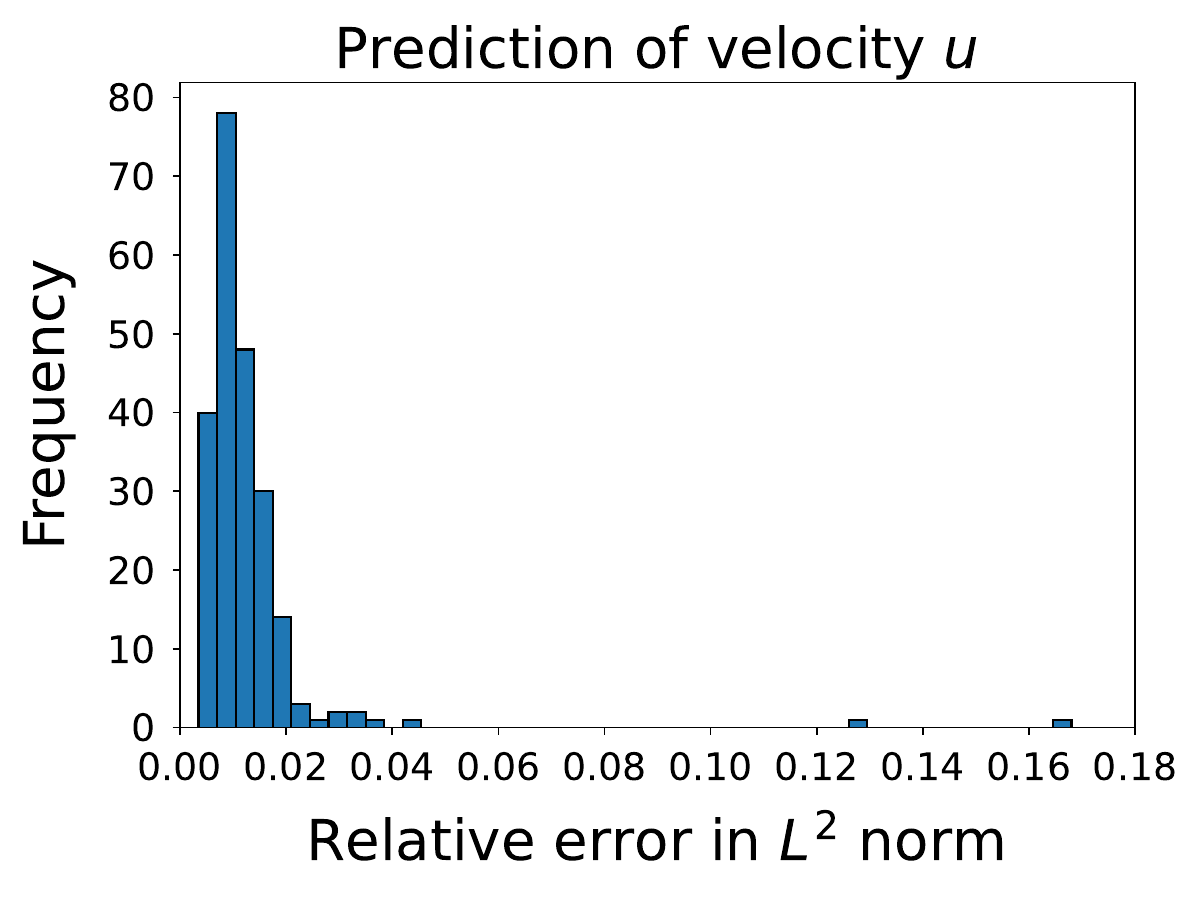}
    \end{subfigure}
    \begin{subfigure}[b]{0.32\textwidth}
    \caption{Diffusion PointNet}
        \centering
        \includegraphics[width=\textwidth]{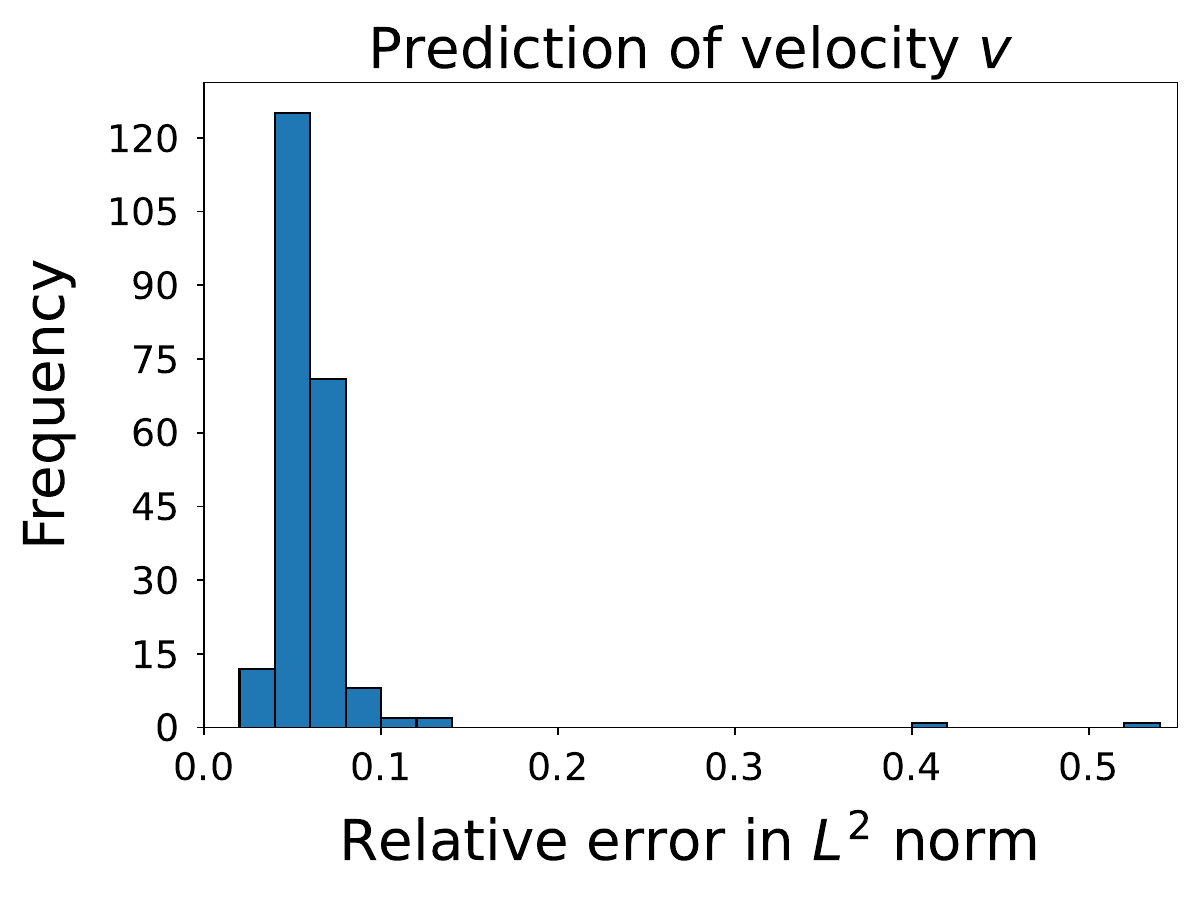}
    \end{subfigure}
    \begin{subfigure}[b]{0.32\textwidth}
    \caption{Diffusion PointNet}
        \centering
        \includegraphics[width=\textwidth]{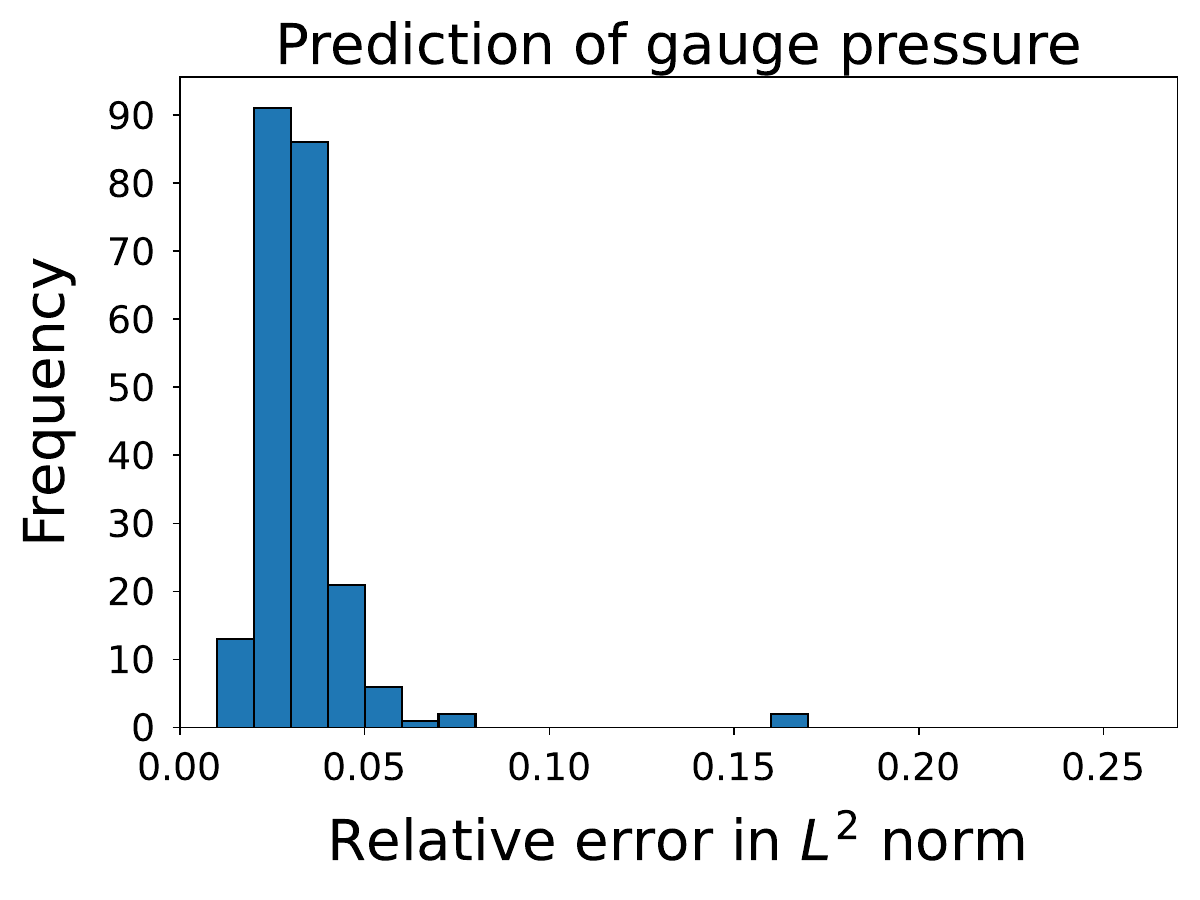}
    \end{subfigure}

        \begin{subfigure}[b]{0.32\textwidth}
    \caption{Baseline PointNet}
        \centering
        \includegraphics[width=\textwidth]{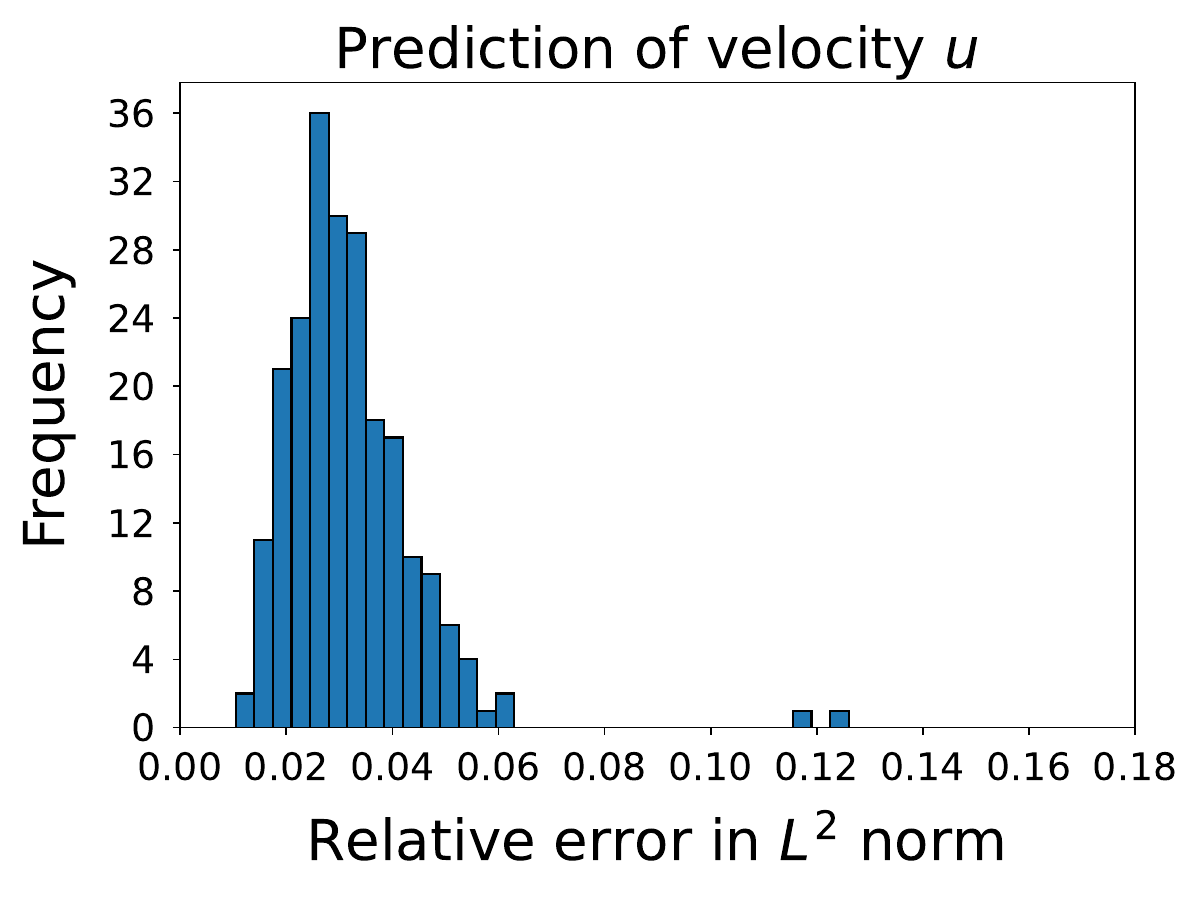}
    \end{subfigure}
    \begin{subfigure}[b]{0.32\textwidth}
    \caption{Baseline PointNet}
        \centering
        \includegraphics[width=\textwidth]{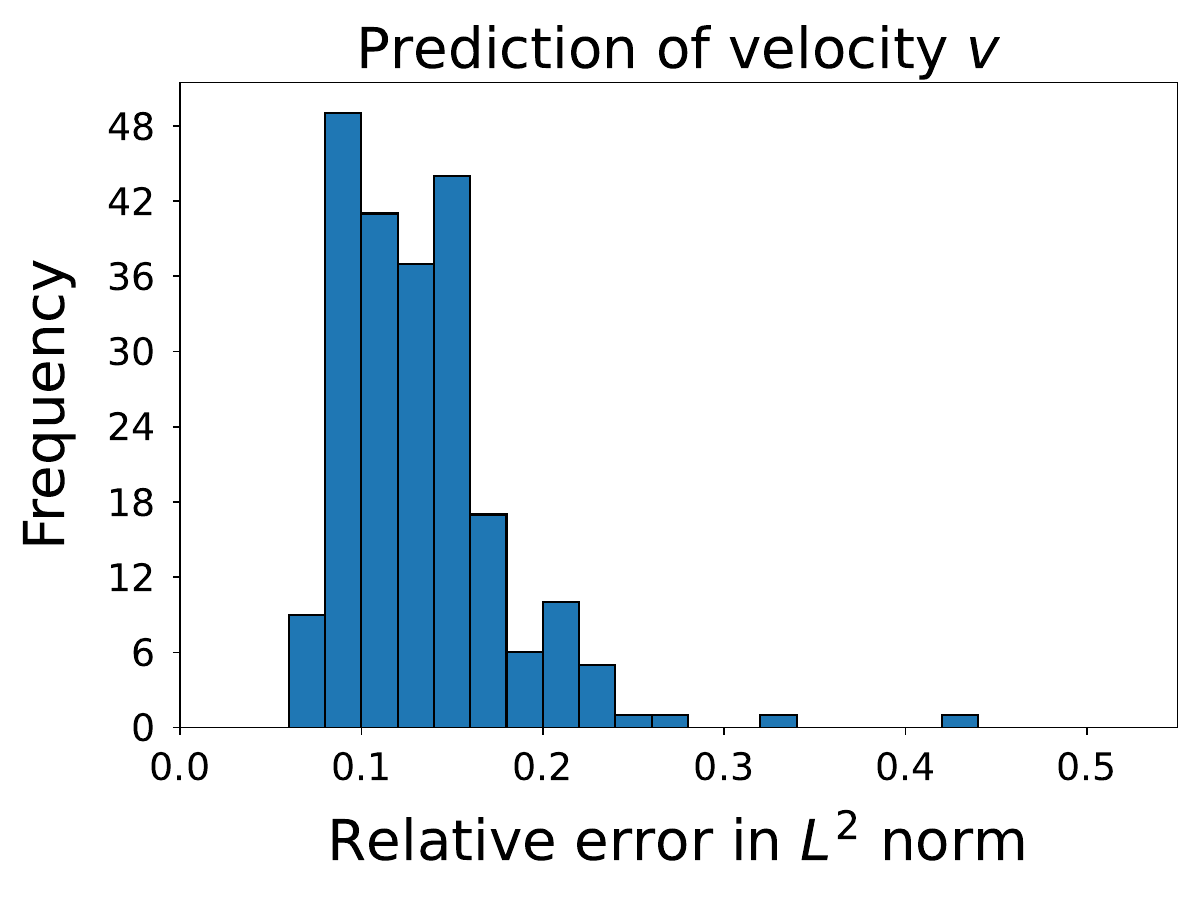}
    \end{subfigure}
    \begin{subfigure}[b]{0.32\textwidth}
    \caption{Baseline PointNet}
        \centering
        \includegraphics[width=\textwidth]{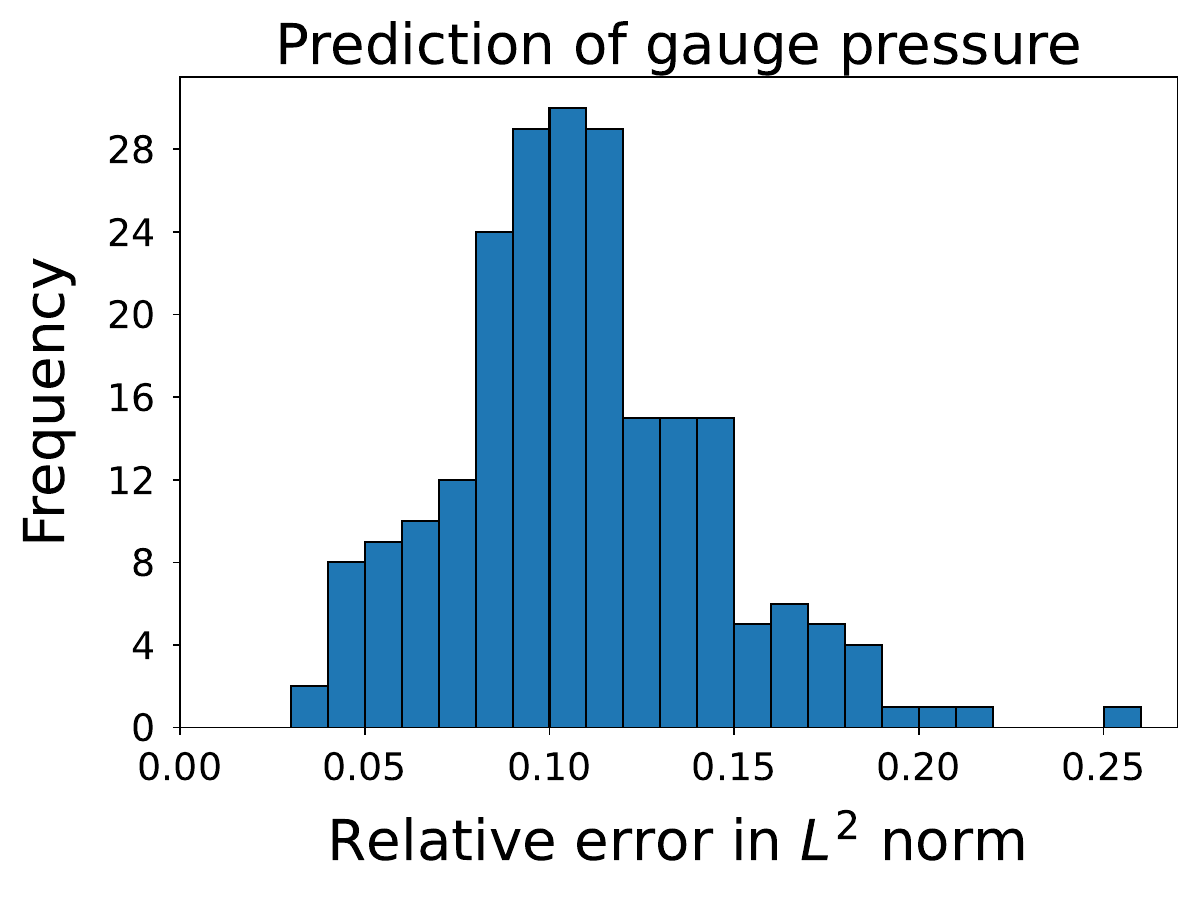}
    \end{subfigure}
    
  \caption{Comparison of histograms of the relative pointwise error in $L^2$ norm for the velocity and pressure fields predicted by the Flow Matching PointNet, Diffusion PointNet, and baseline PointNet models.}
  \label{Fig1C}
\end{figure}


\section{Results and discussion}
\label{Sect4}

\subsection{Overall results}
\label{Sect41}


In this section, we present a comprehensive analysis of the results using tables and histograms. Table \ref{Table5} reports the relative pointwise errors ($L^2$ norm) for predicting the velocity components $u$ and $v$ and the pressure $p$ using the three models: Flow Matching PointNet, Diffusion PointNet, and the baseline PointNet. The results are computed across the test set, which contains 222 unseen geometries. Specifically, for each geometry in the test set, 100 samples are generated using both the Flow Matching PointNet and Diffusion PointNet models, meaning that 100 random noise realizations are conditioned on the same geometry. The reported values in Table \ref{Table5} are expressed in the form $\alpha_1 \pm \alpha_2$, where $\alpha_1$ denotes the mean error and $\alpha_2$ represents the corresponding standard deviation. According to the values listed in Table \ref{Table5}, the average relative pointwise errors ($L^2$ norm) over the 222 geometries for the $u$-velocity component are 1.40\%, 1.26\%, and 3.20\% for the Flow Matching PointNet, Diffusion PointNet, and baseline PointNet models, respectively. For the $v$-velocity component, the corresponding errors are approximately 4.81\%, 6.17\%, and 13.28\%. These results demonstrate that the Flow Matching and Diffusion models, particularly for the $v$ variable, achieve substantially lower prediction errors. In comparison with the baseline PointNet, they are capable of reducing the relative pointwise error ($L^2$ norm) to below 10\%, which is a remarkable improvement from an engineering standpoint. Predicting the \(v\) variable is more challenging than predicting \(u\) because the flow field exhibits both positive and negative velocities in the \(y\)-direction due to vortex formation behind the cylinder. Although the \(u\) variable may also become negative in some regions, its spatial gradients are typically less steep than those of the \(v\) variable.

Examining the results for the pressure variable ($p$) reveals a similar trend. Both the Flow Matching PointNet and Diffusion PointNet models reduce the average relative pointwise error ($L^2$ norm) to below 10\%, whereas the baseline PointNet exceeds this limit. According to Table \ref{Table5}, the relative pointwise errors ($L^2$ norm) for predicting the pressure field are approximately 3.67\%, 3.29\%, and 10.83\% for the Flow Matching PointNet, Diffusion PointNet, and baseline PointNet, respectively. The accuracy of the pressure field is particularly important because it directly influences the computation of lift and drag forces, which will be discussed in detail in Sect. \ref{Sect42}. Furthermore, the uncertainty quantification capabilities of the proposed generative Flow Matching PointNet and Diffusion PointNet models provide valuable insights for practical engineering applications where safety factors must be considered in design processes. Finally, it is worth emphasizing that, as reported in Table \ref{Table5}, the number of trainable parameters, and consequently the model size, is identical across all three frameworks.

In addition to reporting the mean, minimum, and maximum errors across the 222 test geometries in Table \ref{Table5}, the distribution of these errors is illustrated in the histogram shown in Fig. \ref{Fig1C}. As observed in Fig. \ref{Fig1C}, Flow Matching PointNet and Diffusion PointNet demonstrate higher accuracy than the baseline PointNet in predicting the $v$ and $p$ variables. However, in all three models, there exist a few outlier cases for which the prediction error increases noticeably, as can be seen in Fig. \ref{Fig1C}. When comparing the Flow Matching PointNet and Diffusion PointNet in terms of prediction accuracy, the difference between the two is not substantial. Based on the overall results presented in Table \ref{Table5} and Fig. \ref{Fig1C}, the Flow Matching PointNet model predicts the $v$-velocity component on average about 1.3\% more accurately than the Diffusion PointNet model.


\begin{figure}
    \centering

    \begin{tabular}{c c c c}
        & \text{velocity $u$ (m/s)} & \text{velocity $v$ (m/s)} & \text{gauge pressure $p$ (Pa)}\\
        \rotatebox{90}{\ \ \quad \quad \quad \quad \text{$n=0$}} &
        \includegraphics[width=0.32\textwidth]{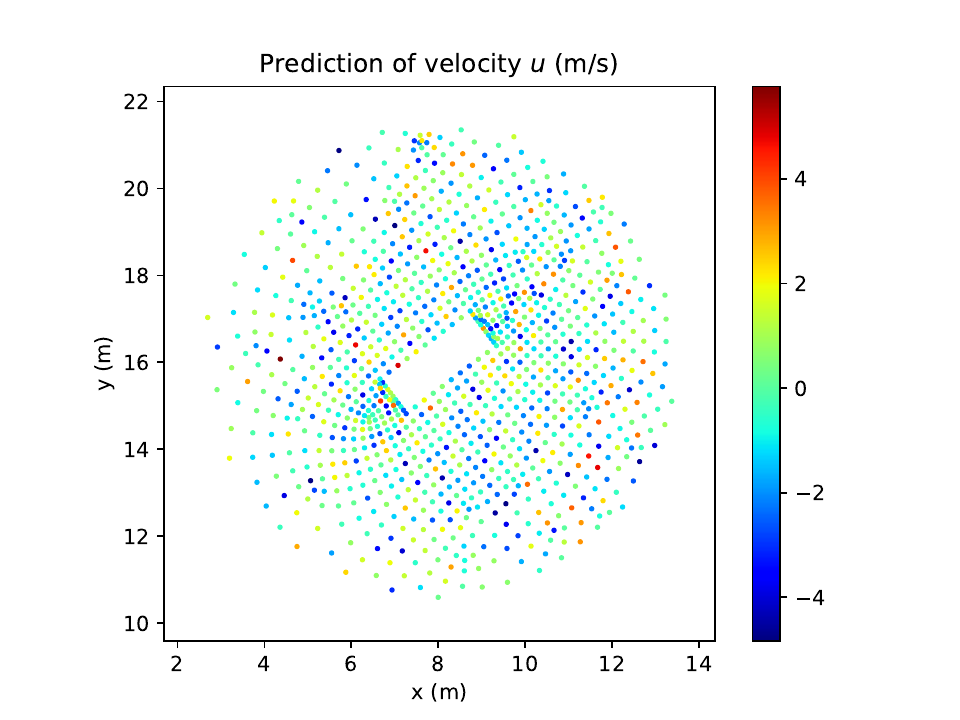} &
        \includegraphics[width=0.32\textwidth]{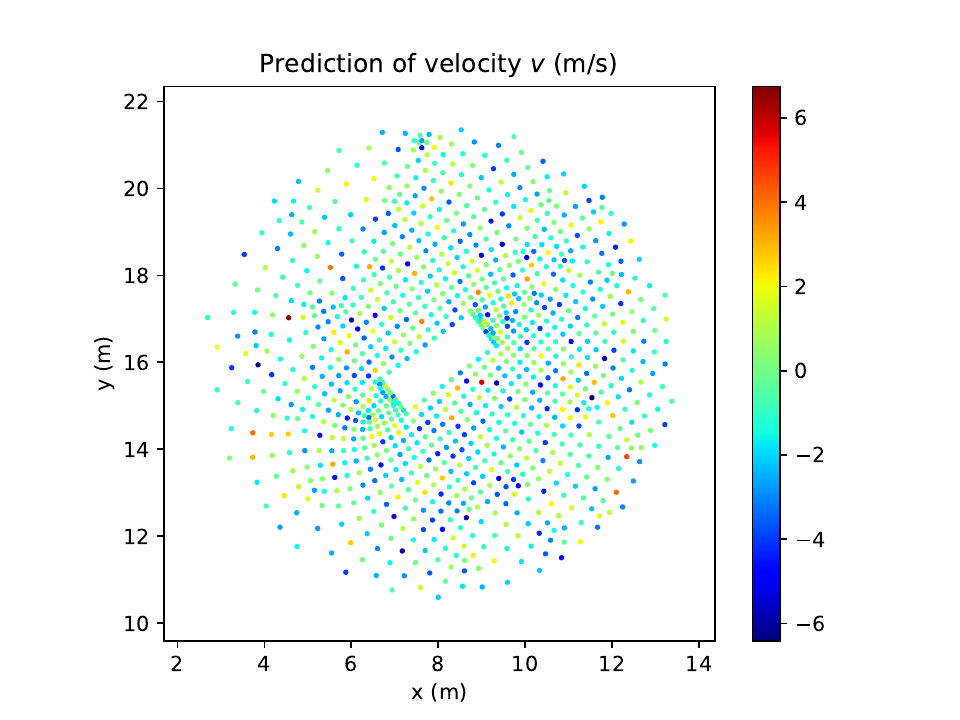} &
        \includegraphics[width=0.32\textwidth]{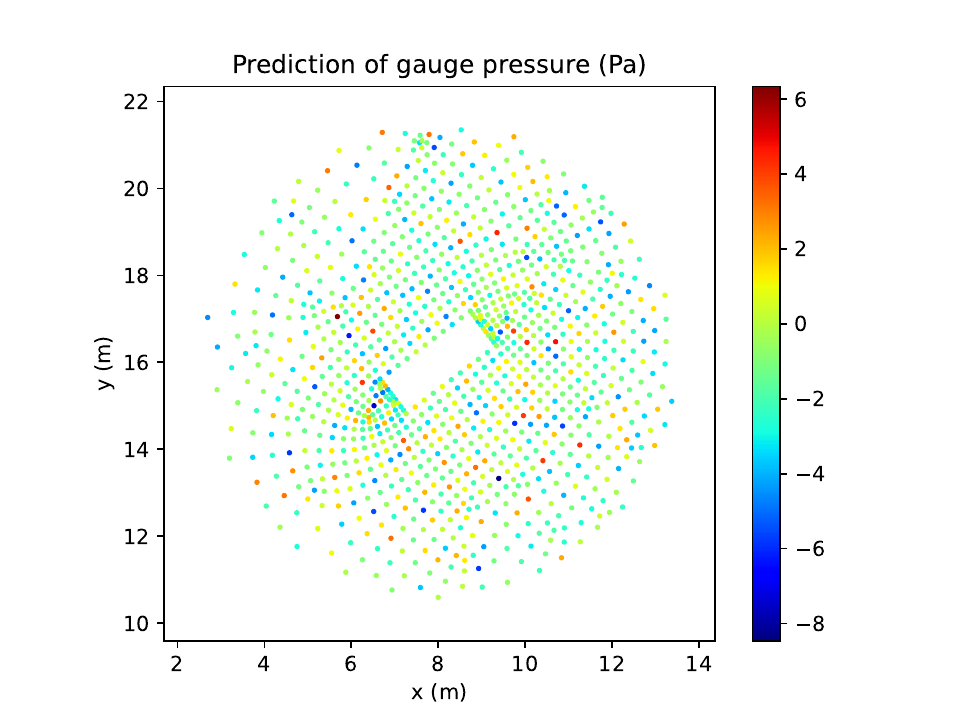} \\

        \rotatebox{90}{\ \ \quad \quad \quad \quad \text{$n=500$}} &
        \includegraphics[width=0.32\textwidth]{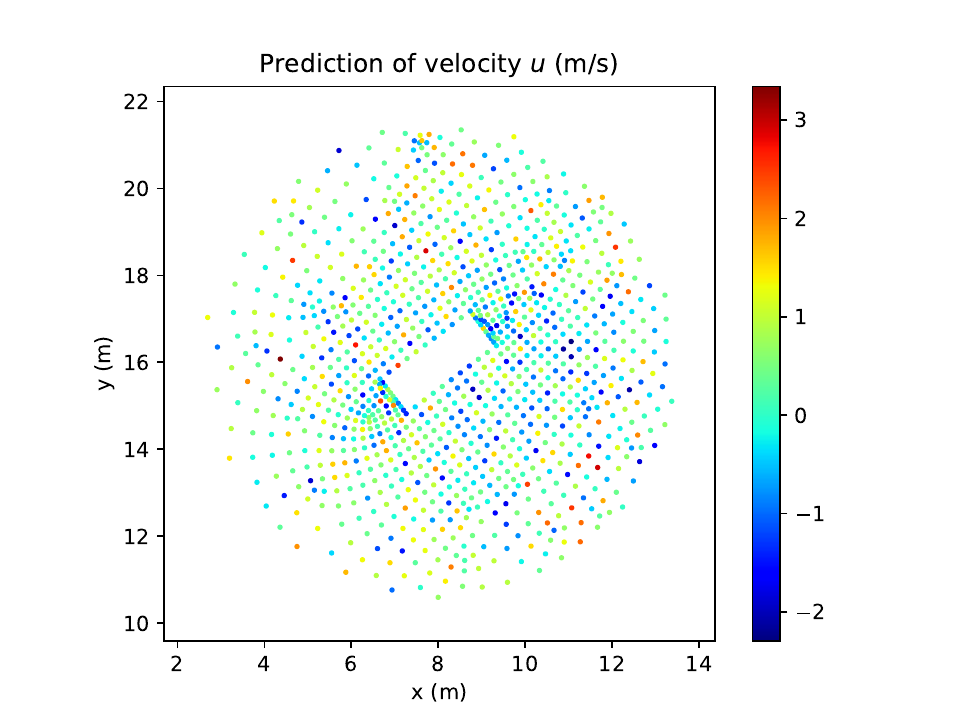} &
        \includegraphics[width=0.32\textwidth]{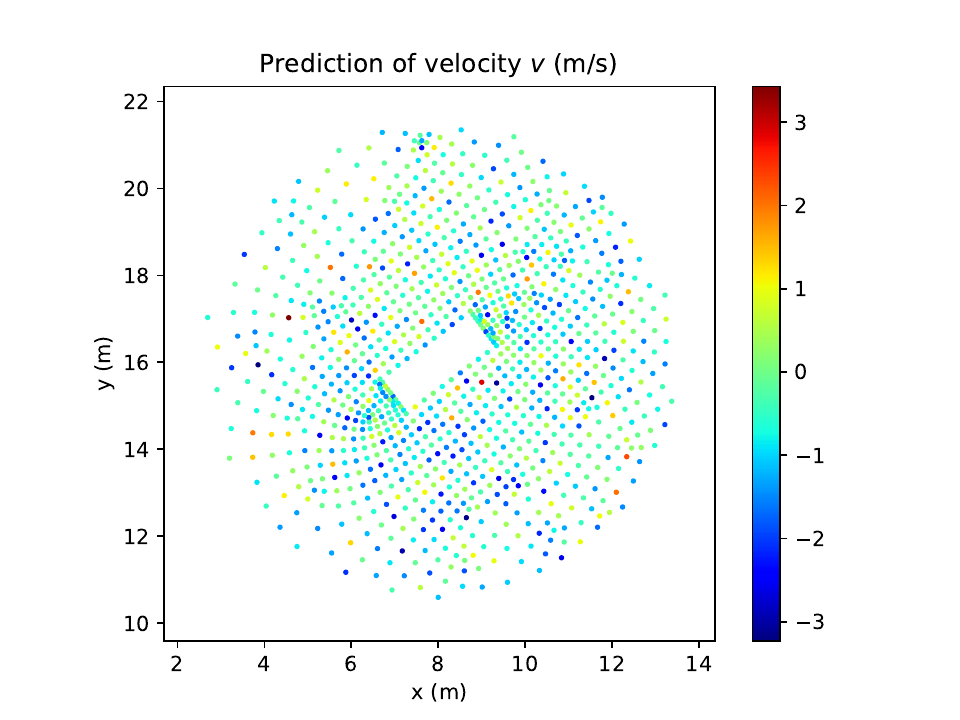} &
        \includegraphics[width=0.32\textwidth]{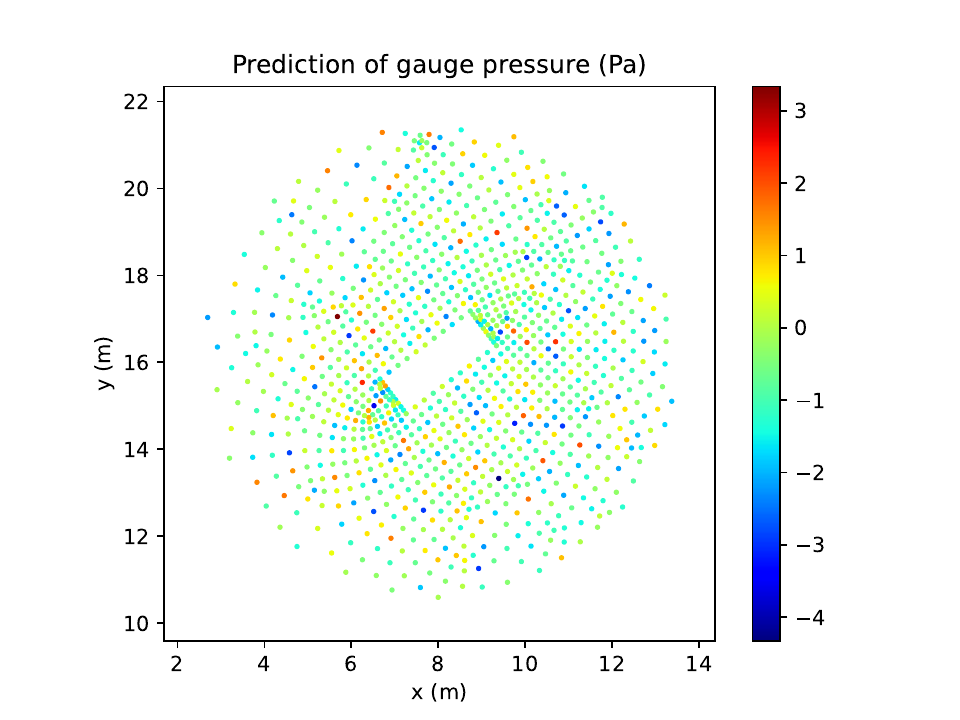} \\

        \rotatebox{90}{\ \ \quad \quad \quad \quad \text{$n=800$}} &
        \includegraphics[width=0.32\textwidth]{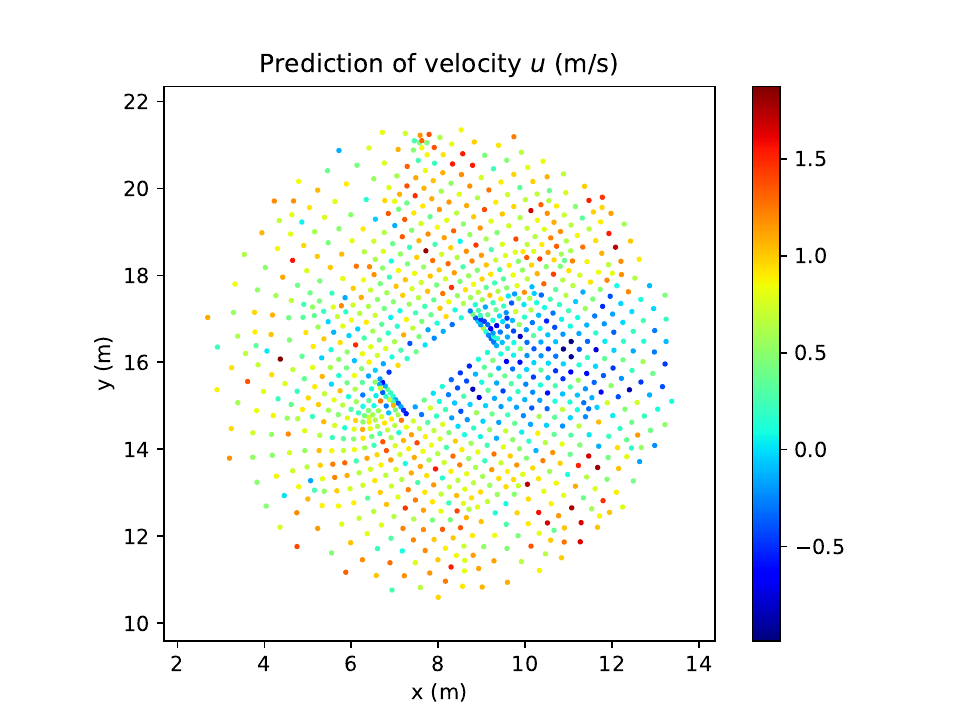} &
        \includegraphics[width=0.32\textwidth]{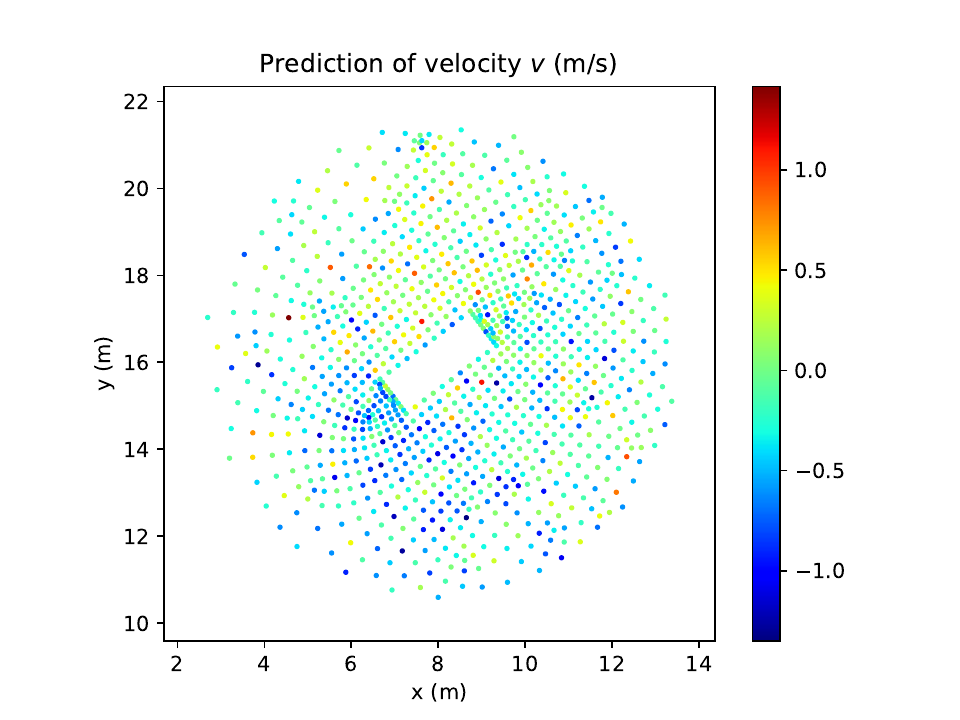} &
        \includegraphics[width=0.32\textwidth]{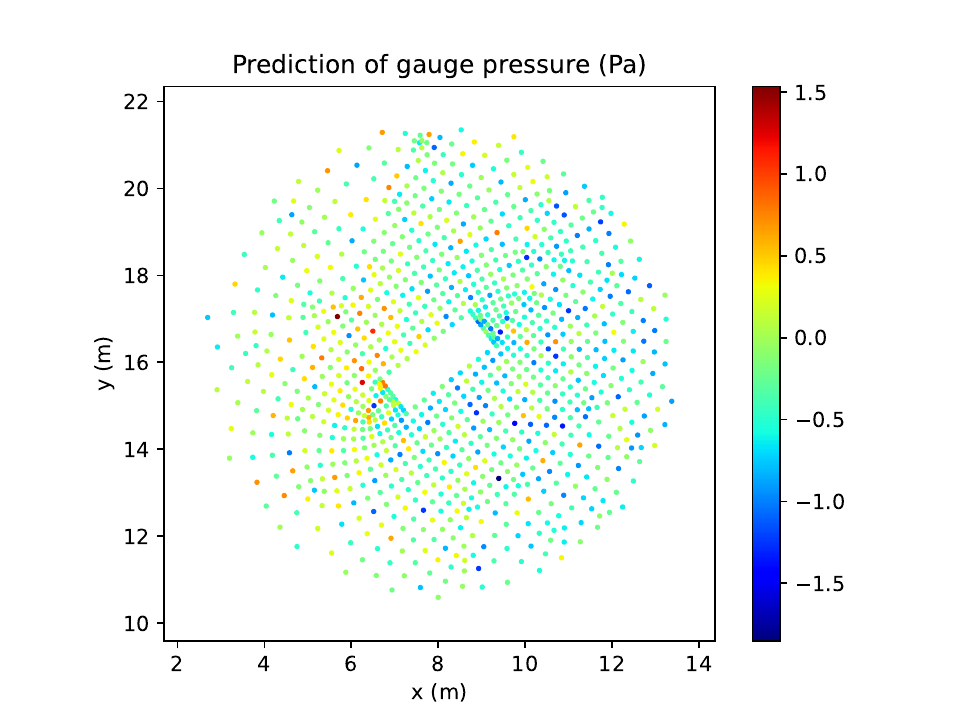} \\

        \rotatebox{90}{\ \ \quad \quad \quad \quad \text{$n=950$}} &
        \includegraphics[width=0.32\textwidth]{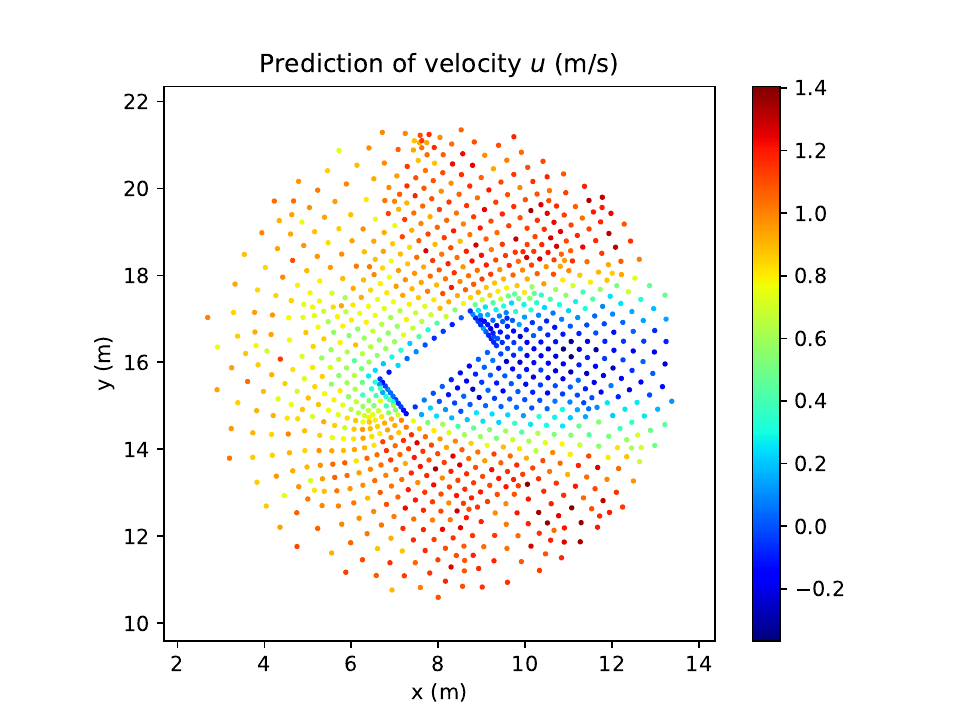} &
        \includegraphics[width=0.32\textwidth]{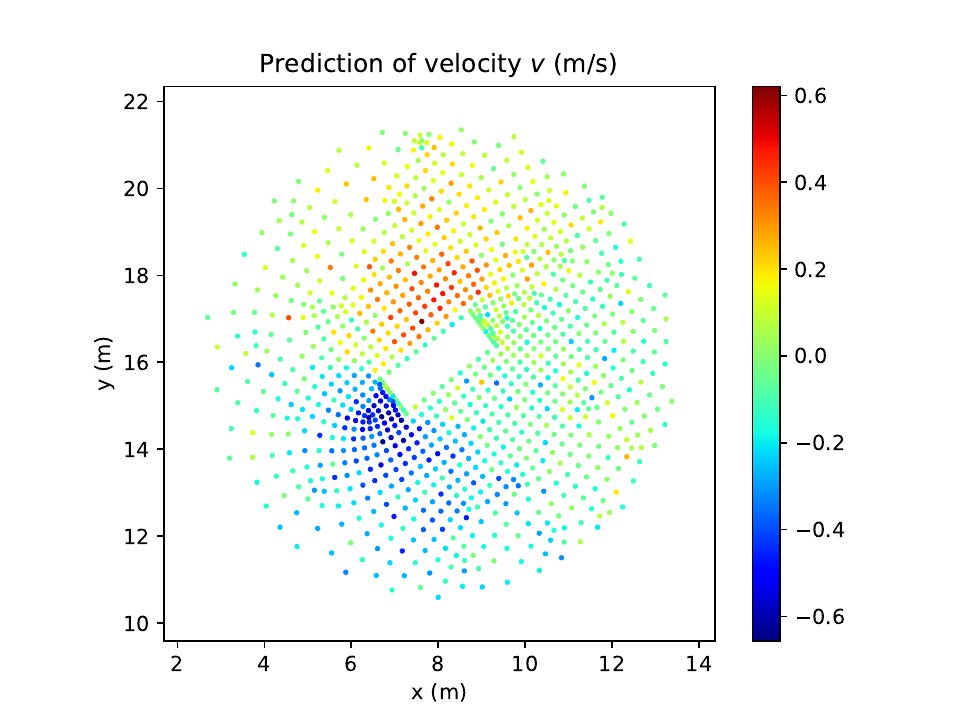} &
        \includegraphics[width=0.32\textwidth]{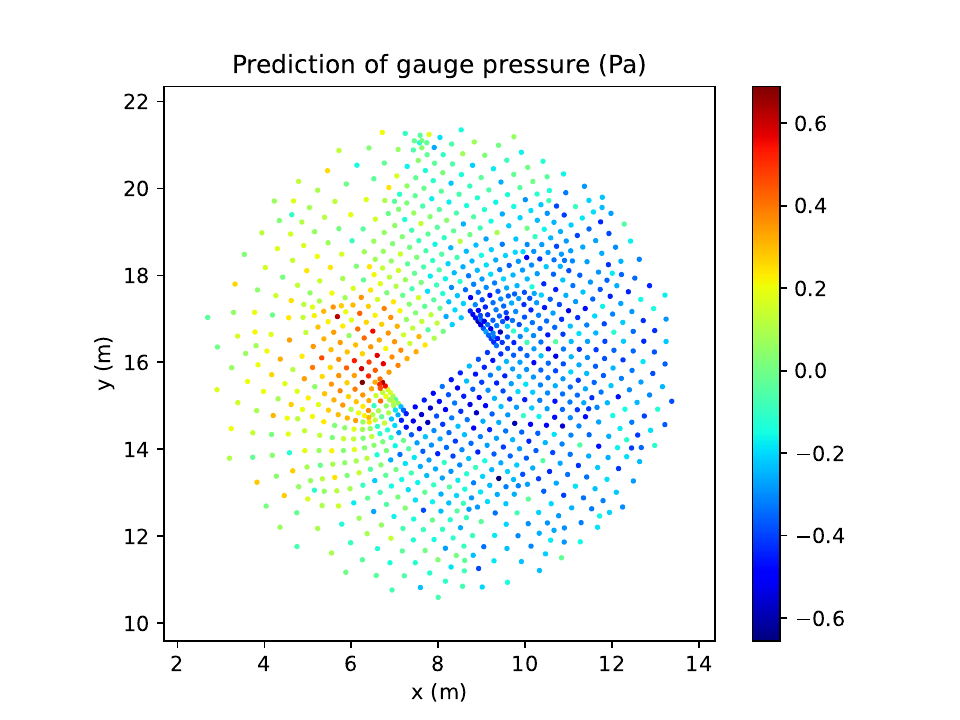} \\

        \rotatebox{90}{\ \ \quad \quad \quad \quad \text{$n=980$}} &
        \includegraphics[width=0.32\textwidth]{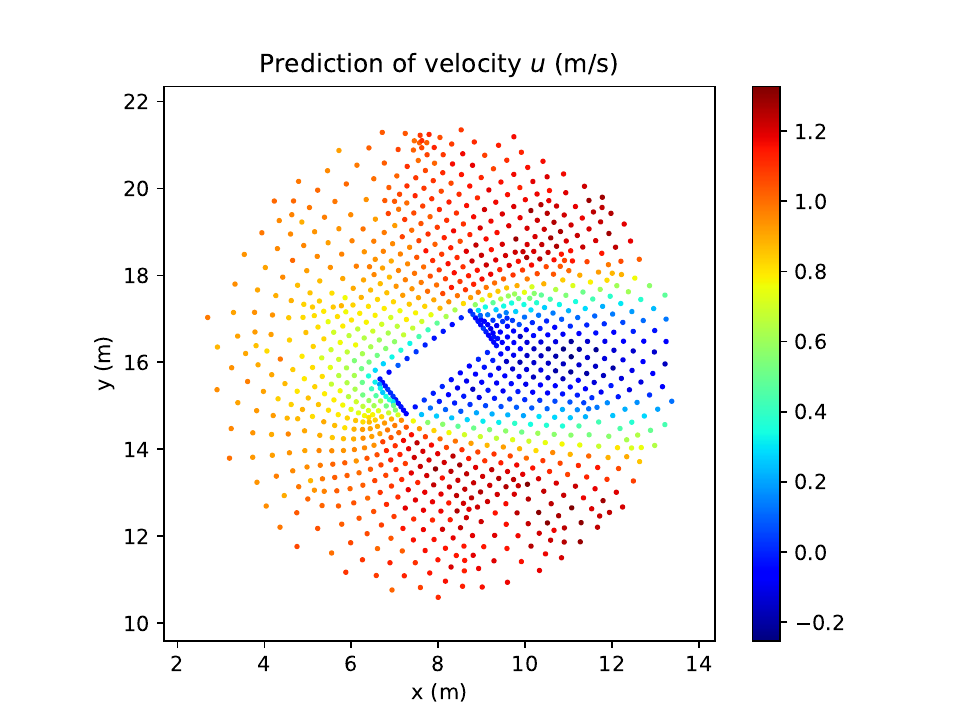} &
        \includegraphics[width=0.32\textwidth]{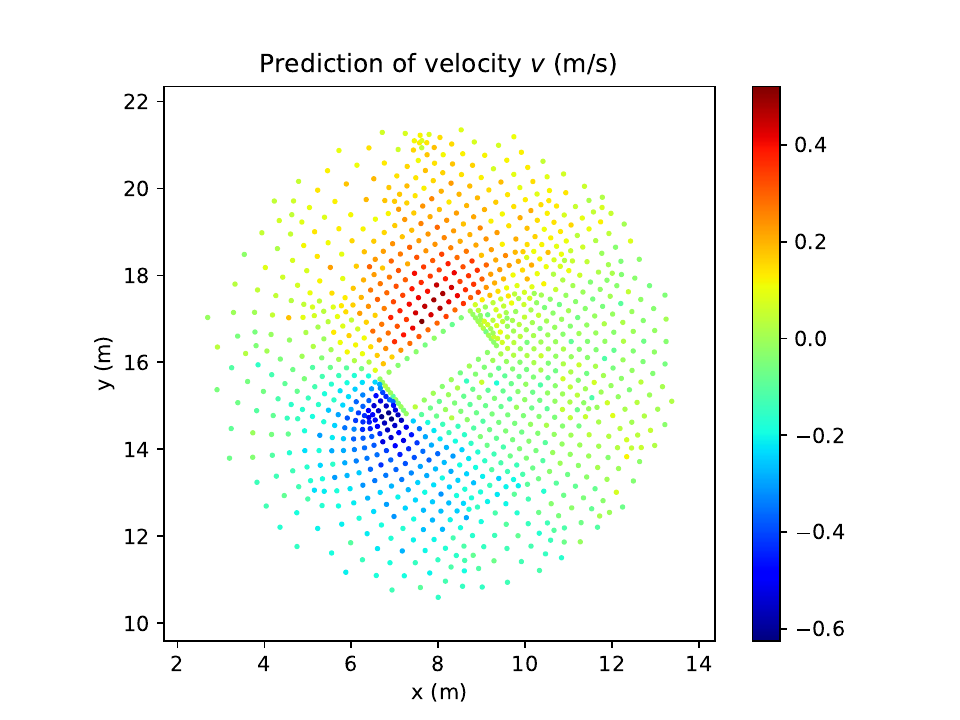} &
        \includegraphics[width=0.32\textwidth]{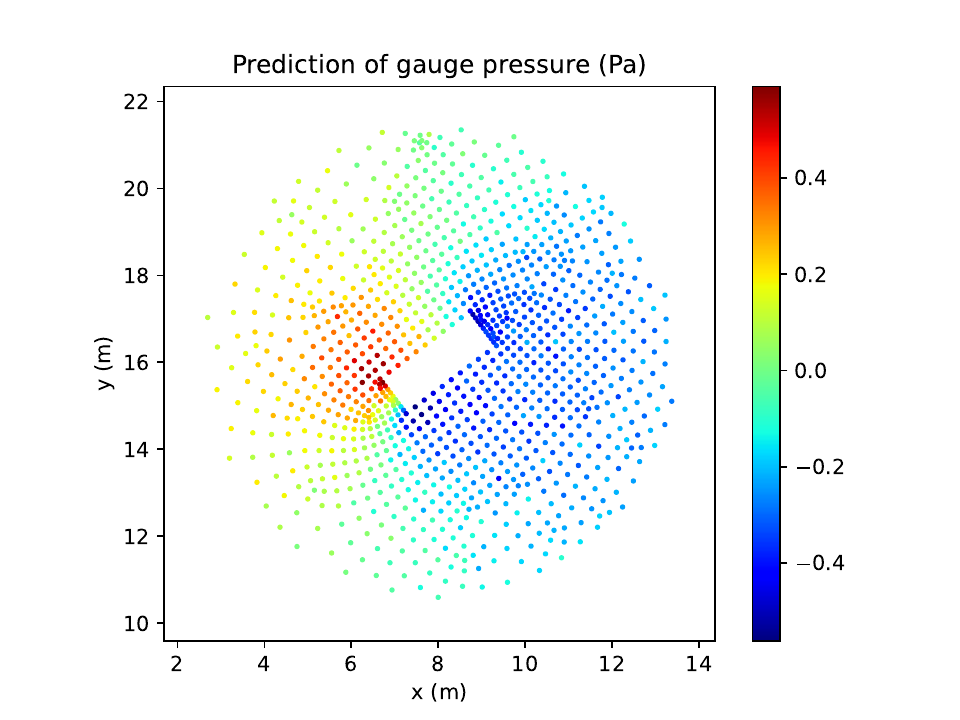} \\

        \rotatebox{90}{\ \ \quad \quad \quad \quad \text{$n=1000$}} &
        \includegraphics[width=0.32\textwidth]{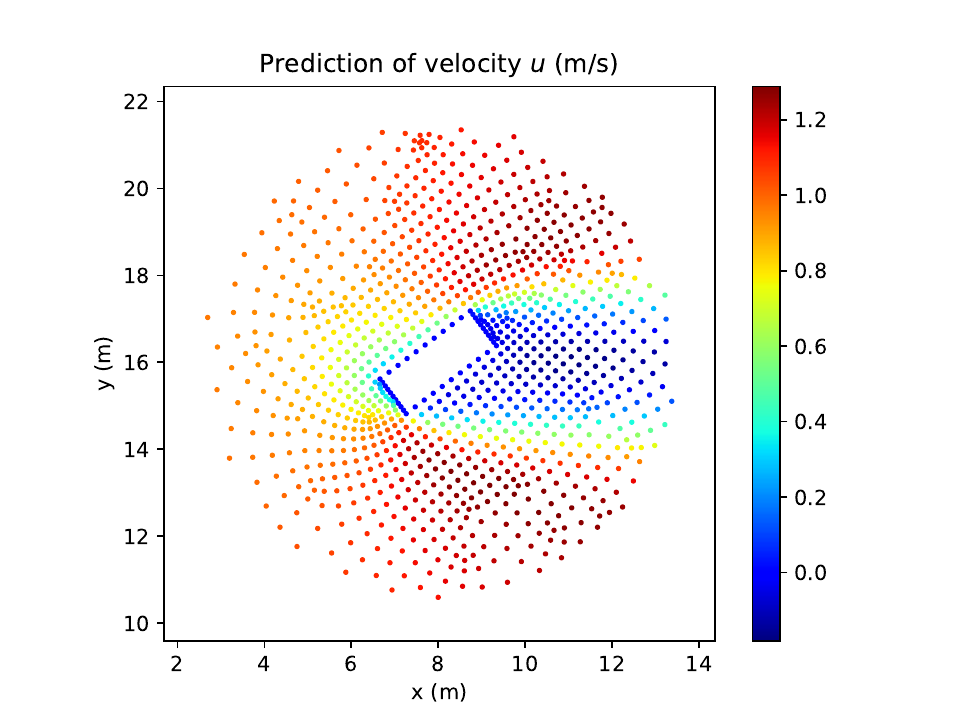} &
        \includegraphics[width=0.32\textwidth]{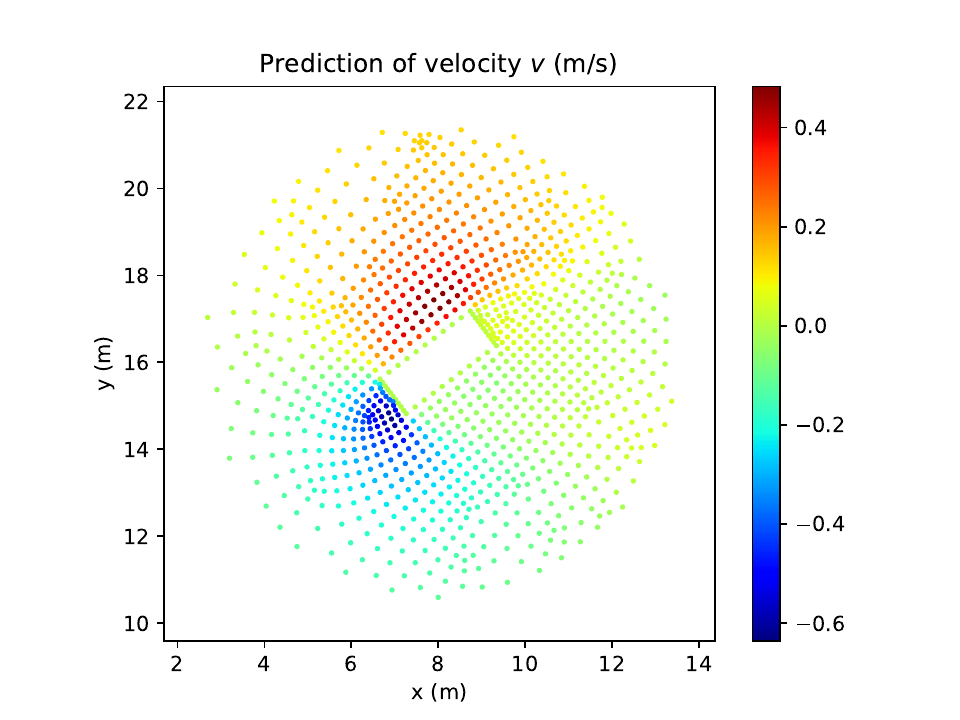} &
        \includegraphics[width=0.32\textwidth]{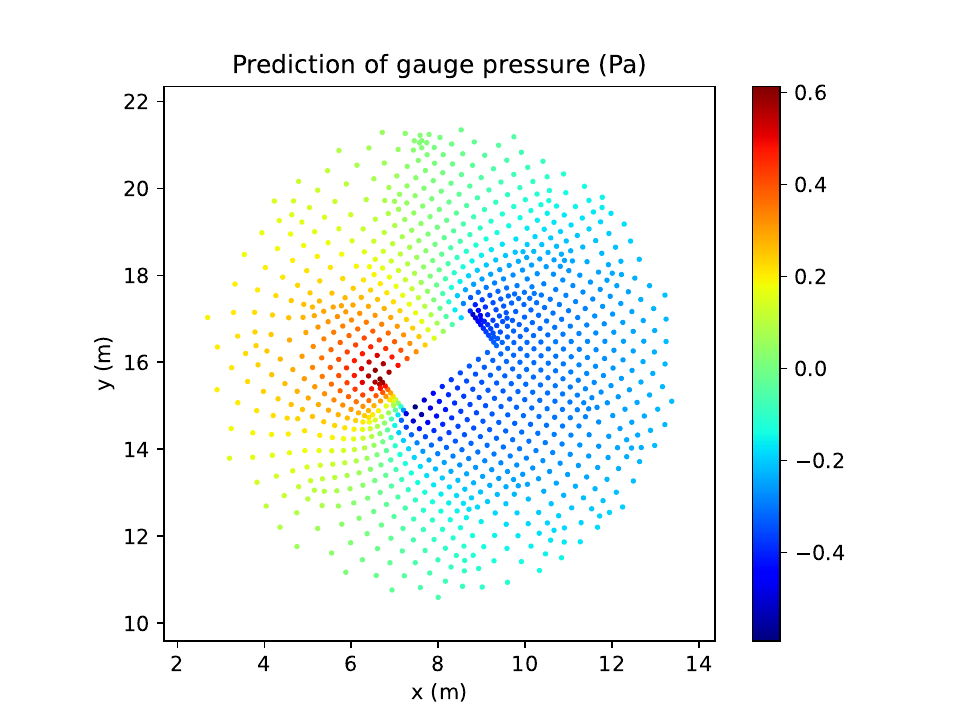} \\
    \end{tabular}

   \caption{Evolution of velocity and pressure fields during the reverse process of Flow Matching PointNet at $n = 0$ (i.e., $n=N_\tau$, standard Gaussian noise), $n = 500$, $n = 800$, $n = 950$, $n = 980$, and $n = 1000$ (i.e., the clean field).}
    \label{Fig1A}
\end{figure}


\begin{figure}
    \centering

    \begin{tabular}{c c c c}
         & \text{velocity $u$ (m/s)} & \text{velocity $v$ (m/s)} & \text{gauge pressure $p$ (Pa)}\\
        \rotatebox{90}{\ \ \quad \quad \quad \quad \text{$t=1000$}} &
        \includegraphics[width=0.32\textwidth]{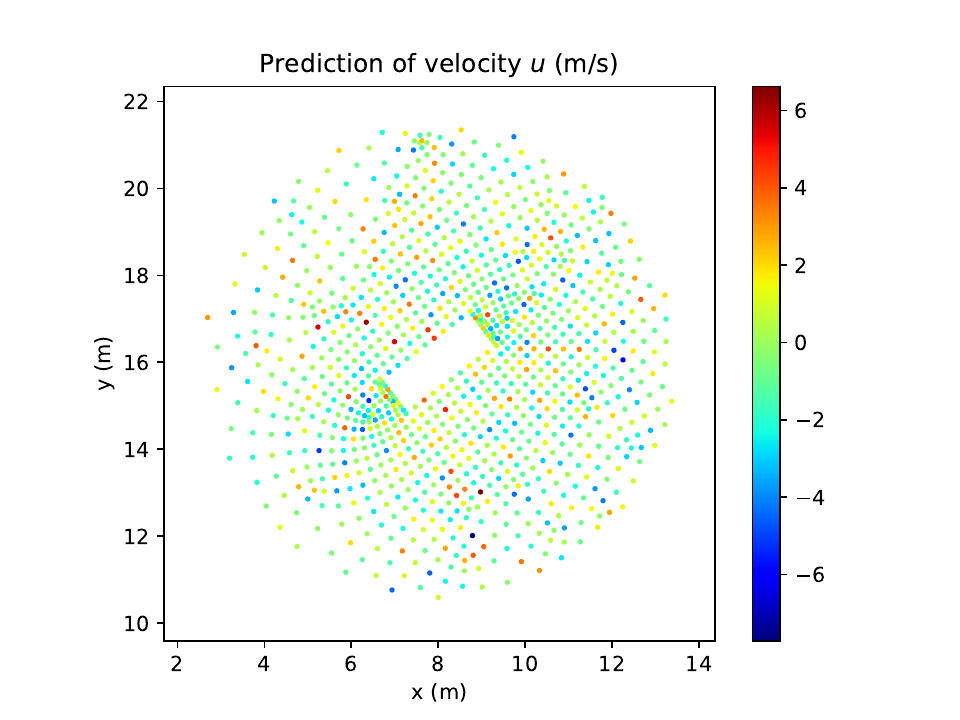} &
        \includegraphics[width=0.32\textwidth]{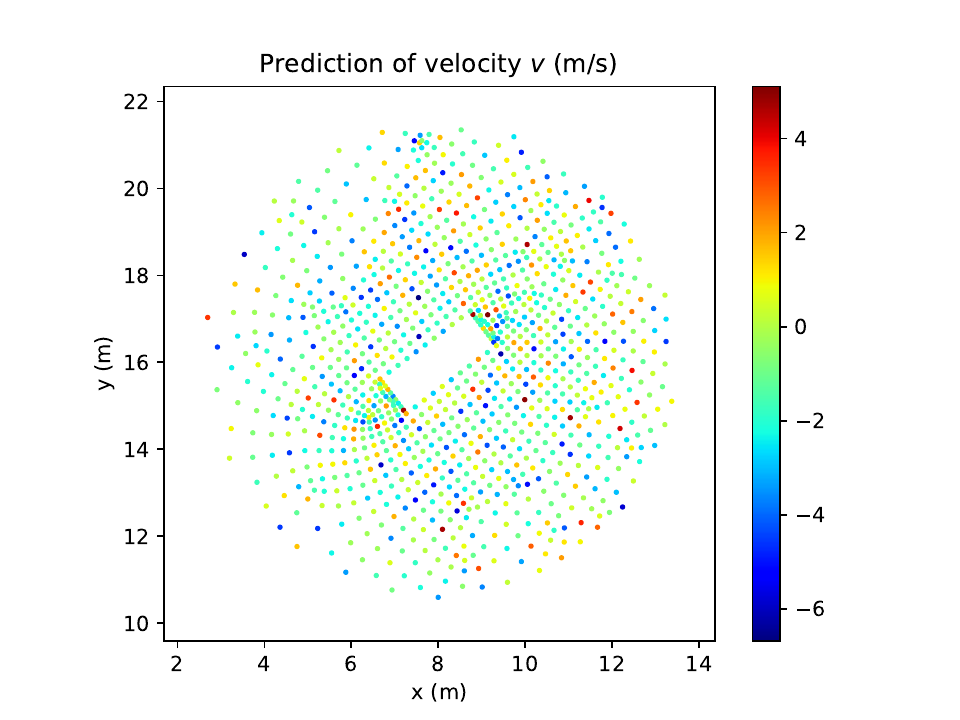} &
        \includegraphics[width=0.32\textwidth]{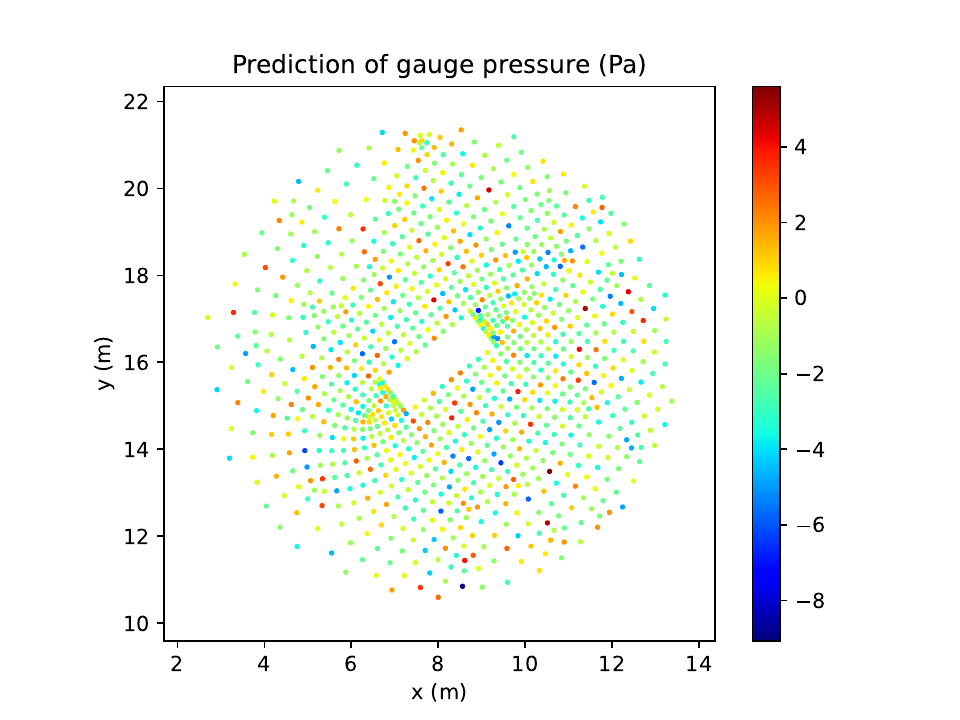} \\

        \rotatebox{90}{\ \ \quad \quad \quad \quad \text{$t=500$}} &
        \includegraphics[width=0.32\textwidth]{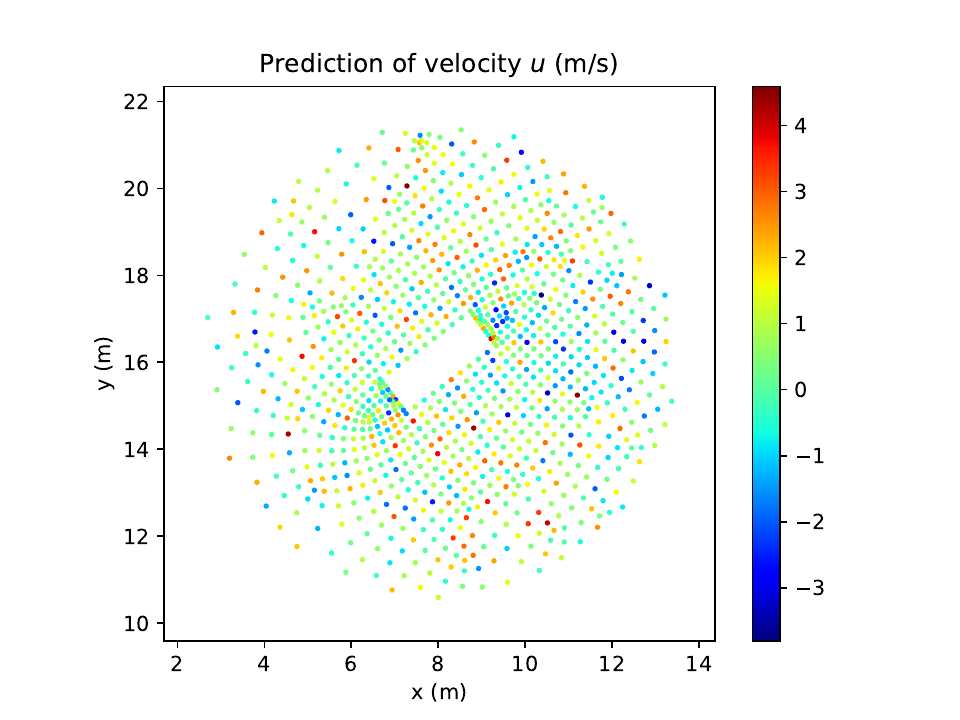} &
        \includegraphics[width=0.32\textwidth]{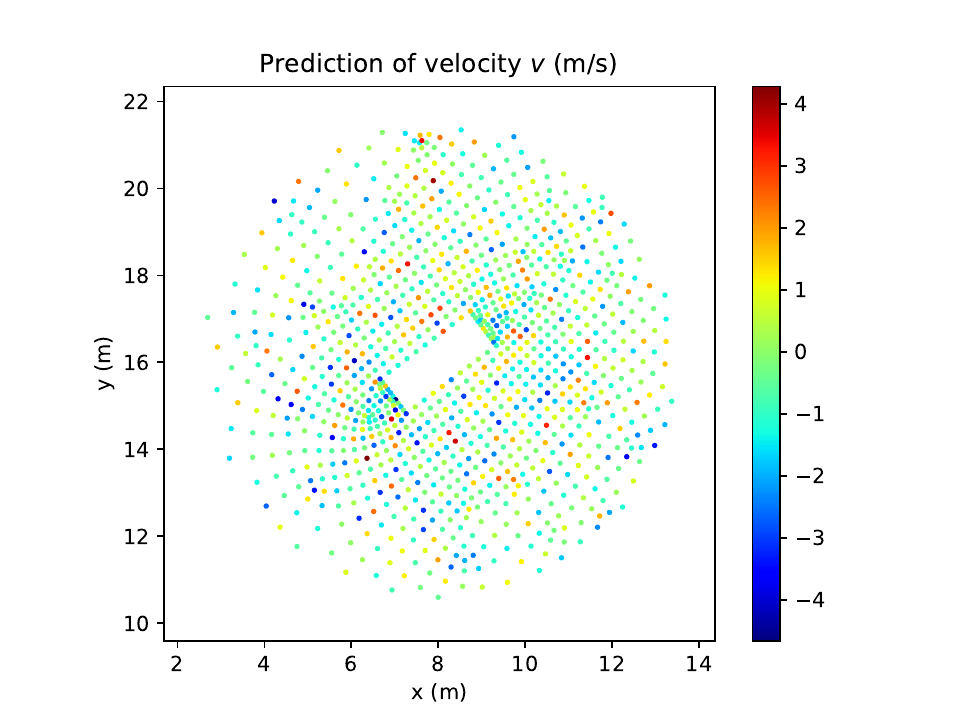} &
        \includegraphics[width=0.32\textwidth]{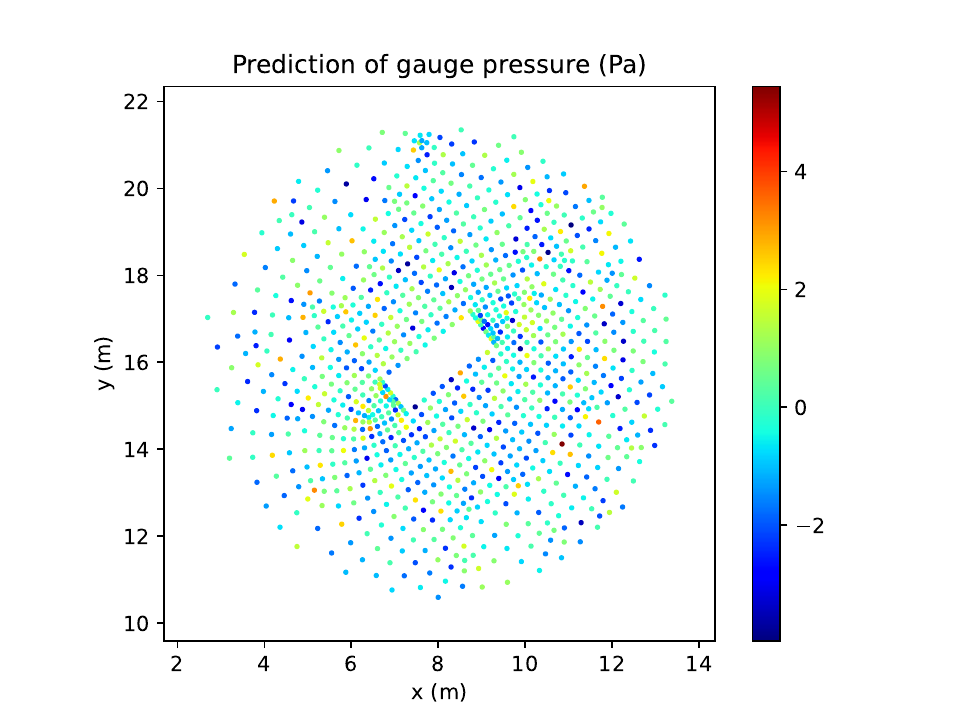} \\

        \rotatebox{90}{\ \ \quad \quad \quad \quad \text{$t=200$}} &
        \includegraphics[width=0.32\textwidth]{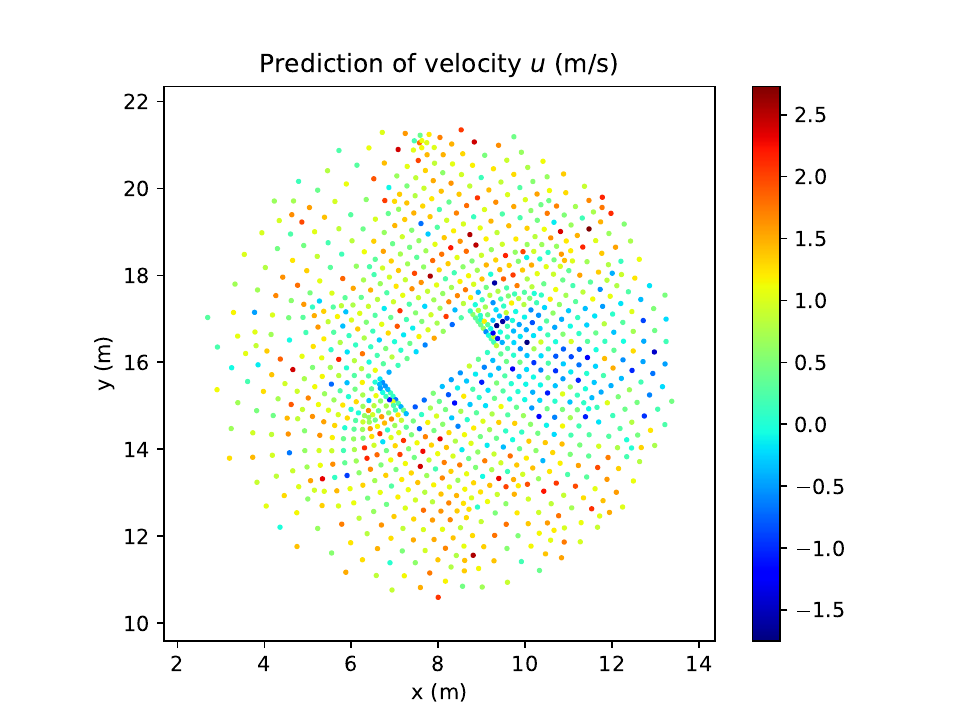} &
        \includegraphics[width=0.32\textwidth]{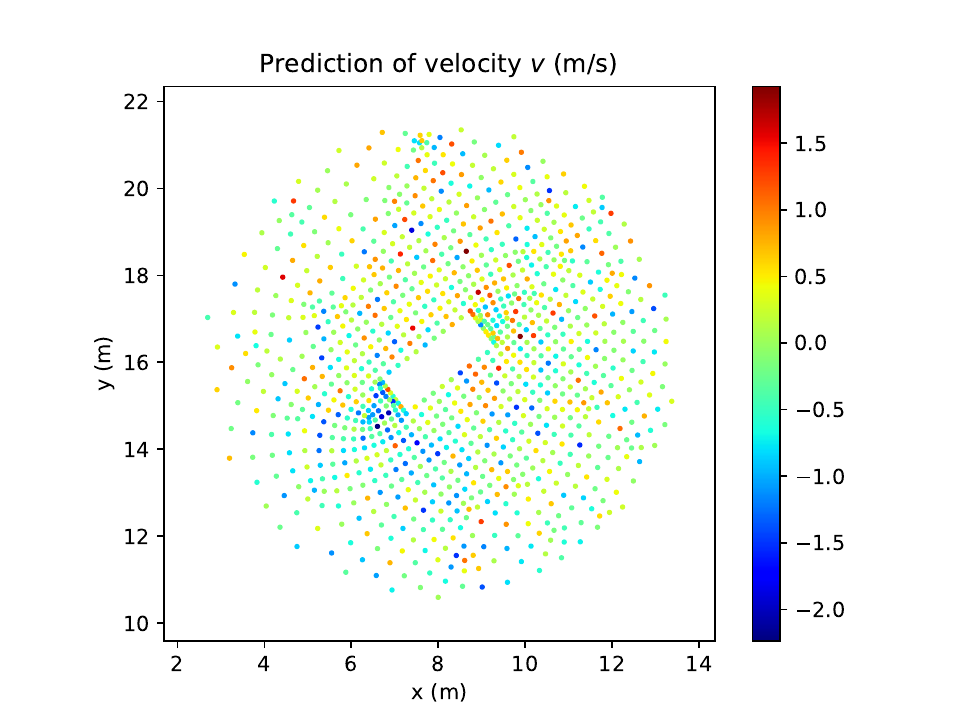} &
        \includegraphics[width=0.32\textwidth]{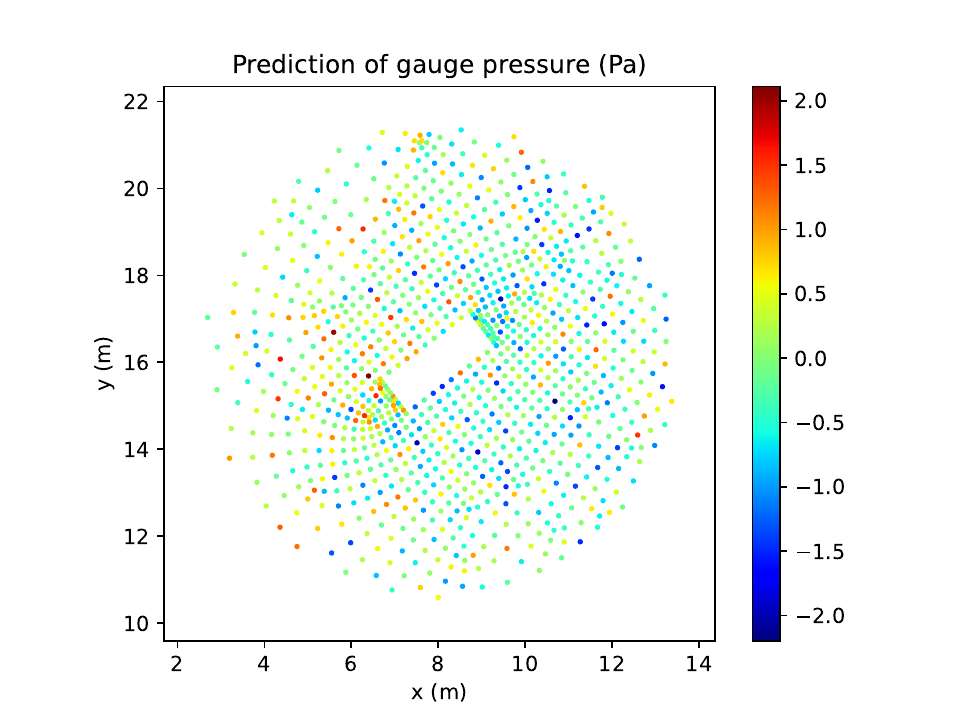} \\

        \rotatebox{90}{\ \ \quad \quad \quad \quad \text{$t=50$}} &
        \includegraphics[width=0.32\textwidth]{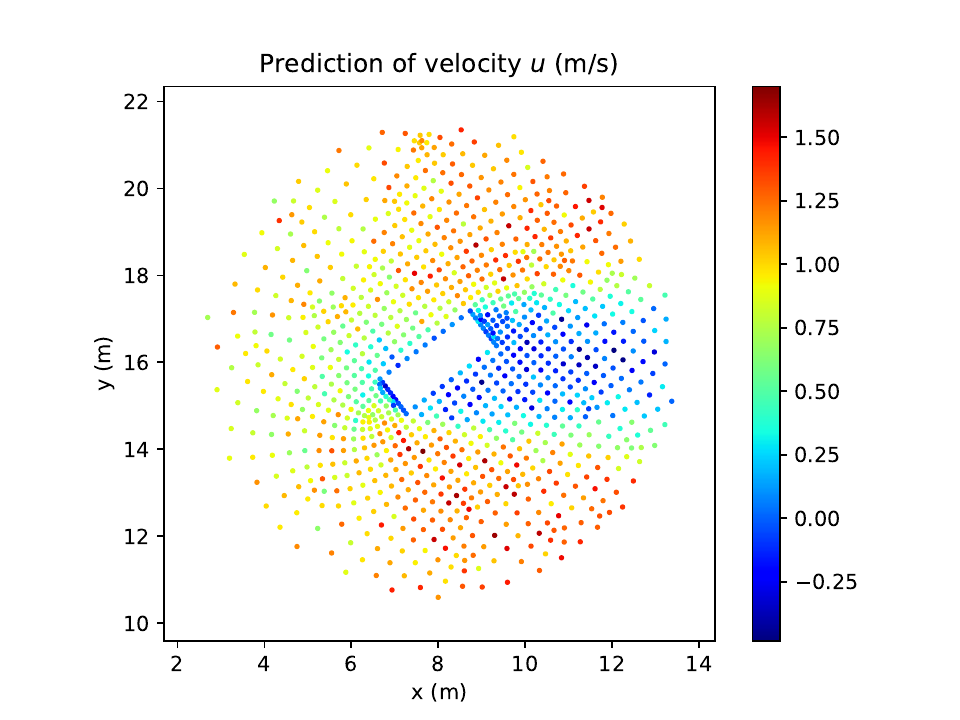} &
        \includegraphics[width=0.32\textwidth]{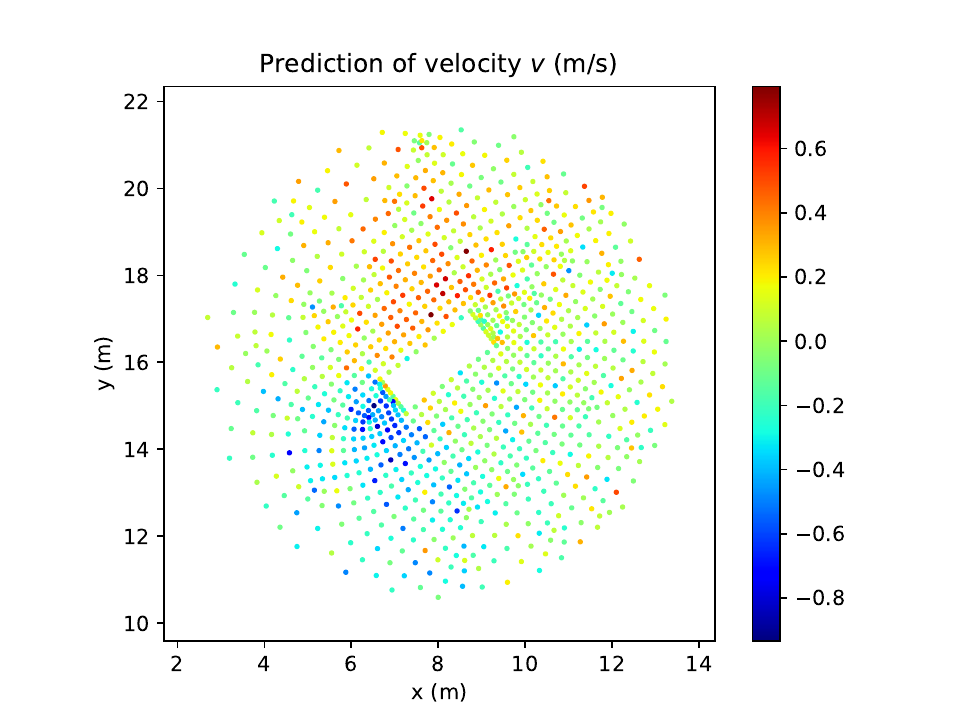} &
        \includegraphics[width=0.32\textwidth]{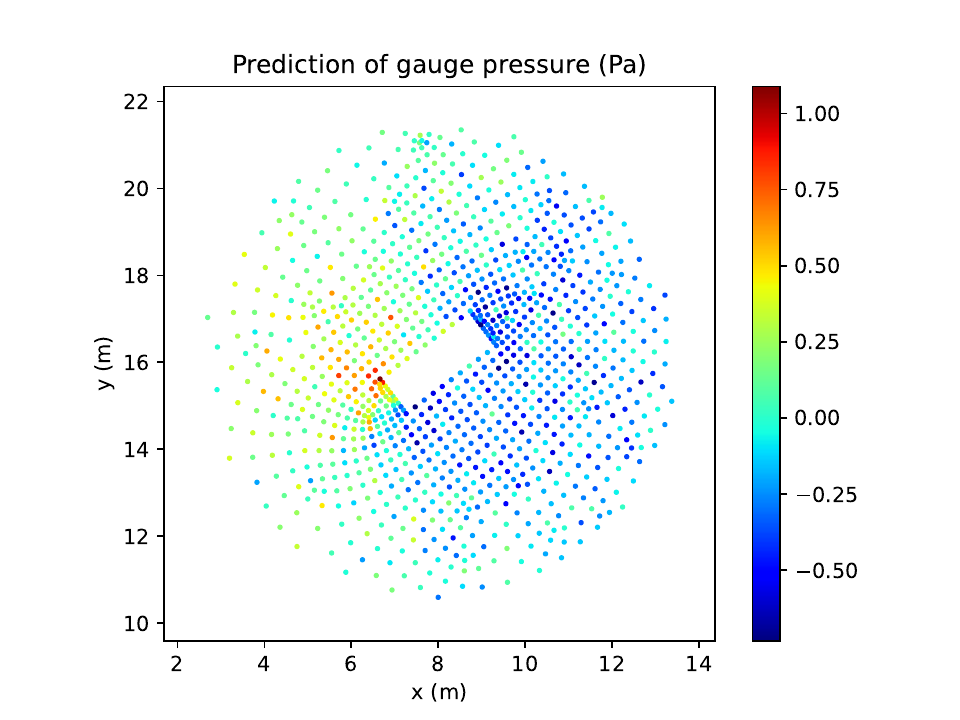} \\

        \rotatebox{90}{\ \ \quad \quad \quad \quad \text{$t=20$}} &
        \includegraphics[width=0.32\textwidth]{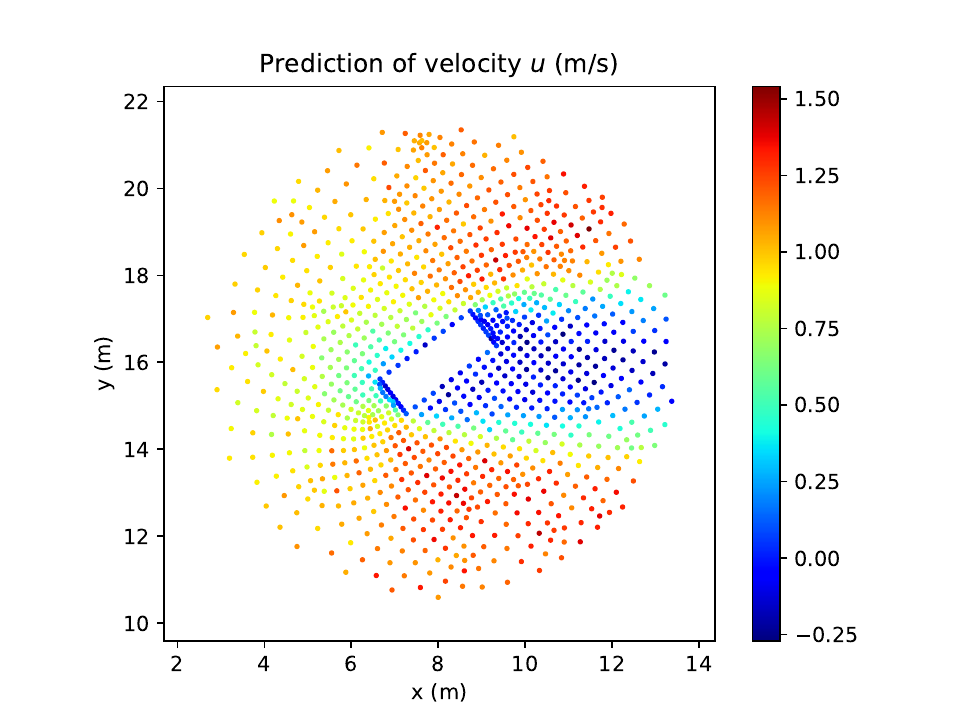} &
        \includegraphics[width=0.32\textwidth]{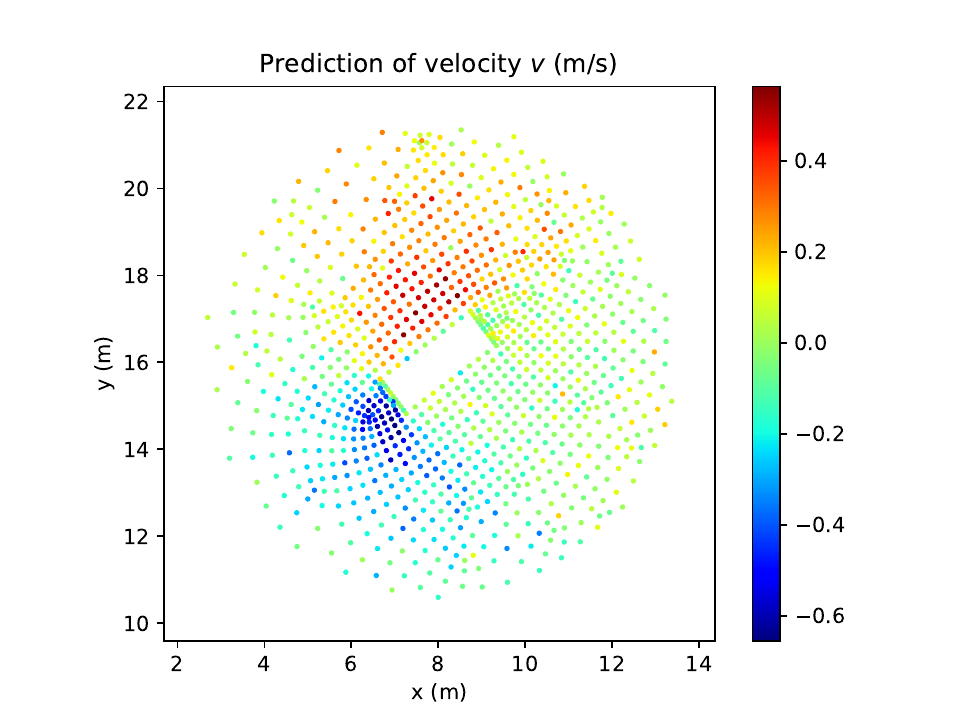} &
        \includegraphics[width=0.32\textwidth]{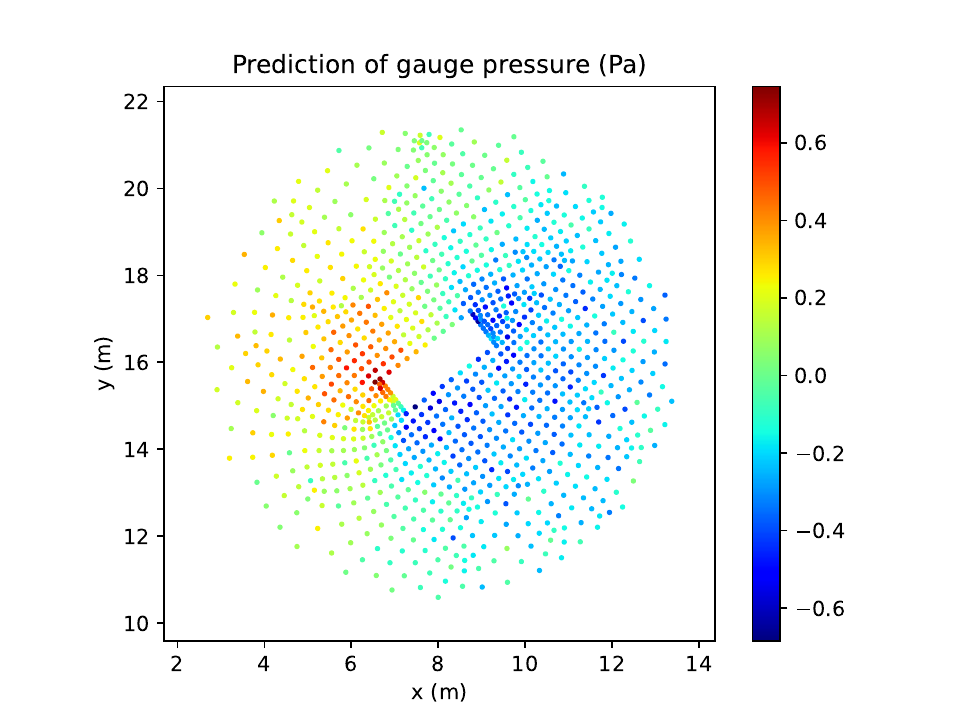} \\

        \rotatebox{90}{\ \ \quad \quad \quad \quad \text{$t=0$}} &
        \includegraphics[width=0.32\textwidth]{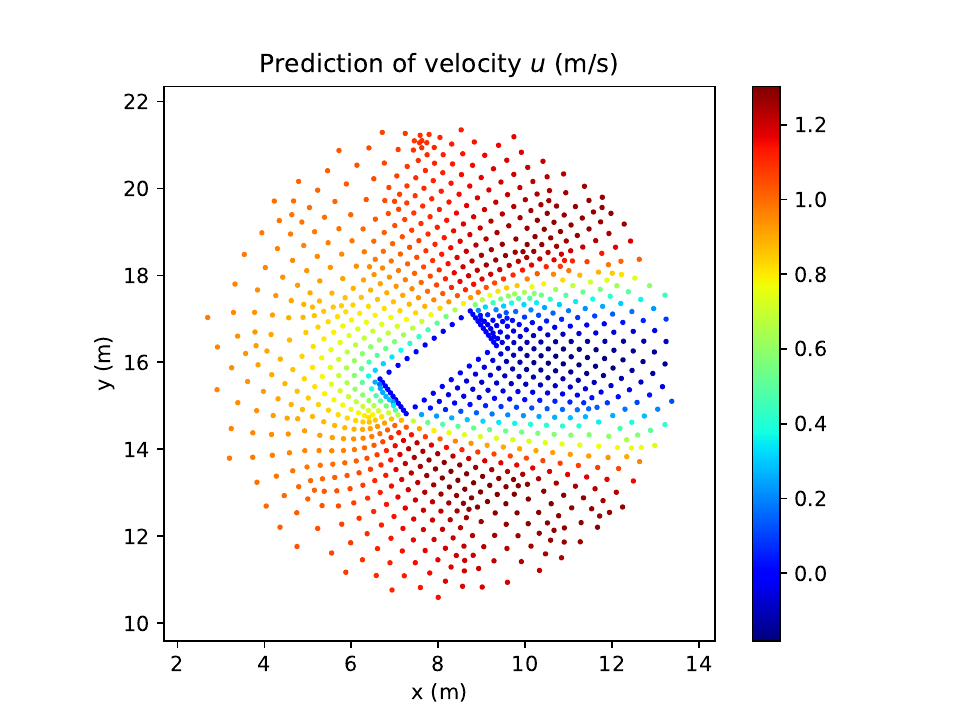} &
        \includegraphics[width=0.32\textwidth]{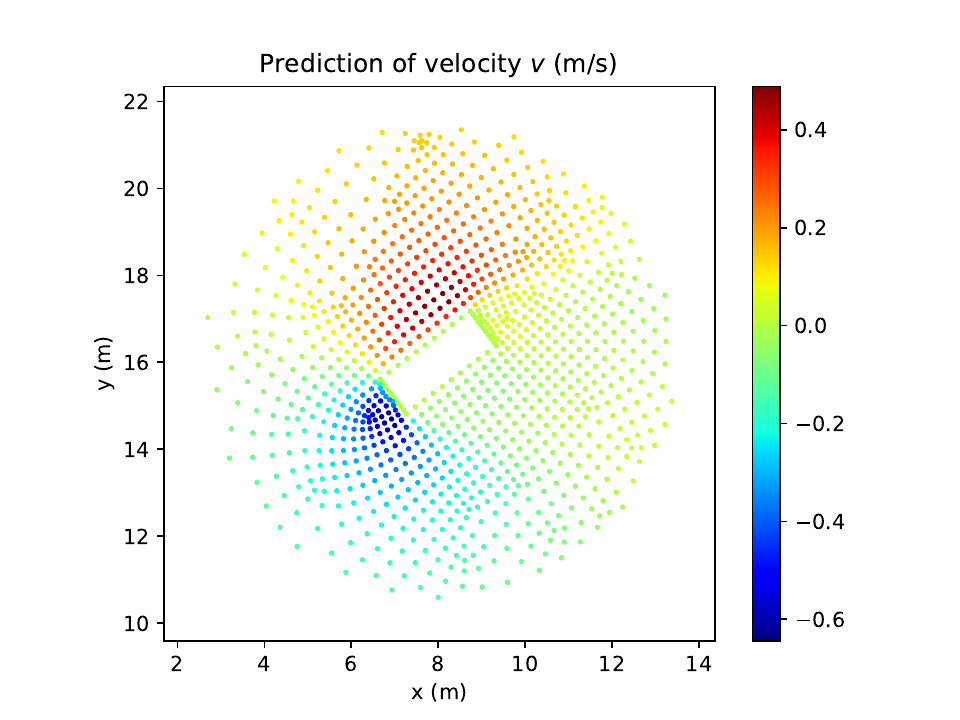} &
        \includegraphics[width=0.32\textwidth]{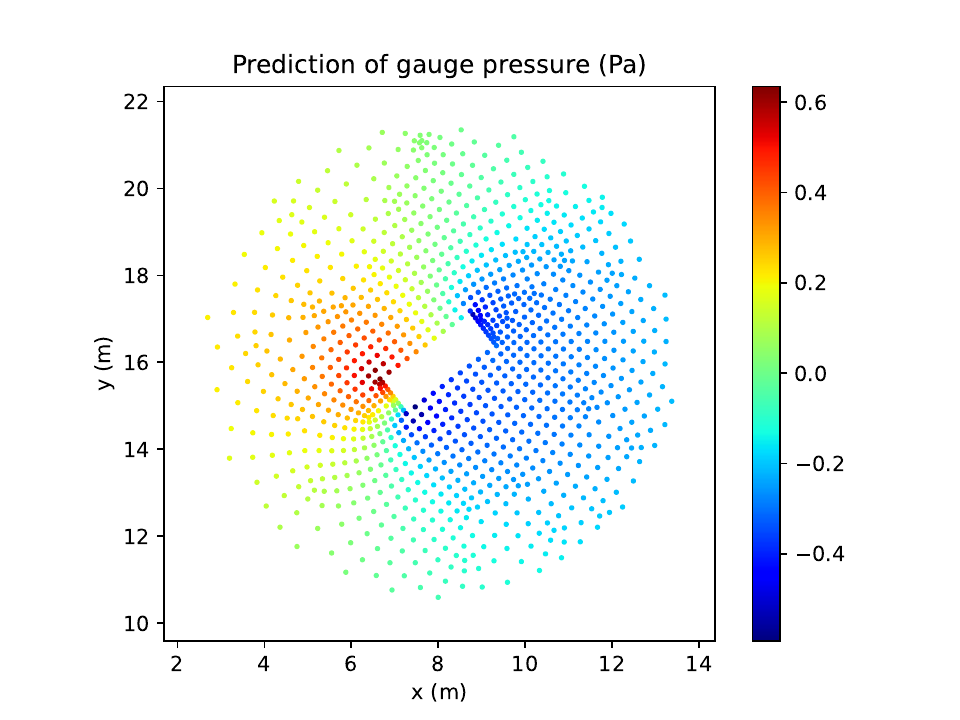} \\
    \end{tabular}

   \caption{Evolution of velocity and pressure fields during the reverse process of Diffusion PointNet at $t = 1000$ (i.e., $t = T$, standard Gaussian noise), $t = 500$, $t = 200$, $t = 50$, $t = 20$, and $t = 0$ (i.e., the clean field).}
    \label{Fig1B}
\end{figure}


\subsection{Reverse generative process}
\label{Sect4BReverse}

Figures \ref{Fig1A} and \ref{Fig1B} respectively illustrate the reverse processes of the Flow Matching PointNet and the Diffusion PointNet for fluid flow fields around a cylinder with a rectangular cross section, shown as a representative example from the test set. More specifically, Figure \ref{Fig1A} presents the velocity and pressure fields at different reverse steps, including $n = 0$ (i.e., where all three fields are sampled from standard Gaussian noise), $n = 500$, $n = 800$, $n = 950$, $n = 980$, and $n = 1000$, which corresponds to the predicted clean velocity and pressure fields. As can be seen from Fig. \ref{Fig1A}, at $n = 500$, the overall structure of the velocity and pressure fields is still largely visually obscured; however, the magnitudes of the variables are approximately within the range of $[-3,\,3]$, compared to the case $n = 0$, where the variables lie roughly in the range of $[-6,\,6]$ due to the standard Gaussian noise. Similarly, at $n = 800$, the noise level decreases such that the flow patterns become partially visible. At $n = 950$, the velocity and pressure fields are largely reconstructed, and during the remaining reverse steps, they are further refined, with the predicted values converging to physically acceptable ranges. Quantitative error metrics have already been reported in Table \ref{Table5}.

Figure \ref{Fig1B} shows similar trends for the reverse process of the Diffusion PointNet, starting from $t = 1000$ (i.e., the standard Gaussian noisy field) and progressing to $t = 0$, which corresponds to the predicted velocity and pressure fields. By comparing the outputs shown in Figures \ref{Fig1A} and \ref{Fig1B}, it can be observed that the underlying mechanisms of the Flow Matching PointNet and the Diffusion PointNet differ. For example, consider the velocity and pressure field predictions obtained by the Flow Matching PointNet at $n = 950$ and by the Diffusion PointNet at $t = 50$, corresponding to the same number of reverse steps in the two models. Given the results shown in Fig. \ref{Fig1A} and Fig. \ref{Fig1B}, a visual comparison indicates that the fluid flow patterns obtained by the Flow Matching PointNet exhibit a lower noise level and more closely resemble the predicted clean flow fields at this step (i.e., $n=950$) than those obtained by the Diffusion PointNet at the corresponding step (i.e., $t=50$). This qualitative observation highlights a fundamental difference in the reverse dynamics of the two models.


\begin{figure}[!htbp]
  \centering 
      \begin{subfigure}[b]{0.27\textwidth}
        \centering
        \includegraphics[width=\textwidth]{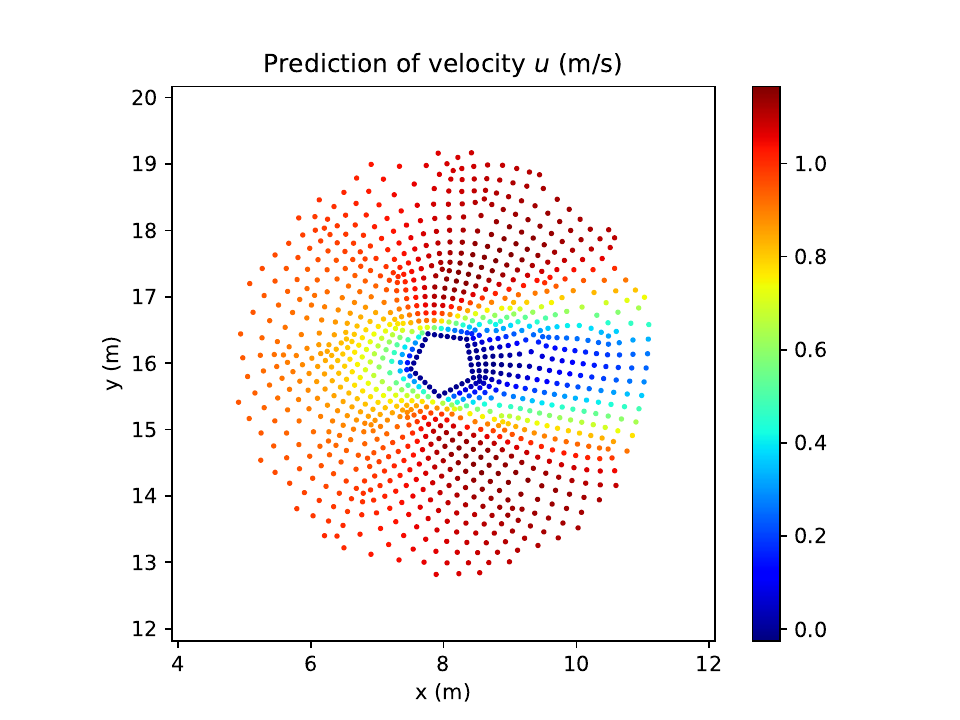}
    \end{subfigure}
    \begin{subfigure}[b]{0.27\textwidth}
        \centering
        \includegraphics[width=\textwidth]{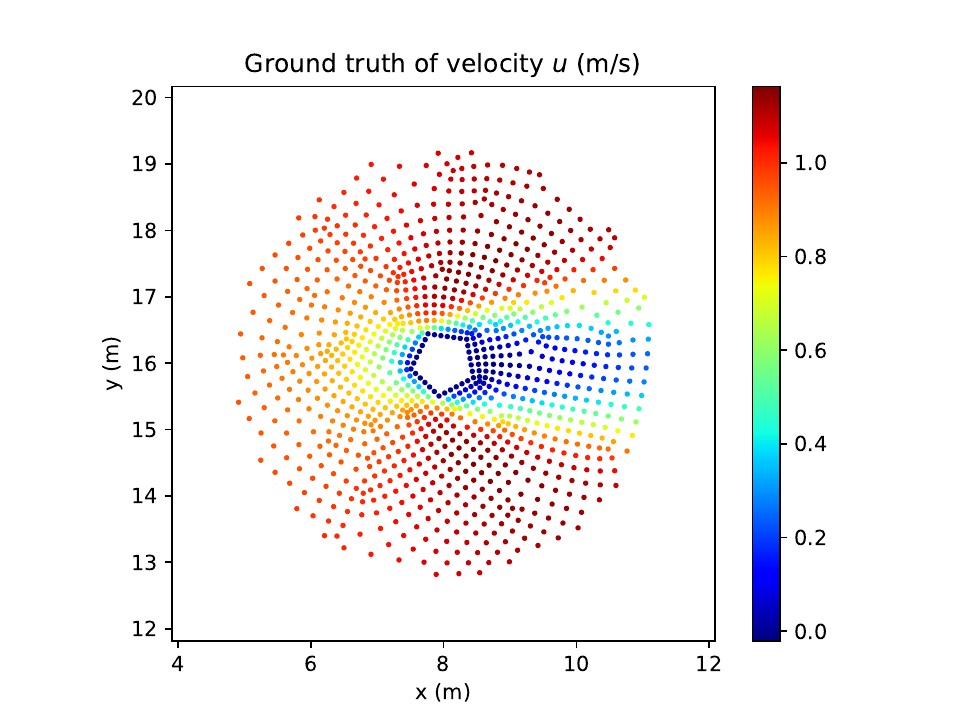}
    \end{subfigure}
    \begin{subfigure}[b]{0.27\textwidth}
        \centering
        \includegraphics[width=\textwidth]{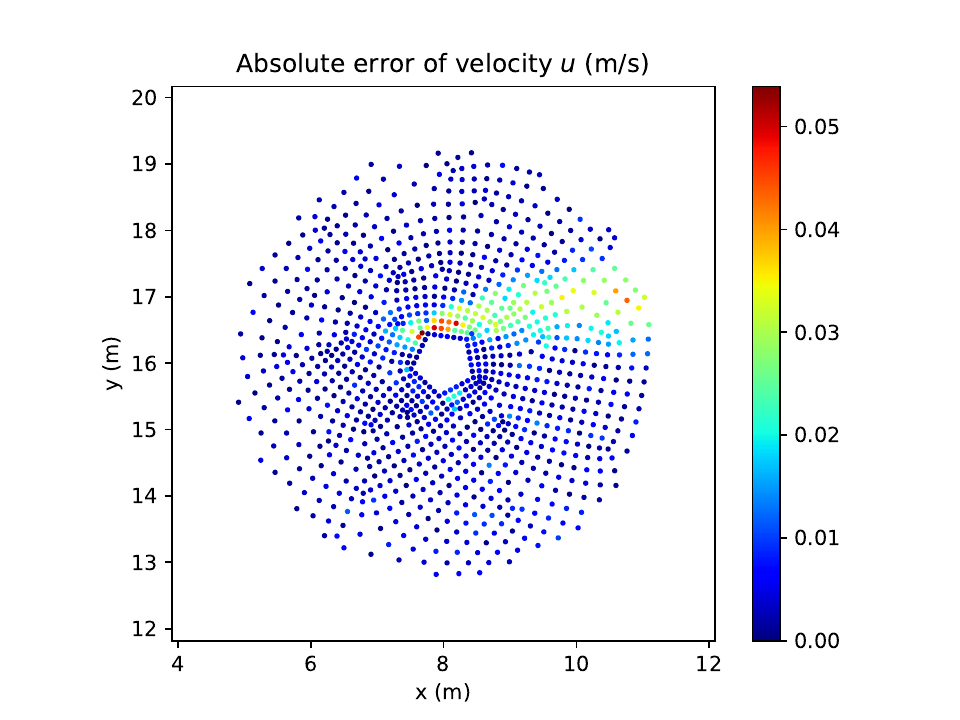}
    \end{subfigure}

    
    \begin{subfigure}[b]{0.27\textwidth}
        \centering
        \includegraphics[width=\textwidth]{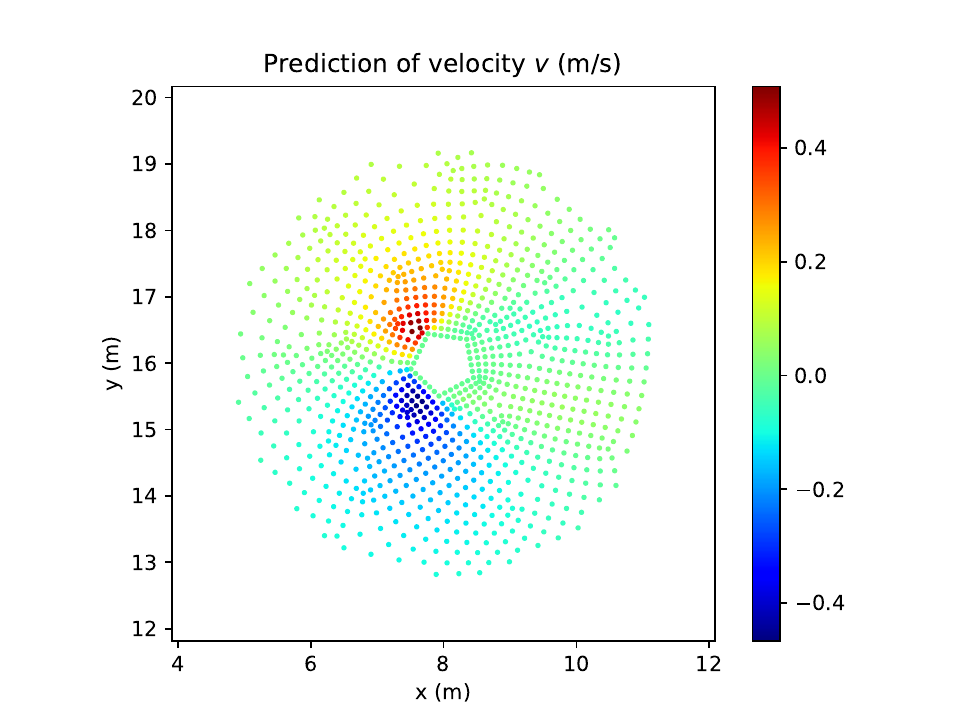}
    \end{subfigure}
    \begin{subfigure}[b]{0.27\textwidth}
        \centering
        \includegraphics[width=\textwidth]{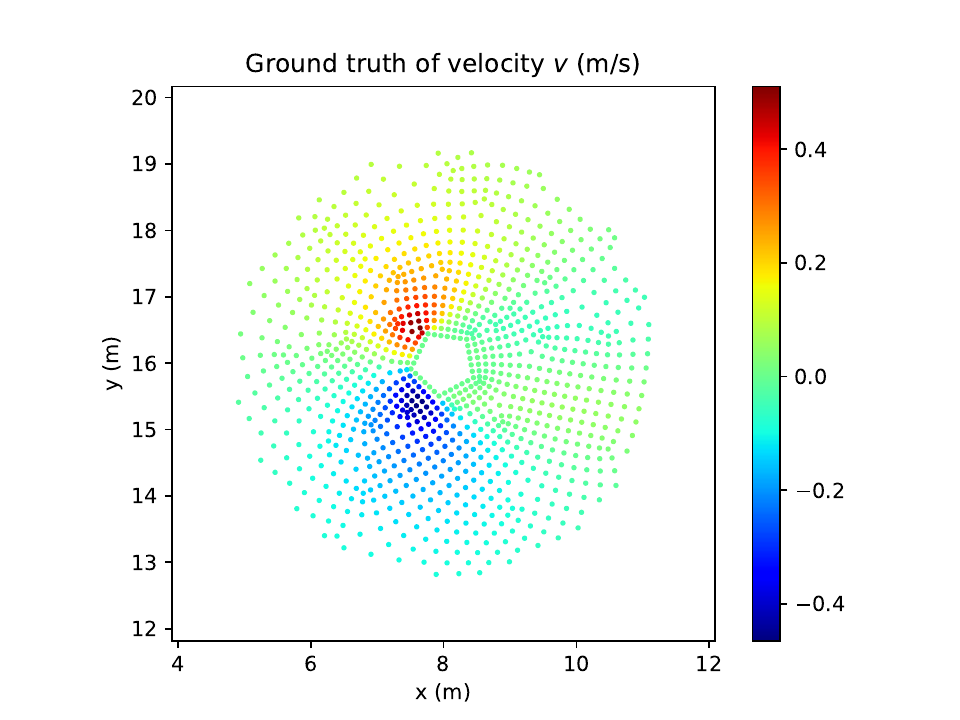}
    \end{subfigure}
    \begin{subfigure}[b]{0.27\textwidth}
        \centering
        \includegraphics[width=\textwidth]{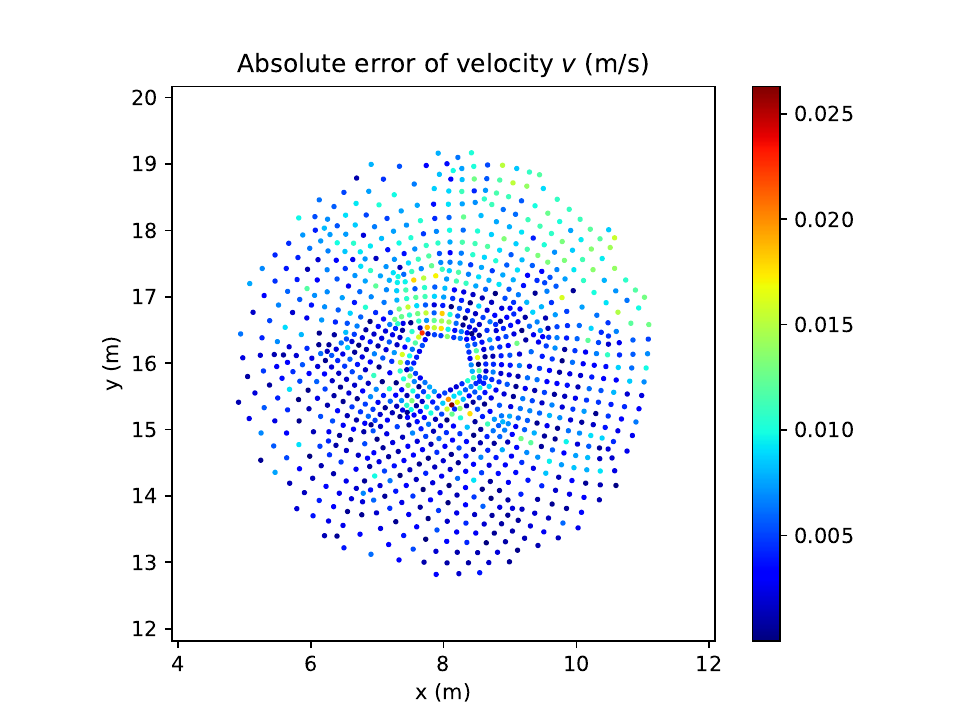}
    \end{subfigure}

    
    \begin{subfigure}[b]{0.27\textwidth}
        \centering
        \includegraphics[width=\textwidth]{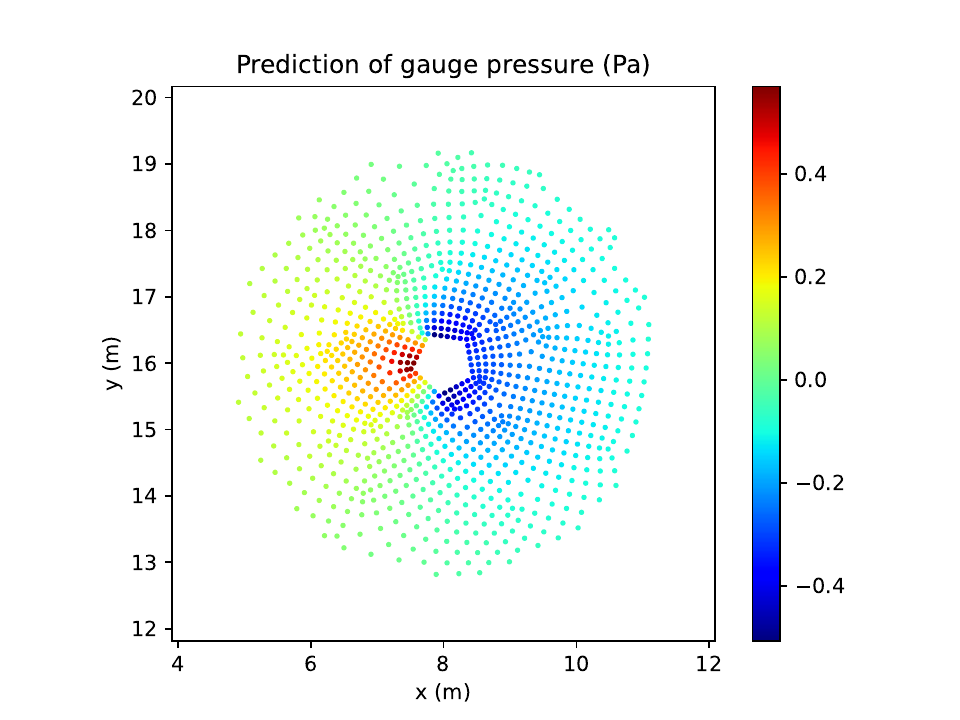}
    \end{subfigure}
    \begin{subfigure}[b]{0.27\textwidth}
        \centering
        \includegraphics[width=\textwidth]{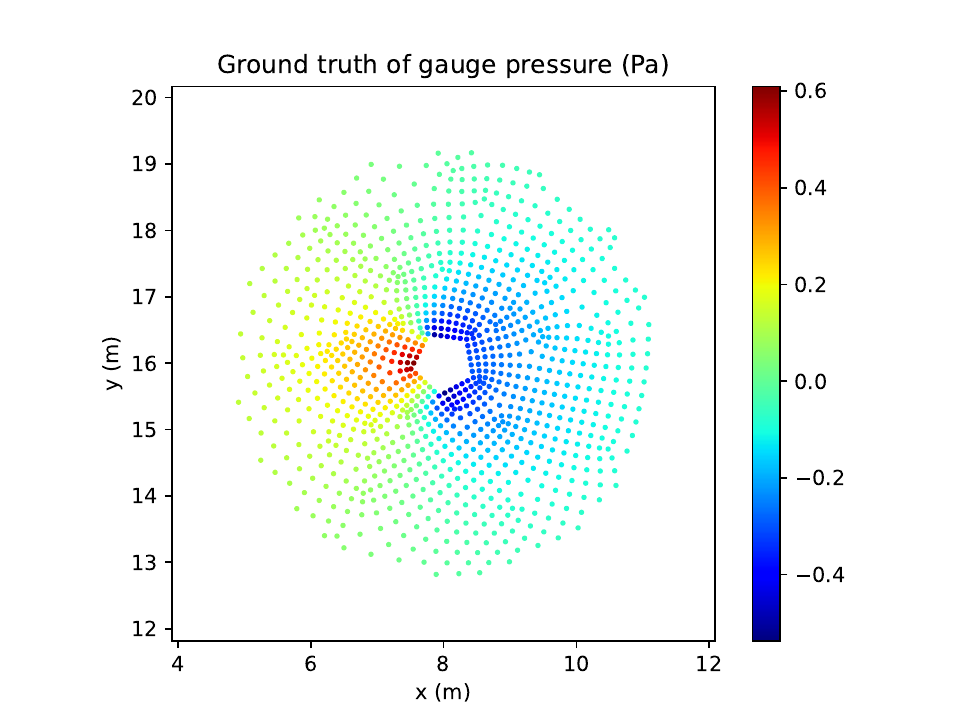}
    \end{subfigure}
    \begin{subfigure}[b]{0.27\textwidth}
        \centering
        \includegraphics[width=\textwidth]{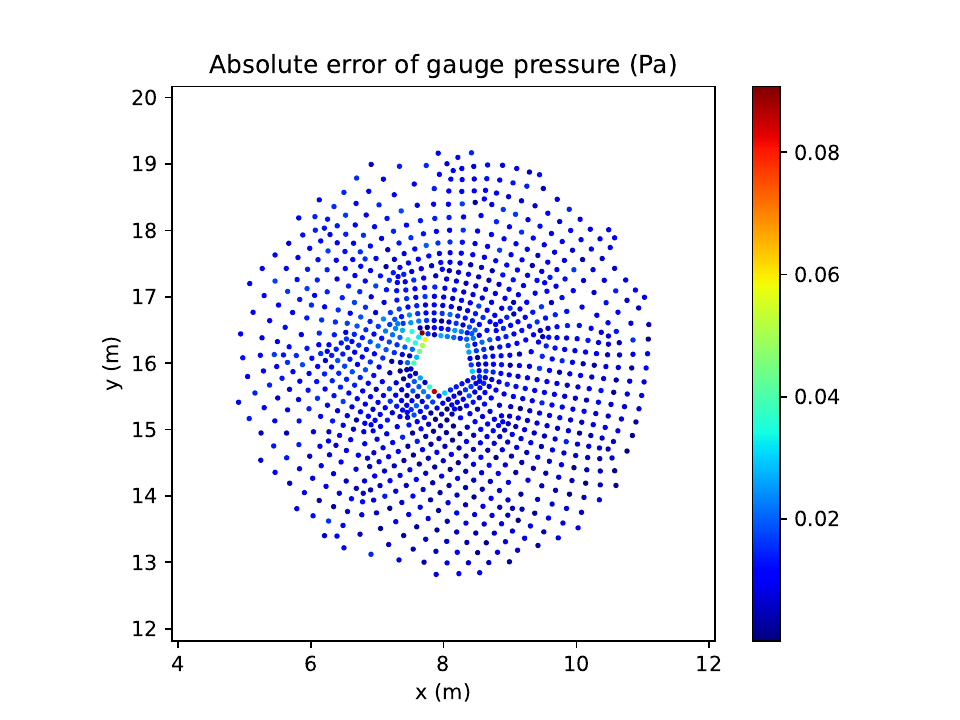}
    \end{subfigure}

  \caption{The first set of examples comparing the ground truth to the predictions of Flow Matching PointNet for the velocity and pressure fields from the test set.}
  \label{Fig9}
\end{figure}


\begin{figure}[!htbp]
  \centering 
      \begin{subfigure}[b]{0.27\textwidth}
        \centering
        \includegraphics[width=\textwidth]{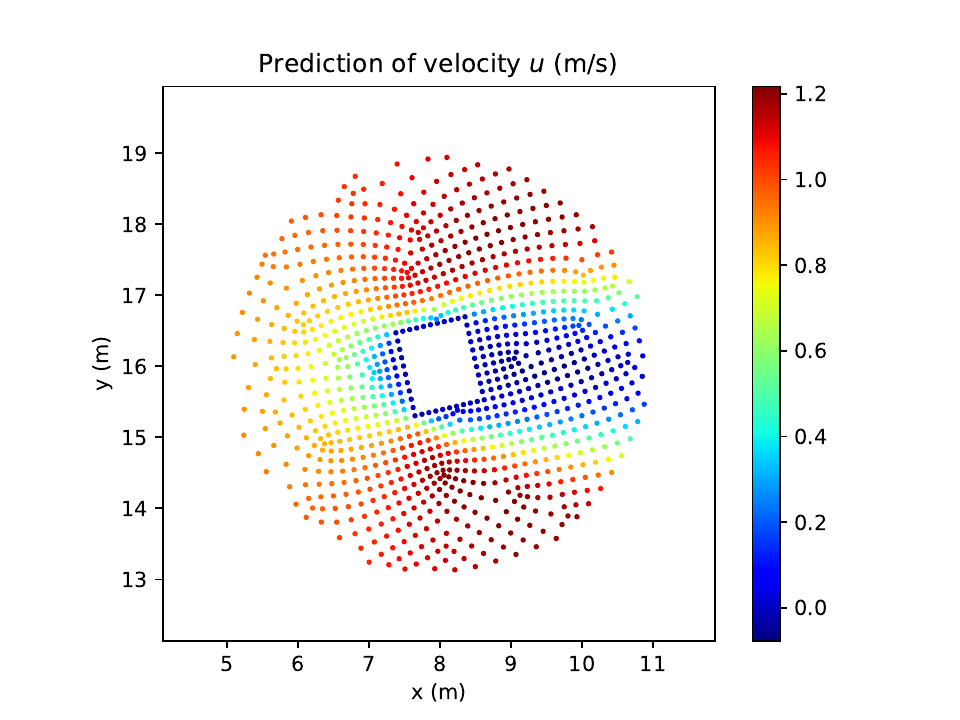}
    \end{subfigure}
    \begin{subfigure}[b]{0.27\textwidth}
        \centering
        \includegraphics[width=\textwidth]{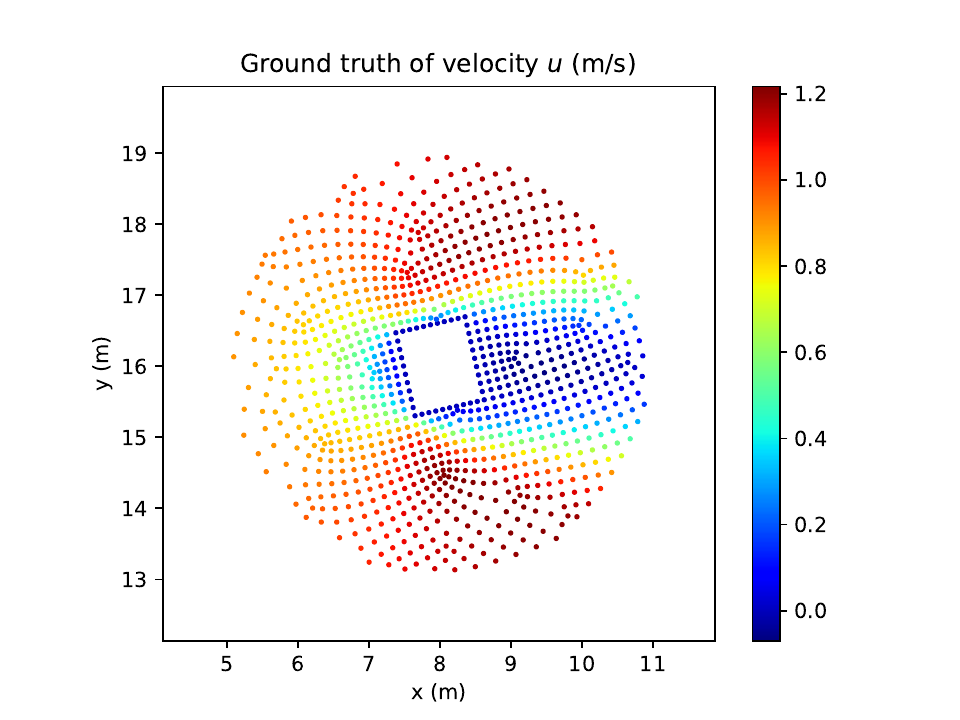}
    \end{subfigure}
    \begin{subfigure}[b]{0.27\textwidth}
        \centering
        \includegraphics[width=\textwidth]{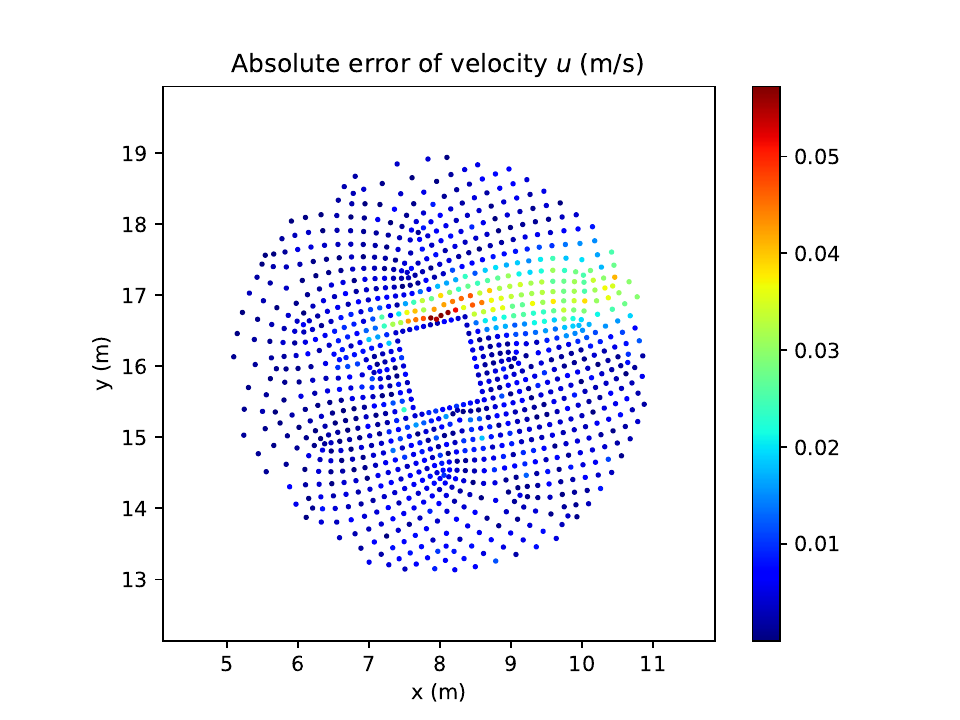}
    \end{subfigure}

    
    \begin{subfigure}[b]{0.27\textwidth}
        \centering
        \includegraphics[width=\textwidth]{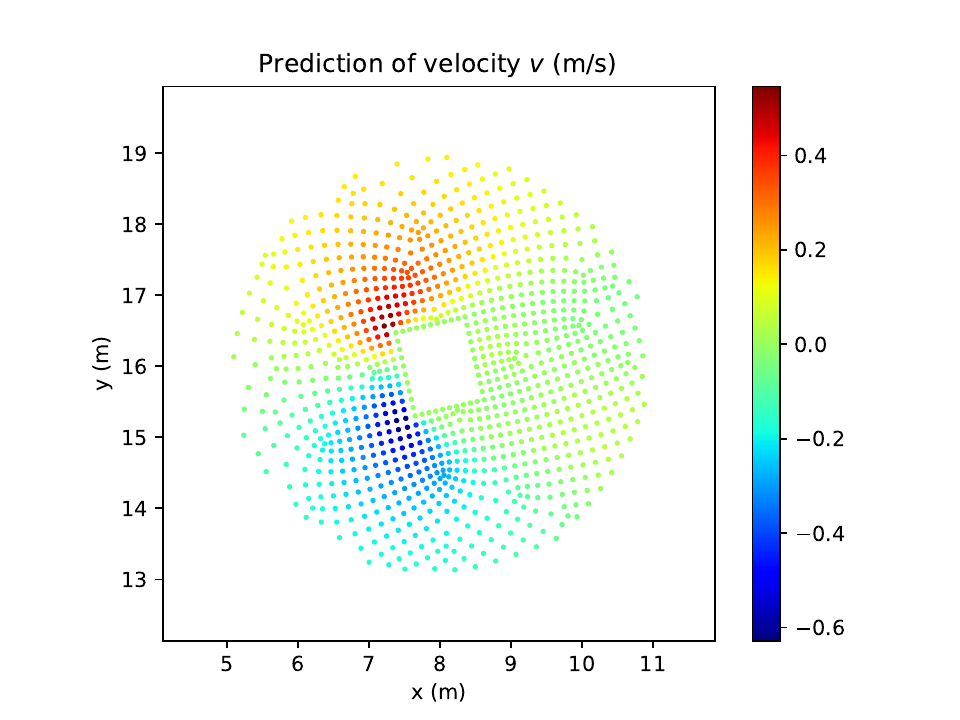}
    \end{subfigure}
    \begin{subfigure}[b]{0.27\textwidth}
        \centering
        \includegraphics[width=\textwidth]{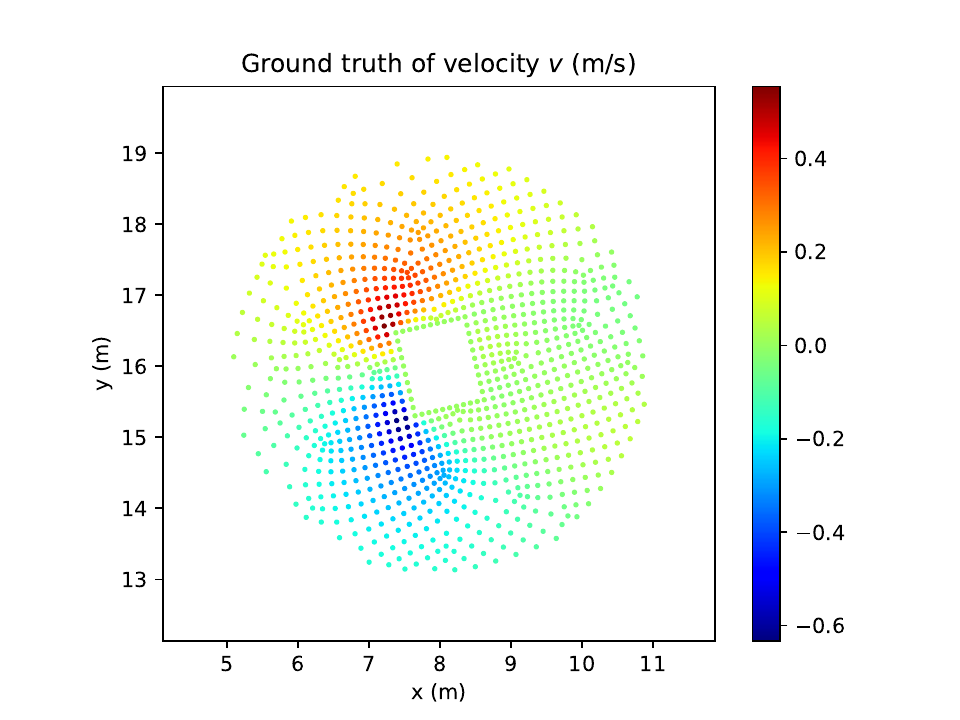}
    \end{subfigure}
    \begin{subfigure}[b]{0.27\textwidth}
        \centering
        \includegraphics[width=\textwidth]{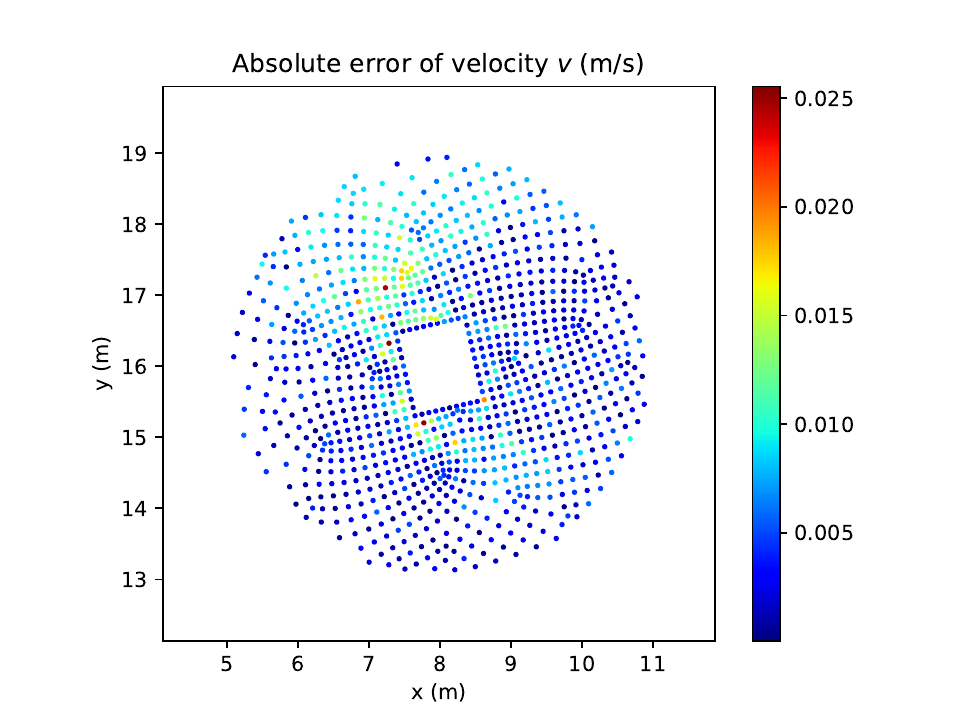}
    \end{subfigure}

    
    \begin{subfigure}[b]{0.27\textwidth}
        \centering
        \includegraphics[width=\textwidth]{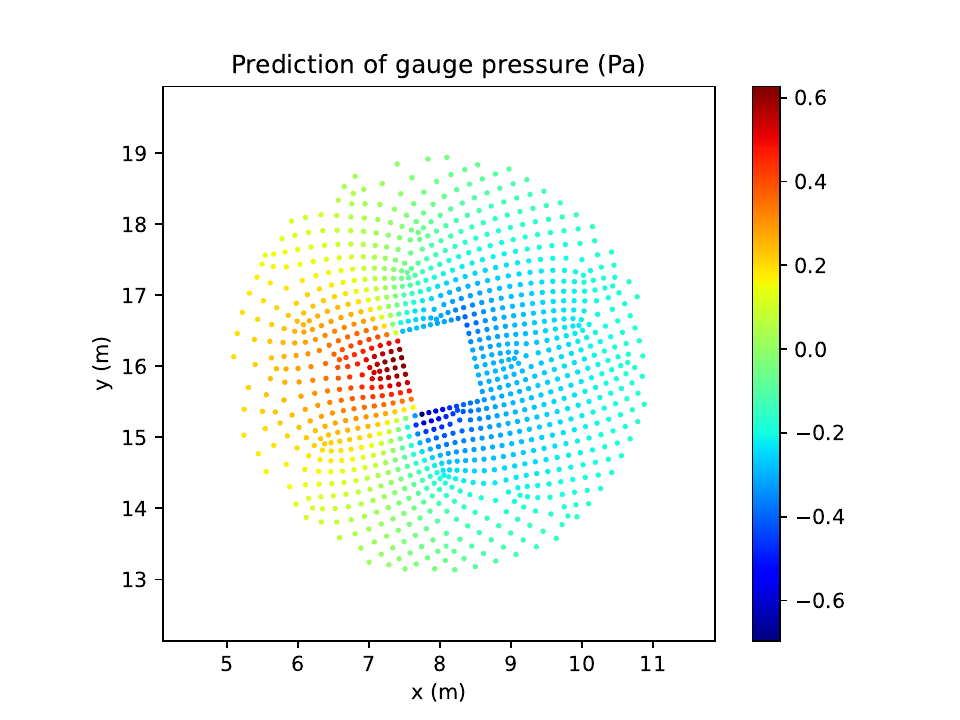}
    \end{subfigure}
    \begin{subfigure}[b]{0.27\textwidth}
        \centering
        \includegraphics[width=\textwidth]{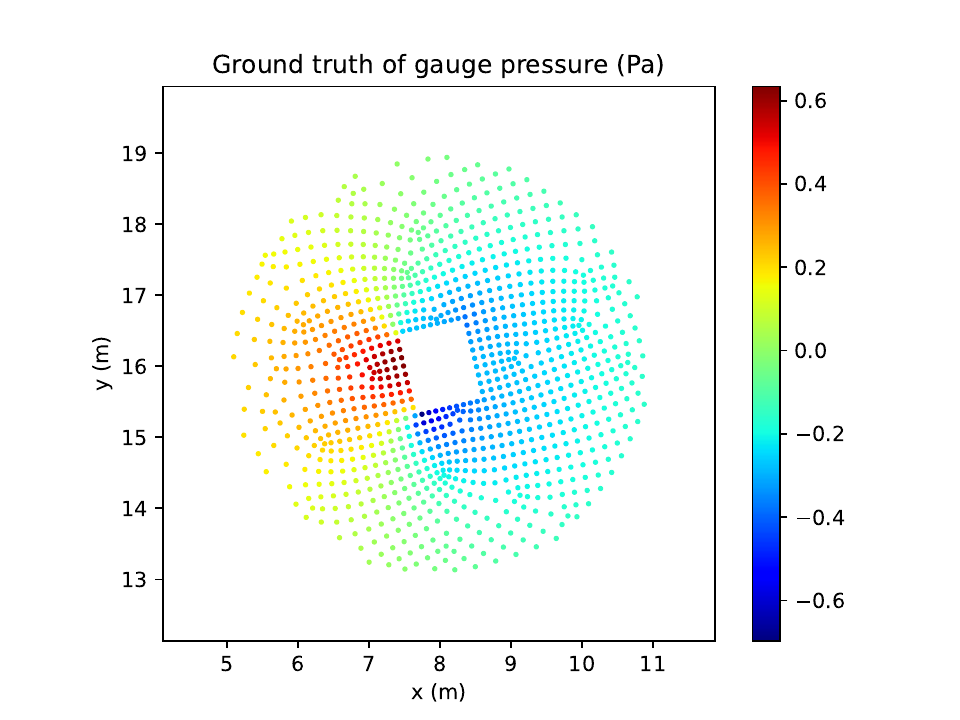}
    \end{subfigure}
    \begin{subfigure}[b]{0.27\textwidth}
        \centering
        \includegraphics[width=\textwidth]{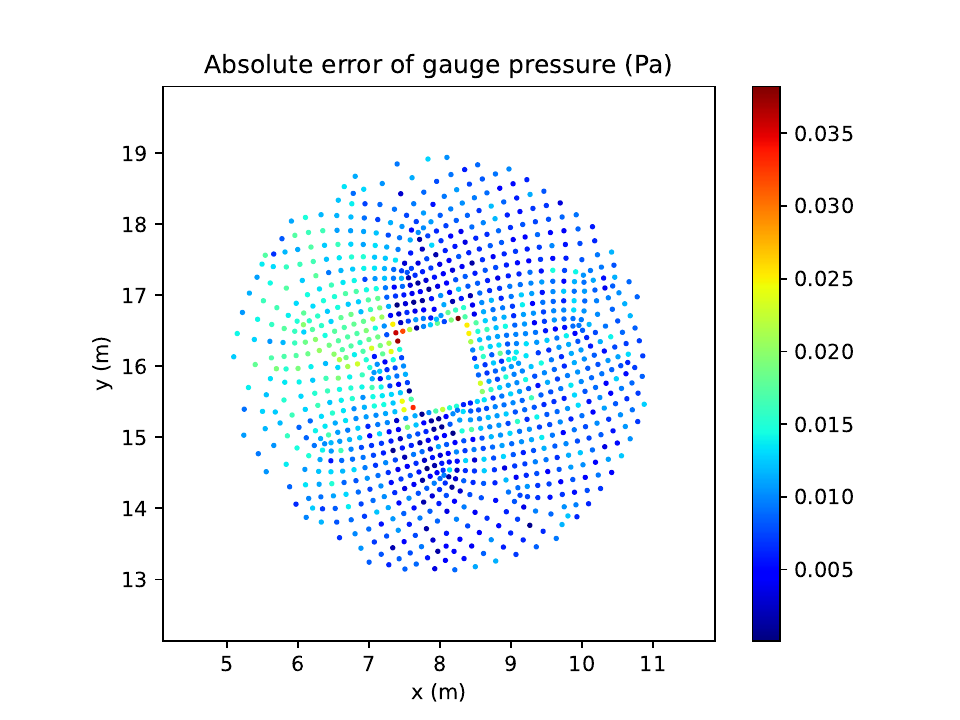}
    \end{subfigure}

  \caption{The second set of examples comparing the ground truth to the predictions of Flow Matching PointNet for the velocity and pressure fields from the test set.}
  \label{Fig10}
\end{figure}


\begin{figure}[!htbp]
  \centering 
      \begin{subfigure}[b]{0.27\textwidth}
        \centering
        \includegraphics[width=\textwidth]{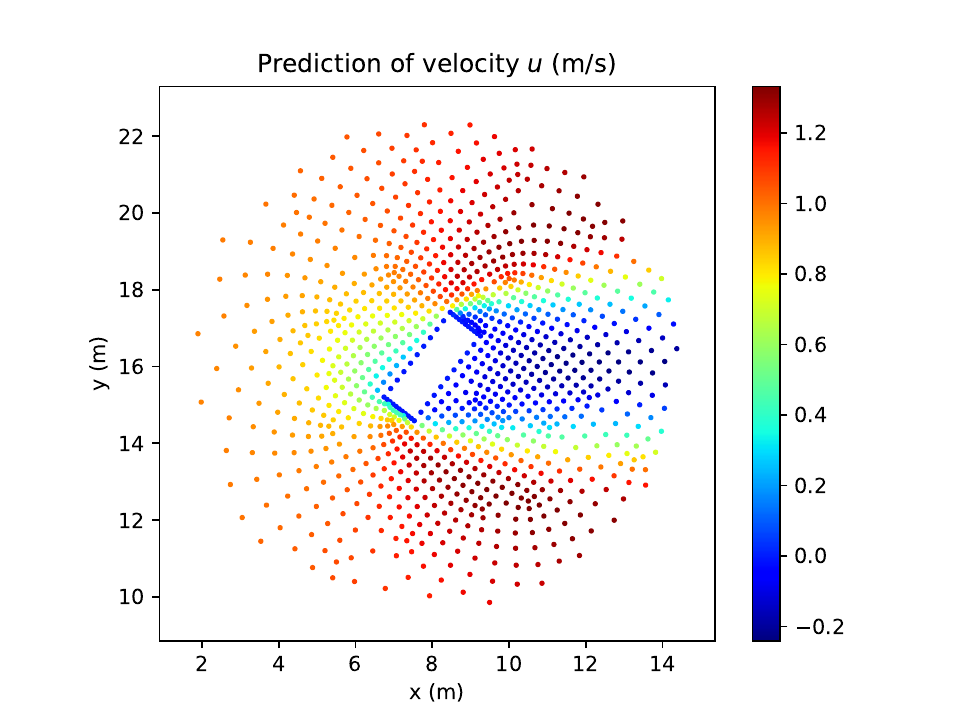}
    \end{subfigure}
    \begin{subfigure}[b]{0.27\textwidth}
        \centering
        \includegraphics[width=\textwidth]{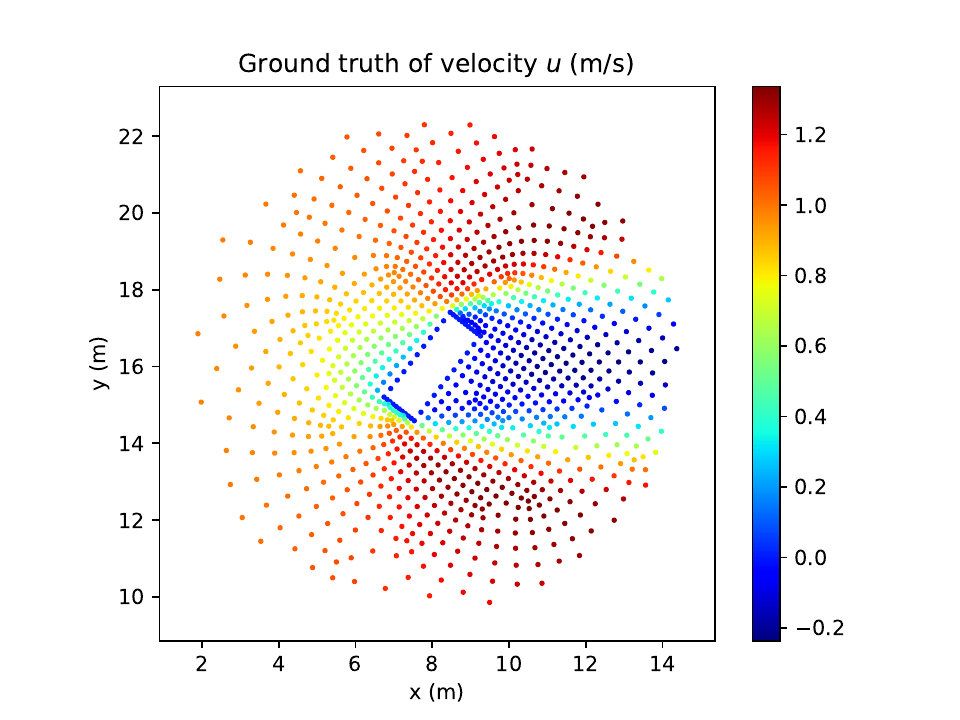}
    \end{subfigure}
    \begin{subfigure}[b]{0.27\textwidth}
        \centering
        \includegraphics[width=\textwidth]{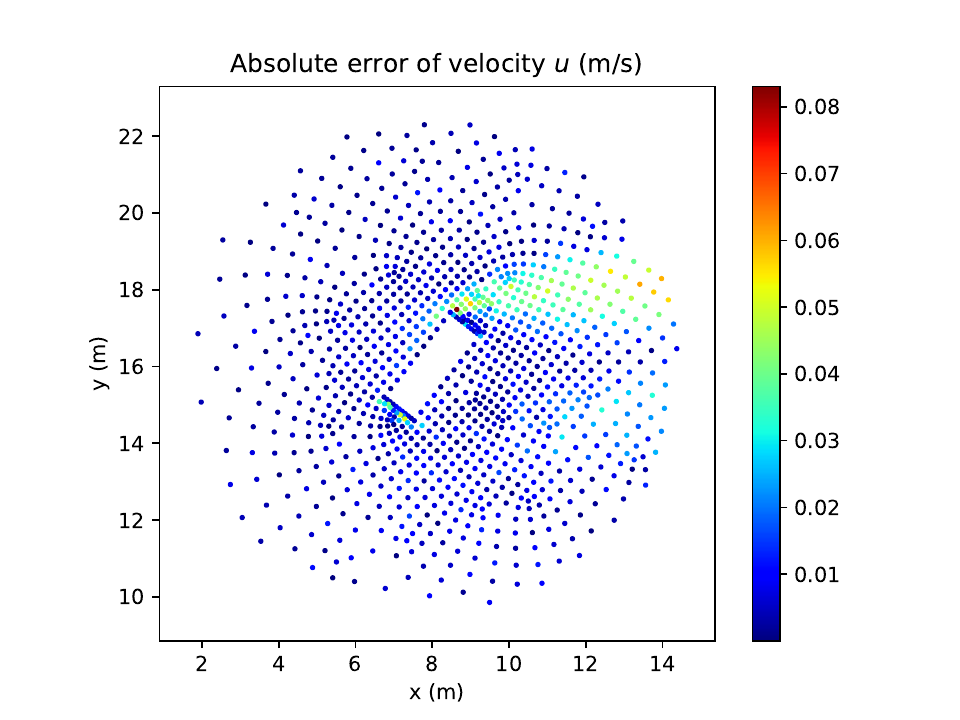}
    \end{subfigure}

    
    \begin{subfigure}[b]{0.27\textwidth}
        \centering
        \includegraphics[width=\textwidth]{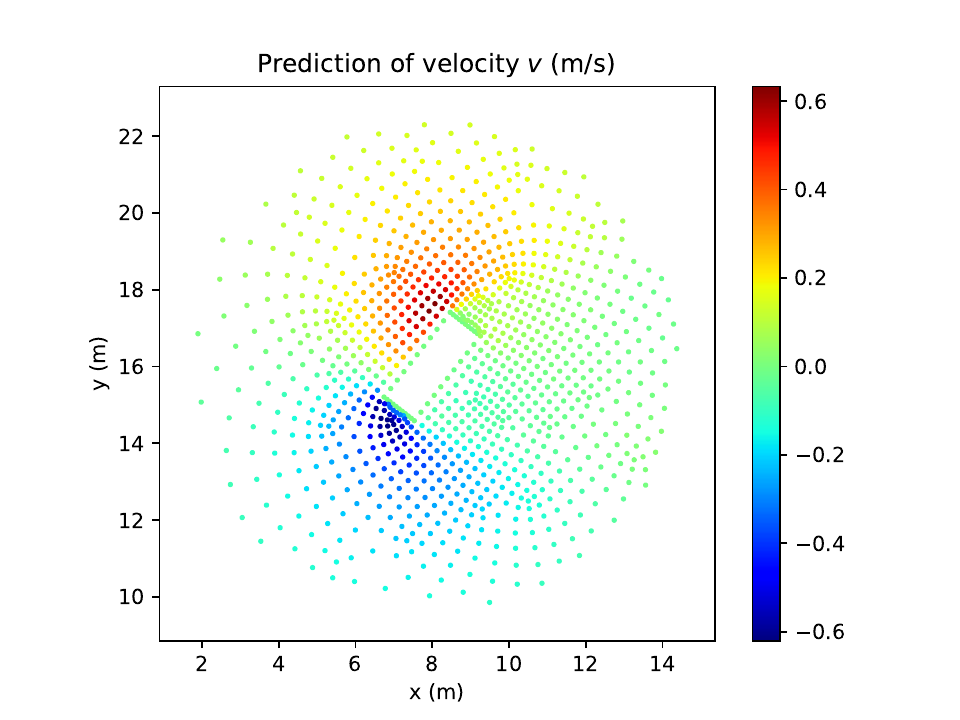}
    \end{subfigure}
    \begin{subfigure}[b]{0.27\textwidth}
        \centering
        \includegraphics[width=\textwidth]{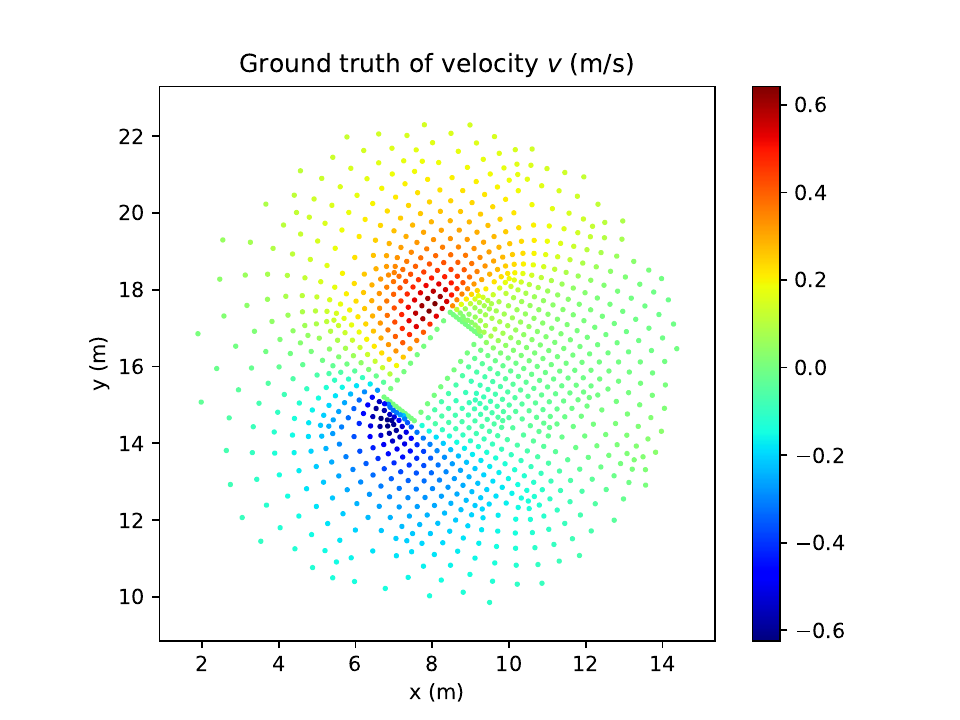}
    \end{subfigure}
    \begin{subfigure}[b]{0.27\textwidth}
        \centering
        \includegraphics[width=\textwidth]{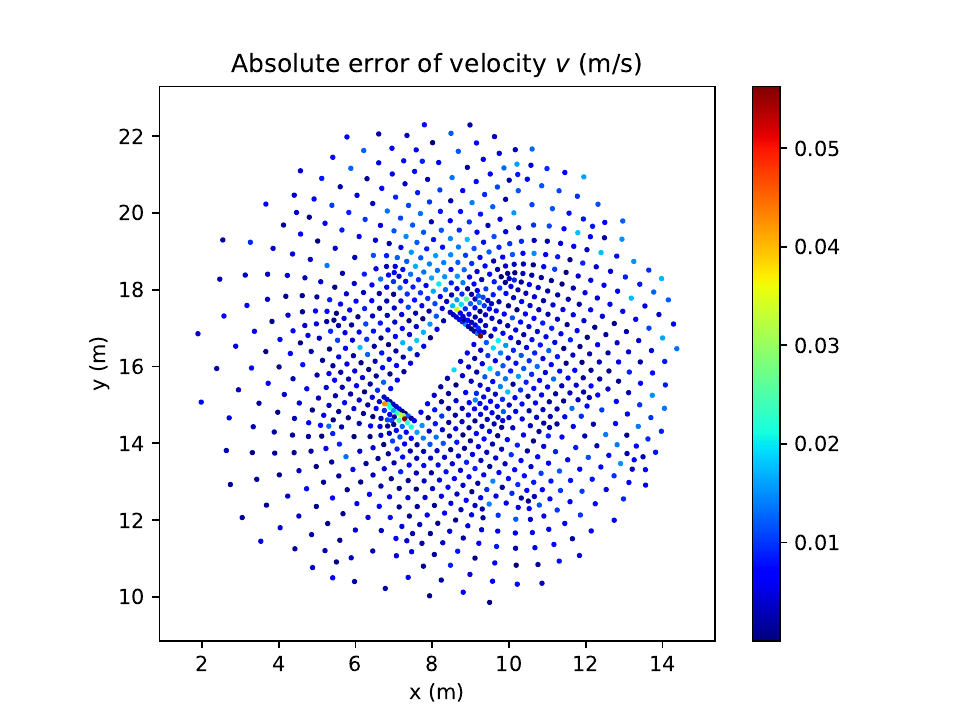}
    \end{subfigure}

    
    \begin{subfigure}[b]{0.27\textwidth}
        \centering
        \includegraphics[width=\textwidth]{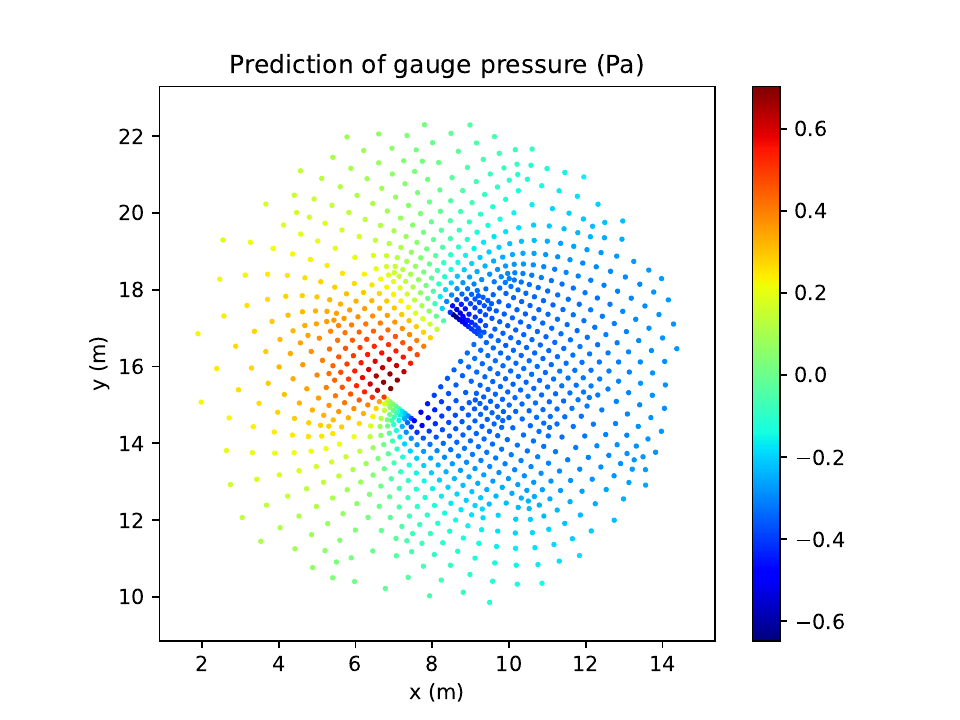}
    \end{subfigure}
    \begin{subfigure}[b]{0.27\textwidth}
        \centering
        \includegraphics[width=\textwidth]{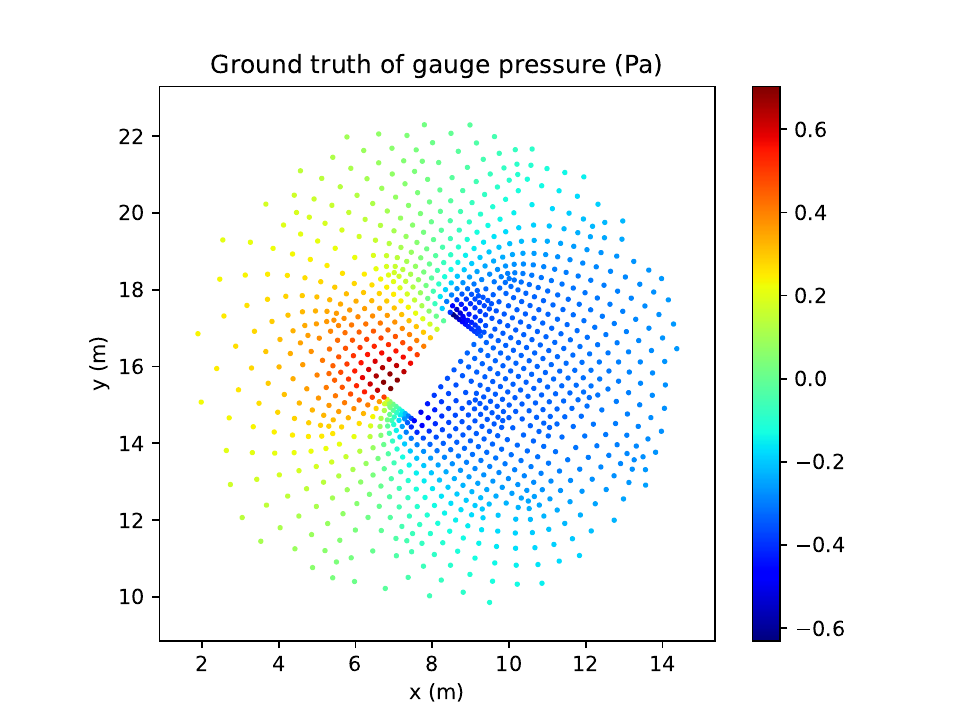}
    \end{subfigure}
    \begin{subfigure}[b]{0.27\textwidth}
        \centering
        \includegraphics[width=\textwidth]{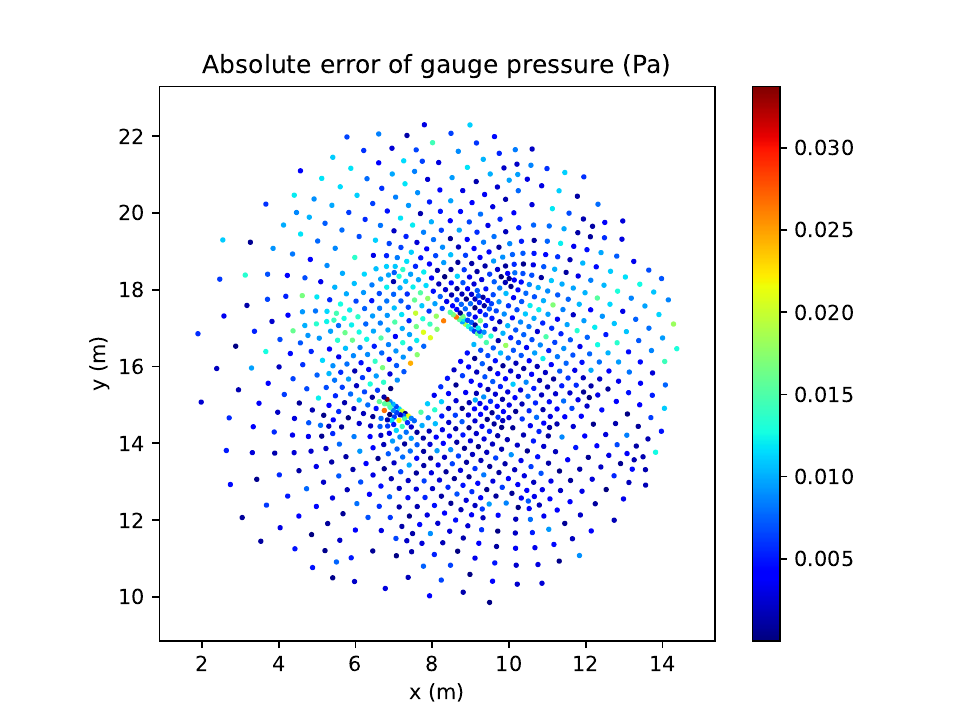}
    \end{subfigure}

  \caption{The third set of examples comparing the ground truth to the predictions of Flow Matching PointNet for the velocity and pressure fields from the test set.}
  \label{Fig11}
\end{figure}


\begin{figure}[!htbp]
  \centering 
      \begin{subfigure}[b]{0.27\textwidth}
        \centering
        \includegraphics[width=\textwidth]{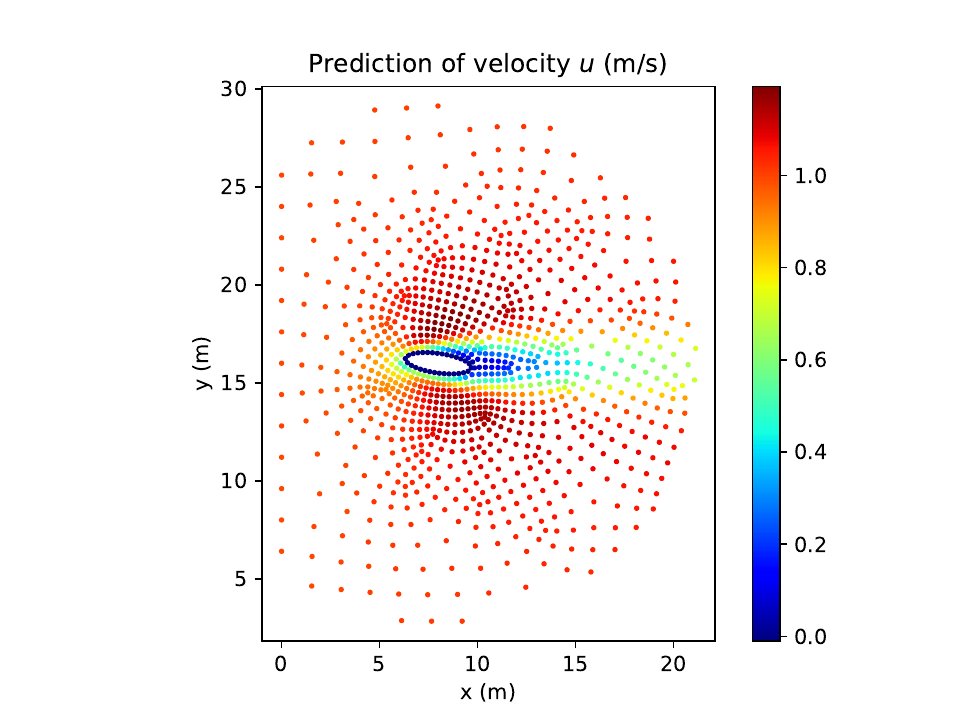}
    \end{subfigure}
    \begin{subfigure}[b]{0.27\textwidth}
        \centering
        \includegraphics[width=\textwidth]{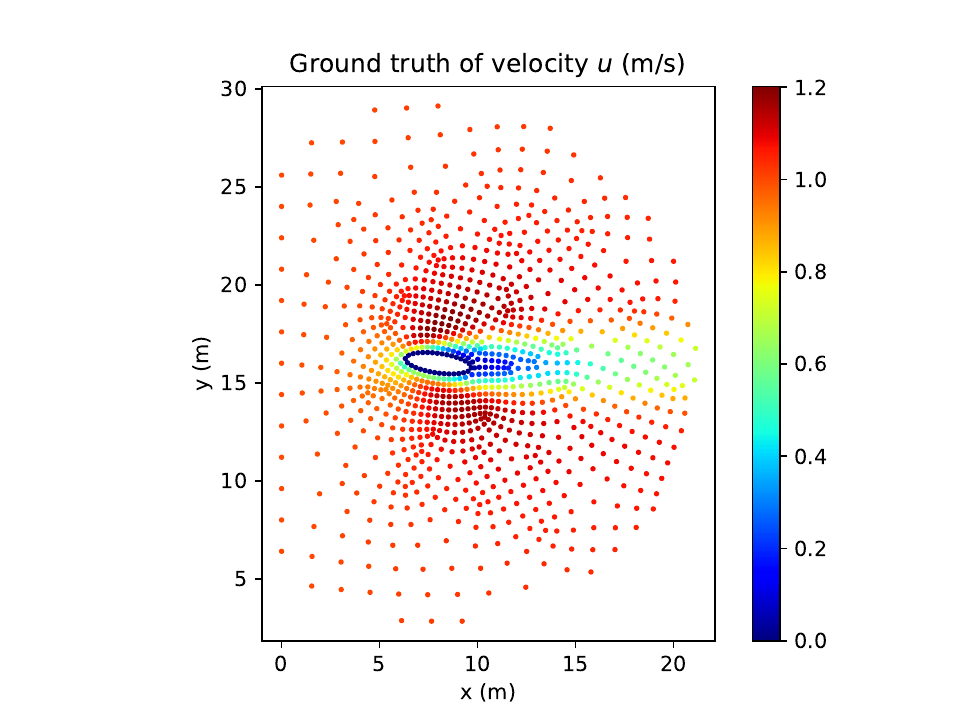}
    \end{subfigure}
    \begin{subfigure}[b]{0.27\textwidth}
        \centering
        \includegraphics[width=\textwidth]{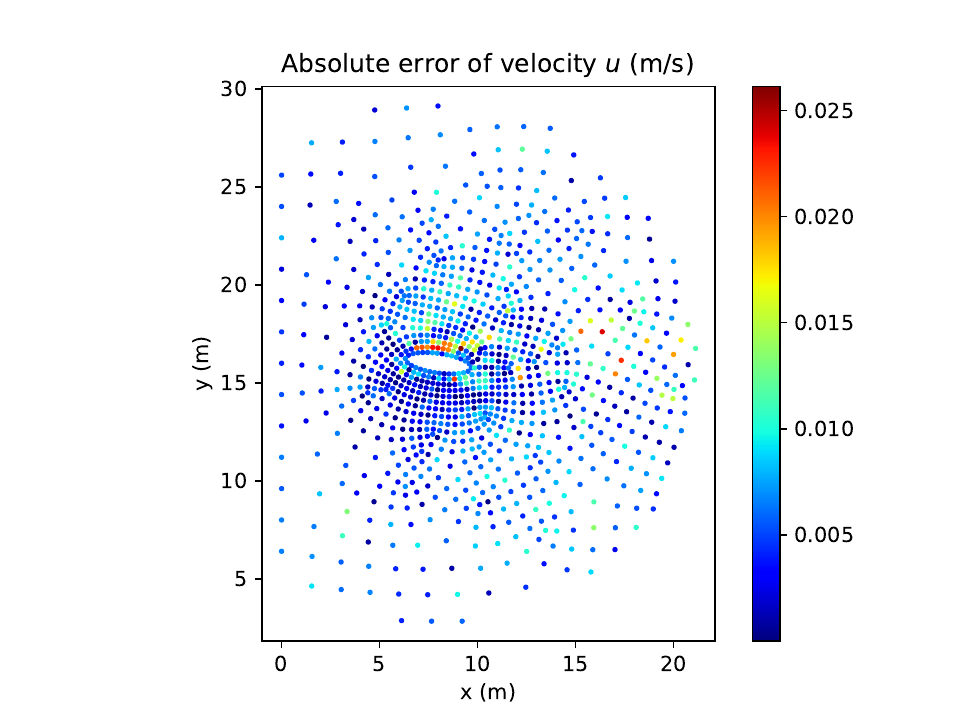}
    \end{subfigure}

    
    \begin{subfigure}[b]{0.27\textwidth}
        \centering
        \includegraphics[width=\textwidth]{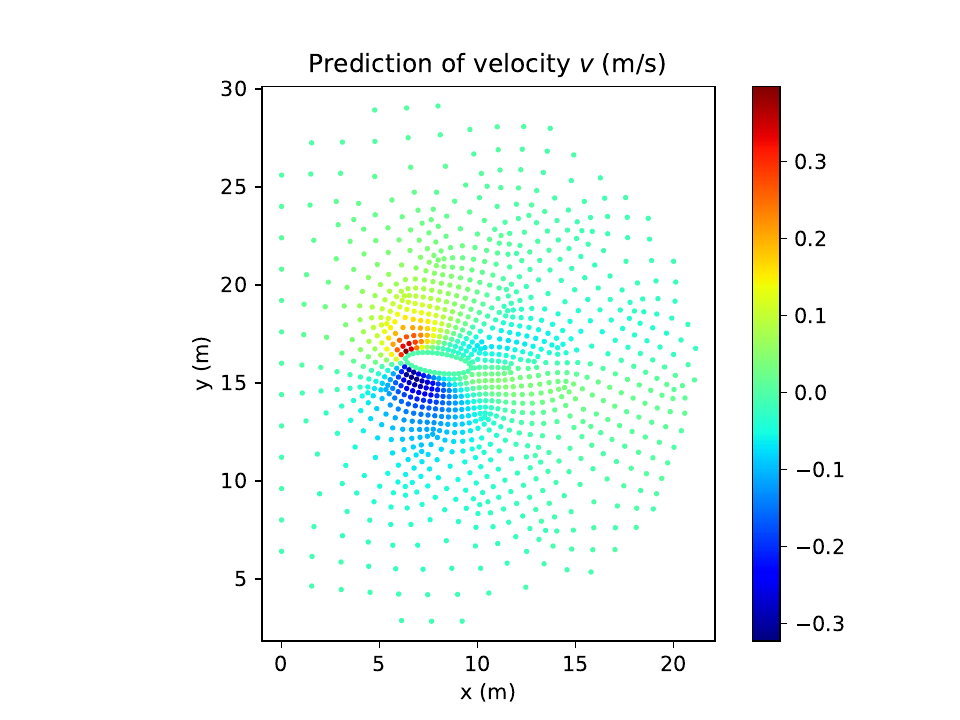}
    \end{subfigure}
    \begin{subfigure}[b]{0.27\textwidth}
        \centering
        \includegraphics[width=\textwidth]{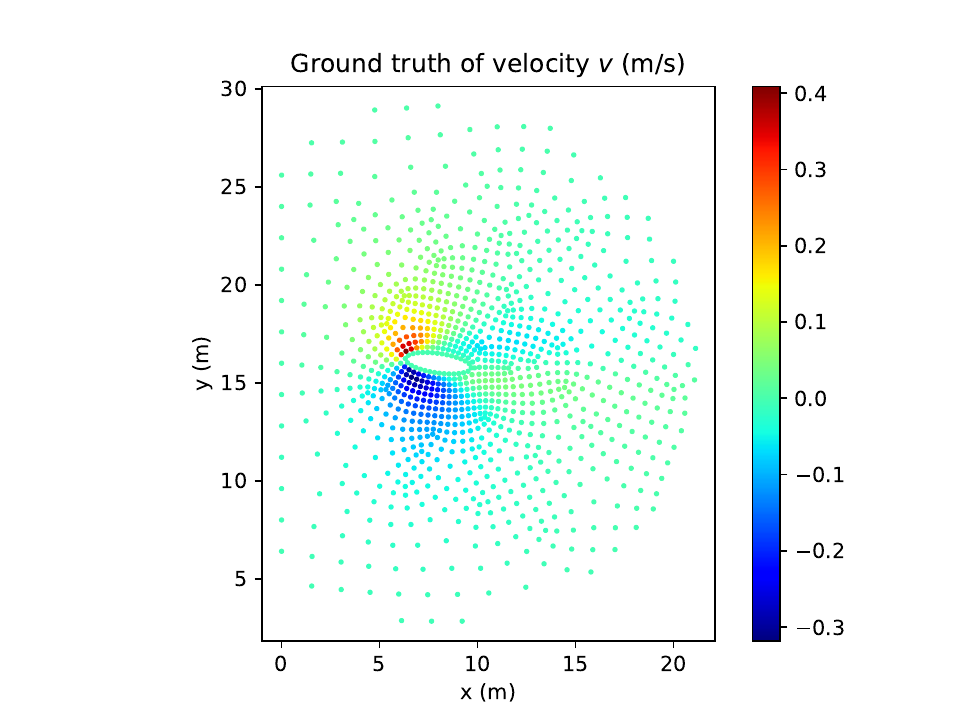}
    \end{subfigure}
    \begin{subfigure}[b]{0.27\textwidth}
        \centering
        \includegraphics[width=\textwidth]{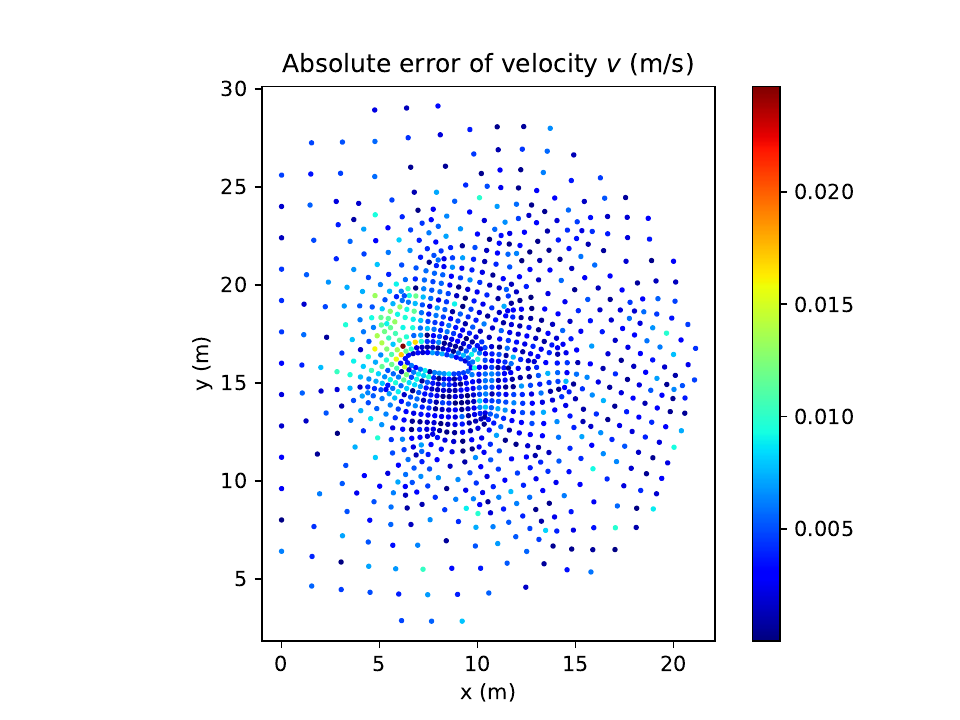}
    \end{subfigure}

    
    \begin{subfigure}[b]{0.27\textwidth}
        \centering
        \includegraphics[width=\textwidth]{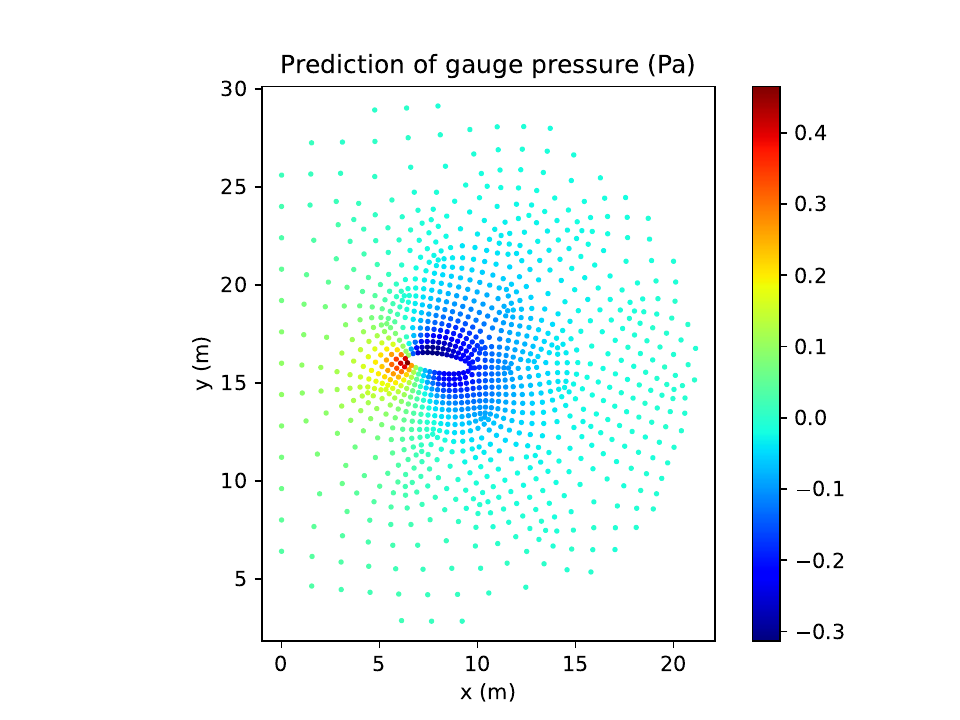}
    \end{subfigure}
    \begin{subfigure}[b]{0.27\textwidth}
        \centering
        \includegraphics[width=\textwidth]{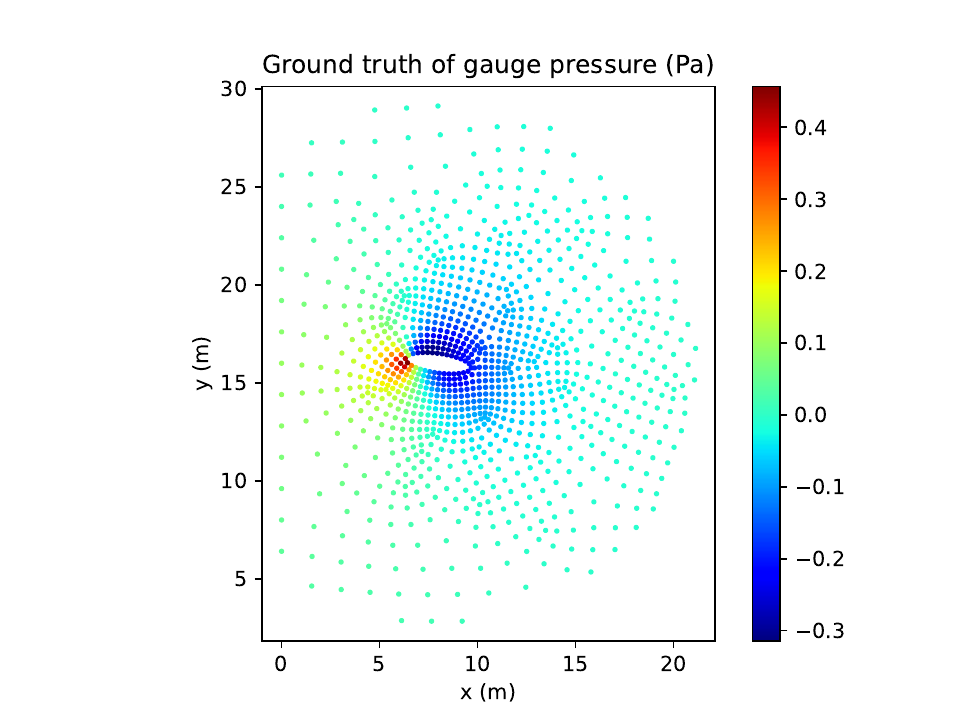}
    \end{subfigure}
    \begin{subfigure}[b]{0.27\textwidth}
        \centering
        \includegraphics[width=\textwidth]{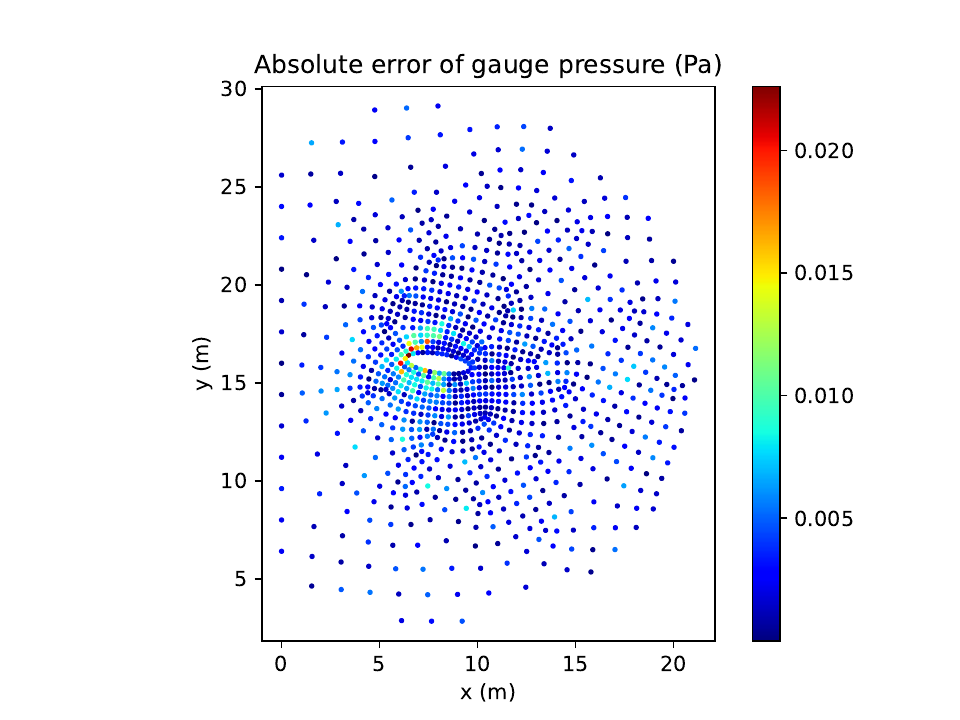}
    \end{subfigure}

  \caption{The fourth set of examples comparing the ground truth to the predictions of Flow Matching PointNet for the velocity and pressure fields from the test set.}
  \label{Fig12}
\end{figure}


\subsection{Visual evaluation}
\label{Sect4CVisual}

\begin{figure}[!htbp]
  \centering 
      \begin{subfigure}[b]{0.27\textwidth}
        \centering
        \includegraphics[width=\textwidth]{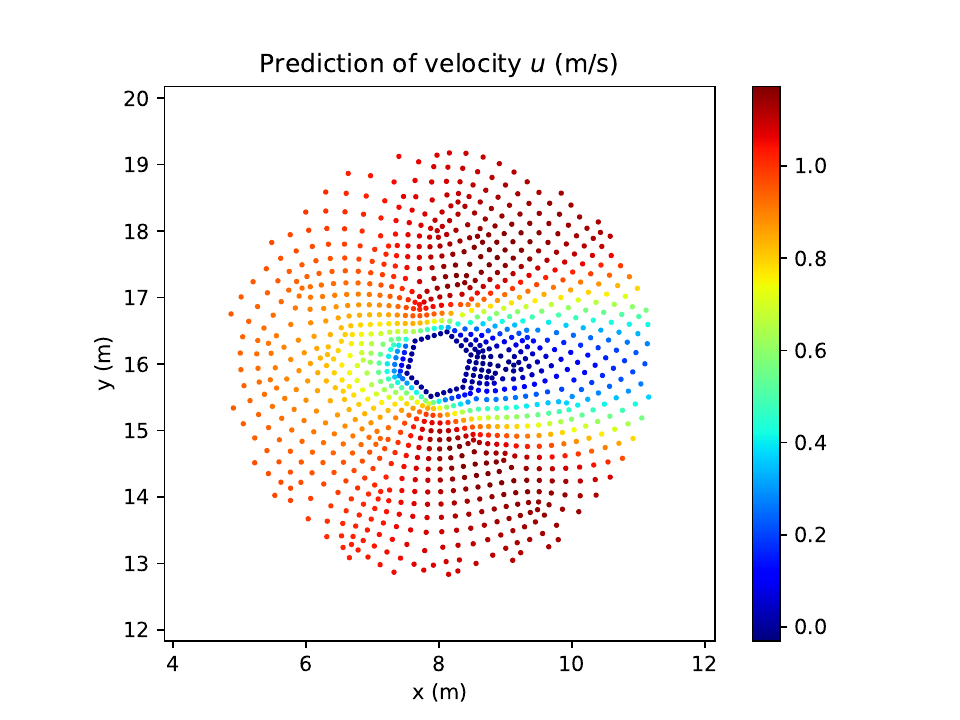}
    \end{subfigure}
    \begin{subfigure}[b]{0.27\textwidth}
        \centering
        \includegraphics[width=\textwidth]{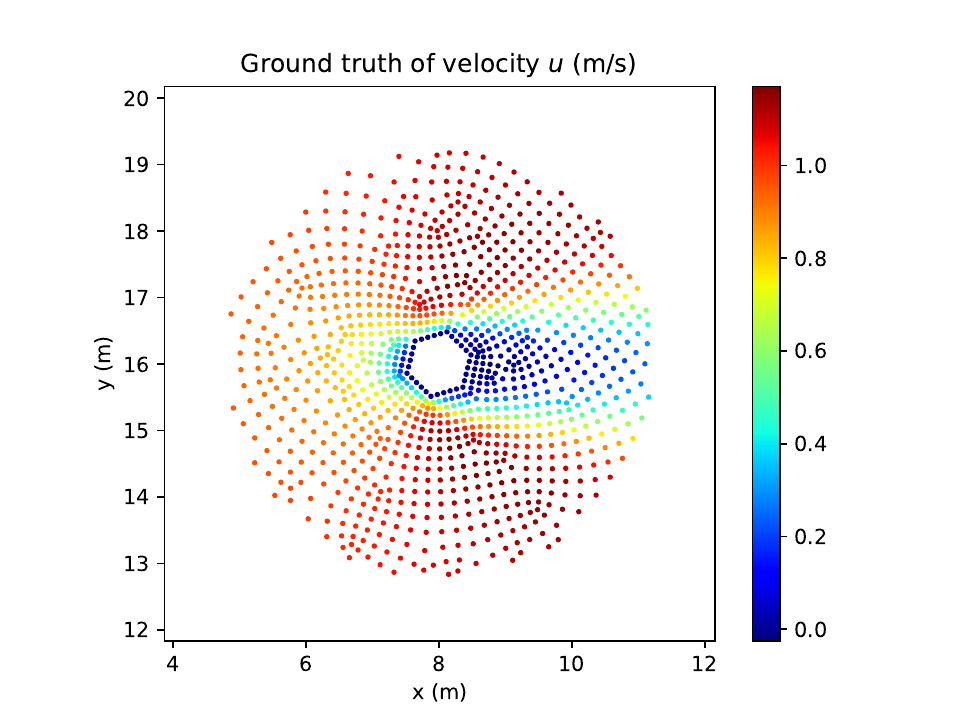}
    \end{subfigure}
    \begin{subfigure}[b]{0.27\textwidth}
        \centering
        \includegraphics[width=\textwidth]{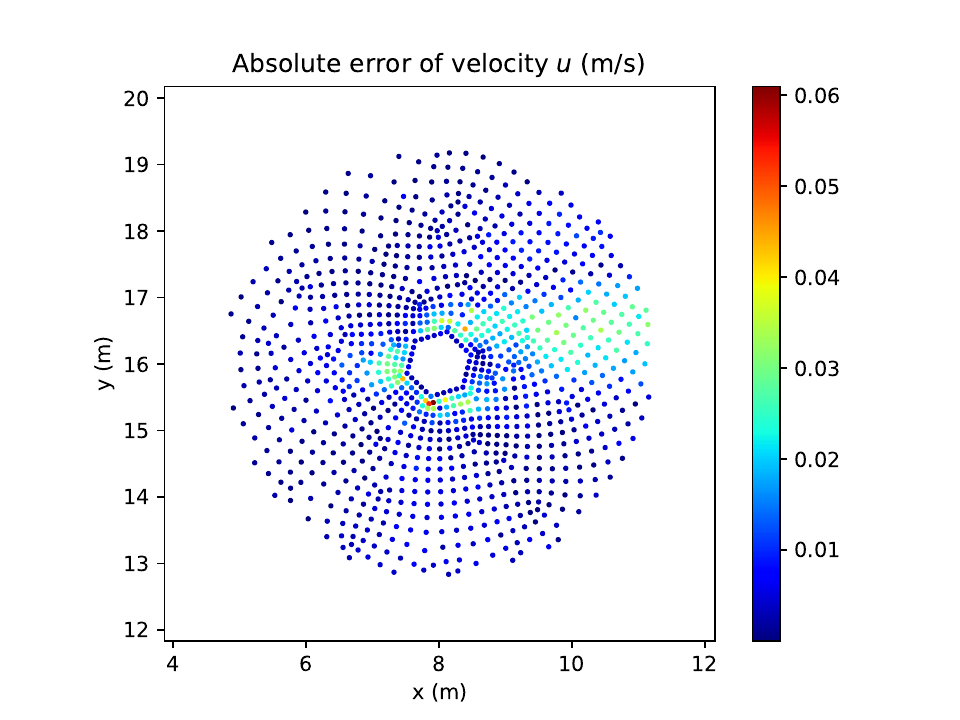}
    \end{subfigure}

    
    \begin{subfigure}[b]{0.27\textwidth}
        \centering
        \includegraphics[width=\textwidth]{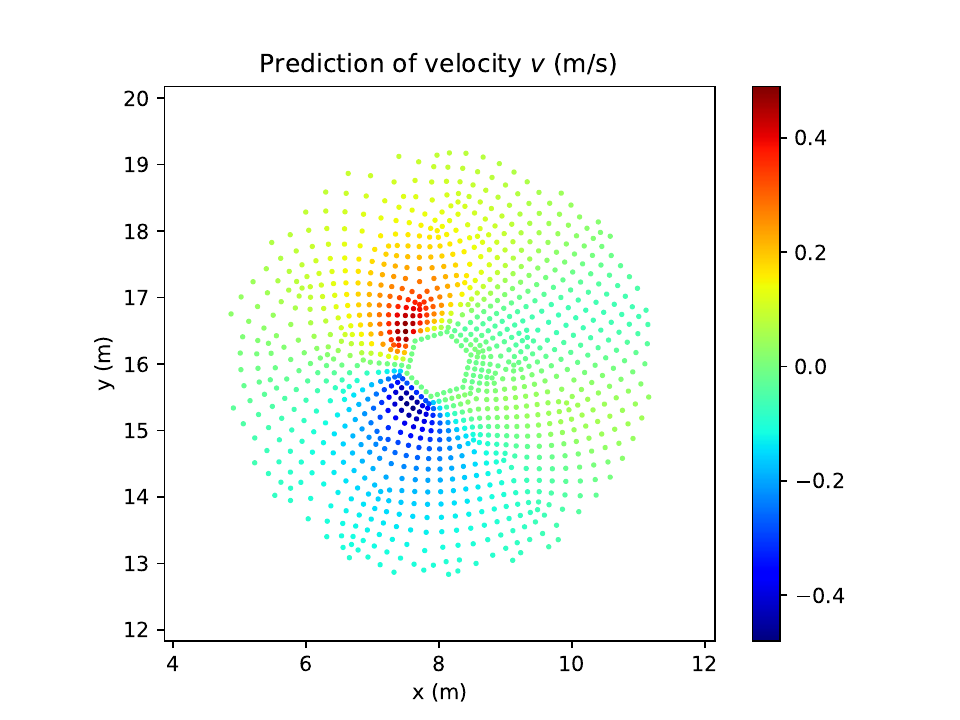}
    \end{subfigure}
    \begin{subfigure}[b]{0.27\textwidth}
        \centering
        \includegraphics[width=\textwidth]{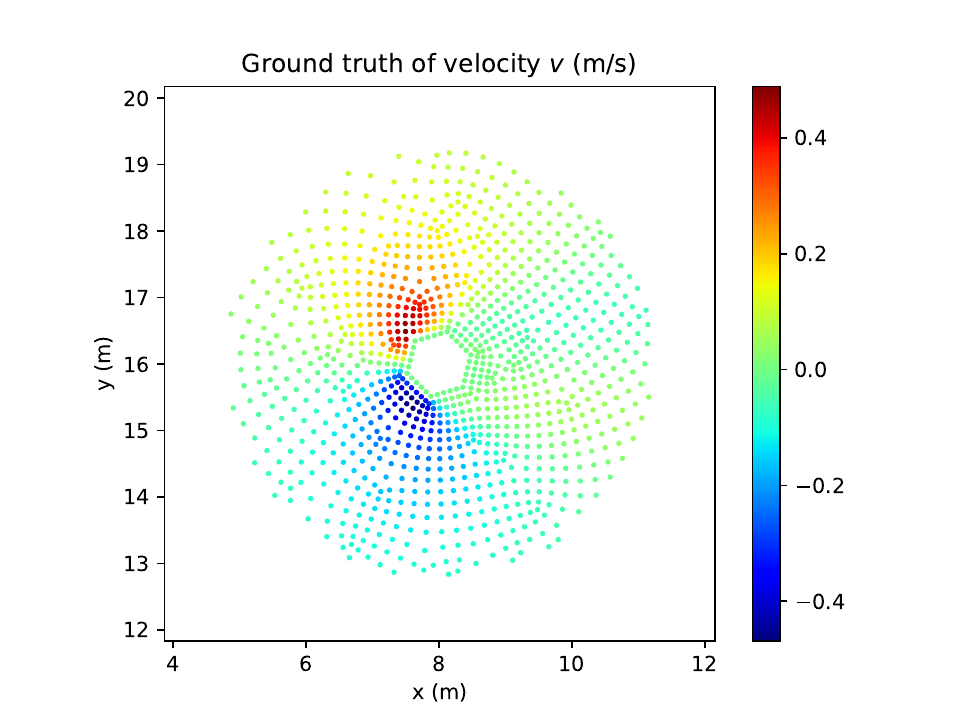}
    \end{subfigure}
    \begin{subfigure}[b]{0.27\textwidth}
        \centering
        \includegraphics[width=\textwidth]{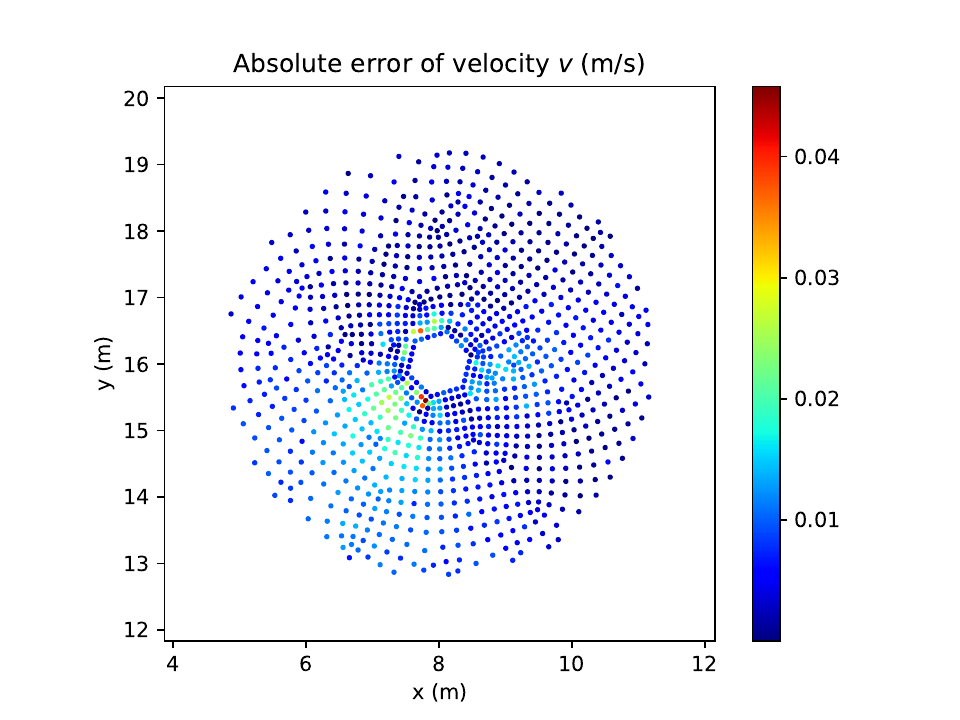}
    \end{subfigure}

    
    \begin{subfigure}[b]{0.27\textwidth}
        \centering
        \includegraphics[width=\textwidth]{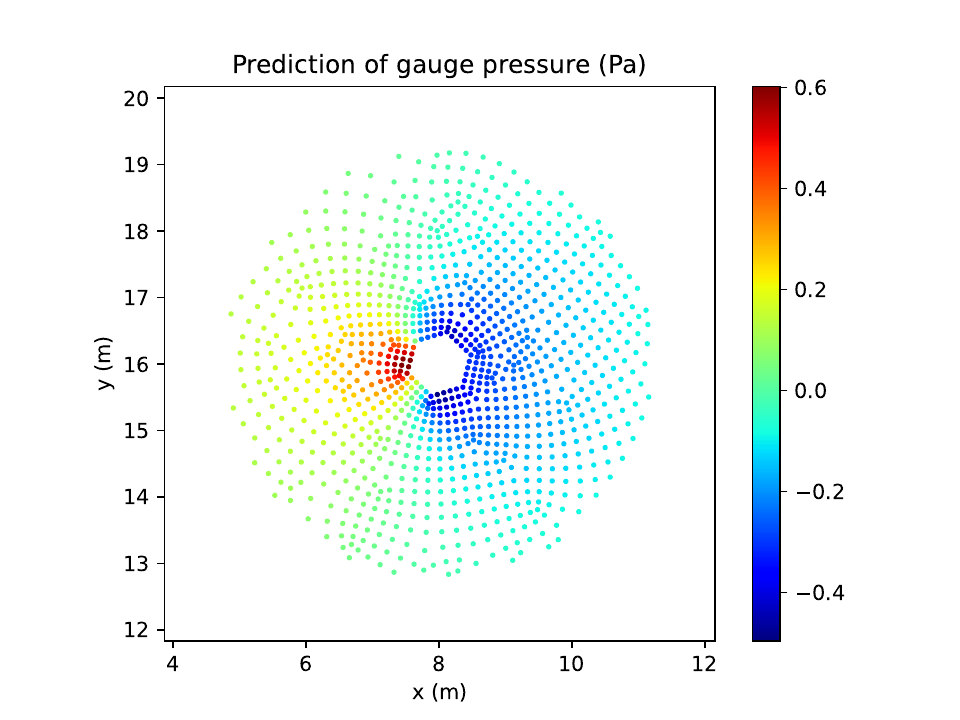}
    \end{subfigure}
    \begin{subfigure}[b]{0.27\textwidth}
        \centering
        \includegraphics[width=\textwidth]{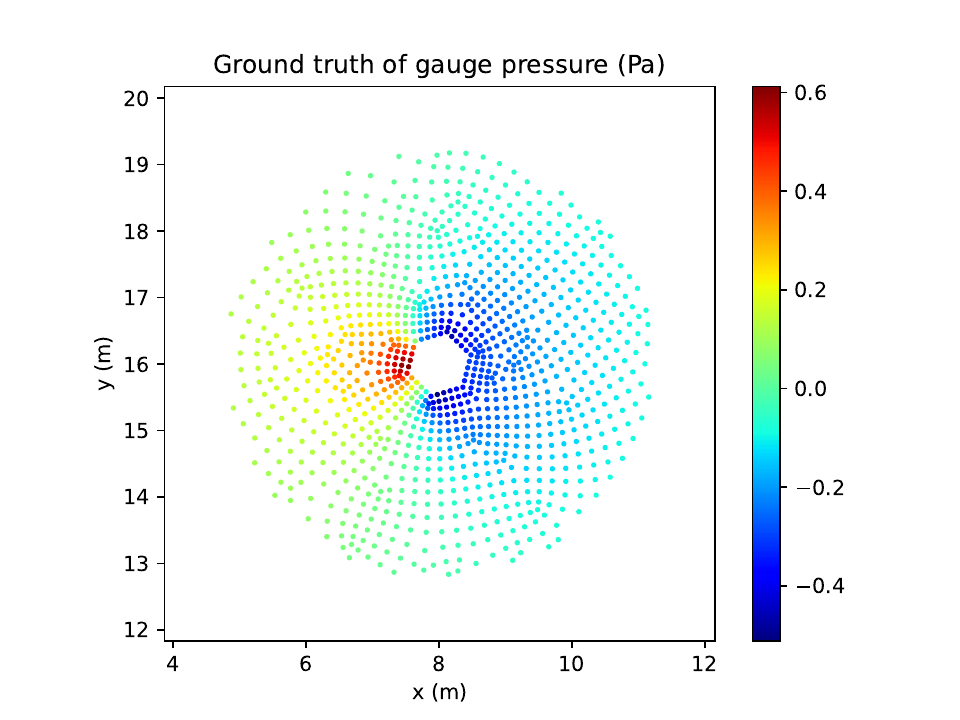}
    \end{subfigure}
    \begin{subfigure}[b]{0.27\textwidth}
        \centering
        \includegraphics[width=\textwidth]{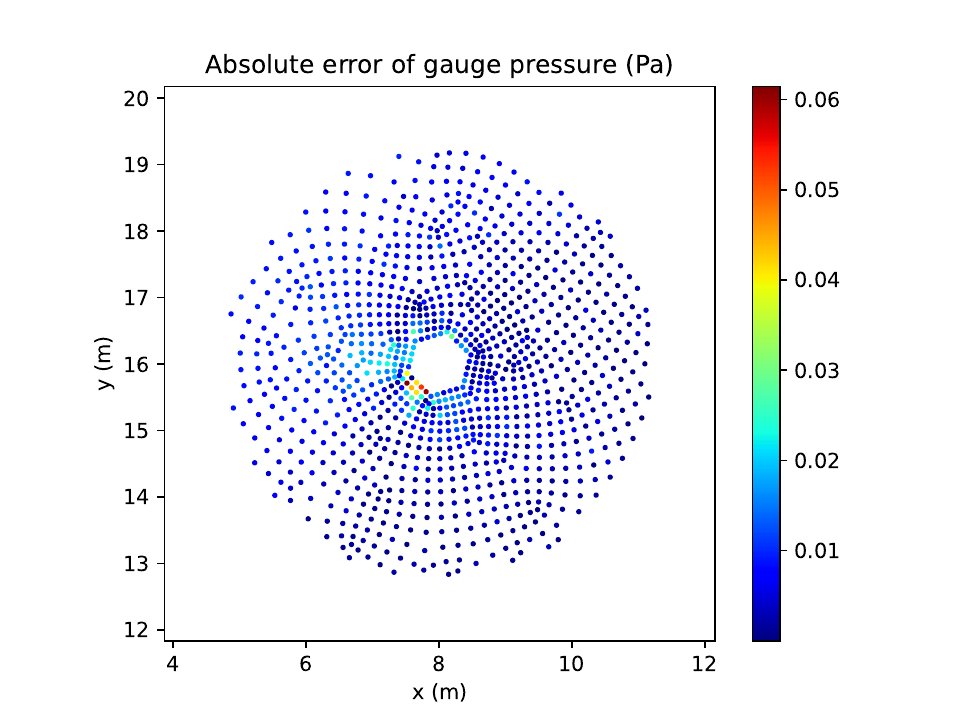}
    \end{subfigure}

  \caption{The first set of examples comparing the ground truth to the predictions of Diffusion PointNet for the velocity and pressure fields from the test set.}
  \label{Fig5}
\end{figure}

\begin{figure}[!htbp]
  \centering 
      \begin{subfigure}[b]{0.27\textwidth}
        \centering
        \includegraphics[width=\textwidth]{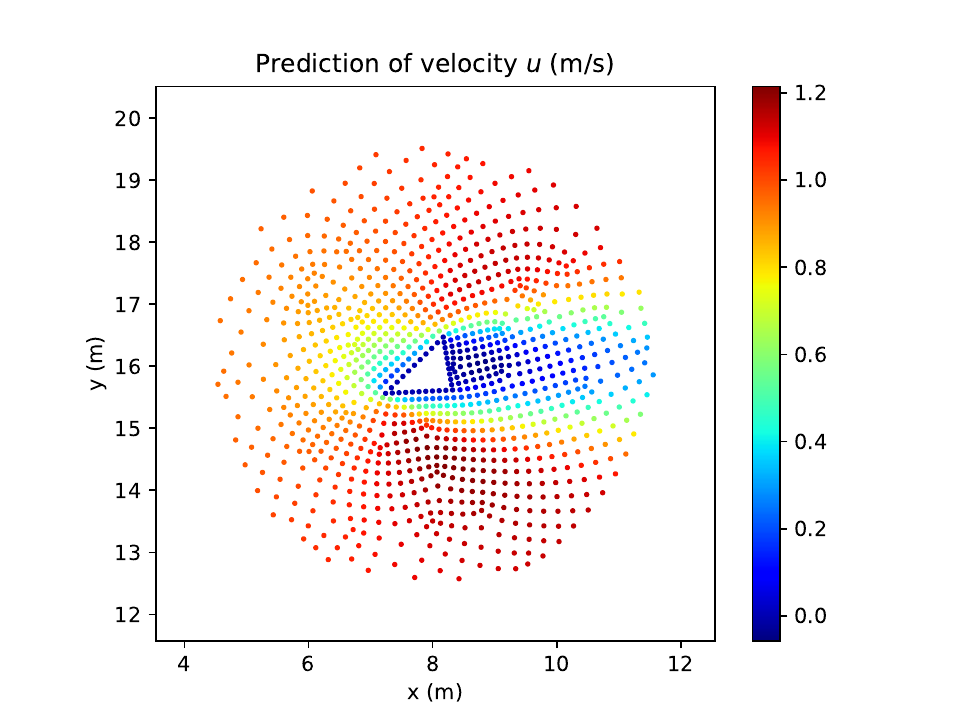}
    \end{subfigure}
    \begin{subfigure}[b]{0.27\textwidth}
        \centering
        \includegraphics[width=\textwidth]{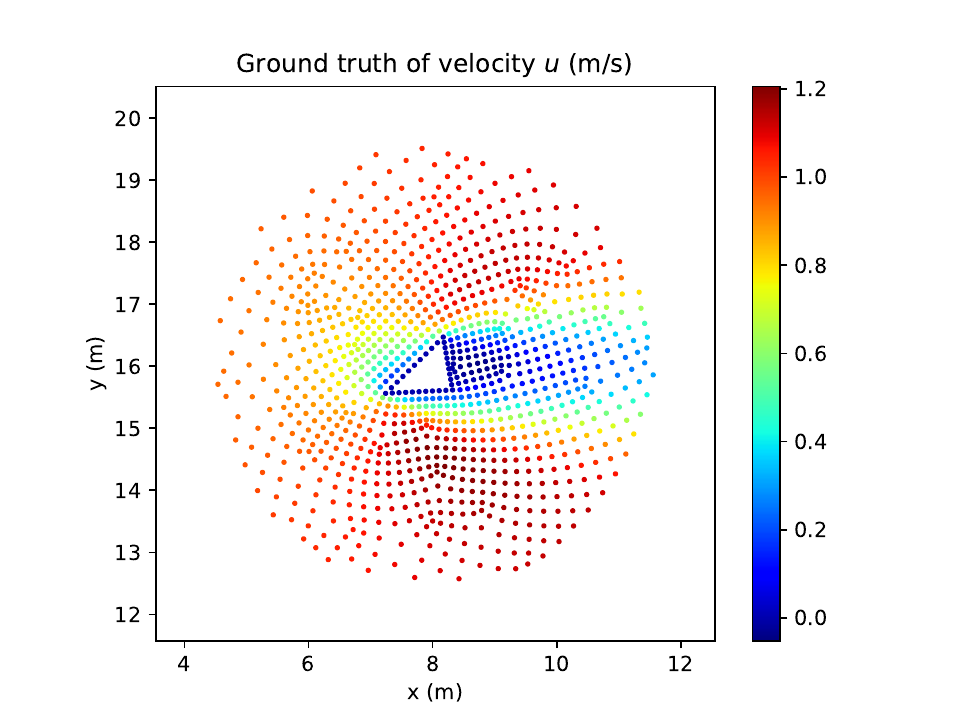}
    \end{subfigure}
    \begin{subfigure}[b]{0.27\textwidth}
        \centering
        \includegraphics[width=\textwidth]{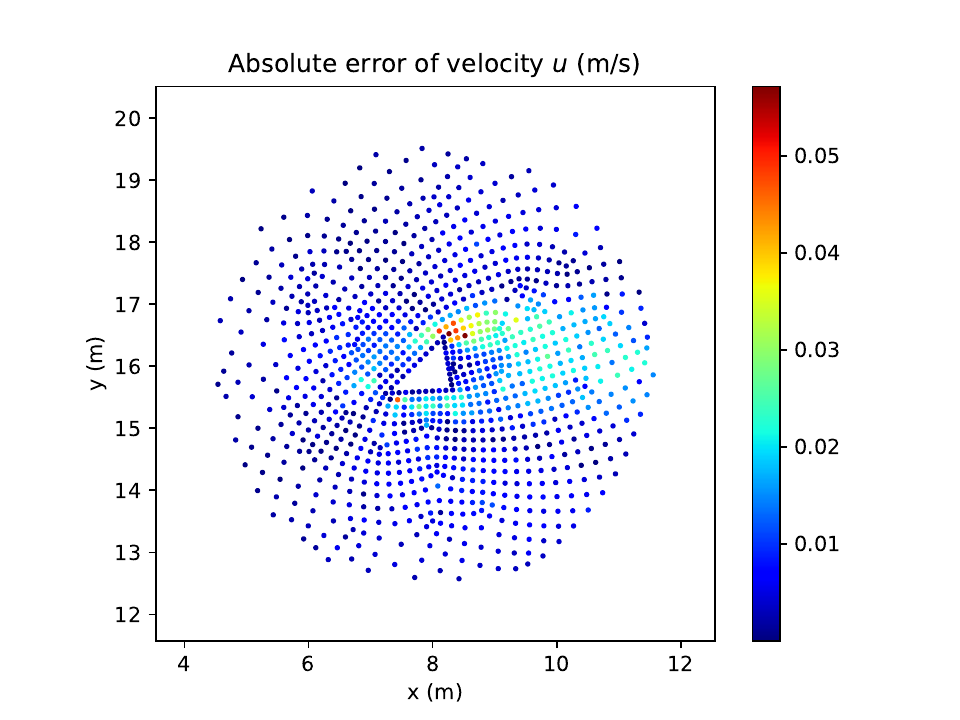}
    \end{subfigure}

    
    \begin{subfigure}[b]{0.27\textwidth}
        \centering
        \includegraphics[width=\textwidth]{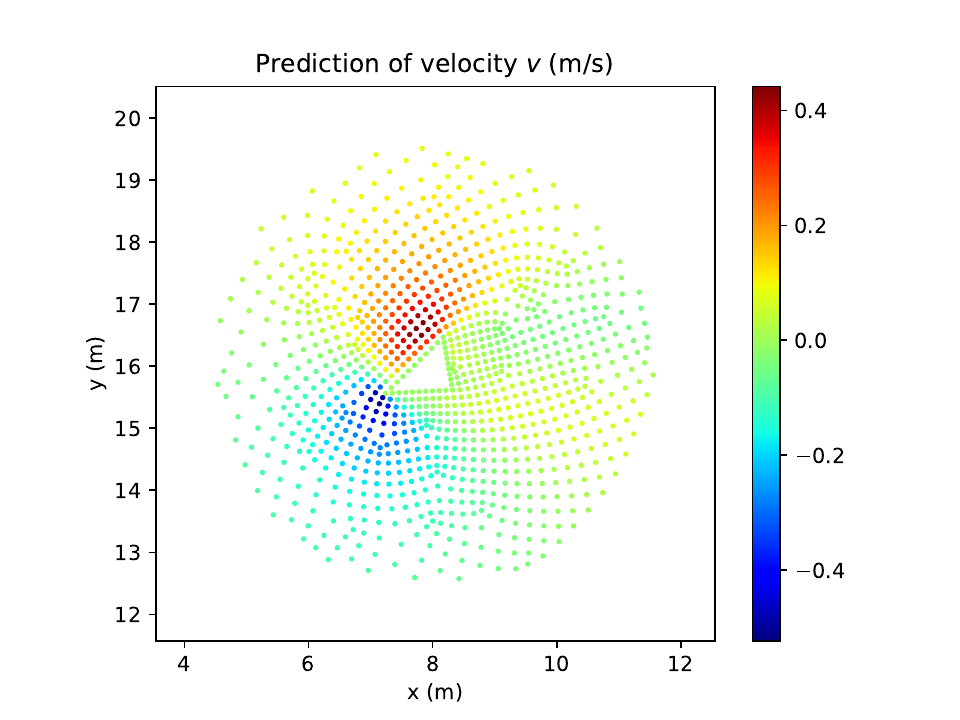}
    \end{subfigure}
    \begin{subfigure}[b]{0.27\textwidth}
        \centering
        \includegraphics[width=\textwidth]{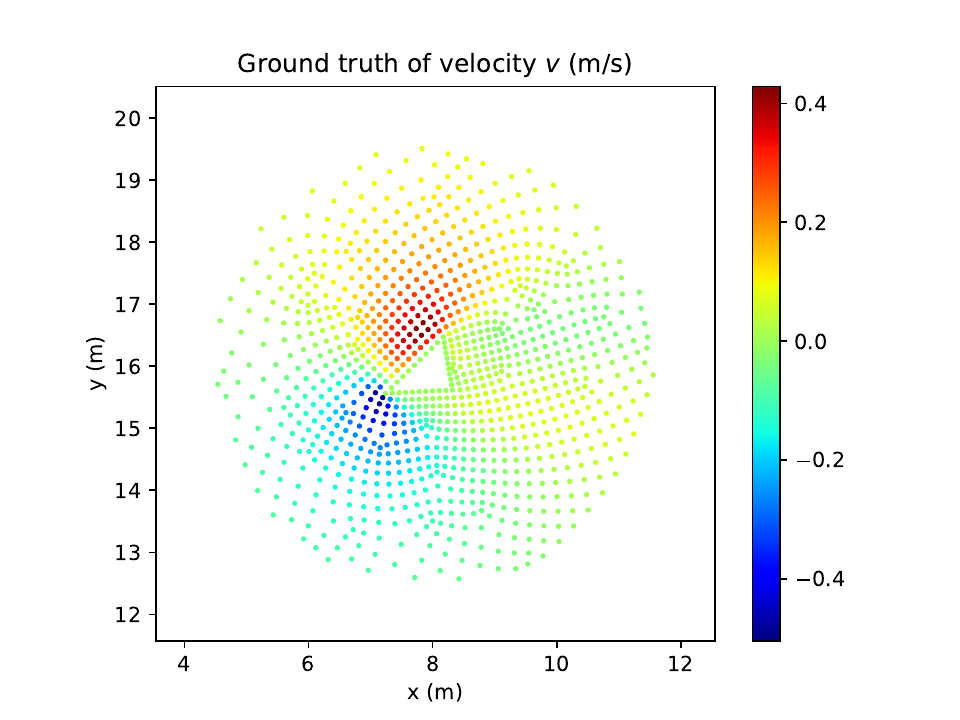}
    \end{subfigure}
    \begin{subfigure}[b]{0.27\textwidth}
        \centering
        \includegraphics[width=\textwidth]{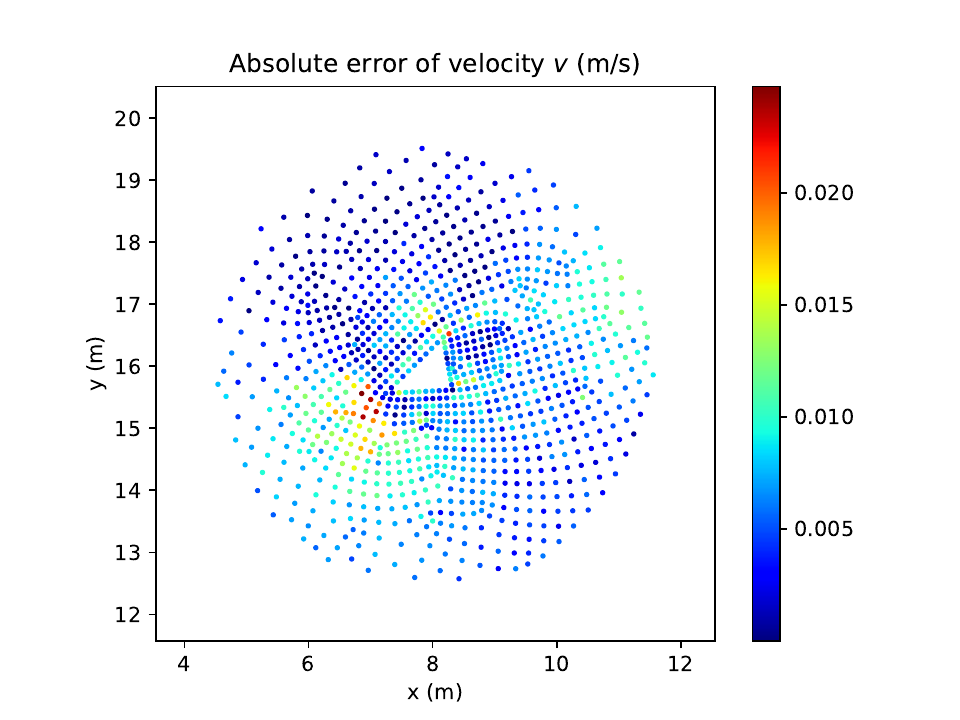}
    \end{subfigure}

    
    \begin{subfigure}[b]{0.27\textwidth}
        \centering
        \includegraphics[width=\textwidth]{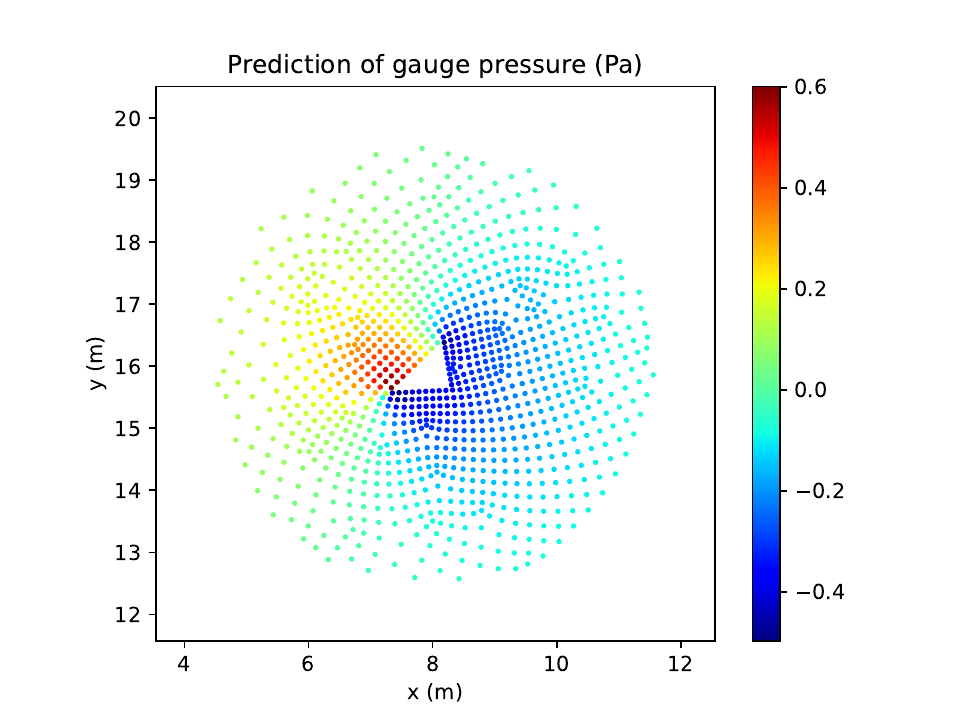}
    \end{subfigure}
    \begin{subfigure}[b]{0.27\textwidth}
        \centering
        \includegraphics[width=\textwidth]{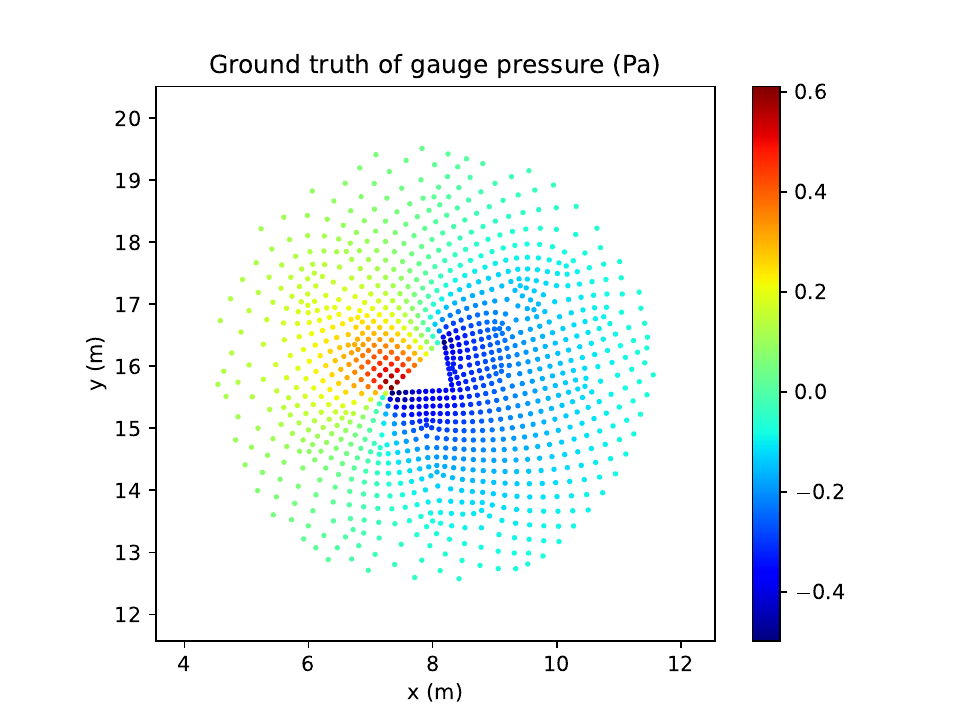}
    \end{subfigure}
    \begin{subfigure}[b]{0.27\textwidth}
        \centering
        \includegraphics[width=\textwidth]{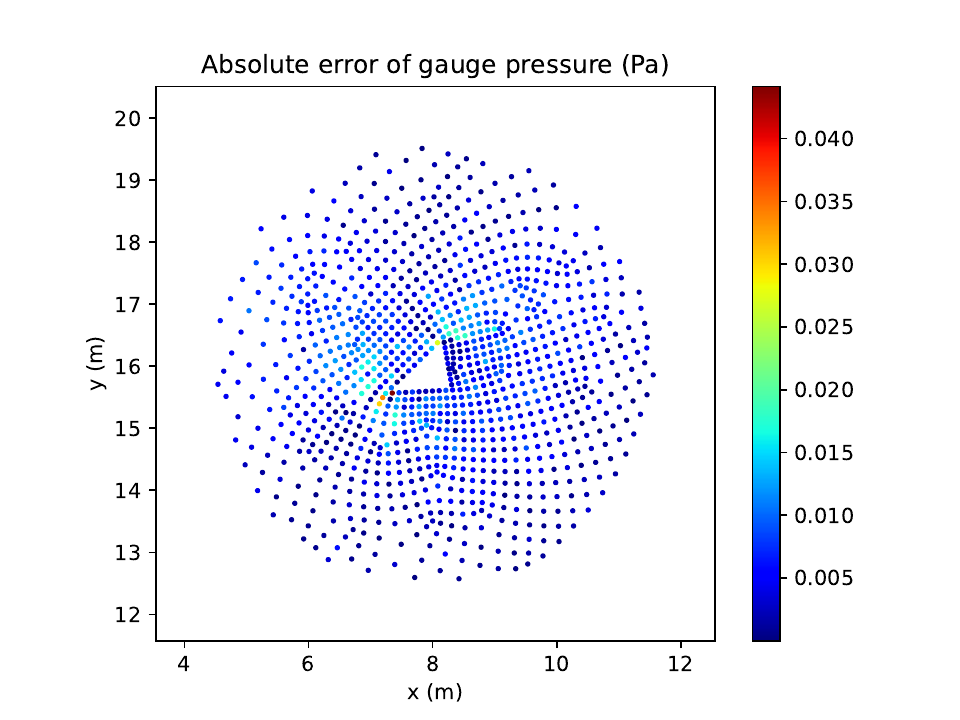}
    \end{subfigure}

  \caption{The second set of examples comparing the ground truth to the predictions of Diffusion PointNet for the velocity and pressure fields from the test set.}
  \label{Fig6}
\end{figure}

\begin{figure}[!htbp]
  \centering 
      \begin{subfigure}[b]{0.27\textwidth}
        \centering
        \includegraphics[width=\textwidth]{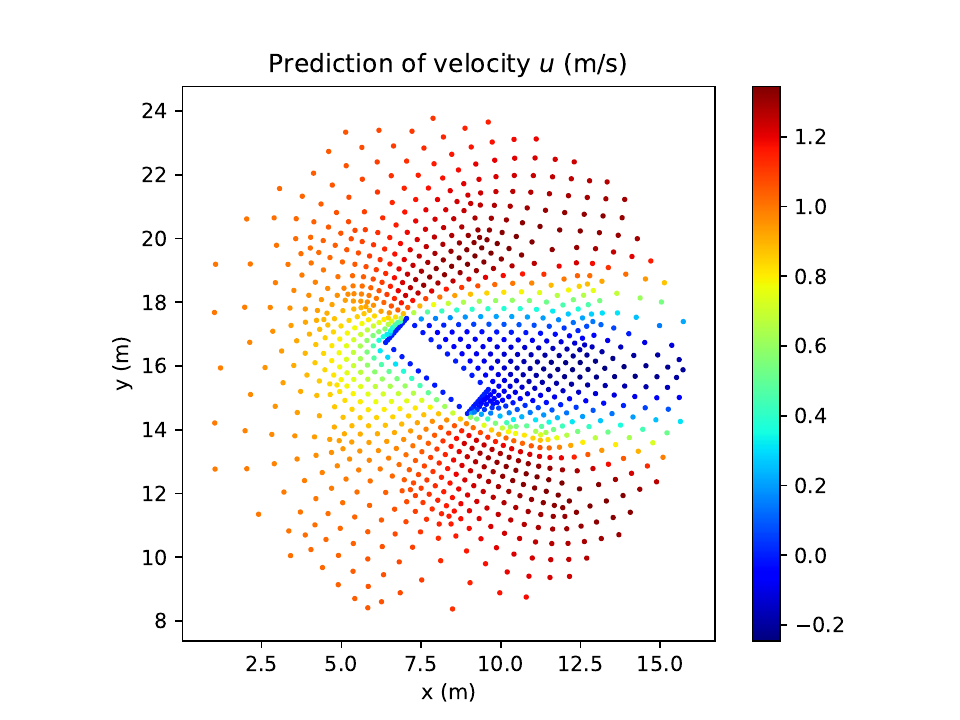}
    \end{subfigure}
    \begin{subfigure}[b]{0.27\textwidth}
        \centering
        \includegraphics[width=\textwidth]{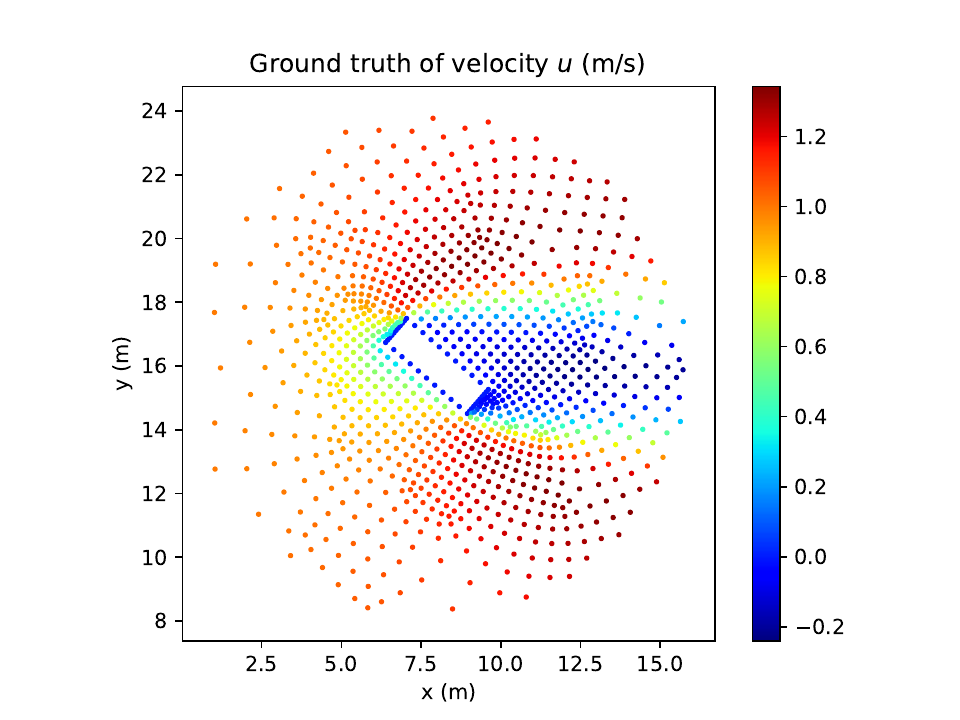}
    \end{subfigure}
    \begin{subfigure}[b]{0.27\textwidth}
        \centering
        \includegraphics[width=\textwidth]{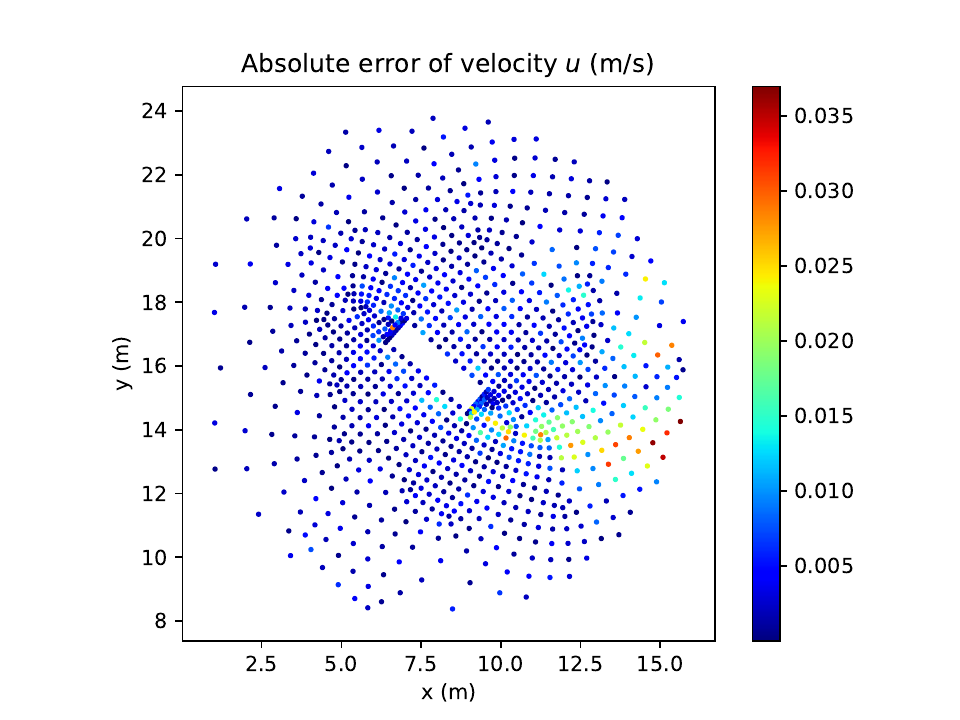}
    \end{subfigure}

    
    \begin{subfigure}[b]{0.27\textwidth}
        \centering
        \includegraphics[width=\textwidth]{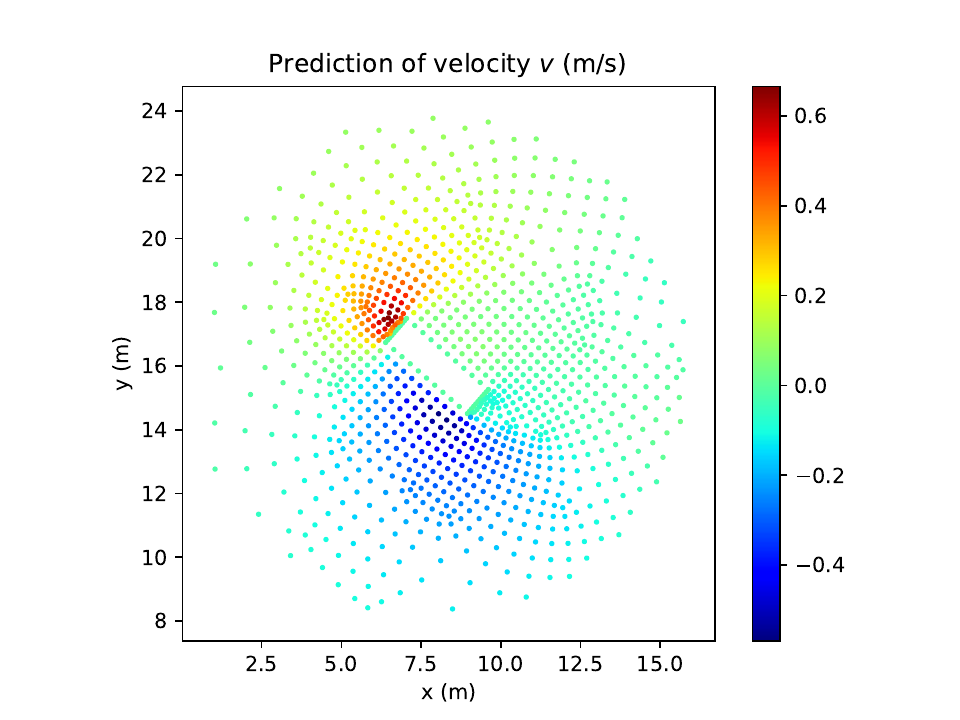}
    \end{subfigure}
    \begin{subfigure}[b]{0.27\textwidth}
        \centering
        \includegraphics[width=\textwidth]{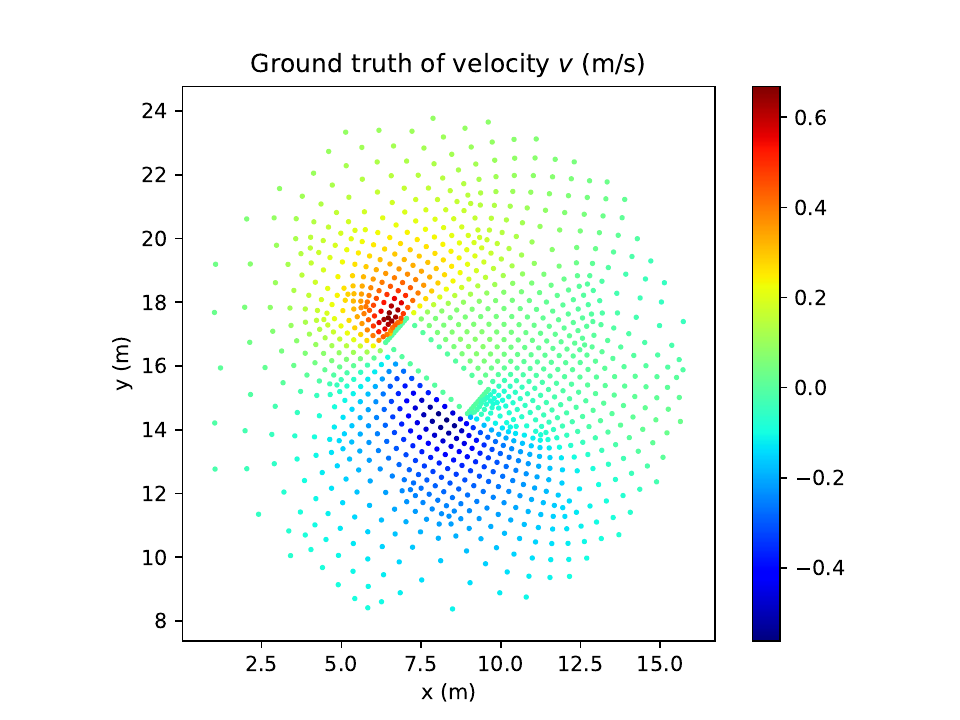}
    \end{subfigure}
    \begin{subfigure}[b]{0.27\textwidth}
        \centering
        \includegraphics[width=\textwidth]{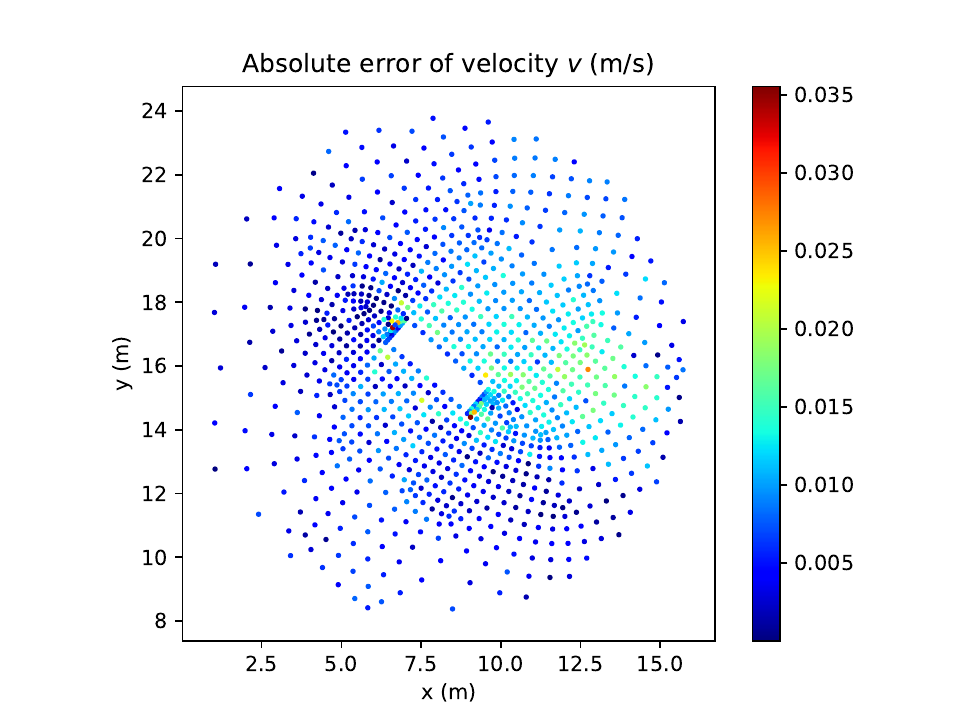}
    \end{subfigure}

    
    \begin{subfigure}[b]{0.27\textwidth}
        \centering
        \includegraphics[width=\textwidth]{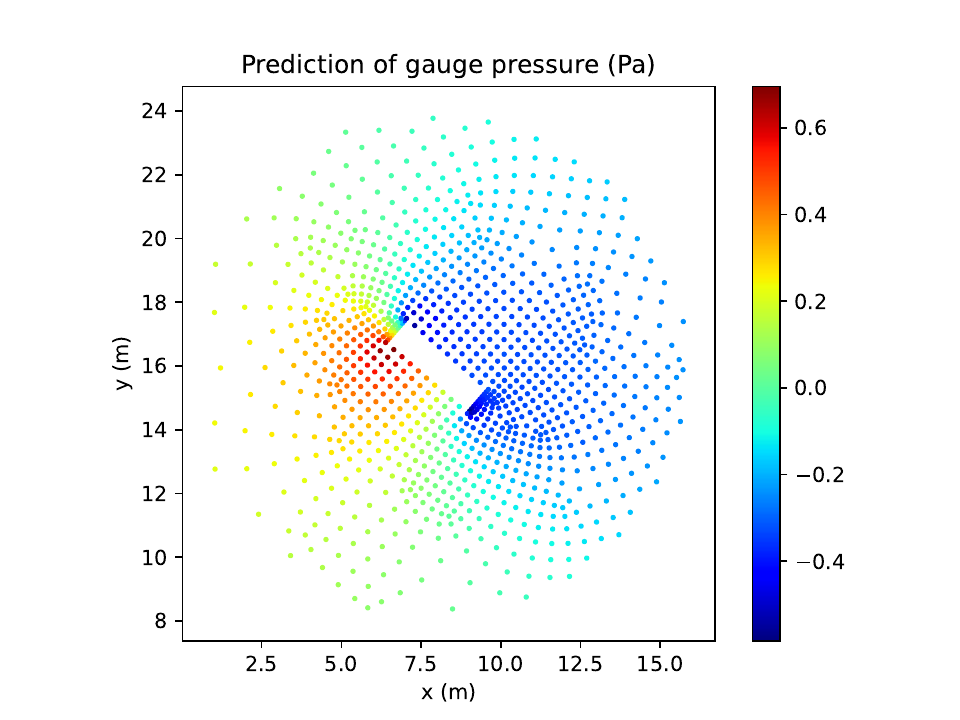}
    \end{subfigure}
    \begin{subfigure}[b]{0.27\textwidth}
        \centering
        \includegraphics[width=\textwidth]{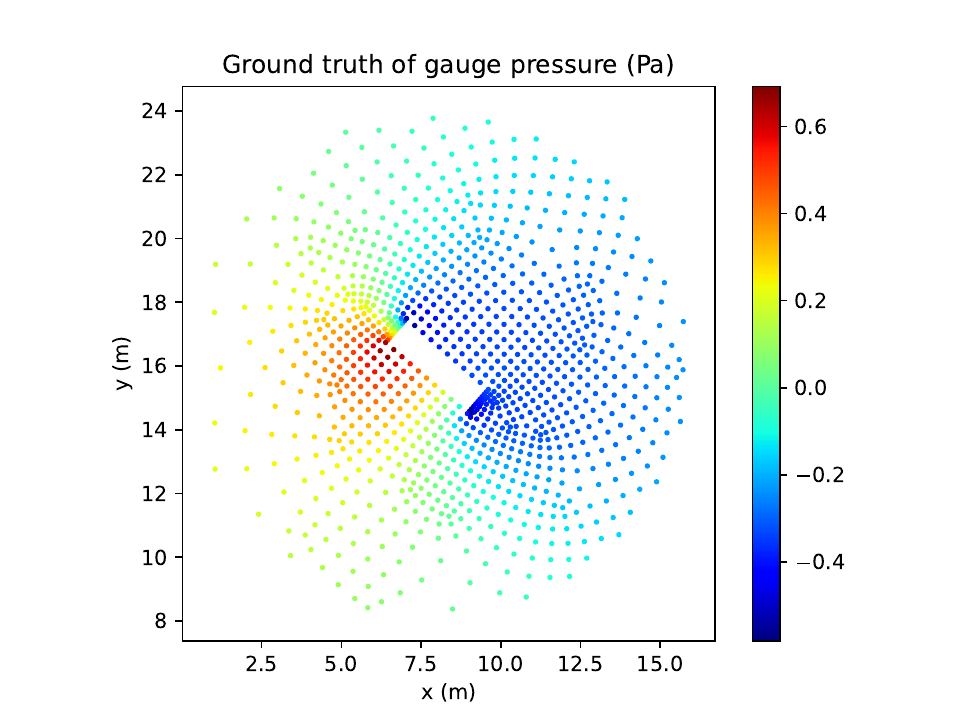}
    \end{subfigure}
    \begin{subfigure}[b]{0.27\textwidth}
        \centering
        \includegraphics[width=\textwidth]{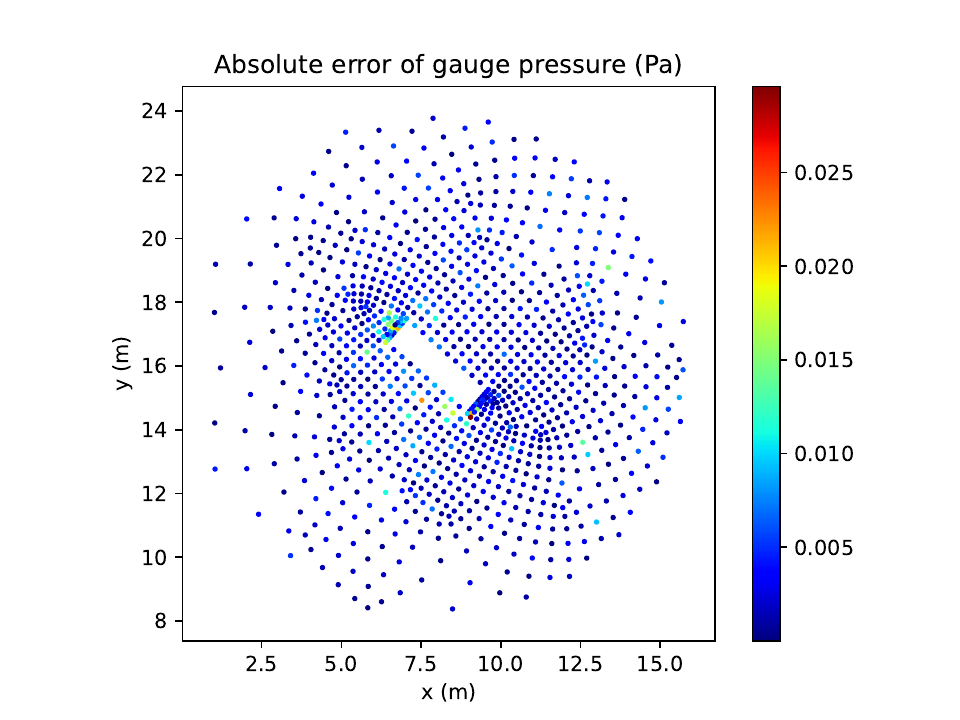}
    \end{subfigure}

  \caption{The third set of examples comparing the ground truth to the predictions of Diffusion PointNet for the velocity and pressure fields from the test set.}
  \label{Fig7}
\end{figure}

\begin{figure}[!htbp]
  \centering 
      \begin{subfigure}[b]{0.27\textwidth}
        \centering
        \includegraphics[width=\textwidth]{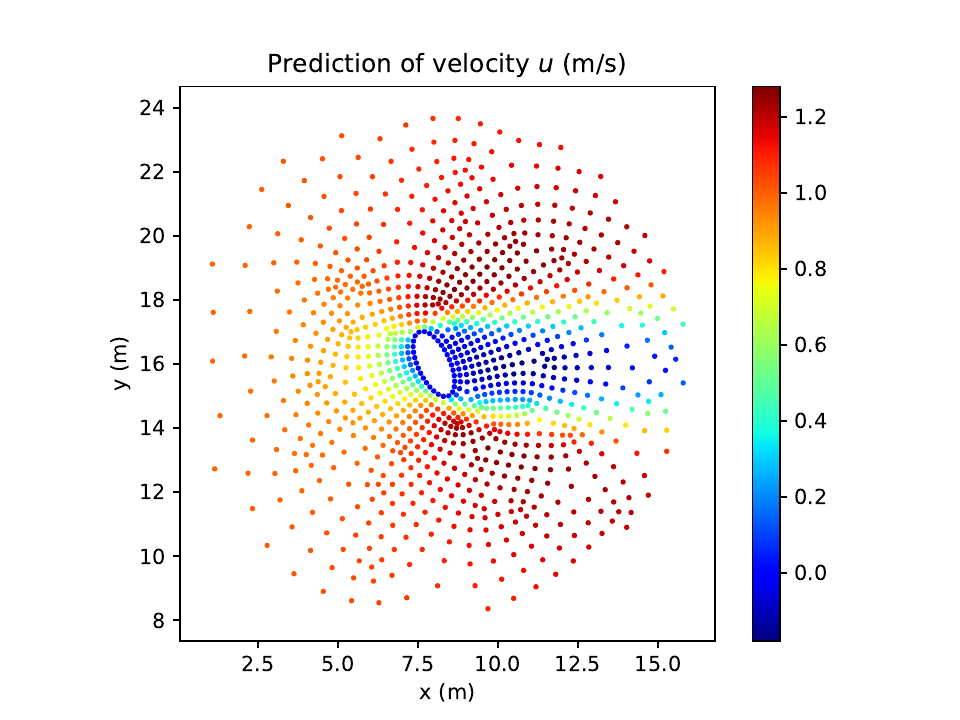}
    \end{subfigure}
    \begin{subfigure}[b]{0.27\textwidth}
        \centering
        \includegraphics[width=\textwidth]{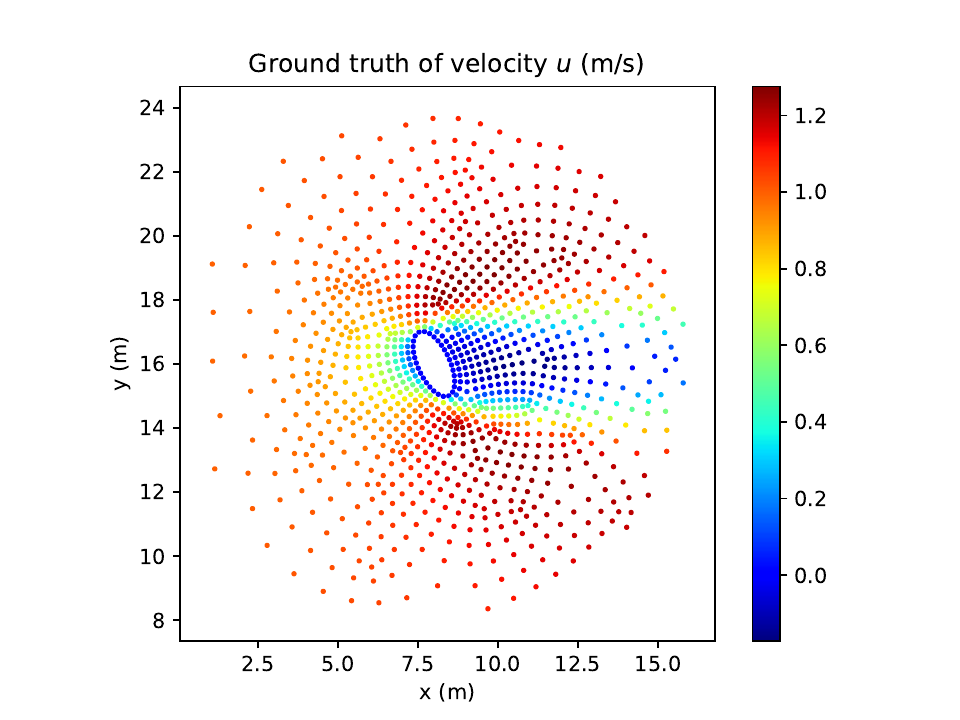}
    \end{subfigure}
    \begin{subfigure}[b]{0.27\textwidth}
        \centering
        \includegraphics[width=\textwidth]{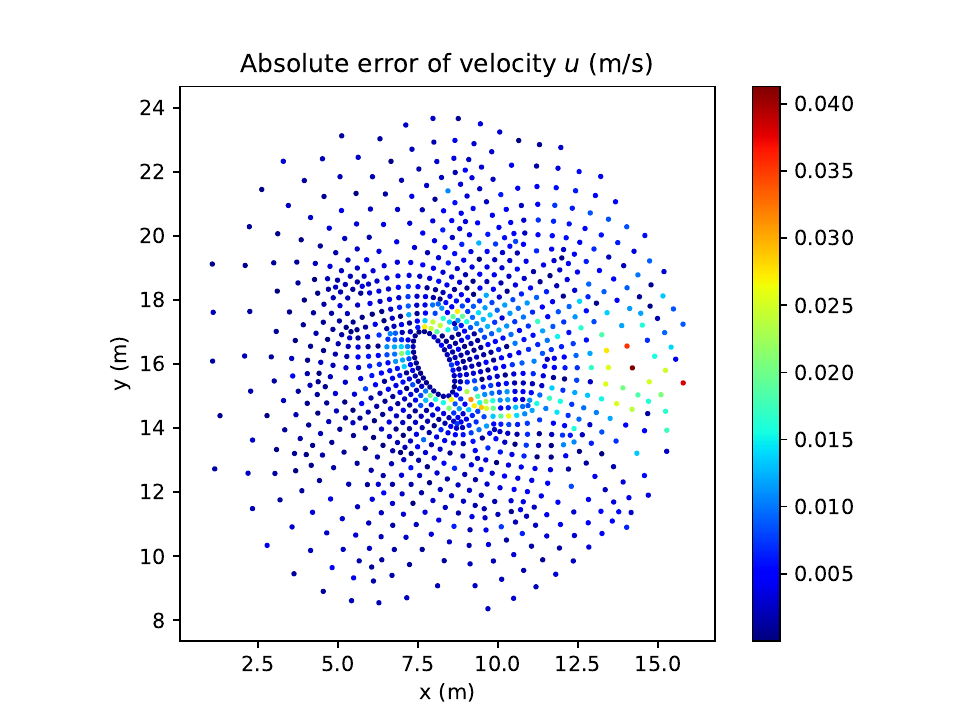}
    \end{subfigure}

    
    \begin{subfigure}[b]{0.27\textwidth}
        \centering
        \includegraphics[width=\textwidth]{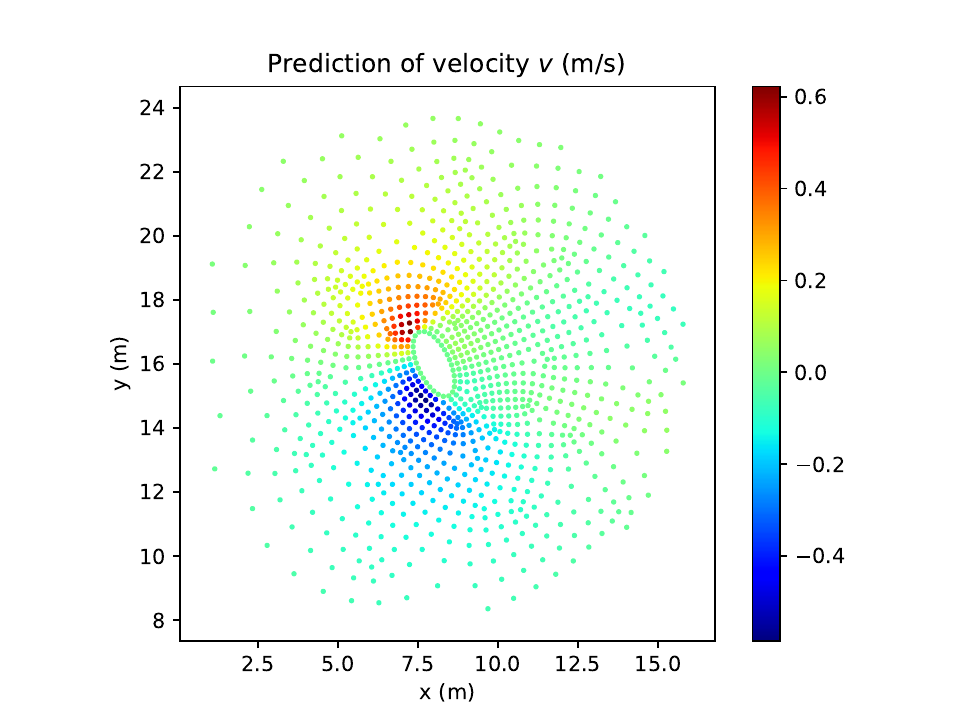}
    \end{subfigure}
    \begin{subfigure}[b]{0.27\textwidth}
        \centering
        \includegraphics[width=\textwidth]{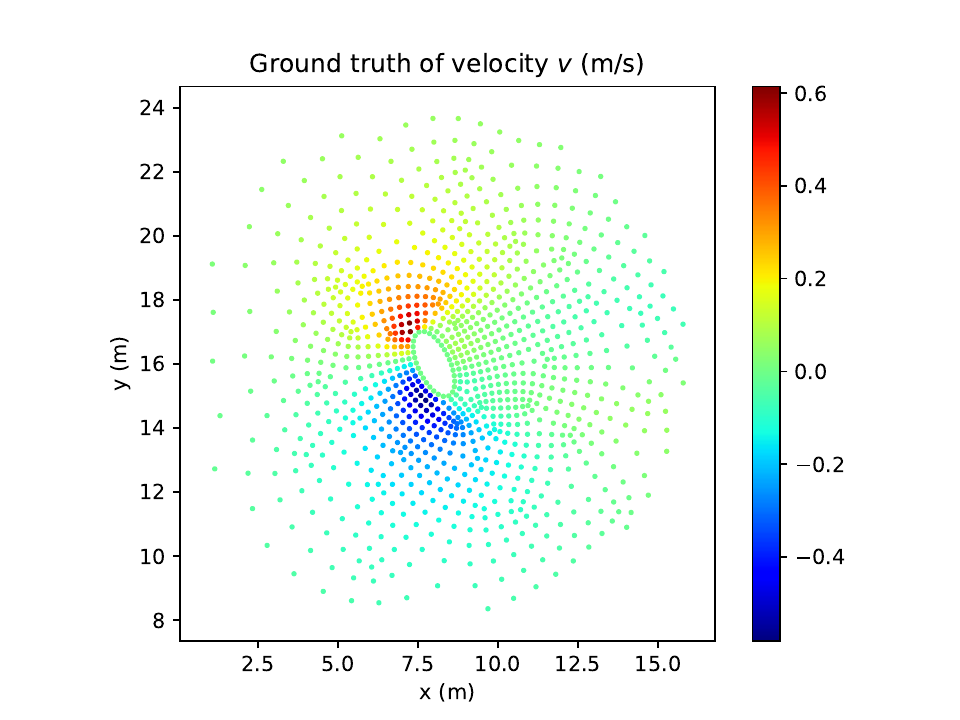}
    \end{subfigure}
    \begin{subfigure}[b]{0.27\textwidth}
        \centering
        \includegraphics[width=\textwidth]{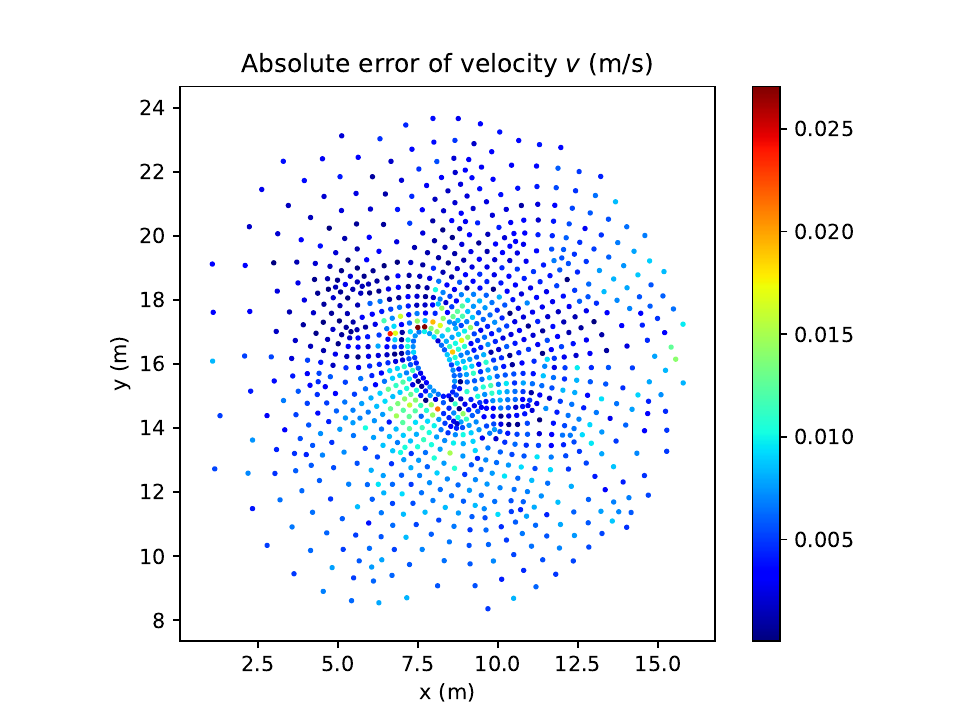}
    \end{subfigure}

    
    \begin{subfigure}[b]{0.27\textwidth}
        \centering
        \includegraphics[width=\textwidth]{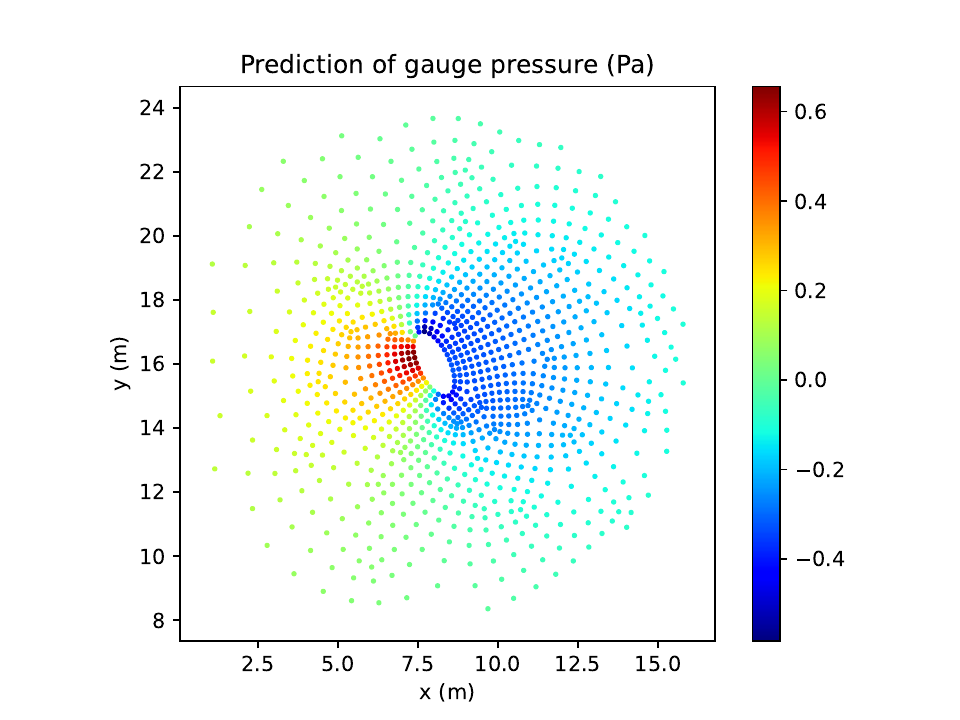}
    \end{subfigure}
    \begin{subfigure}[b]{0.27\textwidth}
        \centering
        \includegraphics[width=\textwidth]{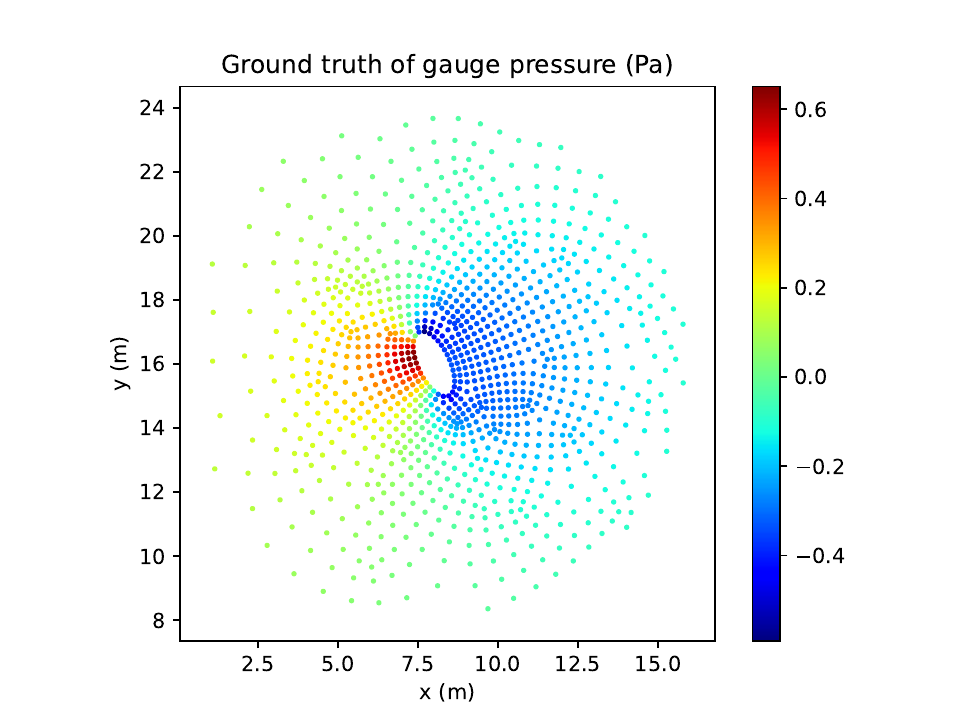}
    \end{subfigure}
    \begin{subfigure}[b]{0.27\textwidth}
        \centering
        \includegraphics[width=\textwidth]{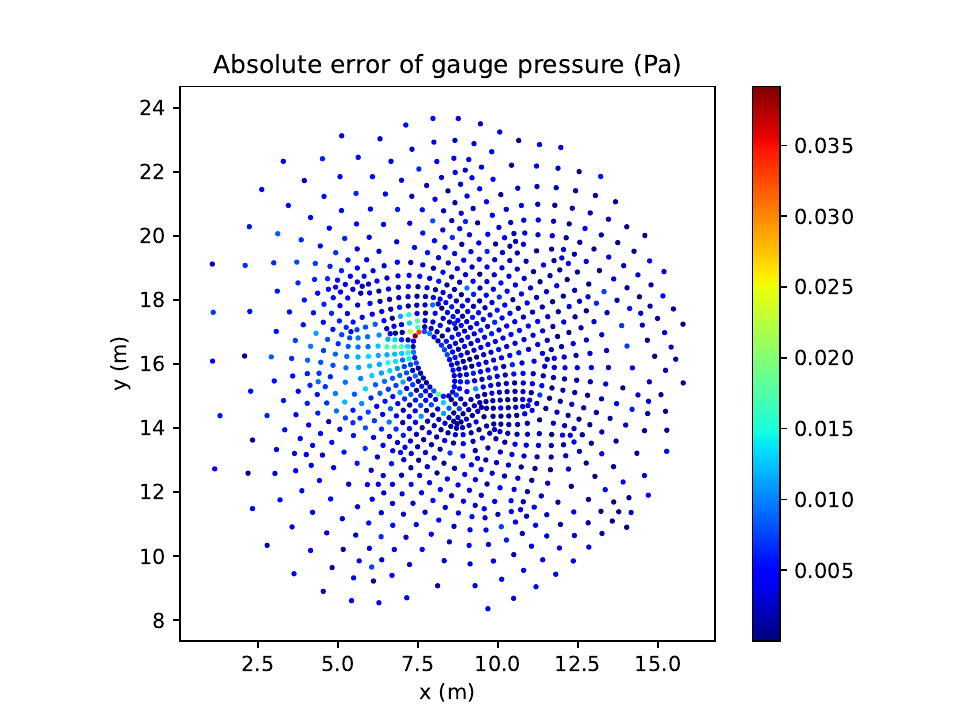}
    \end{subfigure}

  \caption{The fourth set of examples comparing the ground truth to the predictions of Diffusion PointNet for the velocity and pressure fields from the test set.}
  \label{Fig8}
\end{figure}

Following the numerical analysis, a visual comparison of the results is conducted. Figures \ref{Fig9}--\ref{Fig12} present four examples of the Flow Matching PointNet model predictions for velocity and pressure fields, alongside the corresponding ground truth and absolute pointwise error distributions. These figures include cylinders with pentagonal, rectangular, and elliptical cross-sections. Similarly, Figures \ref{Fig5}--\ref{Fig8} illustrate the results of the Diffusion PointNet model for cylinders with hexagonal, triangular, rectangular, and elliptical cross-sections. In all cases, the predictions demonstrate excellent accuracy and a strong agreement with the ground truth. The majority of the absolute errors occur on or near the cylinder surface, where the geometric variations among different test cases are most significant. In Fig. \ref{Fig11} and Fig. \ref{Fig7}, the $x$ component of the velocity vector ($u$) becomes negative, indicating the formation of secondary flow regions, and both the Flow Matching PointNet and Diffusion PointNet models successfully capture these phenomena. It should be noted that the test set is identical for all three models: Flow Matching PointNet, Diffusion PointNet, and the baseline PointNet. For visual diversity, however, we intentionally select cylinders with different cross-sectional shapes to provide a more comprehensive visual representation in Figs. \ref{Fig9}--\ref{Fig8}. These results highlight the models' success in handling diverse geometries with varying shapes, orientations, and sizes. Another notable observation is that the computational domain size differs for each point cloud. For example, as can be seen in Fig. \ref{Fig9}, the approximate minimum and maximum $x$-coordinates are 5 m and 11 m, respectively, while the corresponding $y$-coordinates range from 13 m to 19 m. In contrast, as shown in Fig. \ref{Fig11}, the $x$-coordinates vary approximately from 2 m to 14 m, and the $y$-coordinates from 10 m to 22 m. A similar comparison can be made between Fig. \ref{Fig5} and Fig. \ref{Fig8}. This adaptability reflects a fundamental strength of the PointNet architecture, which can process irregular geometries with varying domain sizes, unlike conventional CNNs that require fixed domain dimensions across the entire dataset. Additionally, the prediction of $\boldsymbol{f}_\text{target}$ by Flow Matching PointNet or $\boldsymbol{y}_\text{noisy}$ by Diffusion PointNet at each point depends on both the spatial coordinates of that point and the global geometry of the point cloud that contains it. Since the spatial ranges of the \(x\) and \(y\) coordinates vary across different point clouds in the dataset, accurately predicting $\boldsymbol{f}_\text{target}$ or $\boldsymbol{y}_\text{noisy}$ is a challenging task. The results presented here demonstrate that the proposed Flow Matching PointNet and Diffusion PointNet models successfully handle this challenge.


\begin{figure}
  \centering 
      \begin{subfigure}[b]{0.32\textwidth}
        \centering
        \includegraphics[width=\textwidth]{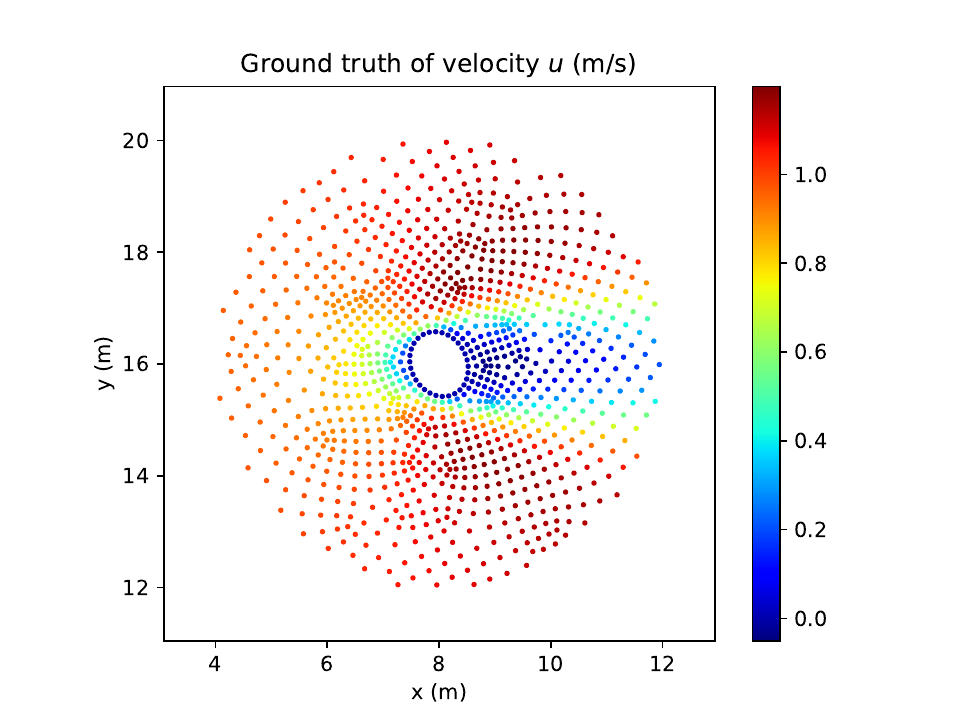}
    \end{subfigure}
    \begin{subfigure}[b]{0.32\textwidth}
        \centering
        \includegraphics[width=\textwidth]{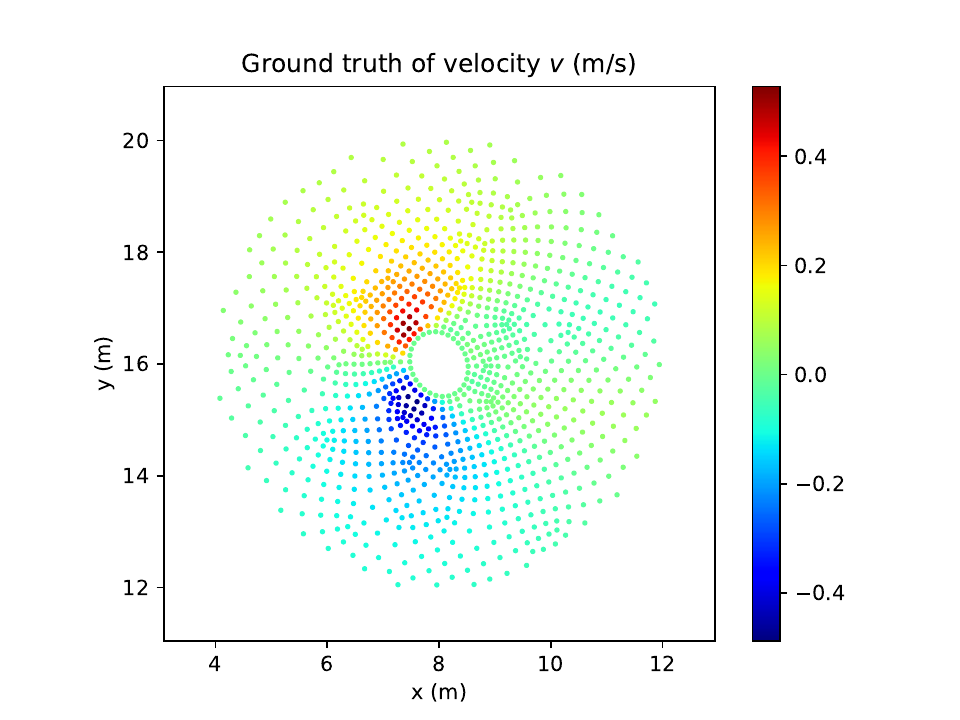}
    \end{subfigure}
    \begin{subfigure}[b]{0.32\textwidth}
        \centering
        \includegraphics[width=\textwidth]{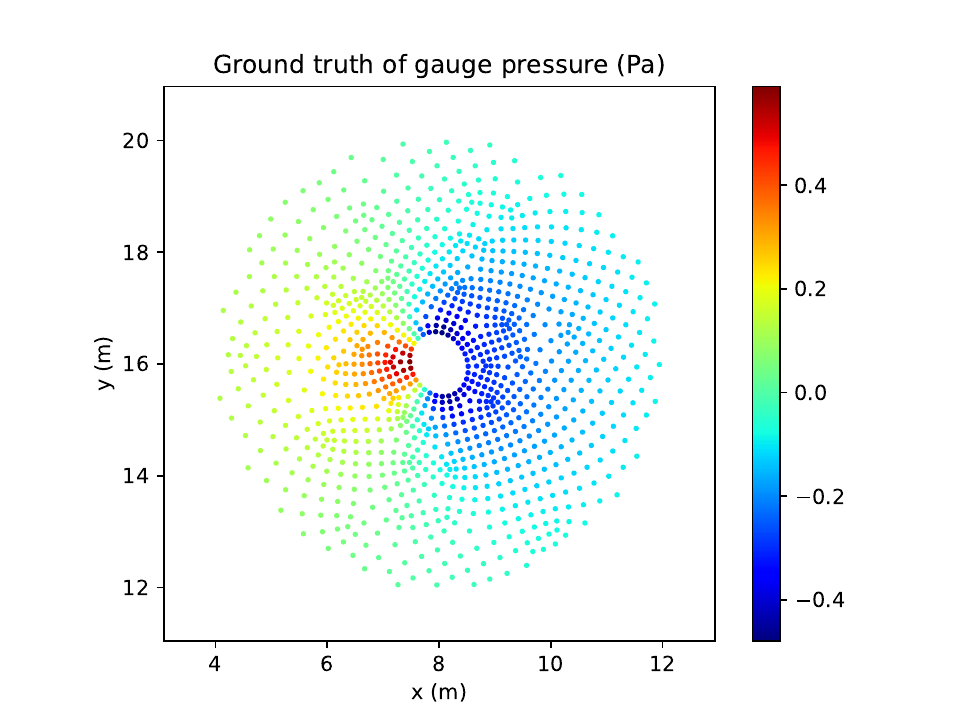}
    \end{subfigure}

    
    \begin{subfigure}[b]{0.32\textwidth}
        \centering
        \includegraphics[width=\textwidth]{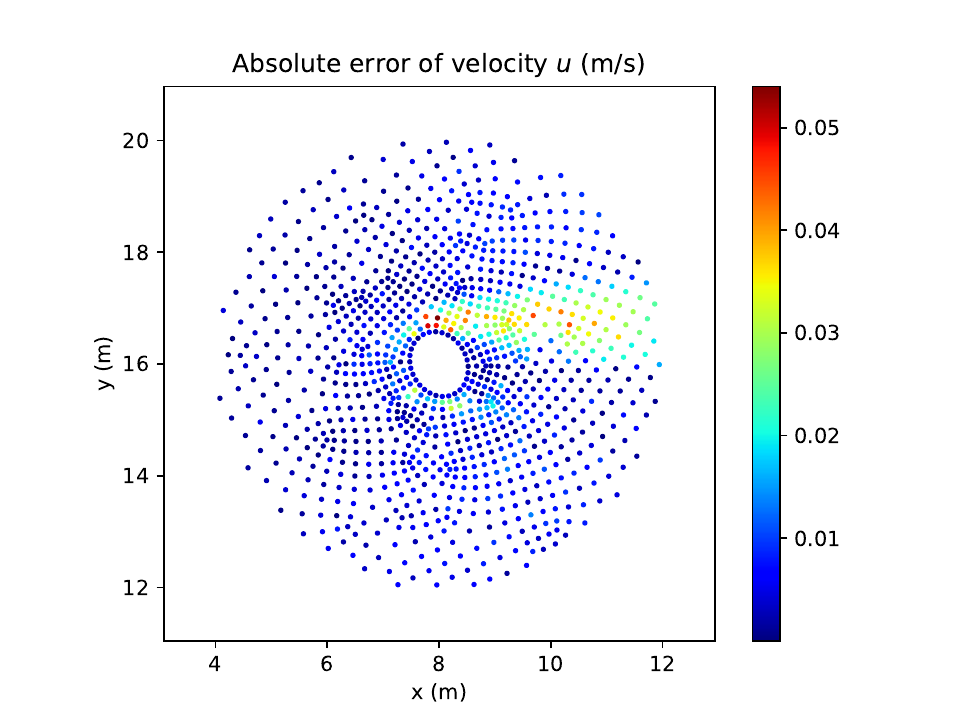}
    \end{subfigure}
    \begin{subfigure}[b]{0.32\textwidth}
        \centering
        \includegraphics[width=\textwidth]{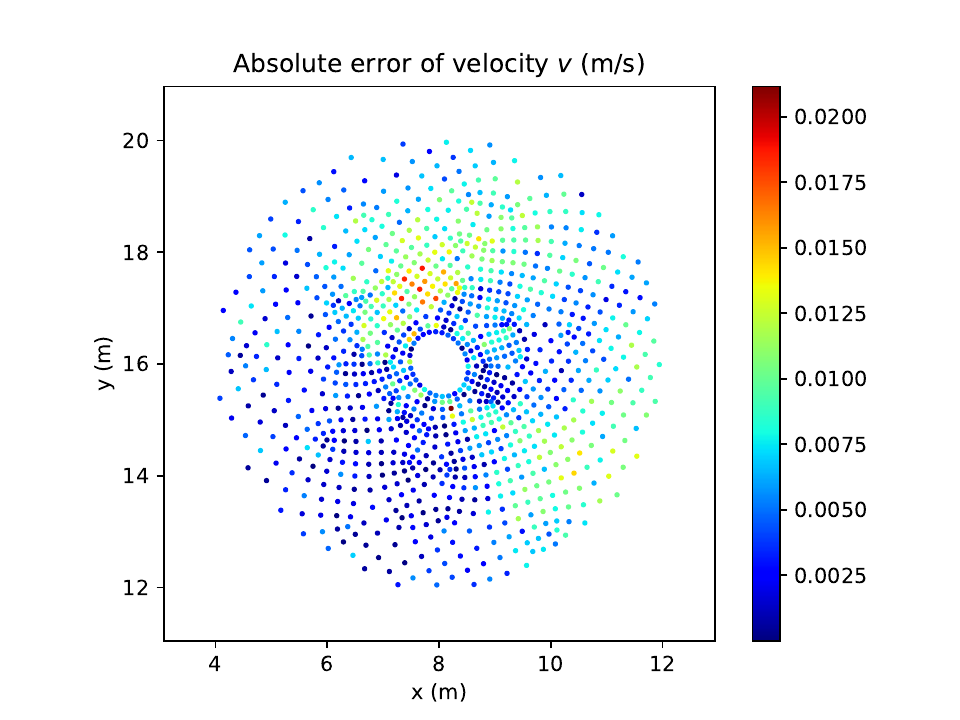}
    \end{subfigure}
    \begin{subfigure}[b]{0.32\textwidth}
        \centering
        \includegraphics[width=\textwidth]{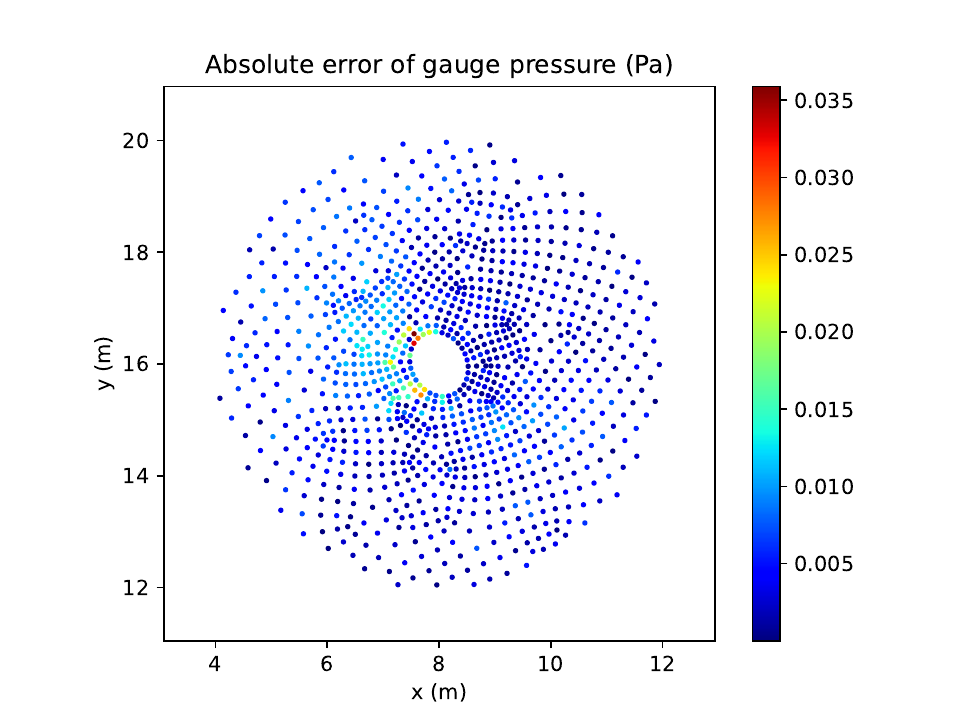}
    \end{subfigure}

    
    \begin{subfigure}[b]{0.32\textwidth}
        \centering
        \includegraphics[width=\textwidth]{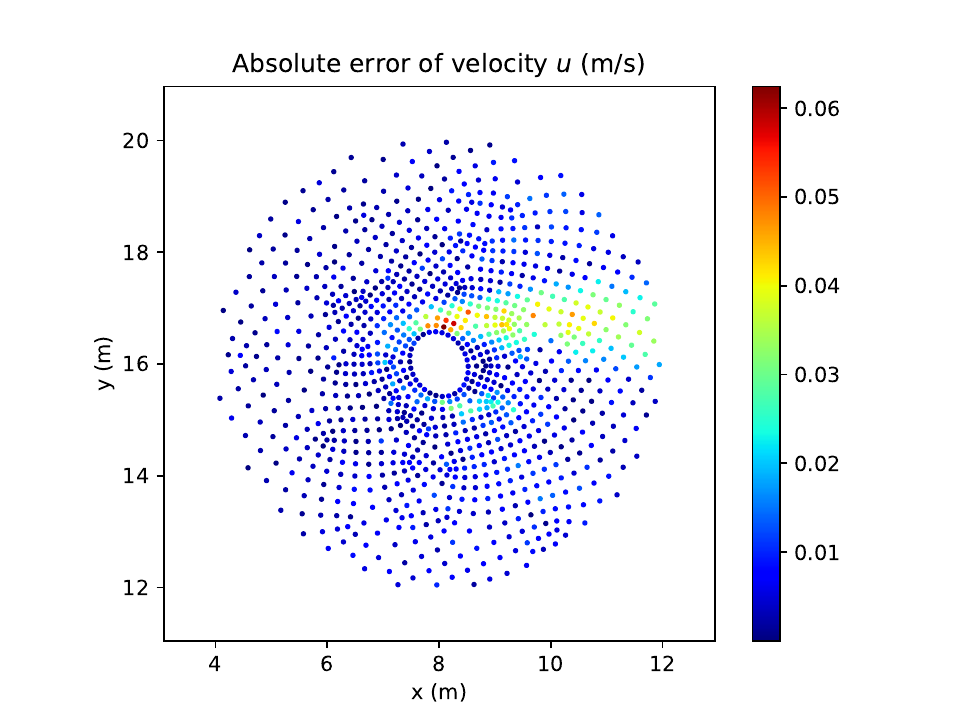}
    \end{subfigure}
    \begin{subfigure}[b]{0.32\textwidth}
        \centering
        \includegraphics[width=\textwidth]{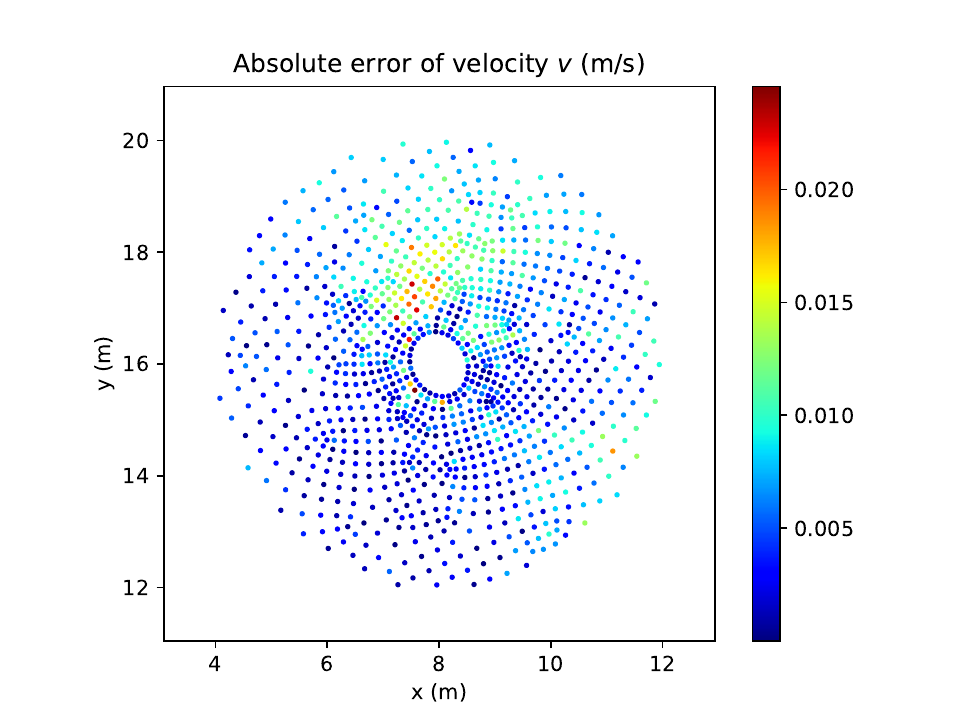}
    \end{subfigure}
    \begin{subfigure}[b]{0.32\textwidth}
        \centering
        \includegraphics[width=\textwidth]{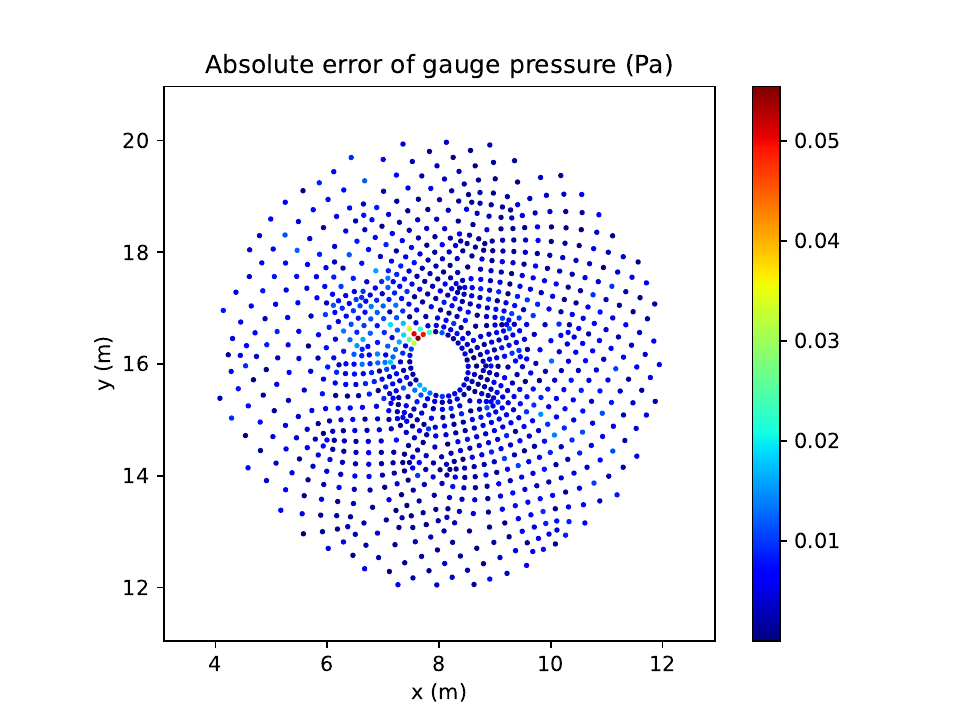}
    \end{subfigure}

 
    \begin{subfigure}[b]{0.32\textwidth}
        \centering
        \includegraphics[width=\textwidth]{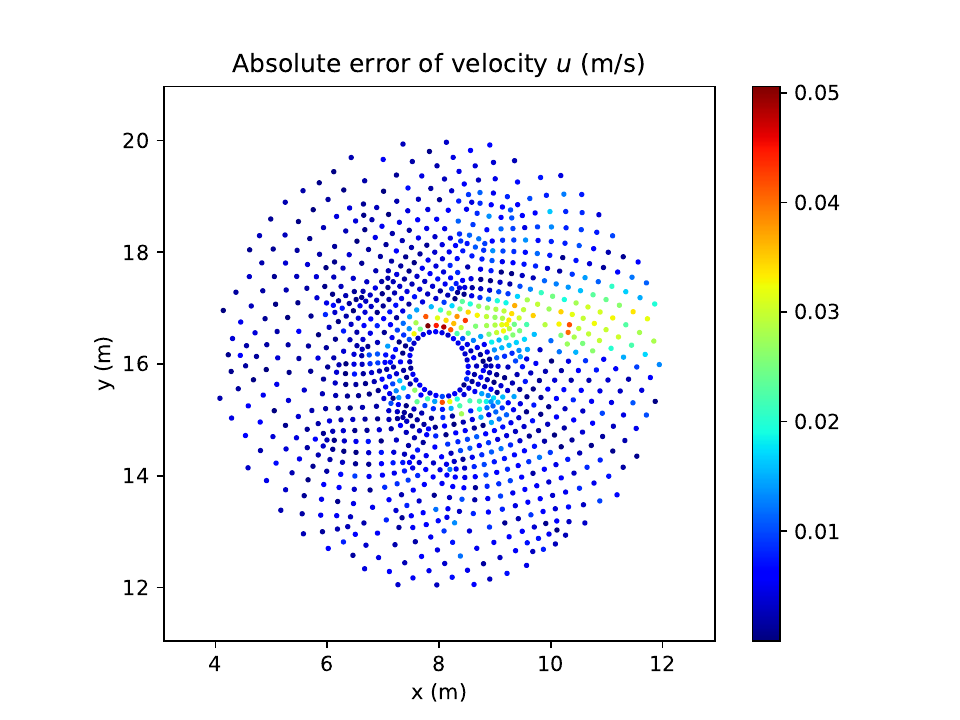}
    \end{subfigure}
    \begin{subfigure}[b]{0.32\textwidth}
        \centering
        \includegraphics[width=\textwidth]{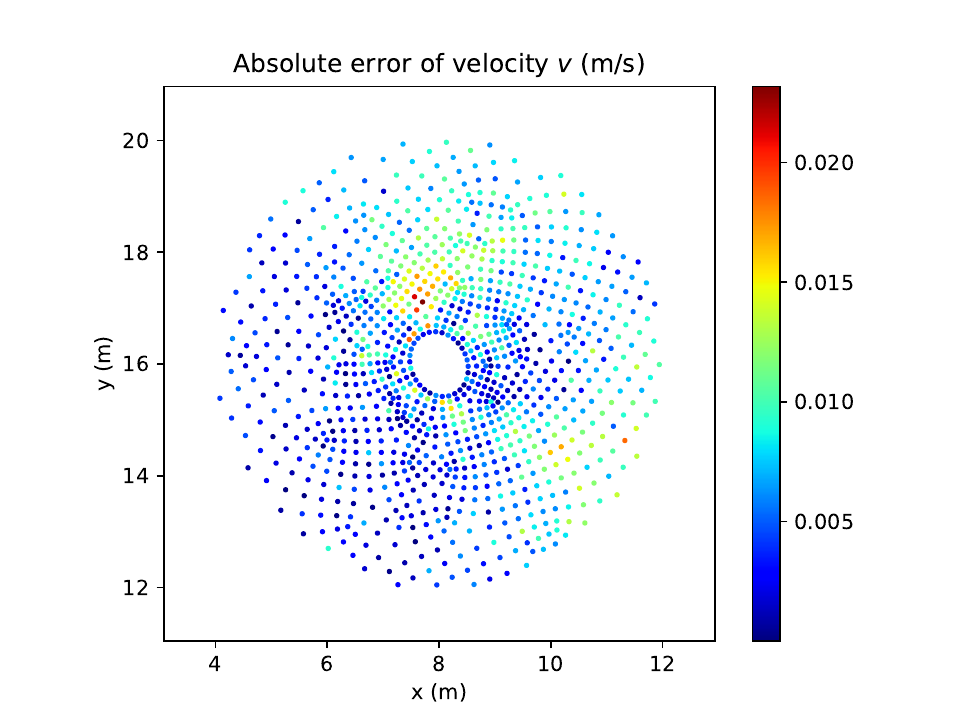}
    \end{subfigure}
    \begin{subfigure}[b]{0.32\textwidth}
        \centering
        \includegraphics[width=\textwidth]{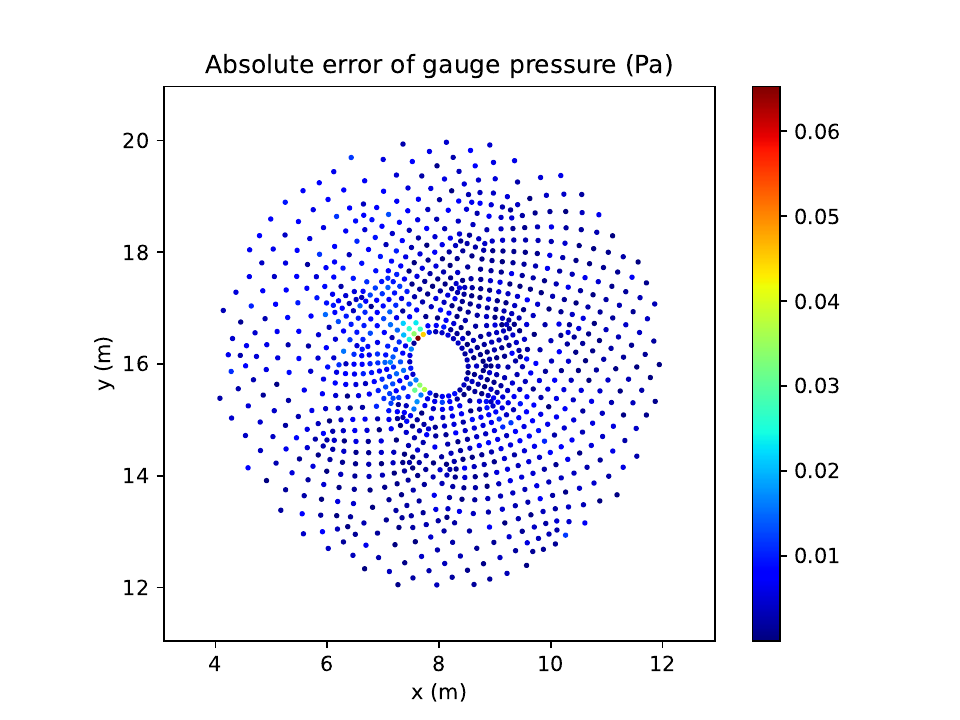}
    \end{subfigure}
      
    
    \begin{subfigure}[b]{0.32\textwidth}
    \centering
        \includegraphics[width=\textwidth]{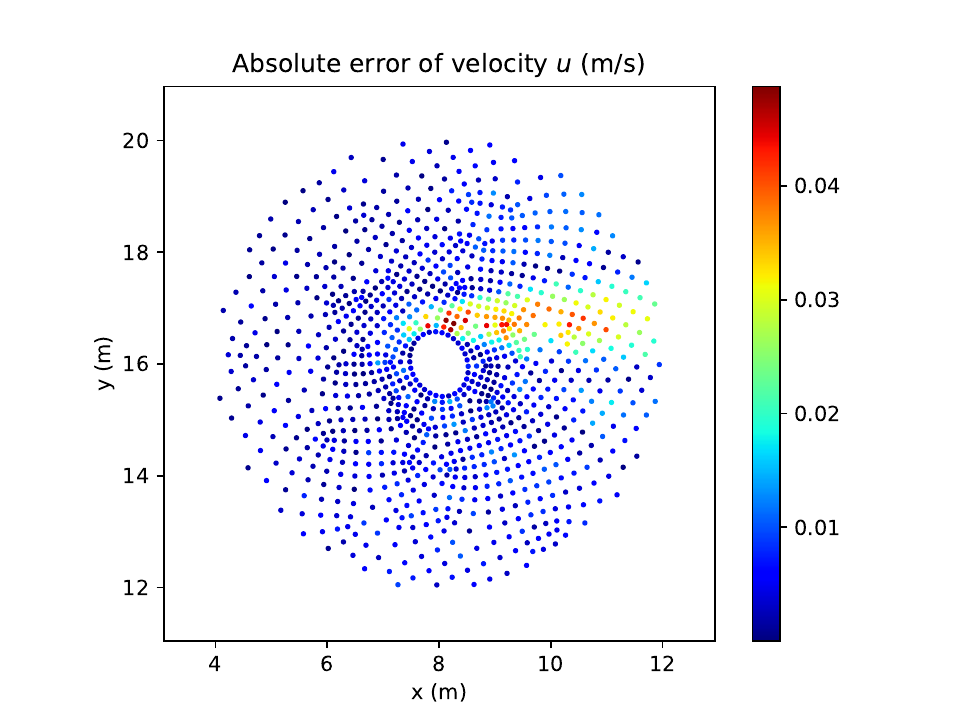}
    \end{subfigure}
    \begin{subfigure}[b]{0.32\textwidth}
        \centering
        \includegraphics[width=\textwidth]{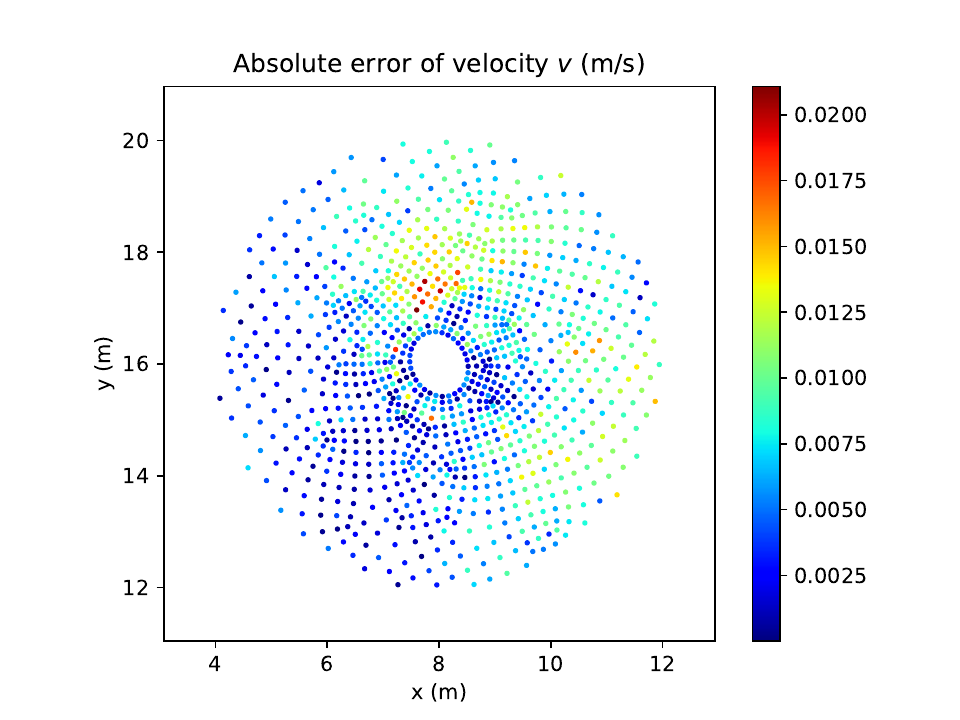}
    \end{subfigure}
    \begin{subfigure}[b]{0.32\textwidth}
        \centering
        \includegraphics[width=\textwidth]{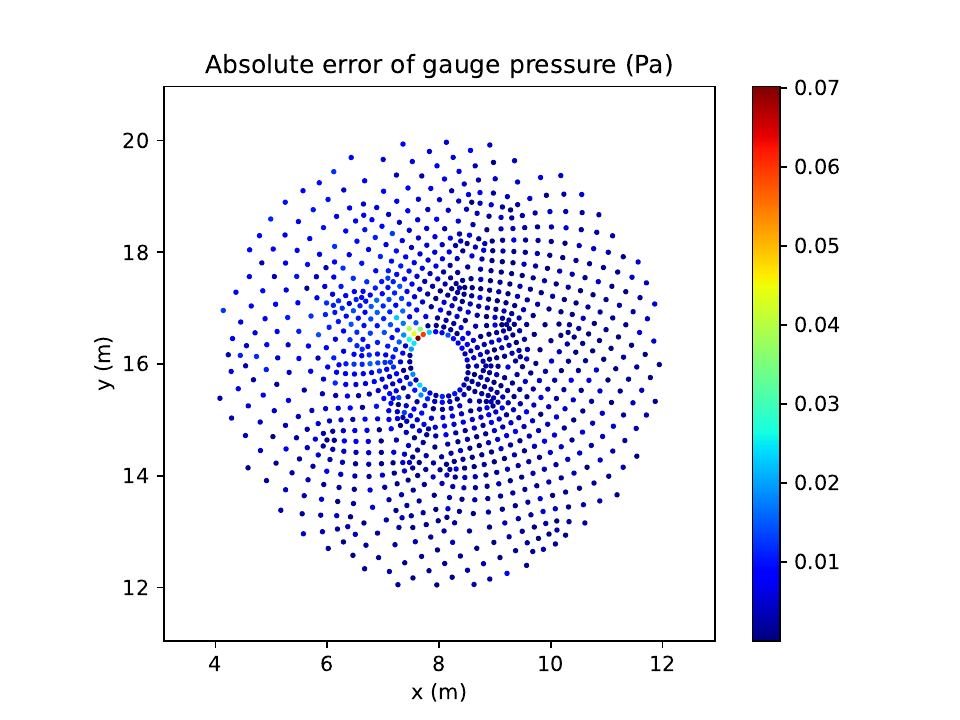}
    \end{subfigure}
    
    
    \begin{subfigure}[b]{0.32\textwidth}
        \centering
        \includegraphics[width=\textwidth]{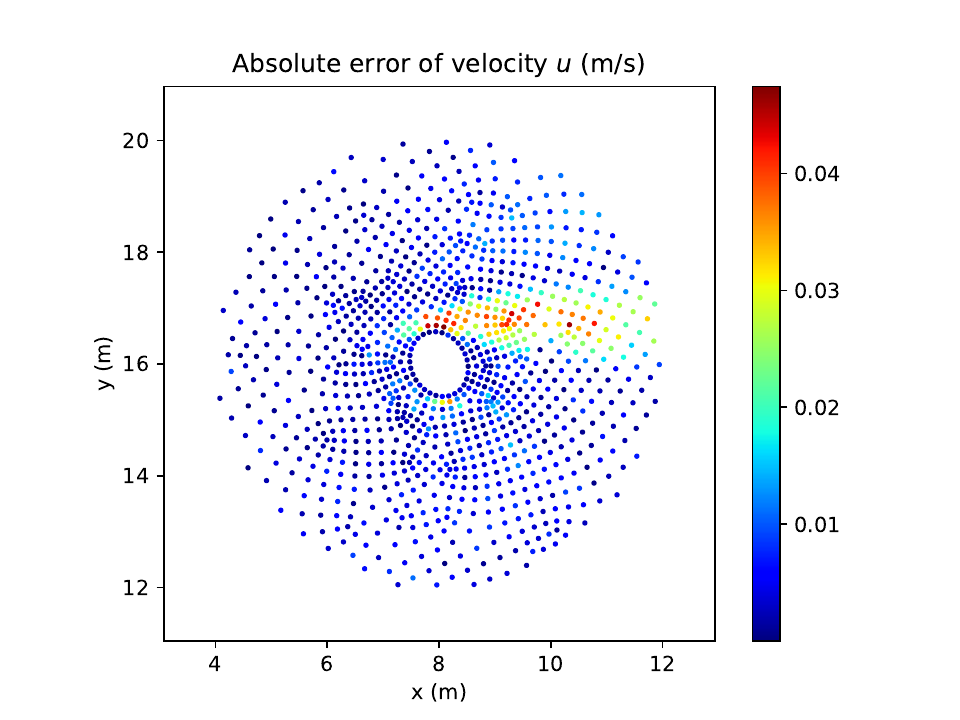}
    \end{subfigure}
    \begin{subfigure}[b]{0.32\textwidth}
        \centering
        \includegraphics[width=\textwidth]{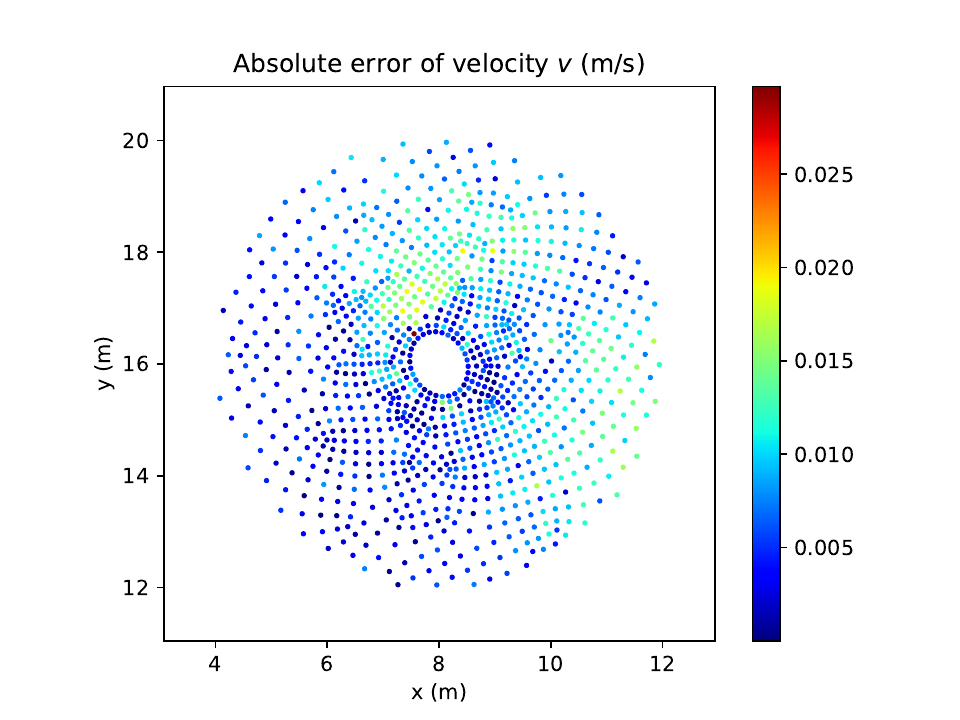}
    \end{subfigure}
    \begin{subfigure}[b]{0.32\textwidth}
        \centering
        \includegraphics[width=\textwidth]{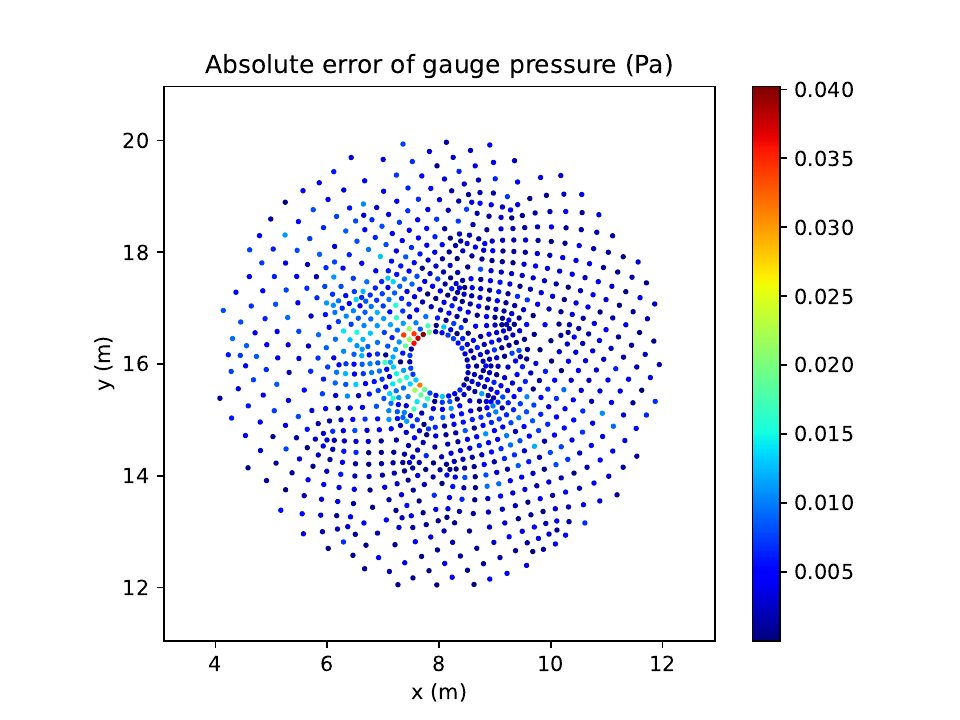}
    \end{subfigure}
    \caption{Flow Matching PointNet realizations; the first row shows the ground truth, and the remaining rows show the absolute pointwise error for five different realizations.}
  \label{Fig20A}
\end{figure}


\begin{figure}
  \centering 
      \begin{subfigure}[b]{0.32\textwidth}
        \centering
        \includegraphics[width=\textwidth]{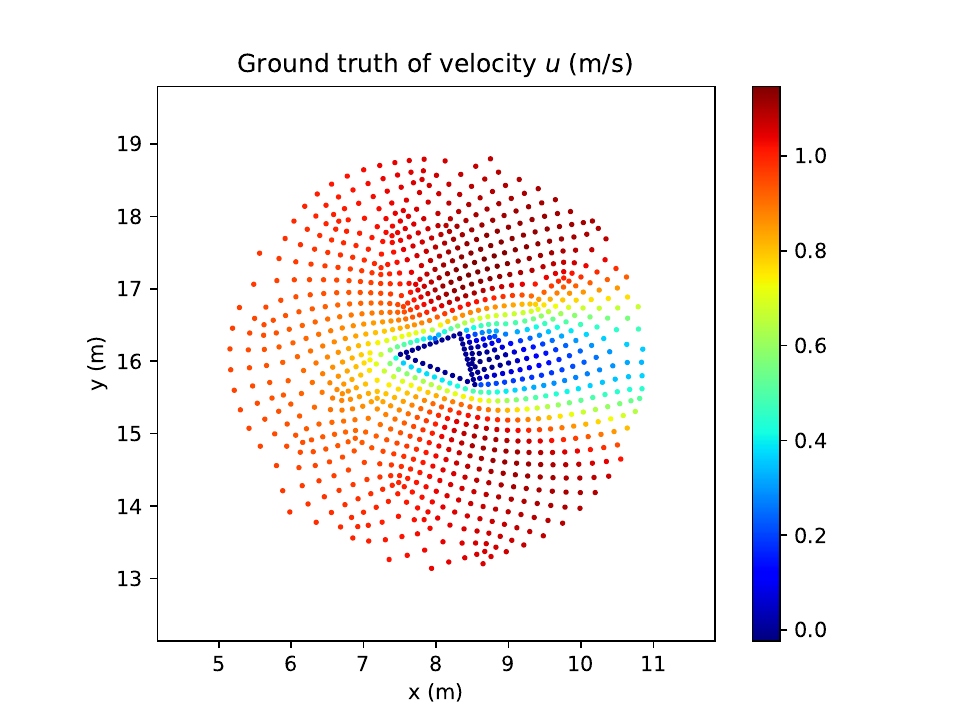}
    \end{subfigure}
    \begin{subfigure}[b]{0.32\textwidth}
        \centering
        \includegraphics[width=\textwidth]{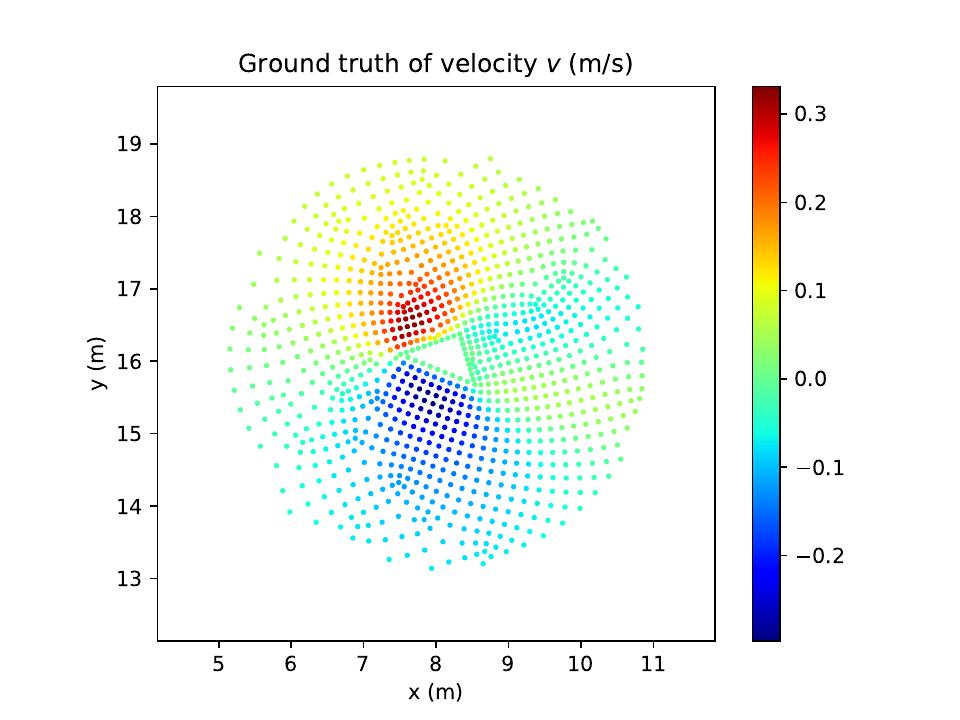}
    \end{subfigure}
    \begin{subfigure}[b]{0.32\textwidth}
        \centering
        \includegraphics[width=\textwidth]{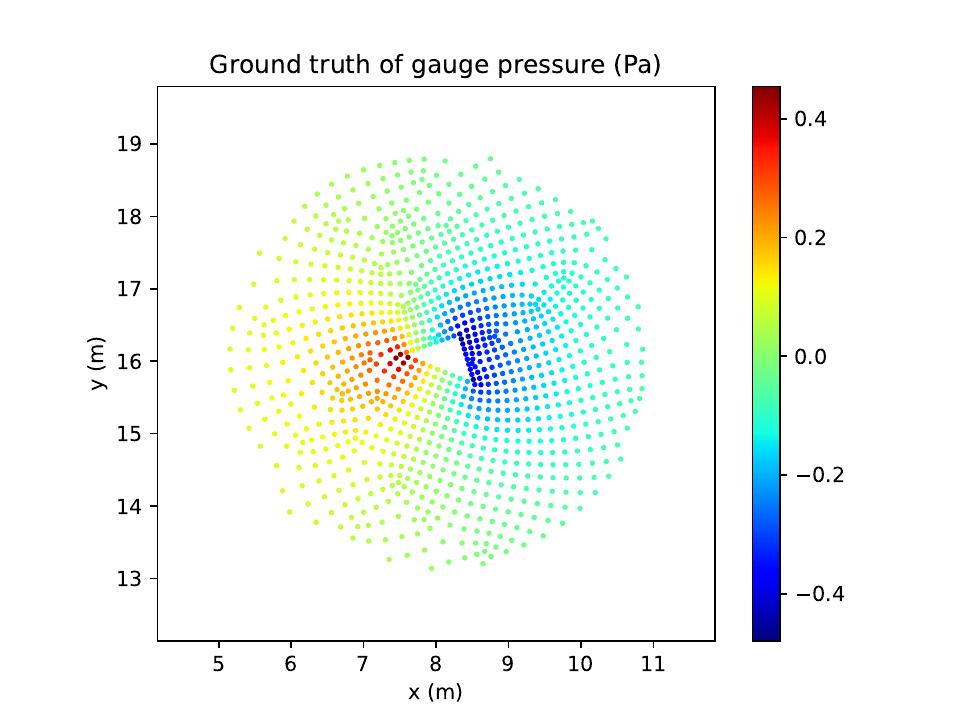}
    \end{subfigure}

    
    \begin{subfigure}[b]{0.32\textwidth}
        \centering
        \includegraphics[width=\textwidth]{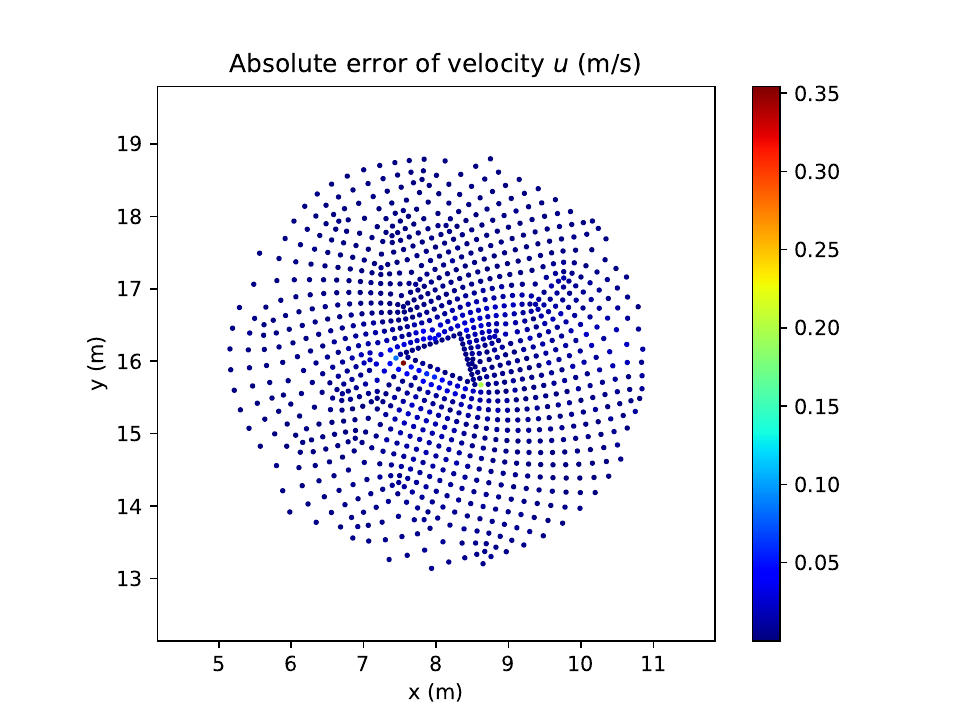}
    \end{subfigure}
    \begin{subfigure}[b]{0.32\textwidth}
        \centering
        \includegraphics[width=\textwidth]{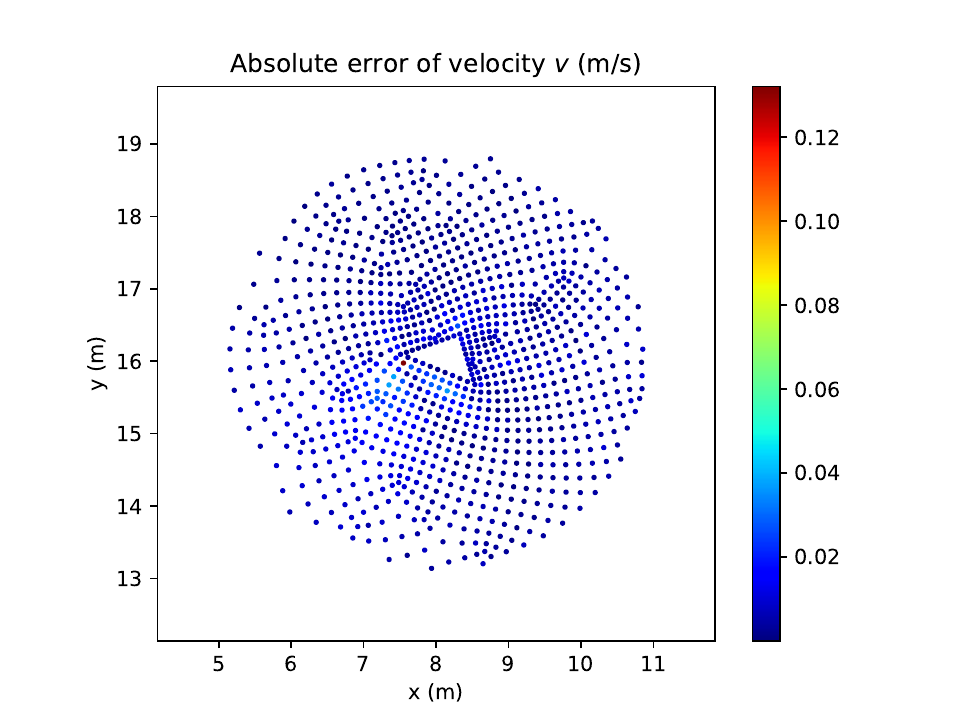}
    \end{subfigure}
    \begin{subfigure}[b]{0.32\textwidth}
        \centering
        \includegraphics[width=\textwidth]{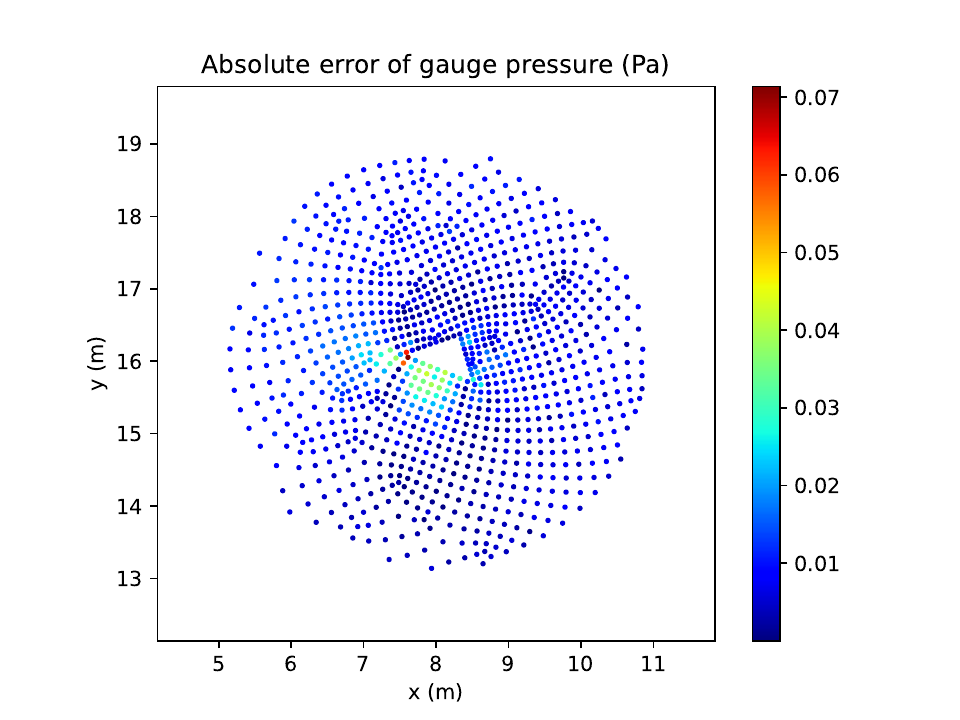}
    \end{subfigure}

    
    \begin{subfigure}[b]{0.32\textwidth}
        \centering
        \includegraphics[width=\textwidth]{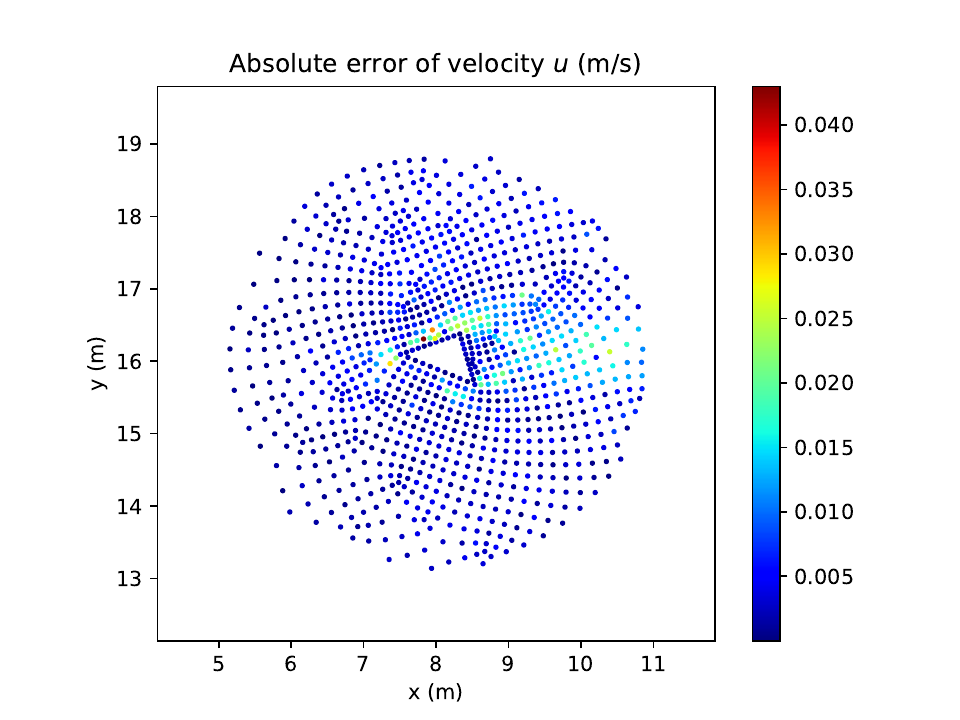}
    \end{subfigure}
    \begin{subfigure}[b]{0.32\textwidth}
        \centering
        \includegraphics[width=\textwidth]{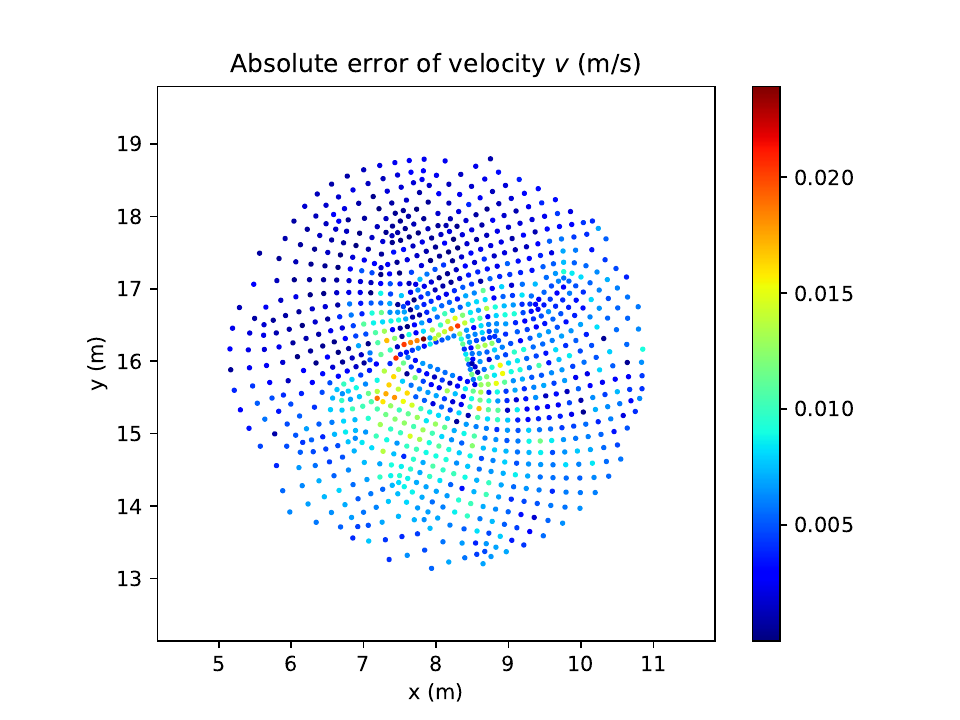}
    \end{subfigure}
    \begin{subfigure}[b]{0.32\textwidth}
        \centering
        \includegraphics[width=\textwidth]{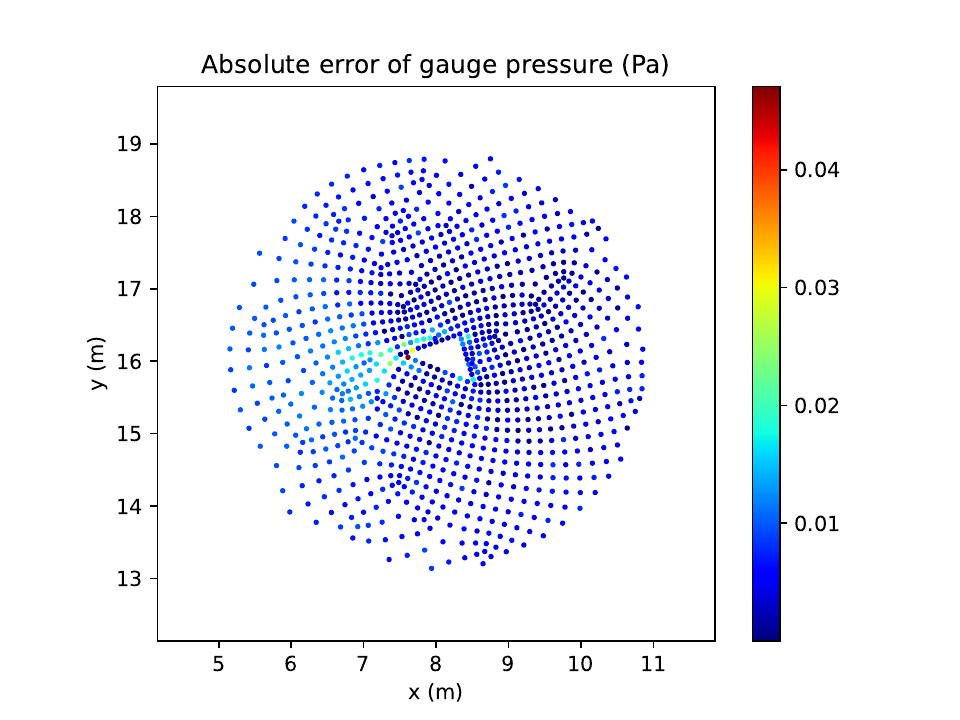}
    \end{subfigure}

     \begin{subfigure}[b]{0.32\textwidth}
        \centering
        \includegraphics[width=\textwidth]{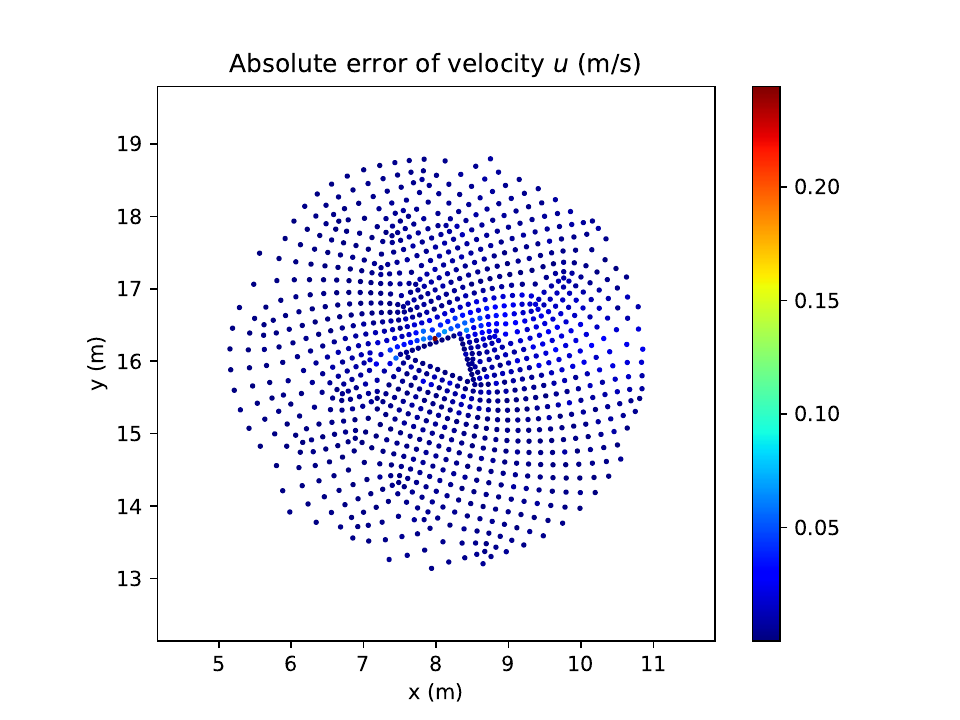}
    \end{subfigure}
    \begin{subfigure}[b]{0.32\textwidth}
        \centering
        \includegraphics[width=\textwidth]{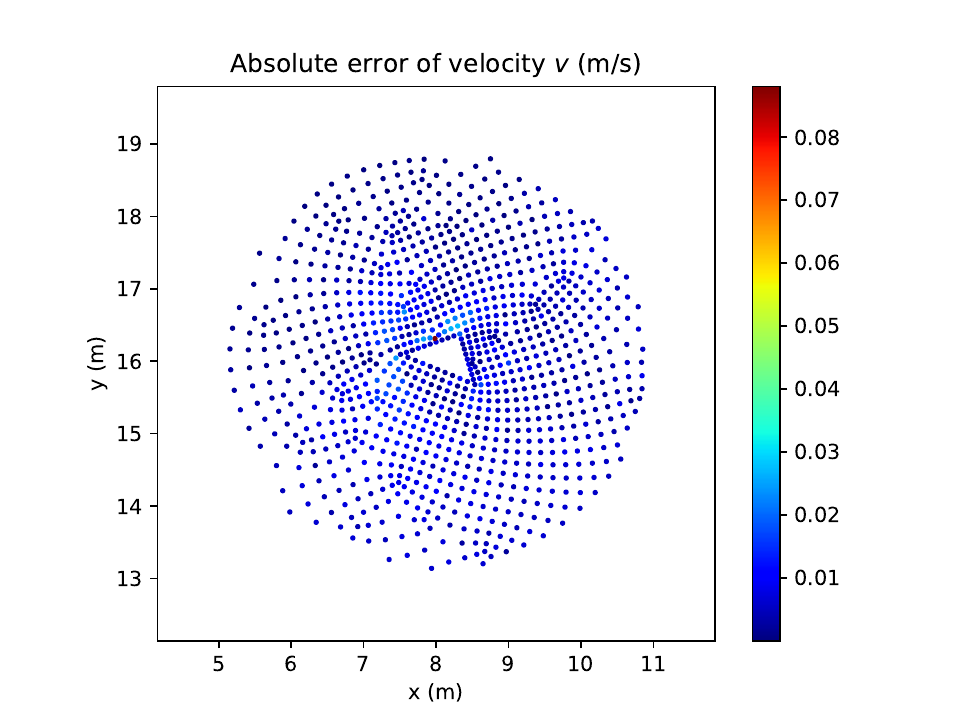}
    \end{subfigure}
    \begin{subfigure}[b]{0.32\textwidth}
        \centering
        \includegraphics[width=\textwidth]{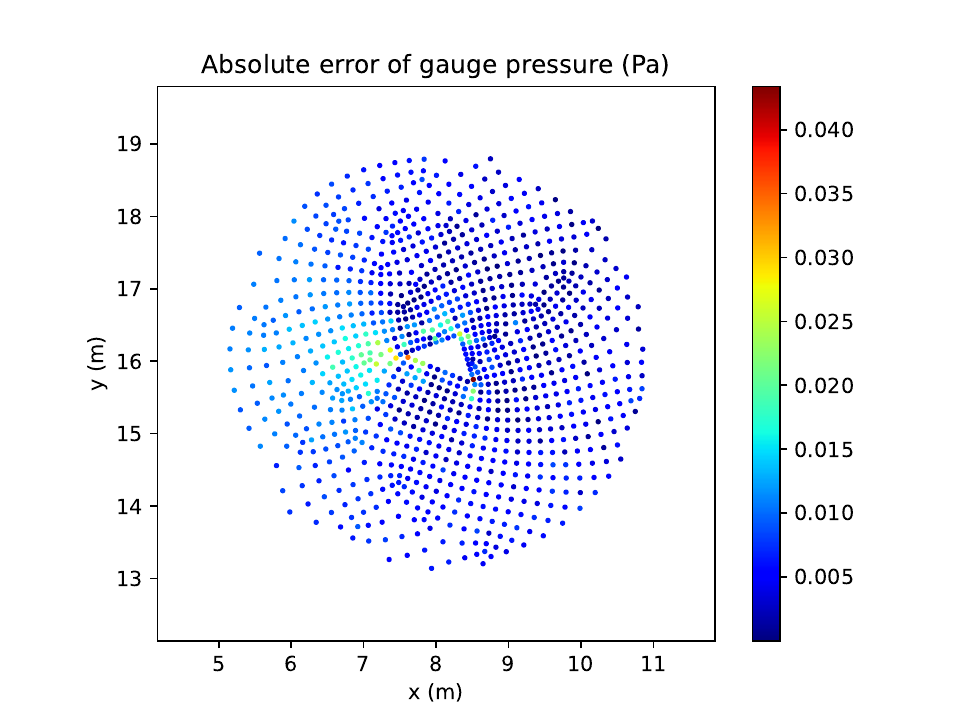}
    \end{subfigure}
      

    \begin{subfigure}[b]{0.32\textwidth}
        \centering
        \includegraphics[width=\textwidth]{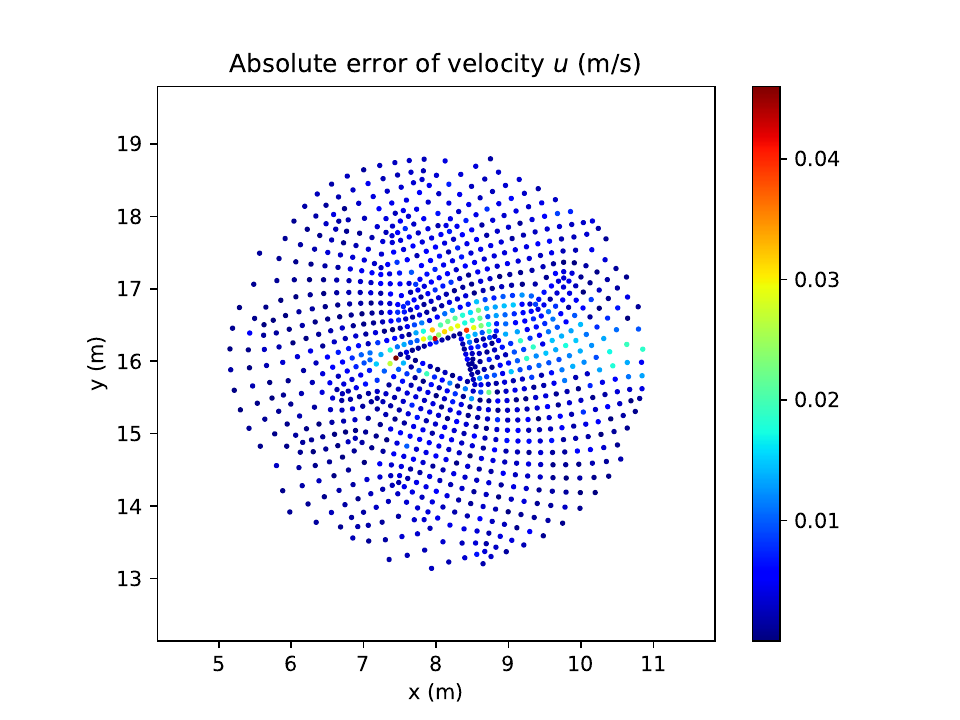}
    \end{subfigure}
    \begin{subfigure}[b]{0.32\textwidth}
        \centering
        \includegraphics[width=\textwidth]{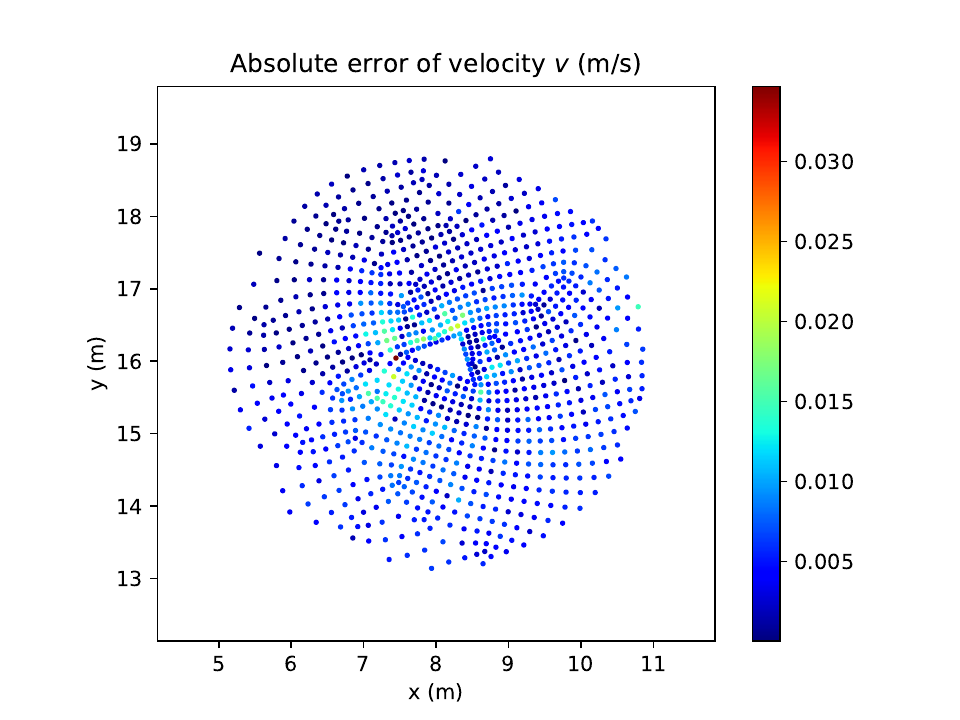}
    \end{subfigure}
    \begin{subfigure}[b]{0.32\textwidth}
        \centering
        \includegraphics[width=\textwidth]{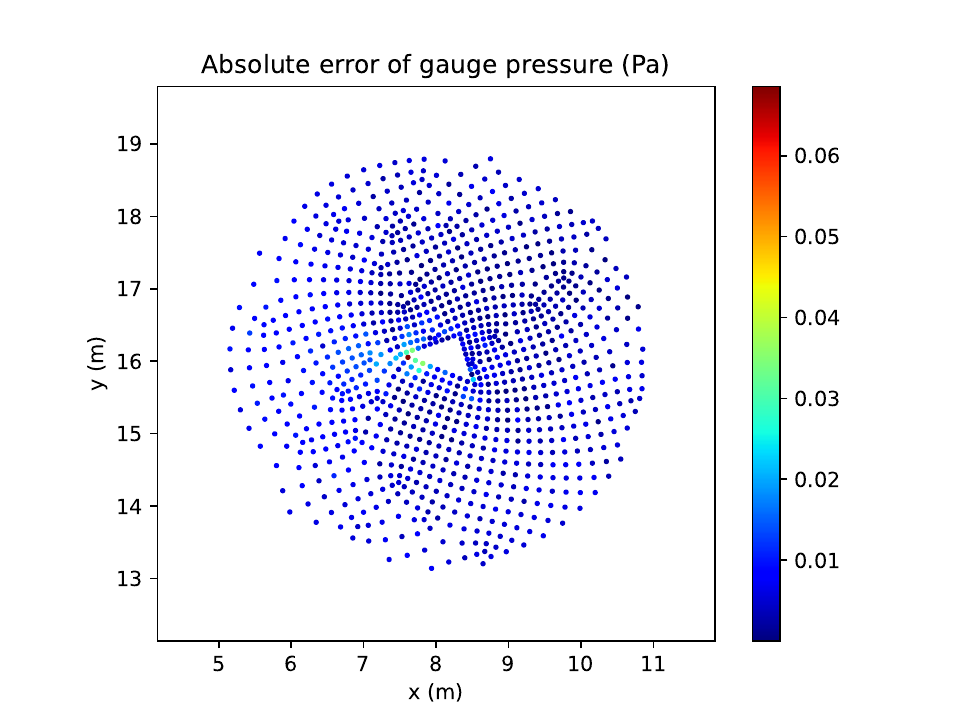}
    \end{subfigure}
    
    
    \begin{subfigure}[b]{0.32\textwidth}
    \centering
        \includegraphics[width=\textwidth]{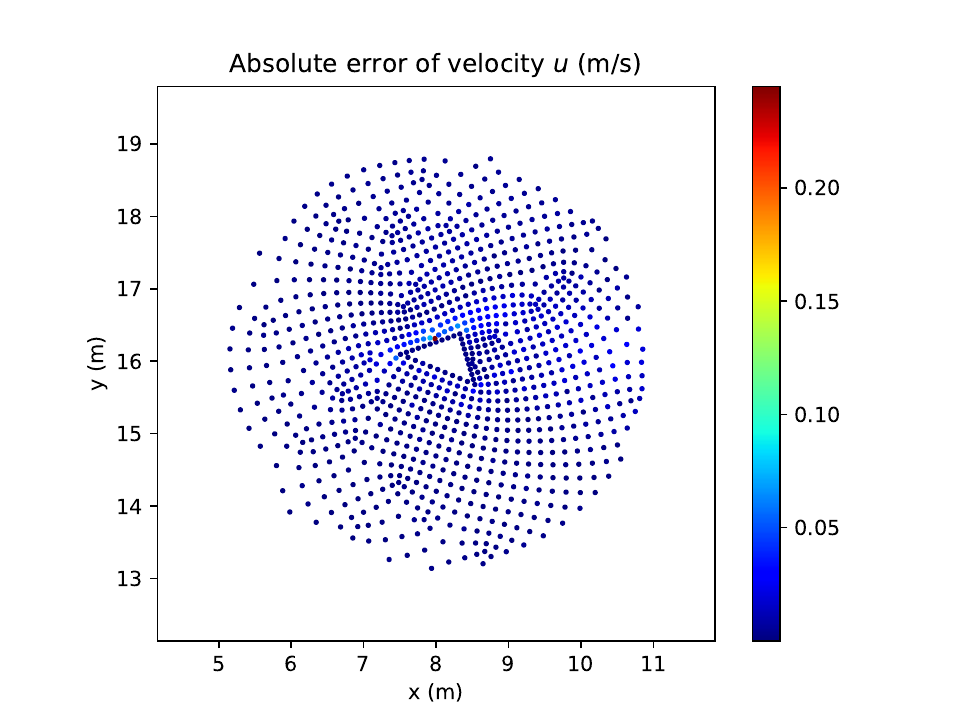}
    \end{subfigure}
    \begin{subfigure}[b]{0.32\textwidth}
        \centering
        \includegraphics[width=\textwidth]{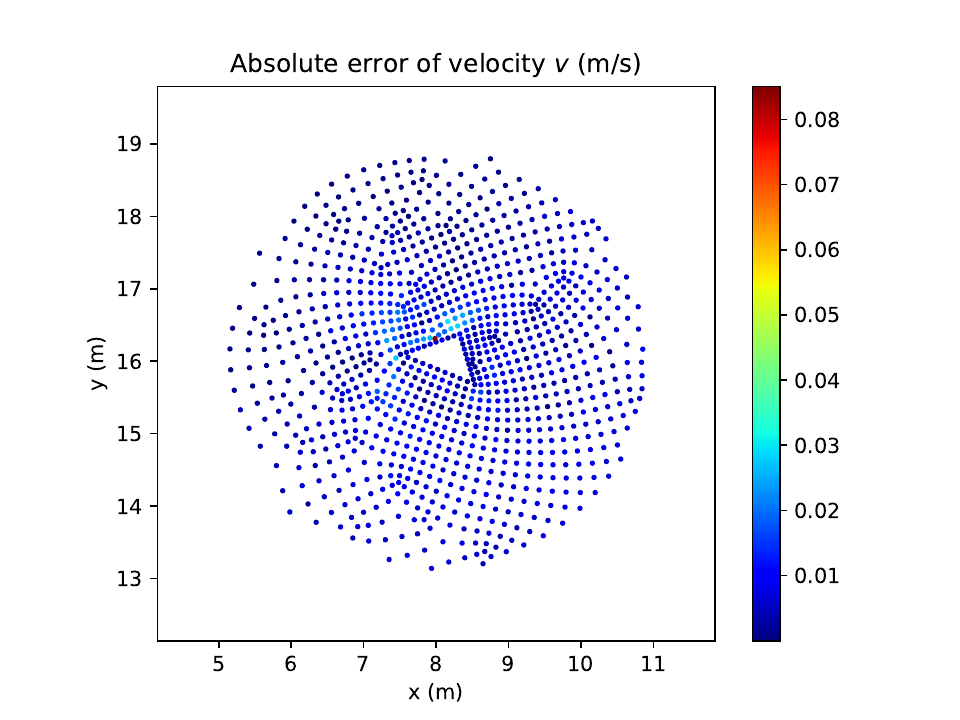}
    \end{subfigure}
    \begin{subfigure}[b]{0.32\textwidth}
        \centering
        \includegraphics[width=\textwidth]{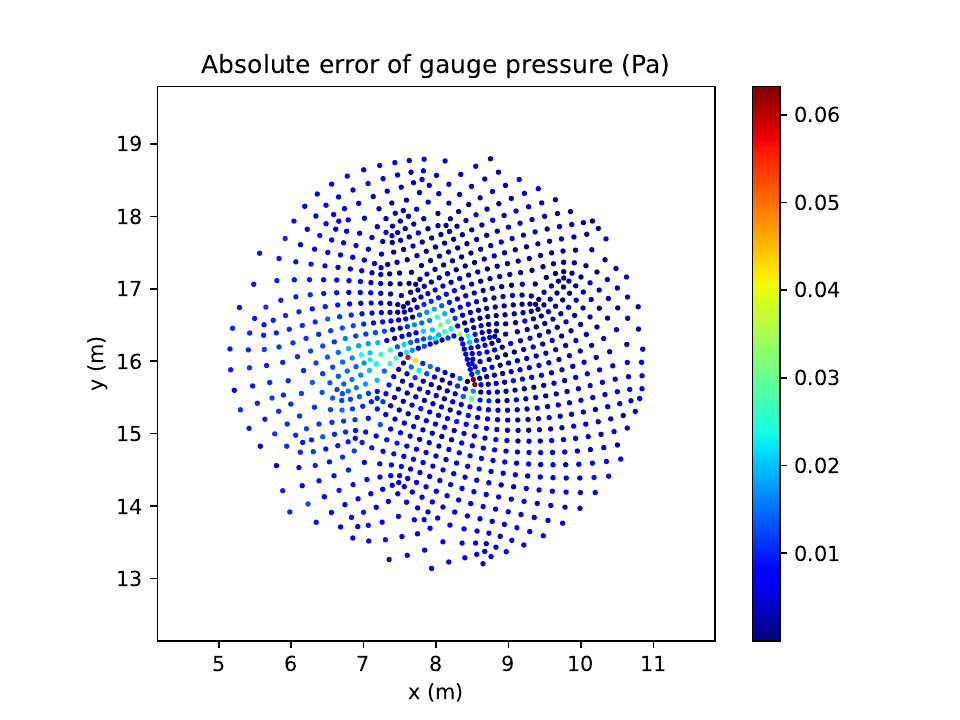}
    \end{subfigure}
    
    \caption{Diffusion PointNet realizations; the first row shows the ground truth, and the remaining rows show the absolute pointwise error for five different realizations.}
  \label{Fig20B}
\end{figure}


\subsection{Sampling diversity}
\label{Sect4ASampling}

As discussed in Sect. \ref{Sect3}, when standard Gaussian noise conditioned on a given geometry (point cloud) is provided as input to the Flow Matching PointNet and Diffusion PointNet models, they generate distinct realizations of the velocity and pressure fields. Here, we compare several of these generated samples. Figure \ref{Fig20A} presents five samples generated by the Flow Matching PointNet model. To better illustrate the differences among the generated samples, instead of visualizing the velocity and pressure fields directly, we show the absolute pointwise error along with the corresponding ground truth. As can be seen from Fig. \ref{Fig20A}, it is evident that each sample differs from the others, even though the regions with the highest errors appear in roughly the same locations. However, the maximum absolute error for the $y$ component of the velocity vector ($v$) and the pressure field ($p$) still varies noticeably across samples. For instance, in the case of the pressure field, the maximum absolute error occurs near the cylinder surface, with approximate values of 0.035, 0.04, 0.05, 0.06, and 0.07 Pa, as visualized in Fig. \ref{Fig20A}. In Sect. \ref{Sect42}, we demonstrate that variations in these pressure values ultimately influence the predicted lift and drag forces induced by the pressure field. By sampling within the frameworks of Flow Matching PointNet and Diffusion PointNet, we estimate the maximum potential prediction error and assess model reliability. Consequently, when these models are applied in engineering design, such uncertainty estimates inform the selection of appropriate safety factors.

Similarly, Figure \ref{Fig20B} visualizes five samples generated by the Diffusion PointNet model. Differences among these samples are evident, particularly in the maximum absolute errors of both the velocity and pressure fields. For example, examining the generated $u$-velocity reveals maximum errors of approximately $0.04$, $0.20$, $0.25$, and $0.35$ m/s. Notably, in the second row of Fig. \ref{Fig20B}, or equivalently the sixth row of the same figure, the maximum absolute error of 0.35 occurs at only a single point among all 1024 points in the point cloud, while the remaining points exhibit near-zero absolute error. From a computer graphics perspective, this phenomenon resembles labeling inconsistencies encountered in part‐segmentation tasks when using PointNet \cite{qi2017pointnet,kashefi2025pointnetKAN}. In such cases, more advanced models such as PointNet++ \cite{qi2017pointnet++}, DGCNN \cite{wang2019dynamic}, and PointMLP \cite{ma2022rethinking} have been proposed and could potentially be employed in future studies for generative modeling in computational physics. These models place greater emphasis on capturing local features, thereby reducing the occurrence of outlier-like labels. Hence, these observations provide motivation for exploring more sophisticated point cloud-based architectures for generative modeling.


\begin{table}[t]
\centering
\caption{Comparison between the performance of the Flow Matching PointNet, Diffusion PointNet, and baseline PointNet models for predicting pressure drag and pressure lift on a test set containing 222 unseen geometries. Lift and drag are computed from the predicted pressure fields. Number of samples is equal to 100. In numbers reported in the form $\alpha_1 \pm \alpha_2$, $\alpha_1$ and $\alpha_2$ show the mean error and the corresponding standard deviation, respectively.}\label{Table8}
\begin{tabular}{llll}
\toprule
 & Flow Matching PointNet & Diffusion PointNet  & Baseline PointNet \\
\midrule
Average absolute error of drag & 1.65256E$-$2 $\pm$ 4.70271E$-$4 &8.82595E$-$3 $\pm$ 3.33781E$-$4 & 1.04599E$-$1 \\
Maximum absolute error of drag & 1.32432E$-$1 $\pm$ 2.13175E$-$2 &1.34004E$-$1 $\pm$ 9.27641E$-$3 & 3.48611E$-$1  \\
Minimum absolute error of drag & 1.40302E$-$4 $\pm$ 1.30478E$-$4 & 4.42719E$-$5 $\pm$ 4.15782E$-$5&  1.06129E$-$2\\
\midrule
Average absolute error of lift & 1.28349E$-$2 $\pm$ 4.41792E$-$4 & 1.06831E$-$2 $\pm$ 3.10677E$-$4&  4.77564E$-$2\\
Maximum absolute error of lift & 6.84403E$-$2 $\pm$ 1.65813E$-$2 &  1.17770E$-$1 $\pm$ 1.18856E$-$2&  1.90848E$-$1\\
Minimum absolute error of lift & 8.33265E$-$5 $\pm$ 9.27361E$-$5 &6.78547E$-$5 $\pm$ 5.77296E$-$5 &  1.75967E$-$4\\
\midrule
Number of trainable & 3554179 & 3554179 & 3554179 \\
parameters &  &  \\
\bottomrule
\end{tabular}
\end{table}


\begin{figure}[htp]
  \centering 
      \begin{subfigure}[b]{0.32\textwidth}
        \centering
        \includegraphics[width=\textwidth]{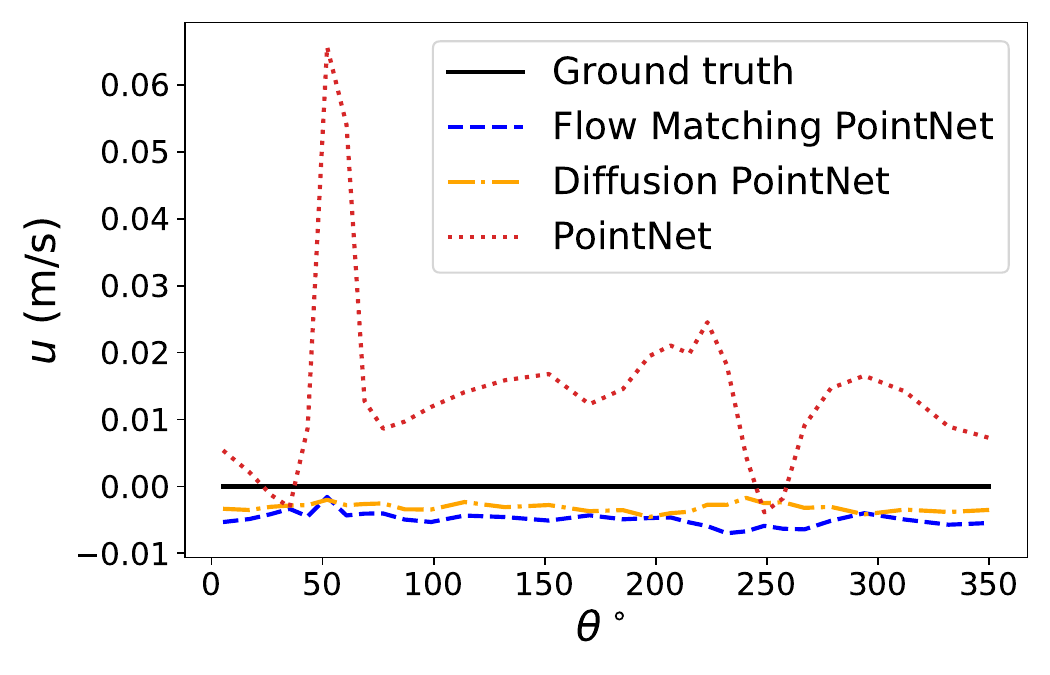}
    \end{subfigure}
    \begin{subfigure}[b]{0.32\textwidth}
        \centering
        \includegraphics[width=\textwidth]{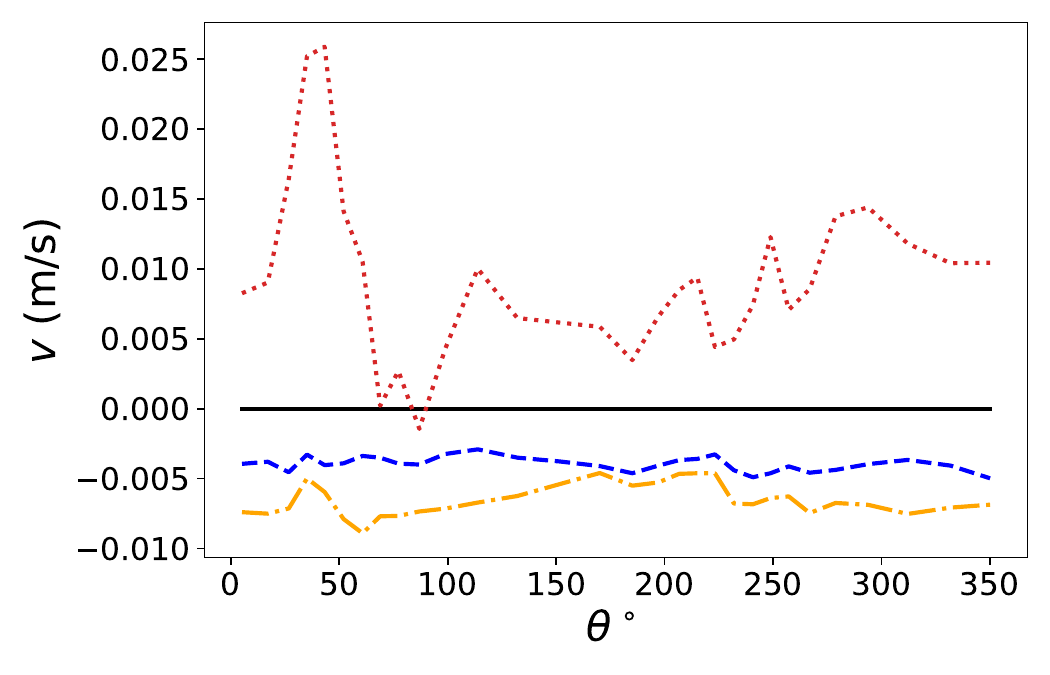}
    \end{subfigure}
    \begin{subfigure}[b]{0.32\textwidth}
        \centering
        \includegraphics[width=\textwidth]{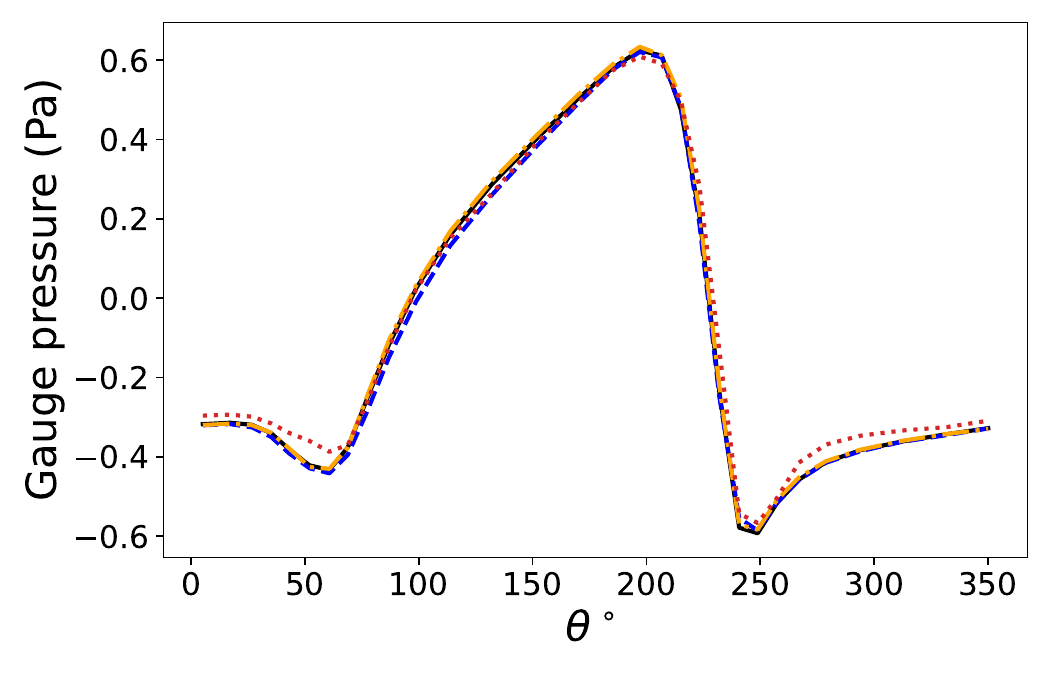}
    \end{subfigure}

  \caption{An example of velocity and pressure distributions predicted by the Flow Matching PointNet, Diffusion PointNet, and baseline PointNet models along the surface of a cylinder with an elliptical cross-section from the test set. The angle is measured counterclockwise from the positive $x$-axis by connecting each surface point to the center of the ellipse at (8, 16).}
  \label{FigVP}
\end{figure}


\subsection{Computation of pressure drag and lift forces}
\label{Sect42}

In addition to performing an error analysis over the entire computational domain, we now focus specifically on evaluating the predictions of the velocity and pressure fields along the surface of the object. A comparative analysis is conducted to determine which of the three models, Flow Matching PointNet, Diffusion PointNet, and baseline PointNet, achieves superior performance. In particular, Figure \ref{FigVP} presents the velocity and pressure profiles on the surface of one test object, which corresponds to a cylinder with an elliptical cross-section. As illustrated in Fig. \ref{FigVP}, since the flow is viscous, the components of the velocity vector (i.e., $u$ and $v$) must vanish at the surface. It is evident from Fig. \ref{FigVP} that the predictions obtained by Flow Matching PointNet and Diffusion PointNet are more accurate than those of baseline PointNet for both the velocity and pressure fields. When comparing Flow Matching PointNet and Diffusion PointNet, both models demonstrate excellent performance; however, the results shown in Fig. \ref{FigVP} indicate that Flow Matching PointNet provides slightly more accurate predictions of the zero-velocity condition in the $y$-direction compared to Diffusion PointNet.

Following the comparison of velocity and pressure distributions on the cylinder surface, performed on one of the test set, we now turn our attention to the prediction of drag and lift forces derived from the pressure field. Accurate estimation of drag and lift forces has always been of great importance in engineering and plays a critical role in aerodynamic applications and geometric design processes. Table \ref{Table8} summarizes a quantitative comparison of the three models of Flow Matching PointNet, Diffusion PointNet, and baseline PointNet. According to Table \ref{Table8}, an examination of the average absolute error in drag reveals that the baseline PointNet model exhibits an error of approximately 10\%, whereas the Flow Matching PointNet and Diffusion PointNet achieve substantially lower errors of below 1.7\% and 0.9\%, respectively. These results indicate that both Flow Matching PointNet and Diffusion PointNet considerably outperform the baseline PointNet model, yielding more reliable predictions for drag-based design tasks. For lift prediction, a similar trend is observed. The Diffusion PointNet achieves the highest accuracy, followed by the Flow Matching PointNet and the baseline PointNet. The Flow Matching PointNet also provides more accurate predictions than the baseline model. In this case, the prediction error of the baseline PointNet for lift is approximately 4.8\%, which is lower than its corresponding drag prediction error.

Finally, as discussed in Sect. \ref{Sect41}, 100 samples are drawn by injecting standard Gaussian noise for each geometry from the test set for both Flow Matching PointNet and Diffusion PointNet. Accordingly, the values reported in Table \ref{Table8} in the form of $\alpha_1 \pm \alpha_2$ indicate $\alpha_1$ as the average value and $\alpha_2$ as the standard deviation. This representation enables the quantification of uncertainty in the predictions, which is particularly valuable for design-oriented applications. For instance, the average over the maximum absolute error of drag for Flow Matching PointNet is approximately $13.25\%$, accompanied by a standard deviation of $2.14\%$, as listed in Table \ref{Table8}. Similar information for the maximum absolute error of lift using Diffusion PointNet is provided in Table \ref{Table8}. In contrast, the baseline PointNet model does not provide such uncertainty intervals for error measurements, which represents a key limitation of the model. In summary, both Flow Matching PointNet and Diffusion PointNet not only deliver more accurate predictions but also offer valuable uncertainty information about the model outputs that can assist design engineers, especially when safety margins or worst-case design scenarios must be considered.

\begin{table}[t]
\centering
\caption{Robustness test of Flow Matching PointNet, Diffusion PointNet, and baseline PointNet by evaluating the error in predicted velocity and pressure fields when a certain percentage of points is randomly removed from point clouds in the test set, containing 222 unseen geometries. The $L^2$ norm is indicated by $||\cdots||$.}\label{Table9}
\begin{tabular}{lllll}
\toprule
Model & Percentage & 5\%  & 10\%  & 15\% \\
\midrule
& Average $||\Tilde{u}-u||/||u||$ & 1.46687E$-$2 & 1.81087E$-$2 &  2.27024E$-$2 \\
Flow Matching PointNet& Average $||\Tilde{v}-v||/||v||$ &  5.25402E$-$2 & 6.84943E$-$2 & 8.20785E$-$2 \\
& Average $||\Tilde{p}-p||/||p||$ & 4.27335E$-$2 & 5.48017E$-$2 & 6.71761E$-$2 \\
\midrule
& Average $||\Tilde{u}-u||/||u||$ & 1.73480E$-$2 & 2.15721E$-$2 & 2.99512E$-$2 \\
Diffusion PointNet& Average $||\Tilde{v}-v||/||v||$ & 7.35410E$-$2 & 8.53514E$-$2 & 1.09470E$-$1 \\
& Average $||\Tilde{p}-p||/||p||$ & 4.15597E$-$2 & 5.14183E$-$2 & 6.95332E$-$2 \\
\midrule
& Average $||\Tilde{u}-u||/||u||$ & 3.32840E$-$2 & 3.59088E$-$2& 4.23352E$-$2 \\
Baseline PointNet& Average $||\Tilde{v}-v||/||v||$ & 1.39502E$-$1& 1.48369E$-$1 & 1.71268E$-$1\\
& Average $||\Tilde{p}-p||/||p||$ & 1.10115E$-$1 & 1.14691E$-$1 & 1.23513E$-$1 \\
\bottomrule
\end{tabular}
\end{table}

\begin{figure}[h]
  \centering 
      \begin{subfigure}[b]{0.32\textwidth}
        \centering
        \includegraphics[width=\textwidth]{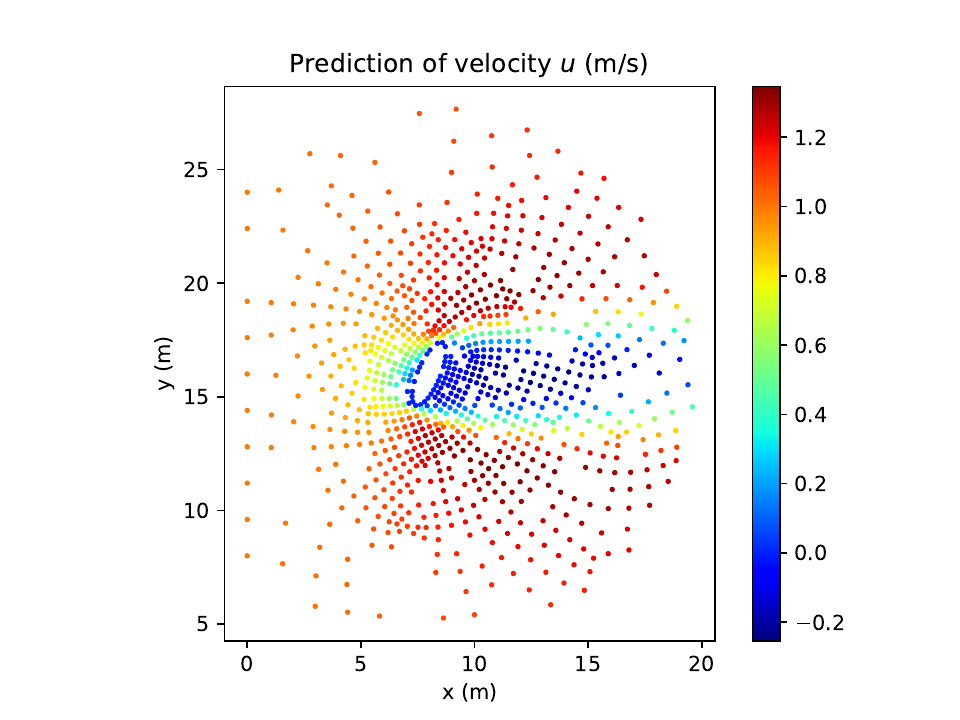}
    \end{subfigure}
    \begin{subfigure}[b]{0.32\textwidth}
        \centering
        \includegraphics[width=\textwidth]{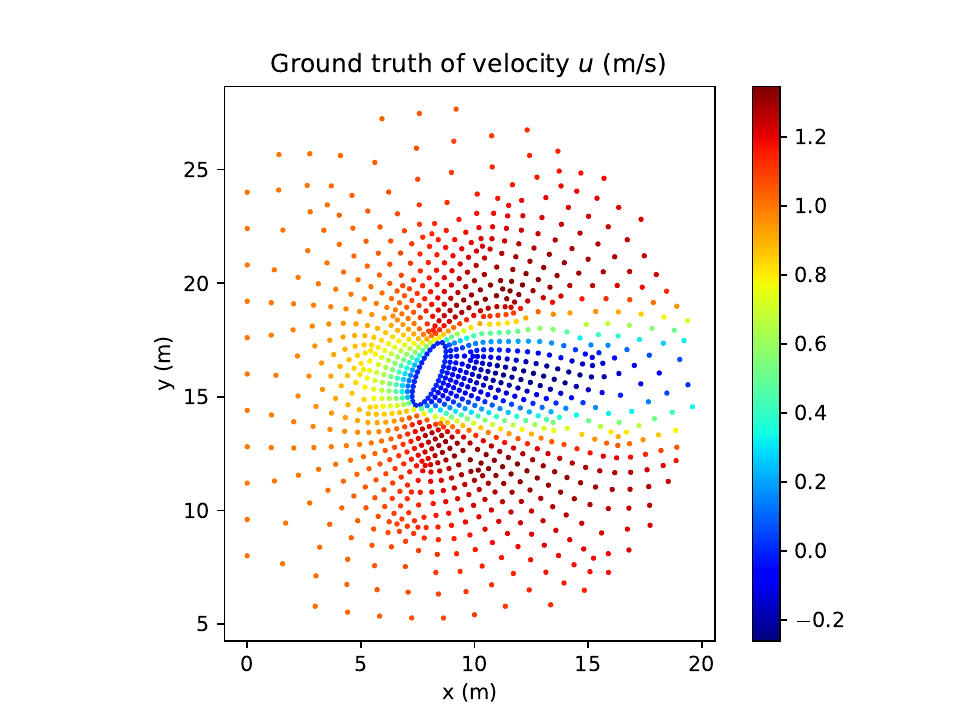}
    \end{subfigure}
    \begin{subfigure}[b]{0.32\textwidth}
        \centering
        \includegraphics[width=\textwidth]{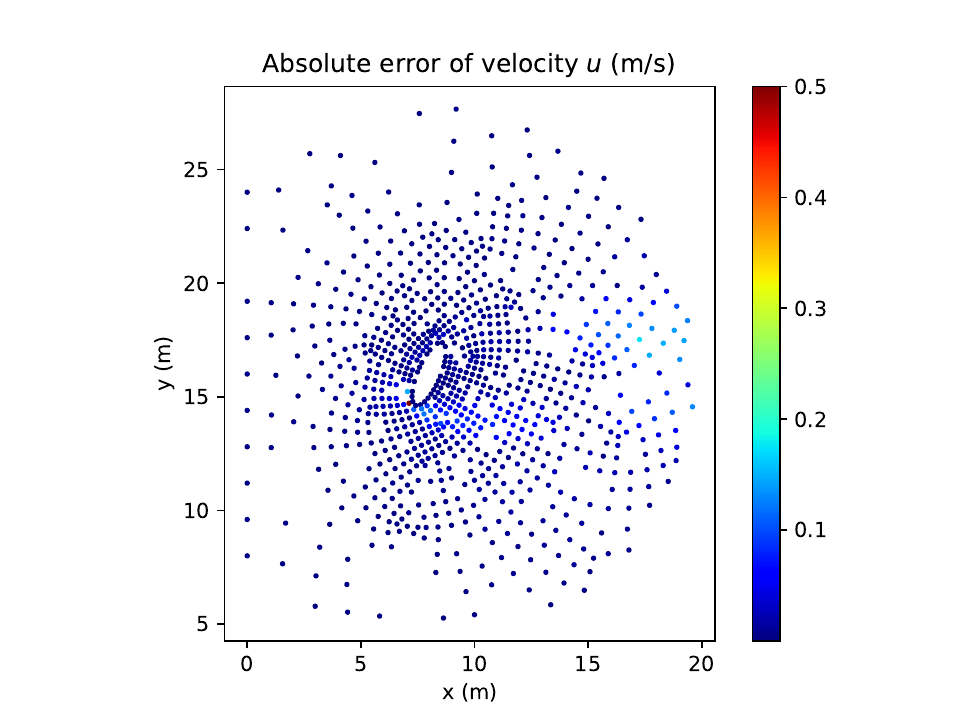}
    \end{subfigure}

    
    \begin{subfigure}[b]{0.32\textwidth}
        \centering
        \includegraphics[width=\textwidth]{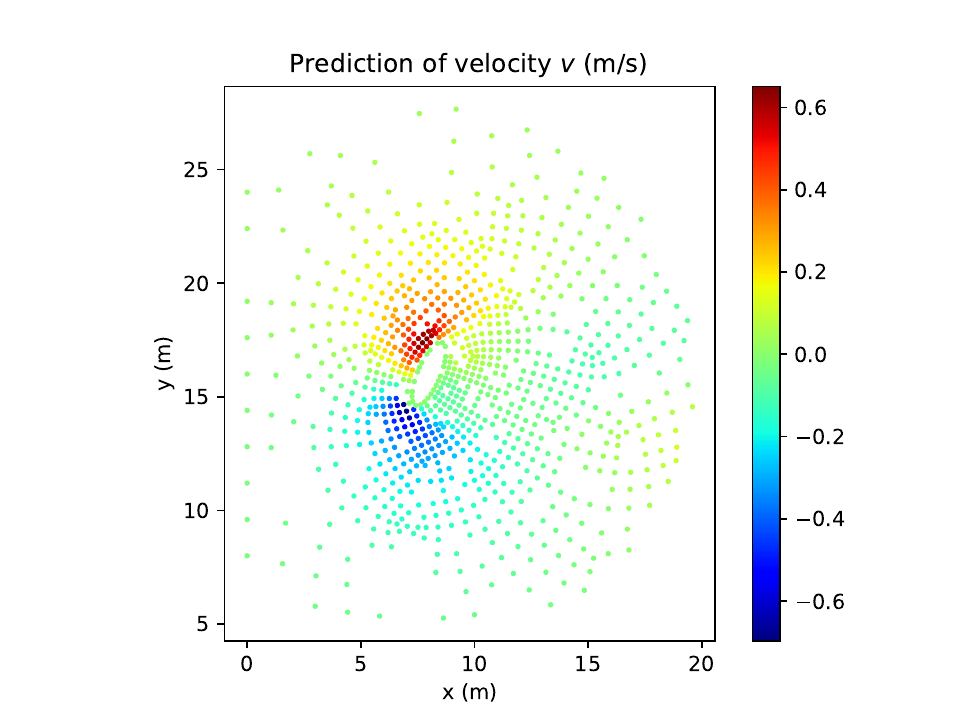}
    \end{subfigure}
    \begin{subfigure}[b]{0.32\textwidth}
        \centering
        \includegraphics[width=\textwidth]{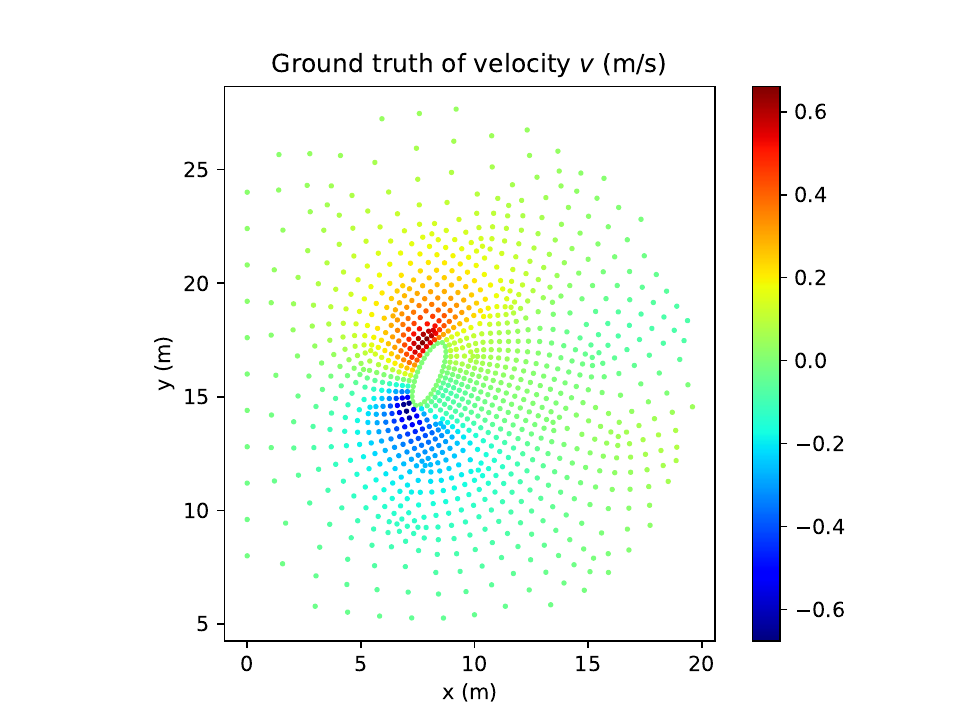}
    \end{subfigure}
    \begin{subfigure}[b]{0.32\textwidth}
        \centering
        \includegraphics[width=\textwidth]{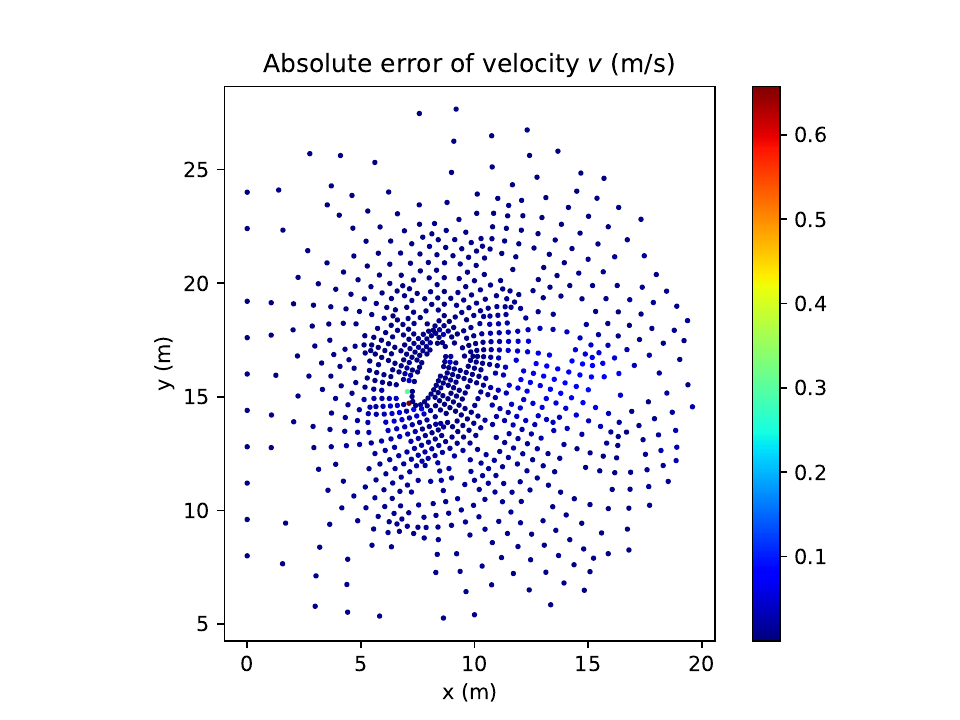}
    \end{subfigure}

    
    \begin{subfigure}[b]{0.32\textwidth}
        \centering
        \includegraphics[width=\textwidth]{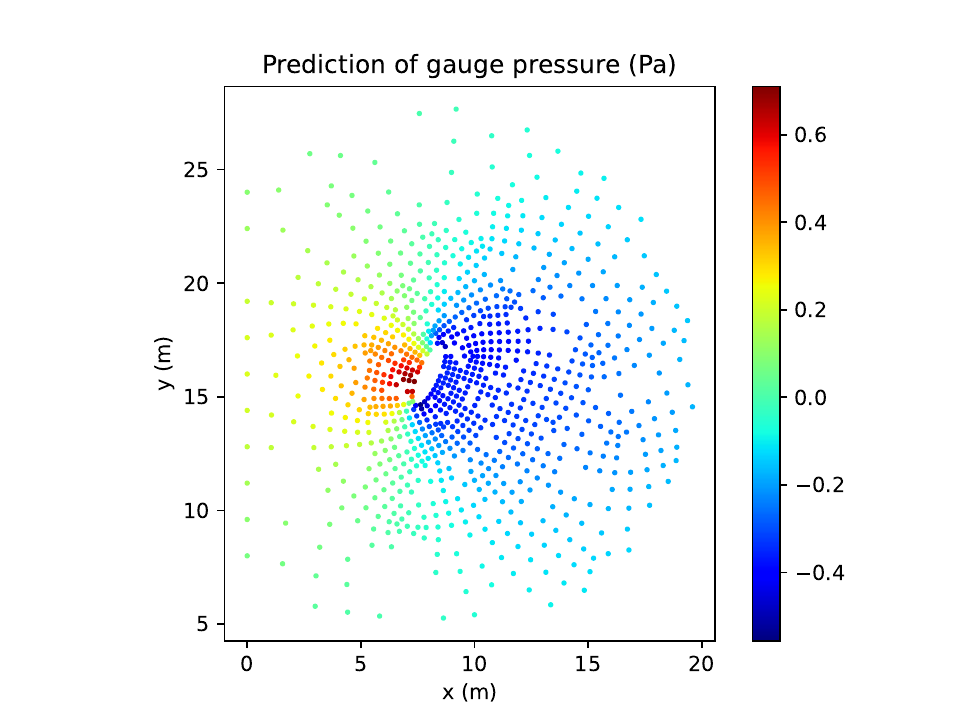}
    \end{subfigure}
    \begin{subfigure}[b]{0.32\textwidth}
        \centering
        \includegraphics[width=\textwidth]{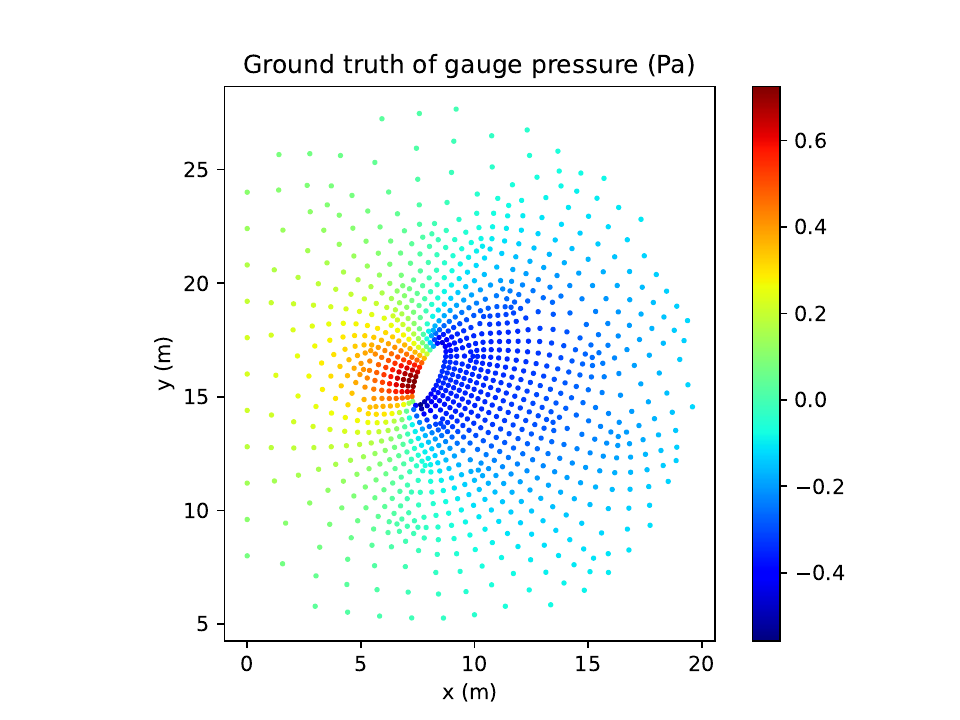}
    \end{subfigure}
    \begin{subfigure}[b]{0.32\textwidth}
        \centering
        \includegraphics[width=\textwidth]{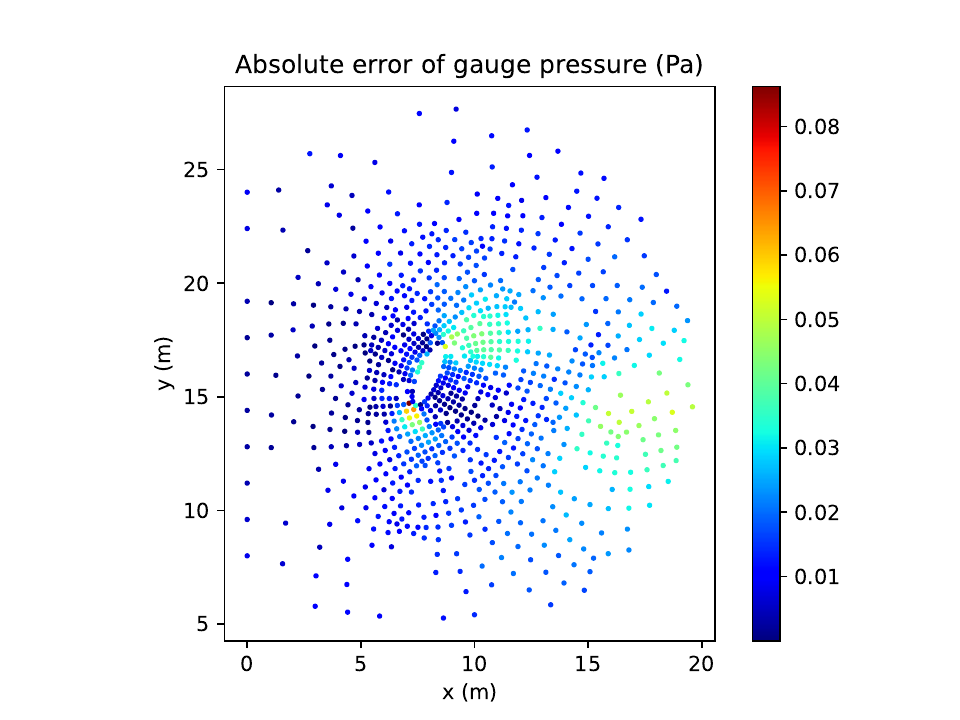}
    \end{subfigure}

  \caption{An example comparing the ground truth to the predictions of Flow Matching PointNet for the velocity and pressure fields from the test set, where 15\% of points are randomly removed from the point clouds, and an imperfect geometry of the test set is fed into the network. Note that the ground truth is shown with all 1024 points.}
  \label{Fig900}
\end{figure}


\begin{figure}[t]
  \centering 
      \begin{subfigure}[b]{0.32\textwidth}
        \centering
        \includegraphics[width=\textwidth]{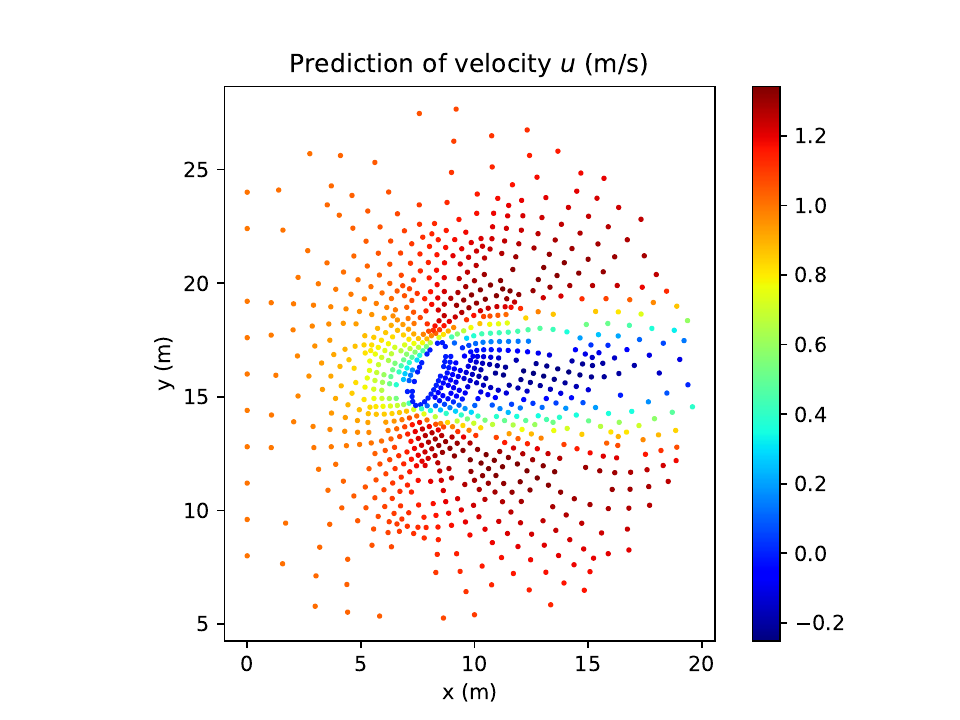}
    \end{subfigure}
    \begin{subfigure}[b]{0.32\textwidth}
        \centering
        \includegraphics[width=\textwidth]{u_truth_test29.pdf}
    \end{subfigure}
    \begin{subfigure}[b]{0.32\textwidth}
        \centering
        \includegraphics[width=\textwidth]{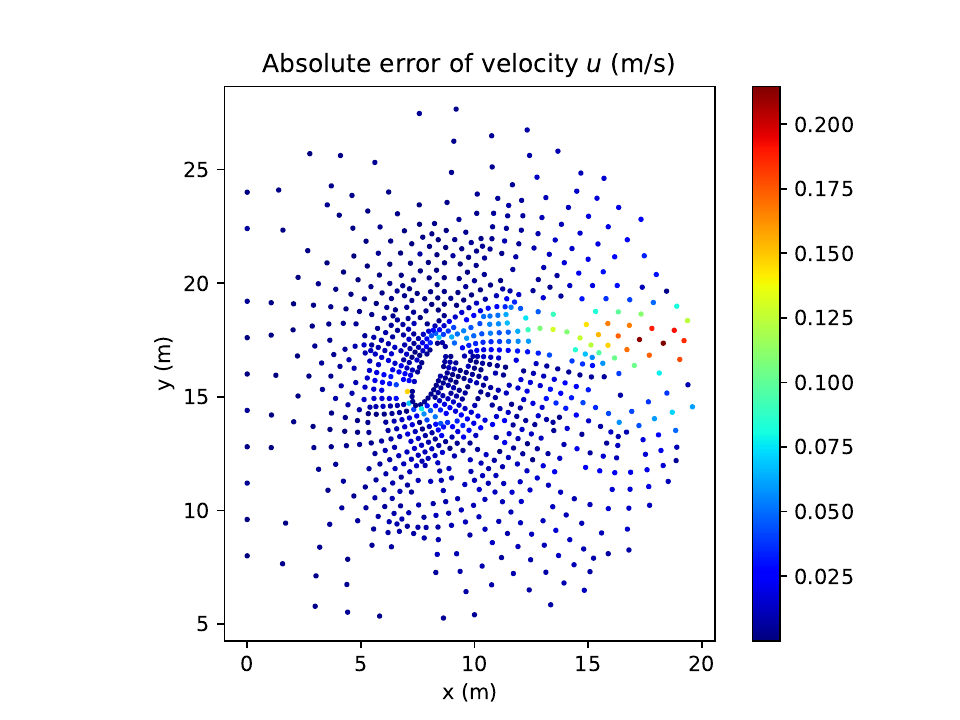}
    \end{subfigure}

    
    \begin{subfigure}[b]{0.32\textwidth}
        \centering
        \includegraphics[width=\textwidth]{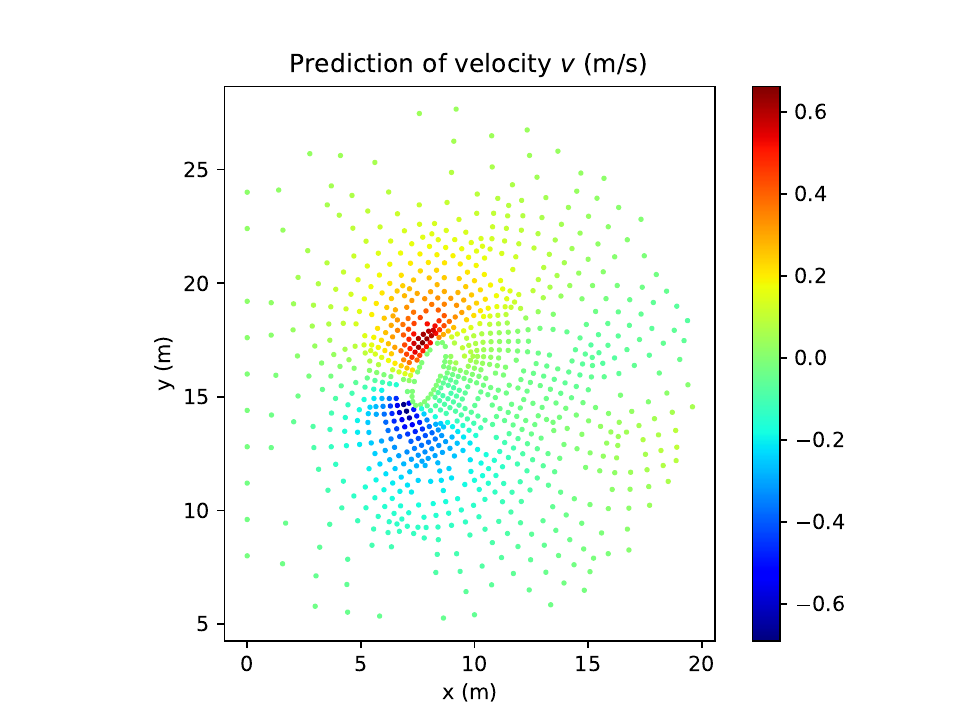}
    \end{subfigure}
    \begin{subfigure}[b]{0.32\textwidth}
        \centering
        \includegraphics[width=\textwidth]{v_truth_test29.pdf}
    \end{subfigure}
    \begin{subfigure}[b]{0.32\textwidth}
        \centering
        \includegraphics[width=\textwidth]{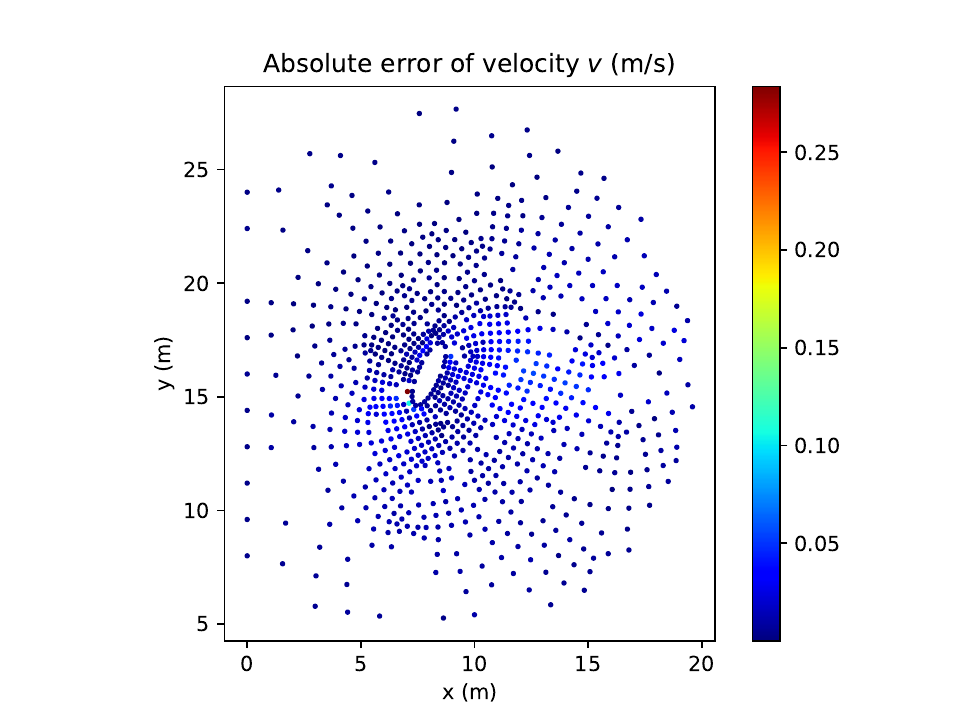}
    \end{subfigure}

    
    \begin{subfigure}[b]{0.32\textwidth}
        \centering
        \includegraphics[width=\textwidth]{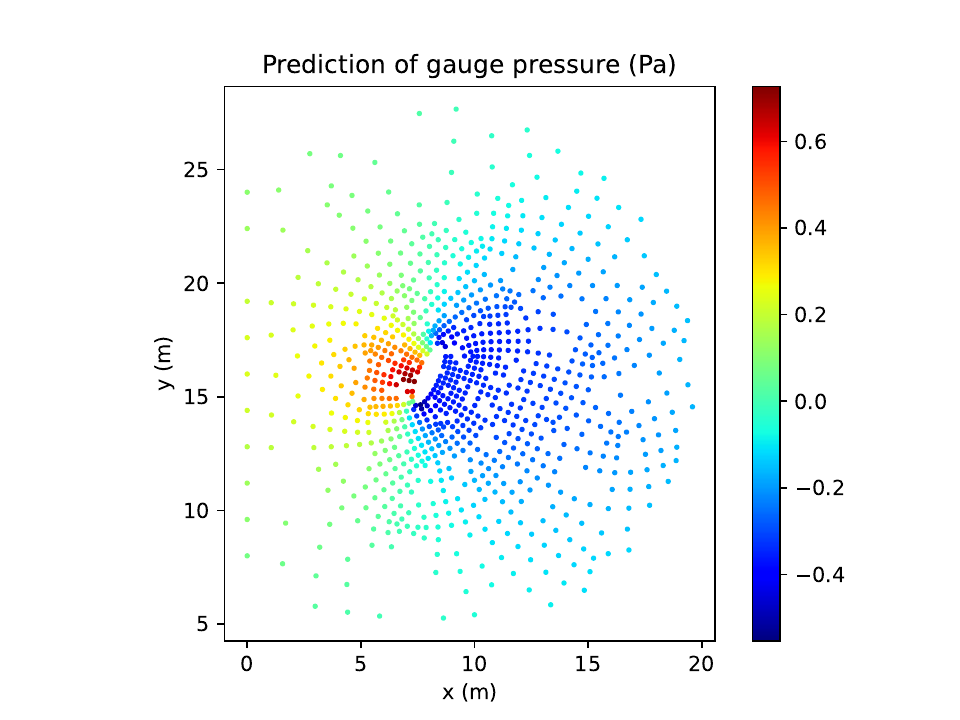}
    \end{subfigure}
    \begin{subfigure}[b]{0.32\textwidth}
        \centering
        \includegraphics[width=\textwidth]{p_truth_test29.pdf}
    \end{subfigure}
    \begin{subfigure}[b]{0.32\textwidth}
        \centering
        \includegraphics[width=\textwidth]{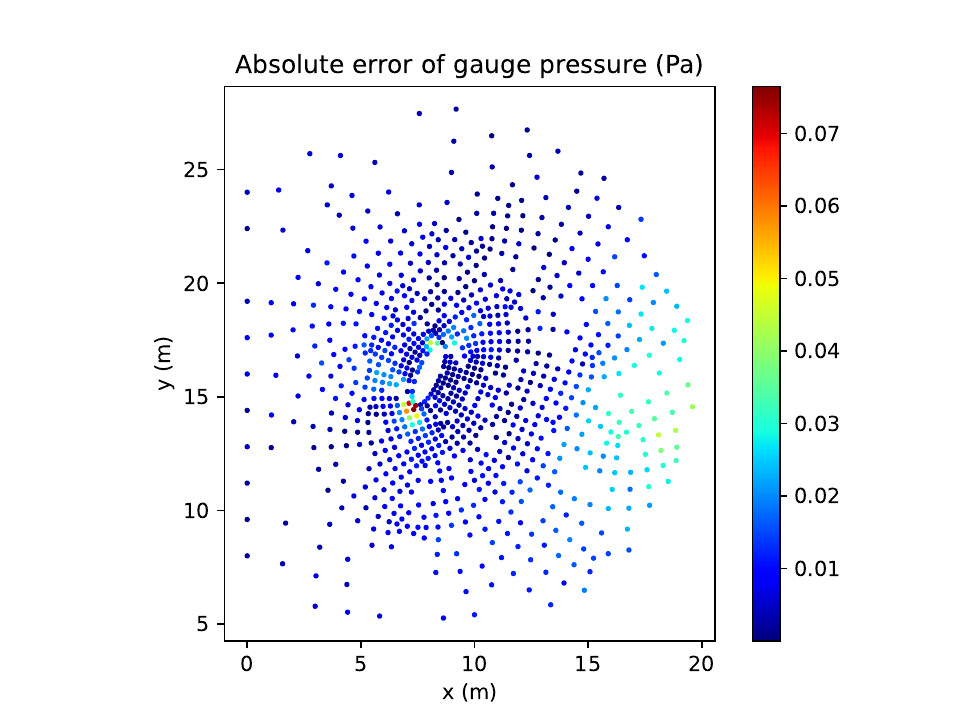}
    \end{subfigure}

  \caption{An example comparing the ground truth to the predictions of Diffusion PointNet for the velocity and pressure fields from the test set, where 15\% of points are randomly removed from the point clouds, and an imperfect geometry of the test set is fed into the network. Note that the ground truth is shown with all 1024 points.}
  \label{Fig100}
\end{figure}


\subsection{Robustness}
\label{Sect43}

Another parameter we aim to investigate is the case in which the target geometry is partially corrupted, meaning that a fraction of the points in a point cloud is missing, such as 5\%, 10\%, or 15\% of the total points. This scenario is realistic from both industrial and experimental perspectives. While training datasets for deep learning models are typically geometrically complete, with all points available or reconstructed prior to training, real-time applications during inference might involve incomplete data.

The objective of this subsection is to evaluate how robust the three deep learning frameworks, Flow Matching PointNet, Diffusion PointNet, and baseline PointNet, are in handling such incomplete data. Table \ref{Table9}, together with Figures \ref{Fig900} and \ref{Fig100}, illustrate the results of this analysis. When 5\%, 10\%, and 15\% of the points in each point cloud (e.g., 15\% of 1024 points) are randomly removed, the corresponding errors in the $x$- and $y$-components of the velocity vector (i.e., $u$ and $v$) and in the pressure variable (i.e., $p$) are listed in Table \ref{Table9}. As expected, increasing the percentage of missing points leads to higher prediction errors. Among the three models considered, the Flow Matching PointNet demonstrates the highest robustness, followed by the Diffusion PointNet, while the baseline PointNet exhibits the lowest robustness. When 10\% of the points are removed, the average relative pointwise error ($L^2$ norm) for predicting the $v$ variable is approximately 6.9\% for Flow Matching PointNet, 8.6\% for Diffusion PointNet, and 14.9\% for baseline PointNet. Notably, the error for baseline PointNet exceeds 10\%, which is generally considered unacceptable in engineering design contexts. When 15\% of the points are removed, the prediction error for the $v$ increases to about 8.3\% for Flow Matching PointNet, nearly 11\% for Diffusion PointNet, and over 17\% for baseline PointNet. Under these conditions, Flow Matching PointNet is the only model maintaining an error below 10\%.

Figures \ref{Fig900} and \ref{Fig100} show, respectively, the predictions obtained by the Flow Matching PointNet and Diffusion PointNet for the $u$, $v$, and $p$ variables when 15\% of the points are removed. It is important to note that some of the missing points are located on the surface of the cylinder. Mathematically, the velocity and pressure fields are functions of the cylinder's cross-sectional geometry. Consequently, the learning process of the deep network inherently depends on this geometry. Kashefi et al. \cite{kashefi2021PointNet} demonstrated that such boundary points are, in fact, critical within the PointNet framework and play an essential role in geometric deep learning. This finding further highlights the importance of robustness and underscores the superior performance of the Flow Matching PointNet in handling geometries with missing boundary points.




\section{Summary and outlook}
\label{Sect5}

In this study, we introduced, for the first time, the Flow Matching PointNet and Diffusion PointNet, which integrated the formulations of flow matching \cite{lipman2023flow} and diffusion models \cite{ho2020denoising} into PointNet \cite{qi2017pointnet,kashefi2021PointNet,kashefi2025kolmogorov} to generate velocity and pressure fields of fluid flows conditioned on the geometry of the computational domain. In these generative frameworks, irregular geometries were represented as point clouds. Independent standard Gaussian noise fields, corresponding to the velocity components in the
$x$- and $y$-directions as well as the pressure field, were concatenated with the spatial coordinates. In addition, sinusoidal positional encodings were concatenated with the spatial coordinates and noise fields to form the input features. A forward processing phase was then initiated, during which the Diffusion PointNet was trained to predict the noise components added to the velocity and pressure fields, while the Flow Matching PointNet, in a similar manner, predicted the derivatives of the noise with respect to a temporal variable. After training, inference was performed via a reverse processing phase, in which independent standard Gaussian noise fields associated with the velocity components and pressure were conditioned on an unseen geometry to generate the corresponding velocity and pressure fields. In Flow Matching PointNet, we used the explicit Euler numerical scheme during the reverse process.

To demonstrate the effectiveness of the proposed frameworks, the classical problem of fluid flow around a cylinder was employed as a benchmark. In this setup, the cross-section of the cylinder varied across different geometries, resulting in corresponding variations in the velocity and pressure distributions. These variations constituted the dataset that was used for training. The analysis of the predictions from the Flow Matching PointNet and Diffusion PointNet models revealed that their performance was significantly superior to that of the baseline PointNet framework. In particular, employing the Flow Matching PointNet and Diffusion PointNet models reduced the average relative pointwise error in the $L^2$ norm to below 10\%, whereas the error for the baseline PointNet model exceeded 10\%. Flow Matching PointNet and Diffusion PointNet demonstrated improved accuracy in predicting pressure-induced lift and drag forces. Furthermore, since the proposed frameworks generated predictions based on standard Gaussian noise inputs, multiple realizations were produced for any given geometry, enabling estimation of the uncertainty in model predictions. This capability was especially valuable for engineering design and safety factor considerations. Specifically, we demonstrated the uncertainty range associated with the maximum possible errors in pressure-induced drag and lift predictions. Additionally, we investigated the robustness of the Flow Matching PointNet and Diffusion PointNet models in cases where the geometry was incomplete (i.e., when a percentage of points in a point cloud was missing). The results showed that both models exhibited greater robustness compared to the baseline PointNet, with the Diffusion PointNet capable of maintaining the average relative pointwise error in the $L^2$ norm of the velocity and pressure fields below 10\% even when up to 15\% of the point cloud data were missing. Finally, although both the Flow Matching PointNet and Diffusion PointNet significantly outperformed the baseline PointNet, the overall difference between the two was not substantial. The Diffusion PointNet performed slightly better in predicting pressure, leading to more accurate pressure-induced drag and lift computations, whereas the Flow Matching PointNet demonstrated greater robustness when the geometry was partially missing.

We provide several directions for future extensions of the current research. One potential avenue is to investigate turbulent flow regimes, where uncertainty in the solution plays a critical role from both scientific and engineering perspectives \cite{liu2024uncertainty,valencia2025learning}. Another direction involves studying time-dependent flows, particularly those in which the geometry of the computational domain itself varies with time. Moreover, the proposed generative models can be further applied to the super-resolution and reconstruction of turbulent flow fields on unstructured grids. Additionally, while we introduced novel flow matching and diffusion models conditioned on the geometry of the computational domain, these models can be further extended to a hybrid conditioning approach, simultaneously conditioned on both geometry and textual prompts, thereby enabling their integration into foundation and large language/vision models \cite{bommasani2021opportunities,ramesh2022hierarchical,achiam2023gpt,kashefi2023chatgpt,team2023gemini,Kashefi2024Misleading} for computational physics.

\section*{Declaration of competing interest}
The author declares that he has no known competing financial interests or personal relationships that could have appeared to influence the work reported in this paper.

\section*{Acknowledgment}
The author wishes to thank the Stanford Research Computing Center.

\section*{Data availability}
The Python codes and data are publicly accessible on the following GitHub repository: \url{https://github.com/Ali-Stanford/Diffusion_Flow_Matching_PointNet_CFD}


\bibliographystyle{unsrt}
\bibliography{sample}


\end{document}